\documentclass[11pt,a4paper,twoside,openright]{report}

\usepackage[british]{babel}  
\usepackage{setspace}         

\usepackage{soul,color} 
\usepackage{booktabs}

\usepackage{paralist}         
\usepackage{colortbl}         
\usepackage[table]{xcolor}

\definecolor{LightCyan}{rgb}{0.88,1,1}
\definecolor{Red}{rgb}{1,0,0}
\definecolor{Green}{rgb}{0,1,0}

\definecolor{DarkGreen}{rgb}{0,0.6,0}

\definecolor{Blue}{rgb}{0,0,1}
\definecolor{Orange}{rgb}{1,0.45,0}

\definecolor{White}{rgb}{1,1,1}

\newcolumntype{w}{>{\columncolor{White}}c}

\usepackage{amsmath}          
\usepackage{amsfonts}         
\usepackage{amsthm}           
\usepackage{bm}               

\usepackage{graphicx}
\usepackage{float}
\graphicspath{{../figures/}{./figures/}{./figures/stochastic/}{../figures/stochastic/}}
\DeclareGraphicsExtensions{.pdf,.jpeg,.png,.eps,.jpg}
\usepackage{placeins}         

\usepackage[font=small]{caption}

\usepackage[T1]{fontenc}      
\usepackage[latin1]{inputenc} 
\usepackage{color}			  
\usepackage{eurosym}					

\usepackage[
pdfpagelabels,                
breaklinks,
plainpages=false,
pdfstartview=FitH,
hyperfootnotes=false,
hidelinks]{hyperref}         
\usepackage{breakurl}         

\usepackage{pdfpages}
\usepackage[printonlyused,nohyperlinks]{acronym}  					

\usepackage[numbers,sort&compress]{natbib}    
\usepackage[notbib]{tocbibind}        				
\usepackage[titletoc,page,header]{appendix}   
\usepackage{makeidx}         									
\usepackage{fancyhdr}            							
\usepackage[sf,bf]{titlesec}         
\usepackage[titles]{tocloft}      

\usepackage{cleveref} 

\usepackage{mathtools}
\usepackage{multirow}
\usepackage{threeparttable}
\usepackage{pifont}
\usepackage{amsfonts,amsmath,amssymb}
\usepackage{stfloats,psfrag}
\usepackage{algorithmic}
\usepackage{algorithm}
\usepackage{eqparbox}
\renewcommand\algorithmiccomment[1]{
	\hfill\ $\triangleright$ \ \eqparbox {COMMENT}\small{{#1}}}

\newcommand{\vcchange}[1]{\textcolor{black}{#1}}
\newcommand{\ehchange}[1]{\textcolor{black}{#1}}

\usepackage[percent]{overpic}
\usepackage[export]{adjustbox}

\usepackage{enumitem}

\usepackage{comment}
\usepackage[a4paper,
bindingoffset=1.8cm,
left=2cm,
top=1.8cm,
headheight=15pt,
hcentering,
vcentering,
includeheadfoot]{geometry}

\usepackage[numbers,sort&compress]{natbib} 
\usepackage[redeflists]{IEEEtrantools}

\definecolor{darkblue}{rgb}{0,0,0.5}
\definecolor{darkgreen}{rgb}{0,0.5,0}
\definecolor{darkred}{rgb}{0.5,0,0}
\hypersetup{
   pdftitle={Your-Name-PhD-Thesis},
   pdfauthor={Your~Name, Supervisor: Prof. Your SupervisorA, Prof. Your SupervisorB},
   pdfsubject={PhD Thesis},
   pdfkeywords={},
   pdfview=FitH,
   pdfstartview=FitH,
   bookmarksnumbered  
}

\usepackage{etex} 
\usepackage{color}
\usepackage{tikz}
\usepackage{chronology} 
\usepackage{units}            
\usepackage{tabularx}
\usepackage{threeparttable}
\usepackage{multirow}
\usepackage{xcolor}

\usepackage{footnote}
\usepackage{hhline}
\usepackage{subcaption}

\DeclareMathOperator*{\argmax}{\arg\!\max}

\usepackage{diagbox}
\usepackage{rotating}

\newcommand{\cmark}{\ding{51}}%
\newcommand{\xmark}{\ding{55}}%

\fancypagestyle{twosidemine}{
\fancyhf{} 
\fancyhead[RE]{\textsf{\nouppercase{\leftmark}}}
\fancyhead[LO]{\textsf{\nouppercase{\rightmark}}}%
\fancyhead[LE,RO]{\thepage}%
\fancyfoot[RE,LO]{\scriptsize Alberto Castagna}%
\fancyfoot[LE,RO]{\scriptsize PhD Thesis}
}

\fancypagestyle{bibmine}{\fancyhf{} 
\fancyhead[LO,RE]{\textsf{\uppercase{BIBLIOGRAPHY}}}%
\fancyhead[LE,RO]{\thepage}%
\fancyfoot[RE,LO]{\scriptsize Your Name}%
\fancyfoot[LE,RO]{\scriptsize PhD Thesis}
}

\fancypagestyle{blank}{\fancyhf{}
}

\makeatletter
\def\cleardoublepage{\clearpage\if@twoside \ifodd\c@page\else%
   \hbox{}%
   \thispagestyle{blank}
   \newpage%
   \pagestyle{mine}
   \if@twocolumn\hbox{}\newpage\fi\fi\fi}
\makeatother

\fancypagestyle{mine}{\pagestyle{twosidemine}}  

\theoremstyle{plain}
\theoremstyle{definition}
\theoremstyle{remark}

\renewcommand{\Gamma}{\varGamma}                   
\renewcommand{\hat}[1]{\widehat{#1}}               

\titleformat{\chapter}
{\sffamily\huge\bfseries}{\thechapter}{1em}{}

\makeatletter
\renewcommand{\@pnumwidth}{1.75em}
\renewcommand{\@tocrmarg}{2.75em}
\makeatother

\graphicspath{
{./}
{figures/}
{../figures/}
}

\makeindex   

\newcommand{\chapstar}[1]{%
\cleardoublepage \phantomsection%
\addcontentsline{toc}{chapter}{#1}%
 \chapter*{#1}%
\markboth{\uppercase{#1}}{\uppercase{#1}}%
\acresetall}

\newlist{objective-list}{enumerate}{1}
\setlist[objective-list,1]{label=\textit{O\arabic*} $-$, leftmargin=*}

\begin{document}
%
\bstctlcite{IEEEexample:BSTcontrol}

\begin{titlepage}

\begin{center}


\includegraphics{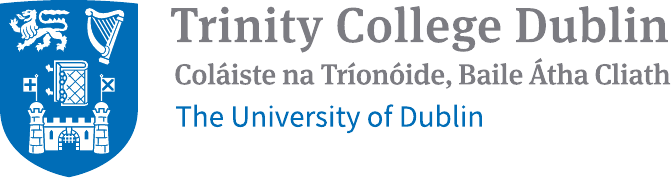}\\[2.5cm]


\textsc{\Large PhD Thesis}\\[0.5cm]

\newcommand{\HRule}{\rule{\linewidth}{0.5mm}} 

\HRule \\[1cm]
{ \huge \bfseries {Expert-Free Online Transfer Learning in Multi-Agent Reinforcement Learning}}\\[1cm]
\HRule \\[1.5cm]

\vspace{2cm}
\begin{minipage}{0.4\textwidth}
\begin{flushleft} \large
\emph{Author:}\\
\textsc{Alberto Castagna}
\end{flushleft}
\end{minipage}
\begin{minipage}{0.4\textwidth}
\begin{flushright} \large
\emph{Supervisor:} \\
Dr.~ \textsc{Ivana Dusparic\\}
\end{flushright}
\end{minipage}

\vfill

{\large \today}

\end{center}

\end{titlepage}
\index{text}

\doublespacing

\fancypagestyle{empty}{\pagestyle{twosidemine}}%
\fancypagestyle{plain}{\pagestyle{twosidemine}} %
\pagestyle{twosidemine}

\renewcommand{\chaptermark}[1]{%
\markboth{\thechapter.~\uppercase{#1}}{\thechapter.~\uppercase{#1}}}
\renewcommand{\sectionmark}[1]{%
\markright{\thesection~\uppercase{#1}}}

\cleardoublepage
\pagenumbering{gobble}

\begin{abstract}
	\ac{rl} enables an intelligent agent to optimise its performance in a task by continuously taking action from an observed state and receiving a feedback from the environment in form of rewards.  \ac{rl} typically uses tables or linear approximators to map state-action tuples that maximises the reward. 
	Combining \ac{rl} with deep neural networks~(\acs{drl}) significantly increases its scalability and enables it to address more complex problems than before. 
	However,~\acs{drl} also inherits downsides from both \ac{rl} and deep learning. 
	Despite~\acs{drl} improves generalisation across similar state-action pairs when compared to simpler~\ac{rl}~policy representations like tabular methods, it still requires the agent to adequately explore the state-action space.  Additionally, deep methods require more training data, with the volume of data escalating with the complexity and size of the neural network. 
	As a result, deep \ac{rl} requires \ehchange{a} long time to collect enough agent-environment samples and to successfully learn the underlying policy.
	\ehchange{Furthermore, often even a slight alteration to the task} invalidates any previous acquired knowledge.

	To address these shortcomings, \ac{tl} has been introduced, which enables the use of external knowledge from other tasks or agents to enhance a learning process\ehchange{. The} goal of~\ac{tl} is to reduce the learning complexity for an agent dealing with an unfamiliar task by simplifying the exploration process. This is achieved by lowering the amount of new information required by its learning model, resulting in a reduced overall convergence time.

	\ac{tl} approaches can be divided in \ac{t2t} and \ac{a2a} transfer. In \ac{t2t}, an agent with expertise in a specific task partially or totally reuses its learning model or belief to address a different novel previously unobserved task. 
	In~\ac{a2a}, an agent transfer part of its knowledge to a target agent addressing an equally defined task, hence with the same state-action domain and alike reward model. Based on the timing of transfer, \ac{a2a} can be further classified into online and offline.
	In online transfer, a novel agent may continuously access knowledge from another agent throughout its entire learning phase. On the other hand, in offline transfer, sharing happens exclusively at initialisation time.

	State-of-the-art approaches in online~\ac{a2a}~\ac{tl} follow teacher-student paradigm, in which expert agents transfer their expertise  to novices during training through advice-sharing following the teacher-student paradigm. 
	Knowledge transferred can influence either the action decision process of an agent or the learnt weight of an action. Having an optimal teacher to provide advice under the teacher-student framework leads to state of the art performance. In fact, effective transfer relies on the degree of expertise of the teacher. As the student improves its policy, the advice provided by the teacher may become outdated and may overwrite better policies learnt by receiving agents, i.e., a certain advice provided by the teacher although initially helpful, might prevent the agent from exploring better actions. 
	To mitigate the outdated advice shortcoming, previous work introduced advising strategies to regulate the transfer process assuming that an expert agent is used as \ehchange{the} source of advice. 	

	This thesis proposes \ac{efontl}, \vcchange{a novel framework for experience sharing. It enables online transfer learning in multi-agent systems enabling mutual online knowledge exchange between the learning agents by selecting the most suitable source of transfer at each time. As a result, each target agent receives a customised stream of knowledge tailored to its specific knowledge gaps. Thus, agents are expected to improve their performances by reducing the exploration phase, leading to faster convergence times. 
	\ac{efontl} is a novel framework for online transfer learning as it facilitates the reciprocal exchange of knowledge across multiple agents without the need for a fixed expert, unlike existing methods that share actions as advice. Furthermore, in \ac{efontl}, the target's performance is not capped at the teacher's expertise.}

	Without the presence of \ehchange{a} fixed known expert, successful transfer relies on agents correctly and fine-grainedly estimating their confidence in the knowledge samples that they do have. To this effect, this thesis also introduces a new epistemic uncertainty estimator \ac{sarnd}: an estimator based on full~\ac{rl} interactions. Compared to a state-visit counter, \ac{sarnd} enables fine-grained estimation during the training phase by taking into account additional information.

	We evaluate~\ac{efontl} across 4 different benchmark environments. 
	First, in three standard~\ac{rl} benchmarks environments of increasingly complexity: Cart-Pole, \acl{pp}, \acl{hfo}, and a real-world simulated environment~\acl{rs-sumo}.  	\ac{efontl} has demonstrated better or equal performance compared to the benchmark \ac{tl} baselines. We have observed that the degree of improvement correlates with the complexity of the environment addressed. In simpler environments, the improvement is relatively modest, while in more complex ones, the improvement is significantly greater.
	
\end{abstract}
\acresetall

\cleardoublepage \pagenumbering{roman}

\chapter*{Declaration}

I declare that this thesis has not been submitted as an exercise for a degree at this or any other university and is entirely my own work.

\vspace{0.5cm}

I agree to deposit this thesis in the University's open access institutional repository or allow the Library to do so on my behalf, subject to Irish Copyright Legislation and Trinity College Library conditions of use and acknowledgement.

\vspace{0.5cm}

I consent to the examiner retaining a copy of the thesis beyond the examining period, should they so wish (EU GDPR May 2018). 

\vspace{2cm}
\begin{tabular}{l}
Signed:\\
\end{tabular}

\vspace{1cm}
\begin{tabular}{l}
\makebox[2.5in]{\hrulefill} \\
Alberto Castagna, \today \\
\end{tabular}

\chapter*{List of Publications}

\begin{itemize}
	
	\item \cite{castagna2020demand} \textbf{Alberto Castagna}, Maxime Gu{\'e}riau, Giuseppe Vizzari, and Ivana Dusparic. Demand-responsive zone generation for real-time vehicle rebalancing in ride-sharing fleets. In 11th International Workshop on Agents in Traffic and Transportation (ATT 2020) at European Conference on Artificial Intelligence ECAI, 2020.
	
	\item \cite{castagna2021demand}\textbf{ Alberto Castagna}, Maxime Gu{\'e}riau, Giuseppe Vizzari, and Ivana Dusparic. Demand-responsive rebalancing zone generation for reinforcement learning-based on-demand mobility. AI Communications , 34(1):73--88, 2021.
	
	\item \cite{Castagna2022MultiAgentTL}\textbf{ Alberto Castagna} and Ivana Dusparic. Multi-agent transfer learning in reinforcement learning-based ride-sharing systems. In International Conference on Agents and Artificial Intelligence (ICAART), 2022.
	
	\item \cite{efontl23} \textbf{Alberto Castagna} and Ivana Dusparic. Expert-free online transfer learning in multi-agent reinforcement learning. European Conference on Artificial Intelligence ECAI, 2023.

	\item \cite{PPCengis} Cengis Hasan, Alexandros Agapitos, David Lynch,\textbf{ Alberto Castagna}, Giorgio Cruciata, Hao Wang, and Aleksandar Milenovic. Continual model-based reinforcement learning for data efficient wireless network optimisation.  European Conference on Machine Learning and Principles and Practice of Knowledge Discovery in Databases, 2023.

\end{itemize}



\chapter*{Acknowledgements}
	
First and foremost, I would like to thank Martina for taking a leap into the unknown with me in a foreign city two weeks before a global pandemic hit, and for her unconditional support throughout the entire journey and beyond. 

I am deeply grateful to my supervisor, Dr. Ivana Dusparic, who is truly one of a kind, for her guidance, encouragement, and availability throughout my doctoral studies. This thesis would not have been possible without her expertise, patience and insightful feedback.

I would like to extend my sincere gratitude to the members of my viva committee, Dr. Enda Howley, Dr. Vinny Cahill and Dr. Siobh\'{a}n Clarke, for their dedication to improving my work and for the pleasant discussion during the viva. I thank CRT-AI for funding my PhD, and Trinity College Dublin, particularly the DSG team, for welcoming and hosting me.

I extend my heartfelt gratitude to my family for believing in me and for their emotional and financial support throughout all these years of my academic journey.

Last but not least, I am thankful to all the people I have met during these years and who generously contributed their time, expertise, and insights to this research endeavour. In particular, I would like to thank Giuseppe, Maxime, Lara, Mohit, Teejay, Jose, Christian, Francesco and Giorgio.

\vspace{0cm}
\begin{center}
Alberto Castagna. \\
\end{center}


\tableofcontents
\listoffigures
\listoftables

\cleardoublepage%
\pagenumbering{arabic}%
\setlength{\parskip}{1ex plus 0.5ex minus 0.2ex} 

\chapter{Introduction}
\label{cpt:introduction}\acresetall%


\ac{rl} \vcchange{is a branch of machine learning that collects a set of algorithms} which enable an intelligent agent to learn an optimal behaviour in an unknown task via trial and error~\cite{sutton2018reinforcement}.
Initially, \ac{rl} \ehchange{was primarily used in academic settings} to demonstrate the feasibility of learning a small unknown task through tabular representation or linear approximators.

\begin{figure}[h!]
	\centering
	\includegraphics[width=0.6\columnwidth]{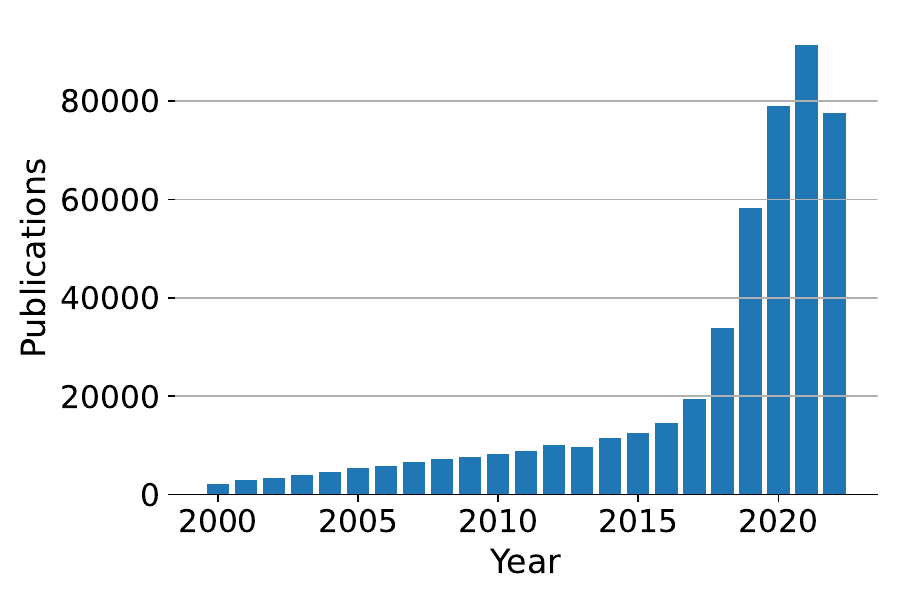}
	\caption{\ac{rl} research work published over the latest years (2000 - 2022) and available through Google Scholar search engine.}
	\label{fig:in_rl_pubs}
\end{figure}

However, with the latest advancements in deep learning, \ac{rl} has seen a sharp increase, as shown in Figure~\ref{fig:in_rl_pubs}, where the number of \ac{rl} publications has grown fourfold between 2017 and 2019,  and this growth trend has continued thereafter.  The figure shows the publication trend of \ac{rl}-based research work when looking for~\textit{"Reinforcement Learning"} through Google Scholar search engine.

In the latest years, \ac{rl} has been used to control agents in a variety of tasks, game-based or real-world tasks. Examples of applications include, robotics~\cite{zhao2020sim, kober2013reinforcement, polydoros2017survey},
financial~\cite{charpentier2021reinforcement, hambly2021recent}, asset management/allocation,
large-scale optimisation such as, traffic~\cite{wiering2000multi, wei2021recent, gueriau2020shared}, 5g networks~\cite{xiong2019deep,sciancalepore2019rl},
medical decisions support~\cite{zhou2021deep, coronato2020reinforcement, jonsson2019deep}, and natural language processing~\cite{sharma2017literature, uc2022survey}.

Despite the remarkable results collected in the aforementioned fields, \ac{rl} still struggles to be efficiently used in certain real world applications \vcchange{ for a variety of reasons, including the complexity of the application domain, settings characterised by non-stationarity, and the continuous evolution of the environment}. Next Section further discusses the main issues behind \ac{rl} applicability.

\section{Motivation}

\ac{rl} enables an intelligent agent to learn an optimal decision policy in an unknown task by evaluating its current status. 
Agent's optimal policy can be defined as a sequence of actions that lead the agent from an initial state to a final state while cumulating the highest possible reward. Such policy is learnt through interactions with an environment. Agent observes the current state, samples an action by following its policy, and finally receives a feedback that expresses how good was the action taken alongside the next observation~\cite{sutton2018reinforcement}. States and actions are well defined over observation space and action space, while reward has a dual impact on agent's policy, making it more challenging to model. Reward expresses the immediate return of a single action taken by the agent based on the current environment state. However, \ehchange{the} agent also needs to learn the long-term effect of a certain action through cumulative reward. For instance, an action might yield a low immediate reward but in the long term might lead the agent to achieve the best possible cumulated reward as the state is followed by other states that yield high rewards~\cite{sutton2018reinforcement}.

A major challenge intrinsic to \ac{rl} is the trade-off between exploration and exploitation~\cite{sutton2018reinforcement}. During exploration, \ehchange{an} agent samples an action randomly and observes the outcome. Exploration enables the agent to discover new state-action pairs that might yield a higher cumulative return. On the other hand, exploitation enables the agent to consolidate its belief by bootstrapping its current policy. The success of an agent lies in achieving the balance between these two stages.

An intuitive family of \ac{rl} algorithms are the \ac{td} learning methods. \vcchange{In these methods, updates to the agent's policy are based on the difference between the estimates at two successive time points}~\cite{sutton2018reinforcement}. 
For \ac{td} based \ac{rl} algorithms, policy can be represented through a table where entries are defined over states and actions.

One of the major shortcoming of \ac{rl} algorithms is the lack of generalisation and flexibility as policy maps state to actions without providing a way of modifying it~\cite{sutton2018reinforcement}, i.e., expanding observation or control space after a policy is learnt. Instead, if a change in the task is required, the agent will need to start the learning from scratch.
Furthermore, when moving towards real-world applications, the state-space complexity increases and two problems arise:

\begin{itemize}
	\item the increasing complexity of state-space makes tabular methods and linear approximators obsolete and moves the research towards the combination of \ac{rl} with non-linear approximators~\cite{sutton2018reinforcement}, such as \ac{nn};
		
	\item the combinatorial complexity of state-action space introduced a need to bootstrapping knowledge to facilitate the exploration process with ad-hoc instructions, aiding agents in converging towards a successful policy.
\end{itemize}

The coupling of \ac{nn} and \ac{rl} enables the applicability of \ac{rl} to more complex tasks, and allows \ehchange{a} generalisation of \ehchange{the} state-space based on past collected experience. As a result, \ehchange{an} agent does not need to explore all the possible state-action combinations to infer an optimal policy\ehchange{,} as it selects action\ehchange{s} based on past information collected across similar\ehchange{ly} encountered states~\cite{sutton2018reinforcement}. The combination of \ac{nn} and \ac{rl} shaped a new family of \ac{rl} algorithms: \ac{drl}.

\ac{drl} improves \ac{rl} thanks to its ability of generalising and approximate state-action space, however, it also inherits deep learning weaknesses.

The deep learning component in \ac{drl} makes the algorithm data hungry as it requires a significant amount of agent-environment interaction to learn effective policies. Furthermore, the architecture underneath might be wide and complex. Thus, a \ac{nn} needs substantial amounts of data to tune layer after layer and to capture patterns. Furthermore, given the amount of data and the number of updating steps to be done on the network, \ac{drl} requires a long time to learn a successful policy. 
Moreover, a change within the task definition invalidates all the acquired knowledge and requires the agent to learn again from scratch.

To overcome these issues, \ac{tl} offers a method of enhancing the traditional learning process with external knowledge. \ac{tl} is a versatile and very broad concept, allowing the integration of external knowledge in various ways within the \ac{rl} framework. Typically, the flow of transferred knowledge goes from a source agent, which provides advice or knowledge, to a target agent, which learns the task by leveraging the shared information from the source agent. Some applications of \ac{tl} include:

\begin{itemize}
	\item using a learnt policy in a novel task~\cite{rusu2016progressive, schwarz2018progress, wang2018deep, d2020sharing, devin2017learning,kulkarni2016deep, kirkpatrick2017overcoming, glatt2016towards, yin2017knowledge, rusu2015policy, traore2019discorl, barreto2017successor};
	\item distilling policies from multiple tasks into a single master policy able to achieve similar performance~\cite{rusu2015policy, traore2019discorl, yin2017knowledge};	
	\item pre-training an agent with selected experiences prior to allowing the agent to interact with the environment online~\cite{hester2018deep, lazaric2008transfer, walke2023don};
	\item enhancing the exploration phase by letting an expert agent act as a teacher to a novice~\cite{taylor2014reinforcement, da2020uncertainty, zhu2020learning,norouzi2021experience,torrey2013teaching, zhu2021q,liang2020parallel,  da2017simultaneously,taylor2019parallel,liu2022,subramanian2022multi};
	\item mitigating the exploration cost by enabling real-time knowledge sharing across multiple agents~\cite{ilhan2019teaching, gerstgrasser2022selectively}.
\end{itemize}

In \ac{rl} the types of knowledge or objects that can be transferred to expedite the learning process, and thus reducing the learning time, depends on the specific transfer objective. For instance, when an expert agent is available to provide on-demand guidance to a novel agent, object transferred is usually an action that is expected to yield a high cumulative reward. Furthermore, the object transferred could be a feedback, given from expert supervisor to the novel agent, which augments the standard reward returned by the environment. The provided signal reflects the expert's estimation of the expected return for a specific action in a given step, based on their knowledge and experience.

In tasks that no agent has previously encountered, there are no experts capable of providing feedback or tutoring novel ones. However, in situations where multiple agents are addressing the same task, they can share selected knowledge, leading to a reduction in the exploration cost for each individual agent.

\subsection{Multi-Agent Systems}
\ac{mas} consists of multiple intelligent agents interacting with an environment while simultaneously learning a policy~\cite{10.5555/520715}. These agents can learn independently or collaborate with others in either a centralised or decentralised manner. Depending on the environment's space and the number of agents involved in the task, the communication may be limited to a certain number of agents, taking into account factors such as their distance and communication cost~\cite{Chu2020marlTrafficSignalControl,das2019tarmac,jiang2018learning, kar2013cal}. In a centralised setting, each agent shares information during training and contributes towards the update of a common policy~\cite{rashid2020monotonic, sunehag2017value, lowe2017multi, matta2019q}. 
When learning independently, each agent optimises its policy merely upon its collected experiences and without any additional interactions with other agents~\cite{tan1993multi, schulman2017proximal}. Independent learning mitigates scalability and communication issues as the learning is distributed and independent, thus reducing the need for synchronisation processes across agents.

On the other hand, \ac{marl} algorithms that enable agents to collaborate achieve usually better performance in multi-agent system at a higher communication cost to collaborate, i.e., by sharing data to be used by a common learning process~\cite{wang2020data,moradi2016centralized, chen2019new, foerster2018counterfactual, matta2019q} or policy parameters to be merged into a centralised one~\cite{khan2018scalable, gupta2017cooperative}.

In real-world scenarios, the n-to-1 communication cost in centralised approaches makes them unsuitable. The time and resources required for agents to communicate and process information centrally result in impractical and lengthy training times to achieve a satisfactory policy. 

\vcchange{
	Communication in \ac{efontl} relies on n-to-n interactions among agents, potentially posing a weak link due to the complexity and resource-intensive nature of such communication requirements.
}

\vcchange{
	This thesis assumes fault-free transmission between agents, disregarding potential breakdowns in communication channels to ensure seamless data exchange and collaboration among all system components. Furthermore, in practice, communication may be restricted to neighbouring agents, significantly reducing the overall communication cost and the likelihood of breakdowns in the transmission channels. Finally, should communication disruptions occur, whether due to faulty agents or unreliable transmission channels, \ac{efontl} is designed to ensure that each agent retains the ability to continue its learning task autonomously.
}

\section{Research Questions}\label{sec:research_questions}
Sharing advice in the form of action from a source agent with an optimal policy, which guarantees the best possible options in any situation, enhances the performance of a target agent and encourages the fulfilment of a goal. 
However, overriding the policy of a target agent might not be the best option when it is not guaranteed that source has an optimal policy or when no expert is available. 

In fact, when a suboptimal agent acts as source of advice, it might limit the target performance on the long term as the provided advice might not lead the target agent to an optimal decision.

Generally, \ac{rl} is applied to unknown tasks, and therefore, it is not very common to have an agent with a policy that achieves a satisfactory level of performance, let alone an oracle agent that consistently outperforms a learning agent at any step of training. 
However, if multiple agents are learning the same task simultaneously, they can share knowledge with each other as they learn, thus improving their performance and leveraging their local expertise.

Ilhan et al.~\cite{ilhan2019teaching} have shown a reduced training time by enabling the sharing of actions in \ac{mas} where no-expert is available. When an agent seeks guidance, other agents provide action-advice based on the target's state and their individual knowledge.

However, on top of communication and synchronisation cost, overriding the target's policy by forcing it to follow an action suggested by others might not be the best option and could result in delaying the target to converge.


To prevent external agents interfering with a target agent exploration process, this thesis investigates the impact of transferring selected agent-environment interactions in \ac{mas}. Agents are homogeneously defined over state and action space with same goal. Furthermore, agents have equal capacity and ability to learn a policy within the given task. When transferring online, no general expert can be identified during the learning cycle. In these settings, overriding the policy of an agent might result in a delay in the learning process. Thus, following the intuition of pre-training a learning model with selected pre-collected agent-environment tuples, which has already shown a reduced overall training time~\cite{hester2018deep, lazaric2008transfer, walke2023don}, this work enables the share of experience batches from a temporary teacher. 
Once teacher is identified, the remaining agents have their own knowledge gaps. As a result, target agents need to isolate incoming experience that is expected to fill their current policy gap.

Therefore, this thesis investigates the transfer of agent-environment interactions as a means to enhance the overall system performance in \ac{mas}, i.e., reducing training time and improve agents performance.  Specifically, the research questions are:

\begin{itemize}

	\item \textbf{RQ1} - \vcchange{\textit{Can, and if so to what extent, online transfer learning through sharing of experience across homogeneous agents with no fixed expert contribute to improving the system performance?}}	
	\item \textbf{RQ2} - \textit{What criteria can agents use to identify the suitable agent to be used as source of transfer?}
	\item \textbf{RQ3} - \textit{What criteria should an agent use to filter incoming knowledge?}
\end{itemize}

To fulfil the research aims, the main objectives of this thesis are the following:

\begin{itemize}
	\item develop a common methodology that enables the selection of an agent, to be used as source of advice, from multiple candidates that has learnt an established policy; 

	\item Investigate multiple criteria that an agent could use to filter incoming knowledge and identify the tuples expected to bridge the gap between the target policy and that of another agent;

	\item investigate the performance impact of transferring agent-environment tuples to a target agent while varying the frequency and the quantity of shared tuples across diverse environments.
\end{itemize}

With solid and well-defined research questions to guide this research work and having identified the primary objectives that drive this thesis, the following section introduces the significant contributions that have resulted from this study.

\section{Thesis Contribution}
The main contribution of this thesis is \ac{efontl}, an expert-free transfer learning framework for online experience sharing between agents in multi-agent context. \ac{efontl} enables \ac{rl}-based agents to reduce exploration complexity and hence learn more quickly by sharing knowledge gathered in different parts of the system. Transferred object consists of a subset of experiences, which are past agent-environment interactions, specifically represented as ($s_t,a_t,r_t,s_{t+1}$) tuples. During each transfer iteration, an agent is selected as source of transfer based on \ac{ss}, and a portion of its collected agent-environment interactions is transferred to other agents. Finally, a target agent samples a batch of shared tuples according to a collectively defined \ac{tcs} criteria computed across source and target agent.

This thesis relies on two criteria to identify worthy experience to transfer, \textit{uncertainty} and \textit{expected surprise}.
Uncertainty is independent from the learning form used and aims to analyse the epistemic confidence of an agent. Assuming that all agents use a common uncertainty estimator methodology, the discrepancy between source and target uncertainty, can then be used as metric to identify relevant tuples.
Therefore, to estimate epistemic uncertainty from a certain agent-environment tuple, this thesis introduces~\ac{sarnd} an extension of~\ac{rnd}~\cite{burda2018exploration}. On the other hand, expected surprise~\cite{white2014surprise} is defined over target agent and is approximated through Temporal Difference error~(TD-error).

\vcchange{
\ac{efontl}, the online transfer learning framework introduced in this thesis, represents a novel approach by enabling the reciprocal exchange of knowledge among multiple agents without the need for a fixed expert. This mutual exchange of knowledge uniquely positions it apart from current state-of-the-art methods that facilitate online transfer learning, which generally do not support such dynamic knowledge exchange. Furthermore, while the sharing of experiences is a common practice in offline transfer learning, online contexts generally prefer the sharing of actions.}

\vcchange{In contrast to traditional action-advice methods that rely on a fixed expert, where the transfer is typically focused on either the source or the target, \ac{efontl} introduces a more adaptive and responsive transfer mechanism. The knowledge assimilated by an agent through \ac{efontl} is precisely tailored to its specific knowledge gaps, as the algorithm dynamically balances between the source's expertise and the target's needs.}

Without the presence of \ehchange{a} fixed known expert, successful transfer relies on agents correctly and fine-grainedly estimating their confidence in the knowledge samples that they do have. To this effect, this thesis also introduces a new epistemic uncertainty estimator \ac{sarnd}: an estimator based on full~\ac{rl} interactions. Compared to a state-visit counter, \ac{sarnd} enables fine-grained estimation during the training phase by taking into account additional information.

\section{Evaluation}

\ac{efontl} impact has been evaluated across 4 different environments of increasing complexity: 
\begin{itemize}
	\item Cart-Pole $-$ to evaluate its impact in a simple minimalistic environment with independent agents~\cite{openAIgym2016};
	\item \ac{pp} $-$ to evaluate its impact into a collaborative but yet competitive \ac{mas} where predators are divided into two teams and a single team is enable to collaborate through experience sharing~\cite{MinigridMiniworld23};
	\item \ac{hfo} $-$ a complex environment with continuous action space where agents have to collaborate to fulfil their team's goal~\cite{hausknecht2016half};
	\item \ac{rs-sumo} with real-world data $-$ a simulation of a real-world case study replicating the ride-requests served by yellow and green taxi within the Manhattan area~\cite{NYC_data}. This environment is unique as long term reward is stronger than immediate and it is hard to establish the quality of an action. Furthermore, the reward function only captures partially the performance of an agent. Thus, an action might seem counterproductive in the short term but might lead the agent to cover more ride-requests while optimising the vehicle usage. 
	The simulator is built upon \ac{sumo} to replicate the same network infrastructure and simulate realistic traffic patterns.
\end{itemize}


As a preliminary case study, this thesis investigates the offline transfer learning scenario in two environments: \ac{pp} and \ac{rs-sumo}. Offline transfer has a reduced set of challenges to be addressed when compared to online~\ac{tl}. The transferred object consists of a sub-set of uncertainty-labelled agent-environment tuples collected by trained agents during their training phase. Subsequently, a new set of agents, referred as target agents, spawn and sample a batch of selected experiences from the given buffer before starting their \ac{rl} training phase. The primary objective of this study is to evaluate the feasibility of positive transfer assuming the availability of an expert and  examining the relation between quality of experience, based upon uncertainty, and the transfer outcome.

After demonstrating the feasibility of positive transfer from expert agent, the focus of this thesis shifts towards~\ac{efontl} in the online scenario, where no fixed expert is available.
To address the limitation of \ac{rnd} in the online scenario, \ac{sarnd} is introduce. Subsequently, an extensive evaluation study is conducted on Cart-Pole and \ac{pp} environments to validate whether the findings from offline~\ac{tl} hold when transitioning to the online setting. The experiments assess various criteria for filtering incoming knowledge on the target agent and investigating the impact of selecting the source of transfer based on different criteria.
Finally, the effectiveness of \ac{efontl} is validated against to selected baseline methods on more complex and real-world scenario, \ac{hfo} and \ac{rs-sumo}.


To benchmark the impact of~\ac{efontl}, this thesis compares the proposed method against the following \ac{tl} baselines:
\begin{itemize}
	\item \textit{No-Transfer} $-$ where agents learn independently without knowledge sharing;
	
	\item \ac{ocmas}~\cite{ilhan2019teaching} $-$ where agents share action-advice. The most uncertain agent seeks guidance from others by asking for an action-advice. Consequently, the advice seeker takes the action by majority voting. This baseline is used to compare the impact of overriding target's policy with actions versus sharing selected experience through~\ac{efontl};
	
	\item \ac{rcmp}~\cite{da2020uncertainty} $-$ where an expert agent is available on-demand to guide a novel agent during its exploration phase. Target agent estimates its epistemic confidence to decide whether to ask for advice.	This baseline is used to assess the impact of transferring from imperfect learning agents, selected dynamically through~\ac{efontl}, versus fixed expert agents.
\end{itemize}

While \ac{ocmas} and \textit{No-Transfer} work as upper and lower limit to benchmark \ac{efontl} against, \ac{ocmas} should achieve similar level of performance. Although, \ac{ocmas} has a greater communication cost as it requires all the agents to synchronise at each decision-step to verify whether an agent is the one with lowest uncertainty within the group.

\section{Dissemination}

As part of the dissemination process, the findings of this research were presented in various venues. 

Starting with Castagna et al.~\cite{castagna2020demand}, presented at the $11^{th}$ internation workshop on Agents in Traffic and Transportation~(ATT 2020) at ECAI 2020 conference, and its extended journal version Castagna et al.~\cite{castagna2021demand}, published in AI Communication. These two publications serve as motivation work and groundwork to investigate the optimisation problem of mapping ride-requests to vehicles while enabling ride-sharing and \ac{tl}.  Specifically, we had $200$ vehicles utilising \ac{rl} agents to learn, which gave rise to intuition that we need a knowledge exchange mechanism for agents to enhance each others learning since all were learning to accomplish the same task. Furthermore, our research  aimed to develop a scalable solution that could allow for the dynamic allocation of vehicles and could be resilient to minor changes in the road-network and demand patterns. This problem sparked our interest towards \ac{tl} to enable the transfer of piece of knowledge or entire policies to new agents to adapt to novel scenarios.

Castagna and Dusparic~\cite{castagna2021multi} in 2022, published in the proceedings of the $14^{th}$ Internation Conference of Agents and Artificial Intelligence~(ICAART), discussed the preliminary evaluation on experience sharing in an offline context, Chapter~\ref{cpt:preliminary_studies} is almost based on this publication. 

The publication that mostly fully encompasses final contributions of this thesis is Castagna and Dusparic~\cite{efontl23}, published in the proceedings of the $26^{th}$ European Conference on Artificial Intelligence~(ECAI 2023). The core contribution of this thesis, \ac{efontl}, and presented a subset of the evaluation reported in Chapter~\ref{cpt:evaluation}. 

Another publication related to \ac{tl} but not specifically \ac{efontl}, resulted from my PhD internship in the Intelligent Automation team at the Huawei Ireland Research Center. This collaboration led to a research paper, Cengis et al.~\cite{PPCengis}, published in the proceedings of the European Conference on Machine Learning and Principles and Practice of Knowledge Discovery in Databases (ECML PKDD 2023). This paper demonstrates a continual learning approach to facilitate policy adaptation in wireless networks with an expanded control space.

\section{Roadmap}
To summarise, this chapter introduced \ac{tl} as a means to mitigate some of the weaknesses typical of \ac{drl} models, i.e., the lack of generalisation and extensive training time required to adequately explore an environment and learn an effective policy. In detail, \ac{efontl} enables the sharing of selected experiences among a set of imperfect agents that are learning a certain task simultaneously. 
The following chapters of this thesis are structured as follows: in Chapter~\ref{cpt:background}, a comprehensive exploration into the background of \ac{rl} and \ac{tl} is given, presenting open challenges and the latest research work in these areas. Subsequently, before introducing the design details of~\ac{efontl}, Chapter~\ref{cpt:preliminary_studies} shows a feasibility study, an offline transfer context, by presenting preliminary results to assess whether experience sharing can effectively improve target performance. Further, Chapter~\ref{cpt:efontl} introduces and discusses \ac{efontl} while providing details about the design choices made in its development. Then, Chapter~\ref{cpt:implementation_details} intricately examines the foundational technologies that underpin~\ac{efontl}, and provides an insightful analysis of the simulators employed to comprehensively evaluate its capabilities. To follow, Chapter~\ref{cpt:evaluation} introduces the evaluation objectives and the evaluation studies. It meticulously analyses the findings in relation to the research questions posed in the thesis. Finally, Chapter~\ref{cpt:conclusion} concludes this thesis by summarising its contributions and outlining open challenges that need to be addressed in future work.

\chapter{Background and Related Work}
\label{cpt:background}\acresetall%

This chapter is organised in three main sections, each covering a specific aspect of the background work addressed in this thesis. Firstly, Section~\ref{sec:back_rl} presents the background of~\ac{rl} and deep \ac{rl} introducing also the challenges and potential limitations when applied to complex tasks. Then, Section~\ref{sec:back_tl} reviews transfer learning to overcome the limitations of deep~\ac{rl}. Finally, Section~\ref{sec:back_uncertainty} concludes the chapter by clarifying the concept of uncertainty in \ac{rl} and reviewing the techniques used to estimate uncertainty in \ac{rl}.

\section{Reinforcement Learning}\label{sec:back_rl}

A \ac{mdp} is a stochastic sequential control process used to model decision-making~\cite{puterman1990markov} based on the Markovian property which states that the probability of going from the current state at time $t$ to another at time $t+1$ is independent by the history of previous visited states: $P(s_{t+1}|s_{t}, s_{t-1},\dots,s_{0})= P(s_{t+1}|s_{t})$.

A \ac{mdp} is defined as a 5-tuple $(S,A,T, R, S_0)$.
\begin{itemize}
	\item $S$ $ - $ is a collection of states that could be visited.  $S$ could be finite or not;
	\item $A$ $ - $ is a set of actions and can be either finite or not;
	\item $T$ $ - $ $T: P(s_{t+1}|s_t, a_t)$ defines the transition probabilities between states, specifically, the probability that an action $a_t$ leads to $s_{t+1}$ from a given state $s_t$;
	\item $R$ $ -$ $R(s_t, a_t, s_{t+1})$ expresses the immediate reward for transitioning from $s_t$ to $s_{t+1}$ through $a_t$;
	\item $S_0$ $ - $ starting state set is the set of possible starting states.
\end{itemize}

Based on the observed state, a \ac{mdp} can be classified as \textit{Fully-Observable}, when the state representation passed to the agent is accurate and reliable~\cite{sutton2018reinforcement}. \textit{\ac{pomdp}}, when an observation is an estimation of the current state~\cite{kaelbling1998planning}.

As a \ac{mdp} is a sequential decision process, the action-decision process over a sequence of consecutive time steps can be defined as~\ac{horizon}~\cite{sutton2018reinforcement}. Based on the length of~\ac{horizon}, a \ac{mdp} can be categorised into three categories: 1) finite-horizon tasks, where the number of action-decision is fixed to a certain number of time steps; 2) indefinite-horizon tasks, where the number of time steps are upper limited by a certain fixed value; and 3) infinite-horizon task, in which interaction is continuous and does not terminate.

For horizon of length $n$, it exists an associated \textit{Discounted Reward~($R_n$)}. $R_n$ is the sum of immediate rewards~$r_k$, received at time step~$k$, weighed by a discount~$\gamma \in [0,1]$, hence: $R_n = \sum_{k=0}^{n-1} \gamma^{k}r_k$. $\gamma$ regulates the impact of future reward versus the greedy reward received at the current time step. 

When $\gamma$ is close to $0$, the immediate reward holds more significance, and the influence of future rewards diminishes significantly. As a consequence the decision process prioritises greedy choices. On the other hand, when $\gamma$ is close to $1$, future rewards weigh more into the decision process and lead to prioritise a long-term planning that aims to maximise the overall return.

A \ac{mdp} provides a mathematical framework to model tasks that require sequential decision-making. Among the requirements to define a task as a \ac{mdp} is the availability of both the transition function~$T$ and the reward function~$R$ that define how the environment responds to the actions taken by an agent. However, transition function is often unknown. To overcome this limitation, \ac{rl} enables an agent to learn the transition function by randomly interacting with the environment. 

\begin{figure}[h!]
	\centering
	\includegraphics[width=0.6\columnwidth]{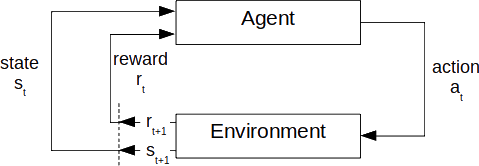}
	\caption{Single step of Reinforcement Learning process~\cite{sutton2018reinforcement}.}
	\label{fig:rw_rl_process}
\end{figure}

\ac{rl} is a cyclic four-step process shown in~Figure~\ref{fig:rw_rl_process}. These four steps are repeated at every time step~$t$:

\begin{enumerate}
 \item Observation $-$ an agent observes the surrounding environment and along with its internal state maps it to the current state~$s_t$;
 
 \item Action Selection $-$ the agent selects an action~$a_t$ to take based on perceived state; 	
 
 \item Actuation $-$ the agent takes a step within the environment by executing~$a_t$ sampled earlier;
 
 \item Evaluation $-$ an agent observes the updated observation  of the state $s_{t+1}$ alongside a signal~$r_{t+1}$ that expresses how good was the action $a_t$ taken from state $s_t$. 
 
 \end{enumerate}

\ac{rl} can be categorised as \textit{model-based} and \textit{model-free}~\cite{sutton2018reinforcement}:
\begin{itemize}[]
	
	\item  model-free $-$ it enables an agent to learn a policy just upon the experience~$(s_t, a_t, r_t, s_{t+1})$ collected by interacting with the environment;
	
	\item model-based $-$ it enables an agent to learn a model of the environment, including the transition dynamics and the reward function, and use that model to plan actions and maximise the expected reward from the taken action. Hence, model-based learns by planning~\cite{sutton2018reinforcement}.

\end{itemize}

This thesis focuses on model-free \ac{rl}.

Learning in \ac{rl} can be represented as \textit{value}-based and \textit{policy}-based.
Value-based~\ac{rl} uses agent-environment interactions to update a single value, e.g., table entry, representing the state-action combination. When exploiting its internal policy, an agent selects the best action as the one with highest value across all actions for the current state.
On the other hand, in policy-based \ac{rl}, an agent uses experience to optimise hidden parameters of a function that maps states to actions through probabilities. Over time, an agent updates these parameters to maximise the expected reward via gradient-based optimisation methods.
For a certain state, an agent will then follow an action with maximum probability.

\vcchange{\textit{Actor-Critic} methods combine both policy and value representations.
These algorithms are composed by two parts: an actor, which maps states to actions and is used to select an action from a state; and a critic, which provides an estimation of the state-value, representing the expected discounted cumulative reward obtained by following the policy from the observed state~\cite{konda1999actor}.}

During the learning process, an agent needs to balance two phases: \textit{exploration} and \textit{exploitation}. While exploring, the agent samples random actions to visit new state-action combinations. On the other hand, exploitation enables the agent to strengthen its belief by leading the agent to follow its policy.
At the initial stage of the learning phase, an agent follows merely a trial-and-error approach. Over time the probability of taking random actions decreases in favour of exploiting the learnt behaviour.


One of the ways to exploring new state-action pairs is by following an $\epsilon$-greedy action selection~\cite{sutton2018reinforcement}, i.e., an agent samples an action by following its policy with a $1 - \epsilon$ probability and a random action otherwise. $\epsilon$ slowly decays to $0$ over time reducing the probability of exploring in favour to exploiting.
$\epsilon$ anneals by the ratio between the difference across starting $\epsilon$~($\epsilon_{start}$) and ending~($\epsilon_{end}$) and the total number of episodes that the agent will undertake, hence $\frac{\epsilon_{start} - \epsilon_{end}}{episodes}$. Generally, $\epsilon$ is initialised to a value close to $1$ and decays towards $0$, e.g, $0.95$ to $0.05$.

For a \ac{mdp} with low-dimensional state and action space, \ac{rl} can be expressed using a tabular representation where rows and columns are defined over possible states and actions. Their intersections are updated based upon the value returned by the environment with Eq.~\ref{eq:rw_ql}.
This family of \ac{rl} algorithms is called \textit{tabular methods} and one popular method is \textit{Q-learning}~\cite{watkins1992q}.

\begin{equation}
\label{eq:rw_ql}
	Q^{}(s_t,a_t) = (1- \alpha) Q(s_t, a_t) + \alpha \Big( r_t + \gamma \max_aQ(s_{t+1},a) \Big)
\end{equation}
\textit{Learning rate} $\alpha$ weights the impact of new knowledge~($ r_t + \gamma \max_aQ(s_{t+1},a)$) against the current agent knowledge ($Q(s_t, a_t)$). $\alpha$ is defined in the interval of $[0,1]$. When $\alpha$ is $0$, the knowledge of the agent is not influenced by incoming information. On the other hand, setting $\alpha$ to $1$ the agent fully overrides its knowledge with the newly information retrieved. 
 $\gamma$ is the \textit{discount factor} and is defined within the $[0,1]$ interval. It weighs the impact of immediate reward and future rewards.

Given a state and an action, its \textit{Q-value}~($Q(s,a)$) is updated by a value computed over the received reward at time $t$ and the maximum possible Q-value from the following state.

\ac{rl} has demonstrated significant potential in tackling various tasks in different domains, including robotics~\cite{schaal1994robot, mahadevan1992automatic, gaskett2002q, mataric1997reinforcement}, game playing~\cite{thiam2014reinforcement,erev1998predicting}, optimisation tasks~\cite{gueriau2018samod, wiering2000multi}. However, there are a number of open challenges that limit the applicability of \ac{rl}. The following section introduces the open challenges in \ac{rl}.

\subsection{Reinforcement Learning - Challenges}
\ac{rl} has proved to be a viable approach to learn a successful policy in tasks where feedback is available to judge single-step action. 
However, a number of challenges remain, some of which are outlined below:
\begin{itemize}
	\item \textit{Exploration-Exploitation trade-off} $-$ Exploration and exploitation are fundamental processes for \ac{rl} and a right balance is required to achieve a nearly optimal policy. The agent needs enough exploration to visit action-state combinations and to observe their outcome. On the other hand, the agent needs to exploit its policy to strengthen the learnt knowledge by taking the sequence of actions, expected to be optimal, and to observe the long term horizon outcome.
	A right trade-off is required depending on the ongoing learning phase. At the beginning, an agent is expected to prefer exploring than exploiting while eventually exploitation takes the lead. Balancing these two phases is challenging as it depends on the complexity of state-action space combination.

	\item\textit{Reward Function Design} $-$ A reward function, or reward model, supports a \ac{rl}-agent to learn a policy by providing feedback after a single-step interaction. In the long-term, an agent is expected to learn a policy that maximises the long-term cumulative reward. Thus, a reward model plays a crucial role towards the definition of optimal policy. 
	While these functions are usually well defined in games and benchmarking environments, i.e., OpenAI Gym~\cite{brockman2016openai}, when it comes to real-world, it is very likely that a reward function is not available. For instance, in a ride-sharing enabled mobility on-demand system, the reward model could  prioritise the revenue maximisation, could punish long delays due to re-routing and enabling ride-sharing or could support the vehicle utilisation. Depending on the desired goal, the reward function should be designed to lead the agent towards the accomplishment of the primary goal while having a fair trade-off on the secondary goals. Thus, designing a proper reward function demands careful attention and deep knowledge of the addressed task~\cite{dulac2021challenges}.
	
	\item\textit{Sparsity of Rewards} $-$ Sparse-reward tasks provide feedback to an agent very infrequently, and in a lot of cases only upon the accomplishment of a goal. Consequently, for most state-action pairs, the agent will not receive any feedback. To expedite the learning process, it may be better to enhance the reward model by providing earlier feedback.
		
	\item\textit{State-Action Dimensional Complexity} $-$ Complexity of a task is directly influenced by its state and action space~\cite{dulac2021challenges}. A larger set of actions, requires the agent to explore an increased number of state-action combinations.
	\item\textit{Lack of Generalisation} - \ac{rl} optimises a policy to a specific task. A change within any of the \ac{mdp} tuple is very likely to invalidate some or even all acquired experiences. Furthermore, agents must learn a generalised policy allowing for various environment configurations. For instance, considering a mobility on-demand context where a \ac{rl} agent controls a vehicle aiming to maximising the number of passengers served. In this scenario, the agent needs to learn a policy that can effectively adapt to different demand patterns, ensuring optimal performance across a diverse range of request distributions and traffic conditions.
	\item\textit{Exploration Cost} $-$  On top of the exploration-exploitation trade-off, exploring towards a real-world task can be dangerous and expensive. When exploring actions in a real-world a physical agent requires certain time to fulfil an action and observe the outcome, introducing a delay,~\cite{dulac2021challenges}. Furthermore, safety concerns need to be taken into account when performing arbitrary actions. A failure might lead the agent to damage its system or external entities~\cite{dulac2019challenges}.
	
	\item\textit{Partial-Observability and Non-Stationarity} $-$ Real-world sensors might provide information that are not fully reliable because of environment conditions or malfunctioning~\cite{dulac2021challenges}. Furthermore, real-world systems are often influenced by external factors or stochastic elements~\cite{dulac2021challenges}. Therefore, the transition dynamics of the environment might change over time leading the task to be non-stationary.	
\end{itemize}

Tabular representation for \ac{rl} limits the applicability to domains with low-dimension state-action spaces due to the combinatorial complexity. However, discretisation functions are often combined with tabular~\ac{rl} to enable the handling of high-dimensional and continuous state-action spaces. As a result, the combinatorial complexity is lowered and tables can be used to fit the state-action combinations. Nevertheless, that might still not be sufficient to learn a successful policy for complex tasks, for example, a self-driving car.

Tabular \ac{rl} lacks generalisation across similar state-action pairs as each update within the model is uniquely related to a specific action and state combination. To overcome the generalisation limit, intrinsic in tabular representations, researchers have focused on combining \ac{nn} and \ac{rl}.

\subsection{Neural Networks}

An Artificial neural network is a learning model inspired by a human brain. The first tracked attempt towards their development is dated back to 1943. Warren McCulloch and Walter Pitts~\cite{mcculloch1943logical} proposed a model of a neuron and demonstrated how multiple neurons could be interconnected to perform simple logical operations. However, the wider development and use of \ac{nn}s started from 1986 when Rumelhart, Hinton and Williams~\cite{rumelhart1986learning} introduced a function to update an internal representation by back-propagating errors. From early 2000s, \ac{nn}s have gained popularity due to their successful applications in various topics, including classification, clustering, patter recognition and prediction across multiple sectors~\cite{abiodun2018state}.

\ac{nn}s are composed of interconnected layers, with each layer containing a specific number of neurons.  Each neuron computes a linear function as follow:
\begin{equation}
	\label{eq:rw-dense_neuron}
	y= w*x + b
\end{equation}

The bias term~$b$ is a constant value that allows the neuron to have an effect even when the input signal $x$ is $0$. $w$ is a scalar value that weighs $x$ and $y$ is the output of the neuron. $w$ and $b$ are adjusted during the training process based on the output of the network. The output of a layer is defined as collection of individual neuron outputs belonging to that specific layer.

The output of a layer is processed through an activation function to introduce non-linearity and enable the network to learn complex relationships. Some of the most common are \textit{ReLU}~\cite{glorot2011deep}, \textit{TanH}~\cite{hinton2006fast}, \textit{Sigmoid}~\cite{lecun1998gradient} and \textit{SoftMax}~\cite{bridle1990probabilistic}. 
Each activator applies a function and decides whether a neuron output is active or not. For instance, the \textit{ReLU} activation function clips any negative value to $0$ and leaves any positive input unchanged.

A \ac{nn} is a sequence of layers and activation functions. Commonly the computational flow follows a sequential direction from first layer to the second and so on to the output layer. That is known as \textit{forwarding} step. Computation can be divided into multiple streams at any point of the network and layers might be connected in different ways according to specific needs.

There are multiple type of layers available based on their inner function. \textit{Linear} or \textit{Dense} layer using Eq.~\ref{eq:rw-dense_neuron}, \textit{Convolutional} layers, that are used to process multi-channel input such as images and finally, \textit{Recurrent} layers, that enable a model to retain additional information by building an inner memory for past events.

Defining the internal \ac{nn} structure is the first step towards deploying a \ac{nn} to tackle a task. Although, defining the architecture along with the activation functions is not enough to design a learning model. All parameters inside the layers need to be adjusted during a training process to learn some high level features and to generate the desired output. 

The process of updating the parameters based on the training data is known as~\textit{back-propagation}~\cite{rumelhart1986learning}. Back-propagation enables a \ac{nn} to learn from the error produced by the model. Back-propagation consists of computing the loss between predicted output and the true outcome. The network error is computed through a \textit{loss function}, such as \ac{mse} defined at Eq.~\ref{eq:rw_mse}.
\begin{equation}
	\label{eq:rw_mse}
	MSE = \frac{\sum_{i=1}^{n} (y_i - \hat{y_i})^2}{n}
\end{equation}
$\hat{y_i}$ is the true outcome and is compared against the network output~$y_i$. 

The computed error is propagated backward through the network, from last layer to the first, to update the parameters using an optimisation algorithm, e.g., \textit{Adadelta}~\cite{zeiler2012adadelta}, \textit{Adam}~\cite{kingma2014adam} or \textit{RMSprop}~\cite{hinton2012neural}.

When a \ac{nn} has $3$ or more hidden layers, it is generally classified as deep~\ac{nn}~\cite{glorot2011deep}. Having multiple hidden layers enables the learning of more complex representations. Thus, deep~\ac{nn} have established a new state-of-the-art performance across multiple fields such as speech recognition, visual object recognition, object detection, drug discovery and genomics~\cite{lecun2015deep}.

The rapid advancement in deep learning and \ac{nn} enabled researchers to shift from \ac{rl} to \ac{drl}. \ac{drl} is the result of merging deep~\ac{nn} with \ac{rl} enabling the tackling of increasingly complex tasks that were previously intractable due to the constraints of combinatorial state-action spaces~\cite{mnih2013playing}.


\subsection{Deep Reinforcement Learning}

\begin{figure}[h!]
	\centering
	\includegraphics[width=0.6\columnwidth]{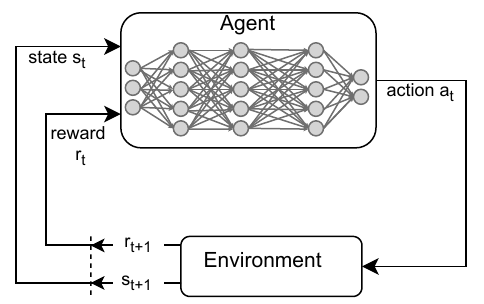}
	\caption{Single step \ac{drl} process with deep~\ac{nn}.}
	\label{fig:dqn_env_interaction}
\end{figure}

\acf{drl} mitigates the challenges of \ac{rl} by using a deep~\ac{nn} to learn the mapping between states and actions. \ac{drl} enables the learning of an optimal policy in tasks with high-dimensional state-action spaces and furthermore, it generalises across similar states and actions. In \ac{drl}, an artificial network is used to learn the mapping between states and actions. Therefore, the agent in Figure~\ref{fig:rw_rl_process} is equipped with a \ac{nn} that is used to learn the high-level features that are used for the action decision process. Figure~\ref{fig:dqn_env_interaction} shows the agent-environment interaction with a \ac{nn}.

\subsubsection{Value-based Methods}
Value-based \ac{drl} methods use a \ac{nn} to approximate a value function used to select the best action to take given a state. 
One of the first algorithms is \ac{dqn}, an evolution of Q-learning where tabular representation is replaced by an artificial neural network.

The first \ac{dqn} version consists of a single deep~\ac{nn} used to approximate the value function, presented in 2013 by Mnih et al.~\cite{mnih2013playing}, to solve seven Atari games using raw pixels as input. 
The use of a single online \ac{nn} to estimate the value from the next state~(Q-value) and select the optimal action resulted in instability during the learning process. Although the algorithm achieved state-of-the-art results in six out of the seven games tested, the training process was not stable. 
Instability is caused by the correlation between consecutive observations and non-stationarity of the environment. In 2015 Mnih et al.~\cite{mnih2015human} overcame the instability issue by introducing an \textit{experience replay buffer}, to break the correlation of consecutive observations, and a \textit{target network}.

An experience replay buffer consists of storing $n$ agent-environment interactions and randomly sampling a subset during training. Furthermore, an experience replay buffer improves sample efficiency by allowing an agent to reuse previously visited interactions multiple times without requiring redundant sampling of the same action within the same state.

The target network shares the same structure of the online network and its parameter are periodically updated based on the online parameters of the network. Two methods, namely \textit{hard-copy} and \textit{soft-copy}, are used to update the target network. In the hard-copy method, the parameters of the online network are copied to the target network after every $n$ learning iterations. On the other hand, in the soft-copy method, the target network's parameters are slowly updated towards the value of the online network parameters during each learning iteration. This update is controlled by a parameter~$\tau$, which regulates the strength of the update. While online network is optimised, the target network is frozen. Having a target network to estimate the Q-values from next state reduces the correlations with the Q-values, on the current state, estimated by the online network and therefore helps achieving a stable learning process~\cite{mnih2015human}. \ac{dqn} loss function~($\mathcal{L}$) using two networks is reported in Eq.~\ref{eq:brw_dqn_loss}, where $Q(s,a)$ is the Q-value obtained using the online network while $\hat{Q}(s,a)$ is the target estimation.

\begin{equation}
\label{eq:brw_dqn_loss}
\mathcal{L} = MSE\Big(Q(s_t,a_t), (r+\gamma \max_{a_{t+1}\in A} \hat{Q}(s_{t+1},a_{t+1}))\Big)
\end{equation}

Despite the improvements brought by the use of a target network and experience replay buffer, \ac{dqn} remains computationally demanding to find an optimal policy. Additionally, while experience replay improves sample efficiency, it relies on random sampling, which may result in insufficiently capturing the crucial experiences. The uniform probability of selecting each tuple means that significant experiences that hold a high impact on learning might not be adequately processed by the underlying learning model.

Furthermore, \ac{dqn} suffers \textit{overestimation bias} as it uses the maximum Q-value over all possible actions in the next state~($\gamma \max\limits_{a_{t+1}\in A} \hat{Q}(s_{t+1},a_{t+1})$). Selecting the maximum Q-value can lead to overestimation of the Q-values where the true Q-values are high~\cite{van2016deep}.

To overcome the aforementioned issues and thus to improve the learning stability of \ac{dqn}, \textit{Double~\ac{dqn}}, \textit{Prioritised Experience Replay} and \textit{Dueling networks} have been introduced.

\textbf{Double \ac{dqn}}  modifies Eq.~\ref{eq:brw_dqn_loss} by using the online network to select the next action $a_{t+1}$ while taking its associated Q-value from the target's. Thus, $\hat{Q}(s,a)$ is estimated as $\hat{Q}\Big( s_{t+1}, \argmax\limits_{a_{t+1}\in A} Q(s_{t+1}, a_{t+1}) \Big)$. Using the online network to select the next action and taking the Q-values estimated by the target network overcomes the overestimation bias~\cite{van2016deep}.

\textbf{Prioritised Experience Replay}  improves the random sampling of experience replay buffer by assigning a weight to each visited agent-environment interaction. Thus, experiences that surprise the most will have a higher probability to be sampled. In the context of \ac{td} learning models, surprise is commonly defined as the \ac{td} error. As the learning progresses, the surprise values are updated based on the training loss.

By prioritising selected experiences that held more information for the learning model, prioritised experience replay leads to accelerated learning and improved performance for the agent~\cite{schaul2015prioritized}.

\begin{figure}[h!]
	\centering
	\includegraphics[width=0.65\columnwidth]{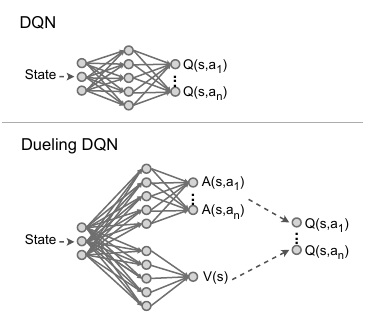}
	\caption{\ac{dqn} architecture at the top and dueling~\ac{dqn} architecture at the bottom.}
	\label{fig:dqn_dueling_dqn_architecture}
\end{figure}

\textbf{Dueling Q-network}  modifies the architecture of \ac{dqn} by forking computation stream into two branches as shown in Figure~\ref{fig:dqn_dueling_dqn_architecture}. The underlying idea is to estimate the intrinsic value~$V(s)$ of being in a certain state~$s$ and the advantage~$A(s,a)$ of taking a certain action~$a$ from $s$. The decomposition of Q-values into state-value and action advantages, allows the agent to identify states that are valuable and to learn the advantage brought by each action. The decoupling improves the learning efficiency and stability by enabling  more efficient exploration and exploitation of the environment. 
However, forking the architecture into two streams is not sufficient to decouple the two estimators. To ensure that the model will learn the decoupled representations of~$V(s)$ and $A(s,a)$, the mean value of advantages has to be zero. Such constraint need to be included within the loss function, usually done through advantage normalisation, to ensure that the model focuses on the relative differences between actions. For dueling Q-network, the Q-values can be defined as the sum between normalised advantage and state value as reported in Eq.~\ref{eq:brw_dueling_dqn_loss}~\cite{wang2016dueling}.
\begin{equation}
\label{eq:brw_dueling_dqn_loss}
Q(s,a) = V(s) + A(s,a) - \frac{1}{|A|} \sum_{a_i\in A} A(s,a_i)
\end{equation}

\subsubsection{Policy-based Methods}
Policy-based methods focus on learning a policy, which is a mapping function from states to actions. The learning model outputs a probability distribution over actions from a given state. When exploiting the policy, an agent takes the action with highest probability in a certain state. The distribution is given by the output layer and therefore is affected by the network parameters. Policy-based methods enable the agent to explore based on the probability distribution defined over the actions from states.

Policy-based methods are generally optimised using policy gradient methods. The gradient~($\nabla J$), computed with Eq.~\ref{eq:brw_policy_gradient},  identifies the direction in which parameters should be optimised to maximise the expected cumulative reward.

\begin{equation}
\label{eq:brw_policy_gradient}
\nabla J = \mathbb{E}[Q(s,a) \nabla \log \pi(a|s)]
\end{equation}

Policy-based methods can be divided into \textit{Off-policy} and \textit{On-policy} based on how they handle the collected agent-environment interactions. 
On-policy methods, update the learning model based on most-recent gathered experiences collected by the latest version of the policy. A batch of samples is collected, used for updating the policy, and then discarded. The constraint of using collected experience only once can be a limitation of on-policy methods given that they cannot improve sample efficiency with a replay buffer. 


\subsubsection{Actor-Critic Methods}
Actor-Critic combines value-based and policy-based methods. These algorithms are composed by two components, an actor which is learning a policy, and a critic, estimating a value function. The two most common Actor-Critic algorithms in \ac{drl} are \ac{ppo}~\cite{schulman2017proximal} and \ac{ddpg}~\cite{hausknecht2015deep}.

\textbf{\ac{ppo}} is an actor-critic on-policy algorithm that updates the policy gradient using the trust region policy optimisation method. The trust region method ensures that the probability ratio~($r_t$), defined as the ratio~$r_t(\theta)$ between the probabilities of selecting a certain action under the new updated policy~$\pi_\theta$ and the old policy~$\pi_{\theta_{old}}$ ($r_t(\theta) =\frac{\pi_\theta(a_t|s_t)}{\pi_{\theta_{old}}(a_t| s_t)}$), lies within a certain interval.  
This constraint prevents sudden changes between the updated and old policies, ensuring stable and controlled policy updates.

A common approach to enforce the interval constraint on~$r_t(\theta)$ in \ac{ppo} is by using the clipped version defined in Eq.~\ref{eq:brw_clip_ppo}. This function is used to compute the clipped loss~$L^{CLIP}$ during the training process\cite{schulman2017proximal}.

\begin{equation}
\label{eq:brw_clip_ppo}
L^{CLIP}(\theta) = \hat{\mathbb{E}}_t \Bigg[\min \Big( r_t(\theta) \hat{A}_t, clip( r_t(\theta), 1-\epsilon, 1+\epsilon) \hat{A}_t \Big)\Bigg]
\end{equation}

$\hat{A}_t$ is an estimator of the advantage function at time $t$ that expresses how good was a certain action taken in a given state.
The first term inside the min $(r_t(\theta) \hat{A}_t)$, is to encourage the policy~$\pi_\theta$ to take actions that maximises the expected return. However, as $(r_t(\theta) \hat{A}_t)$ might lead the updated policy to shift too much from the old one, the second term constrains $r_t(\theta)$ within an $\epsilon$-interval. Here, $\epsilon$ is the clipping factor, which ensures that the updates to the parameters $\theta$ are kept within a controlled range during training, ensuring stability in the learning process.

\begin{figure}[h!]
	\centering
	\includegraphics[width=0.8\columnwidth]{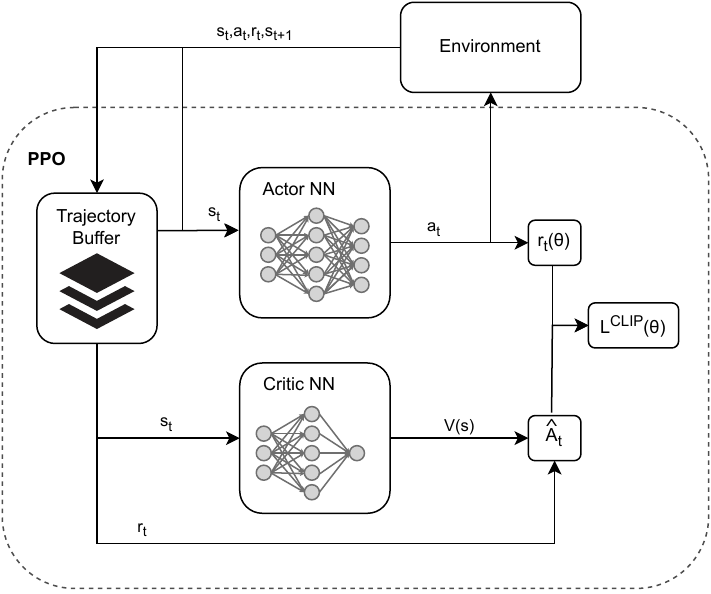}
	\caption{\ac{ppo} architecture~\cite{schulman2017proximal}.}
	\label{fig:ppo_architecture}
\end{figure}

The \ac{ppo} clipped version objective function balances between improving the policy and preventing sudden changes that may lead to instability or divergence. 
While the \ac{ppo} actor is updated using a surrogate function, i.e., Eq.~\ref{eq:brw_clip_ppo}, the critic is updated by \ac{mse} between the predicted state-value and the true discounted rewards obtained from the collected agent-environment data. Figure~\ref{fig:ppo_architecture} shows \ac{ppo} architecture with actor and critic networks.

\textbf{\ac{ddpg}} is an actor-critic off-policy method used for tasks with continuous control space. Both actor and critic are implemented through a deep model. Actor is used to estimate an action that maximises the estimated expected reward from a certain state. Critic estimates the state-action value used to evaluate the actor performance. The output of the actor is the mean value of a Gaussian distribution that is used to generate the action to take from a given state. To ensure exploration, action sampling is enriched by a noise value~$\mathcal{N}$ sampled by a distribution
: $\mu(s_t) = \mu(s_t|\theta_t^\mu) + \mathcal{N}$.
First part of the sum is the output of the actor for state $s_t$ and with the current network parameters used to compute the mean. 
\ac{ddpg} is off-policy as it uses an experience replay buffer to gather experience and sample afterward a batch of tuples to break the correlation between consecutive samples and improve the learning stability~\cite{lillicrap2015continuous}. 

Although \ac{ddpg} is a popular algorithm for continuous control tasks, there are improved versions that might be more suitable for different tasks requirements.
\begin{figure}[htb!]
	\centering
	\includegraphics[width=.8\columnwidth]{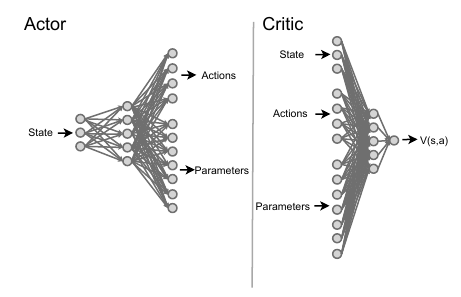}
	\caption{\ac{paddpg}~\cite{hausknecht2015deep} Actor-Critic architecture.}
	\label{fig:paddpg_architecture}
\end{figure}

In scenarios where a single action controls multiple underlying parameters, which are often defined within a continuous domain, \textit{\acf{paddpg}}~\cite{hausknecht2015deep} extends \ac{ddpg} for such tasks with parameterised action space. \ac{paddpg} actor has two output layers and selects simultaneously a discrete action and the parameters needed to parametrise the chosen action, as shown in Figure~\ref{fig:paddpg_architecture}. The critic computes state-action value from visited state and the outputs from the actor network~\cite{hausknecht2015deep}.

\subsection{Deep Reinforcement Learning - Challenges}

\ac{drl} enables an agent to tackle more complex tasks and perform as well as traditional \ac{rl} methods when applied to simple tasks. 
Therefore, \ac{drl} overcomes the state-action dimensional complexity and mitigates the lack of generalisation by generalising across similar state-action tuples. However, \ac{drl} leaves unsolved the other \ac{rl} weaknesses, introduced earlier in Section~\ref{sec:back_rl}. The unsolved challenges that remain to be addressed are the following: 1) exploration-exploitation trade-off; 2) reward function design; 3) sparsity of rewards; 4) lack of generalisation, across similar but different tasks; 5) exploration cost; 6) partial-observability and non-stationarity.

Furthermore, the use of deep~\ac{nn} to learn the state-action mapping introduces two additional challenges to \ac{drl}.
\begin{itemize}
	\item The capacity of a deep \ac{nn} learning model is influenced by the number of parameters present within it.
Increasing the number of neurons and layers expands the model ability to learn a wide range of features. Although, it is important to note that while larger networks are capable of addressing more complex problems, an agent may not need that much capacity for a certain task. In general, larger \ac{nn} require a larger amount of data compared to smaller networks in order to achieve similar level of performance for a given task. Hence, a trade-off is required to ensure sufficient granularity while keeping the total number of neurons as low as possible.  In addition to network width and depth, the choice of layer types used within the network also impact the network size. Different layers require different amount of parameters to be optimised during the learning phase.

\item Deep~\ac{nn}s are generally known to be data-hungry as they require enormous amount of data to optimise the inner parameters. Furthermore when coupled with \ac{rl}, impact of new knowledge is usually scaled by the learning rate and that amplifies the neediness of data.

\end{itemize}

To address the limitations of \ac{drl}, one effective approach is to grant the learning agent access to  an external source of knowledge. This process is known as \ac{tl} and,  when applied to \ac{drl}, some of its benefits are the following:
\begin{itemize}
	\item \ac{tl} improves the sample efficiency by transferring crucial experiences across multiple agents;
	\item \ac{tl} improves the generalisation lack by re-using policies across similar tasks;
	\item \ac{tl} lowers the learning complexity related to deep models by adapting previous acquired policies;
	\item \ac{tl} lowers the exploration cost by guiding the exploration of an agent.
\end{itemize}

The next section delves into the state of the art of \ac{tl} applied to \ac{drl} to overcome its limitations.

\section{Transfer Learning}\label{sec:back_tl}
\ac{tl} is a branch of \ac{ml} that focuses on the reusability of knowledge acquired in various contexts, tasks, or by different agents. 
When applied to \ac{rl}, it can reduce the learning time to learn an effective policy by enabling knowledge reuse. The broad idea behind \ac{tl} applied to \ac{rl} is to optimise resources consumption, i.e., computation power and time required to learn a policy, by re-using existing knowledge rather than learning from tabula rasa and updating the knowledge by re-sampling agent-environment interactions.

In \ac{tl}, the communication can be conceptualised as a one-way channel from a \textit{source} agent to a \textit{target} agent. The goal of \ac{tl} is to use source knowledge to speed-up the learning process of a target by reducing, and in some cases removing, the exploration complexity.

Transferring knowledge from source to target can result into two possible outcomes. \textit{Positive transfer} occurs when the target agent shows an improvement upon some metrics compared to tabula rasa learning. On the other hand, \textit{negative transfer} occurs when the use of external knowledge delays or hinders the learning process of the target agent. This negative transfer might happen when the transferred knowledge is outdated or when the source and target task are defined on incompatible \ac{mdp}s.

Transferring in \ac{rl} can be divided in two categories: \textit{\ac{a2a}} and \textit{\ac{t2t}}. 
In \ac{a2a} \ac{tl}, both the source and targets agents address tasks defined over the same \ac{mdp}. Thus, no mapping is needed to transfer knowledge from one agent to another.
On the other hand, \ac{t2t} \ac{tl} occurs when the source and target tasks are defined over different \ac{mdp}s. In \ac{t2t}, a mapping function is required to enable the transfer of knowledge across the tasks or different agents.

\subsection{Task-to-Task Transfer Learning}
Transferring from a known source task to a previously unseen task is a pillar concept towards the development of a \textit{lifelong learning} model where an agent has to adapt throughout time to fit an evolving task. For instance, a task might evolve on the control space by expanding the action space.

The objective of \ac{t2t} \ac{tl} is to fill the lack of generalisation typical of \ac{rl} approaches. When applied to \ac{drl}, the high data dependency is reduced and the training time, required to learn an effective policy, is reduced.

The complexity of transferring across tasks results from their \ac{mdp}s discrepancy. Tasks might differ by the state and action domains, the environment dynamics $T$ or the reward model~\cite{lazaric2012transfer}. As examples of these challenges,

\begin{itemize}
	\item $S_s \neq S_t$ $-$ Two tasks can have a different representation of the same state-space~\cite{rusu2016progressive, schwarz2018progress, d2020sharing, devin2017learning, silva2019playing,lazaric2008transfer, taylor2008transferring}, e.g., graphical representation vs raw feature vectors, or even different sensors. A policy may need to be resilient to sensors failure or upgraded sensors.	
	\item $A_s \neq A_t$ $-$ Over time a task may need to adjust the control space dimension. As a result, tasks might have a different set of actions available to the agent.~\cite{yin2017knowledge,rusu2015policy, traore2019discorl,rusu2016progressive, schwarz2018progress,   devin2017learning}. As an example, $A_s \subset A_t$ or $A_t \subset A_s$ or $A_s \cap A_t \neq \emptyset$.
	
	\item $R_s \neq R_t$ - Two tasks can share the same state-action space but have distinct reward structures or even divergent ultimate objectives~\cite{barreto2017successor, torrey2005using, rusu2016progressive, schwarz2018progress, wang2018deep, d2020sharing, devin2017learning,kulkarni2016deep, kirkpatrick2017overcoming}.
	
	\item $T_s \neq T_t$ - Two tasks defined within the same domain can differ in their probability transition function~$T$. Consequently, an action from a particular state may result in a different next state for each task.
\end{itemize}

An effective approach for \ac{t2t} knowledge transfer involves leveraging an existing learning model for a new task. Thus, the model can be utilised partially or entirely for the new task~\cite{rusu2016progressive, schwarz2018progress, wang2018deep, d2020sharing, devin2017learning,kulkarni2016deep, kirkpatrick2017overcoming, glatt2016towards, yin2017knowledge, rusu2015policy, traore2019discorl, barreto2017successor}.

When it is not possible to re-use the model but sampled agent-environment interactions are available or provided by a human demonstrator, a mapping function can be defined to map incompatible features of source and target task~\cite{lazaric2008transfer, taylor2008transferring, torrey2005using, taylor2011integrating, taylor2018improving, brys2015reinforcement}. Thus, samples gathered in one task can be adapted and re-used in a novel task to pre-train a learning model. Another application of \ac{t2t} \ac{tl} is when a task can have multiple different representation forms for the current state, a latent space can be learnt to enable cross-modal input~\cite{silva2019playing}.

\subsection{Agent-to-Agent Transfer Learning}

Multiple agents addressing the same task, or tasks defined over the same \ac{mdp}, are generally defined as homogeneous agents. An example of a system with multiple homogenous agents are multi-agent systems where multiple agents interact with the same environment to achieve a shared goal. 
This section overviews challenges and methodologies to enable knowledge transfer across homogeneous agents.

The primary objective of \ac{a2a} transfer learning is to expedite the learning process for an agent by integrating external knowledge, thus reducing the overall learning time compared to learning without additional input. To achieve this objective, \ac{a2a} \ac{tl} aims to minimise exploration complexity and enhance the overall performance of the system.

A common trait observed in \ac{a2a} \ac{tl} is the use of external expertise to influence the learning or action decision process of a target agent, as opposed to adapting an existing optimal model as in \ac{t2t}.
The external input can be provided by either a human or an artificial agent, and the knowledge provided can be integrated into the exploration process or can be used as initial phase prior to exploration.


An approach to categorise \ac{a2a} work is by analysing the shared object, which can be in the form of the following:

\begin{itemize}

	\item \textit{Action} - an agent provides on-demand advice to an exploring novel agent. Advised action overrides the target policy and ideally leads the novel agent towards the expected objective~\cite{torrey2013teaching, liu2022, subramanian2022multi, zhu2020learning, taylor2014reinforcement, da2020uncertainty, ilhan2019teaching, norouzi2021experience, da2017simultaneously, wang2018interactive, wang2017improving}.
	
	\item \textit{Feedback} - an expert provides feedback, in the form of a reward signal, on decisions made by a target agent during its learning process. The expert signal extends the environment reward and is used by the target agent to update the policy~\cite{taylor2018improving}. 
	
	\item \textit{Q-values} - when agents are based on a \ac{td} learning model, expert agent provides Q-values to a target agent. These values are then integrated into the action decision process of the target agent to influence the final decision~\cite{liu2022, taylor2019parallel, zhu2021q, liang2020parallel, chen2020active}.
	
	\item \textit{Experience} - a target agent receives agent-environment interactions obtained by a source agent. Such experience can be used to pre-train the target learning model before exploring, or to extend the standard \ac{rl} exploration phase~\cite{hester2018deep, nair2018overcoming, vecerik2017leveraging, rajeswaran2017learning, cruz2017pre, gabriel2019pre, gao2018reinforcement, gerstgrasser2022selectively}. 
\end{itemize}

Moreover, a learning model can still be entirely re-used and potentially fine-tuned as suggested for \ac{t2t} \ac{tl}.

The majority of \ac{a2a} \ac{tl} research work is based on the teacher-student framework initially proposed by Torrey and Taylor~\cite{torrey2013teaching}. In this framework, a teacher guides a student agent during the learning phase, and their interaction is often constrained by a budget.

\subsection{The Teacher-Student Framework}\label{ss:teacher-student-fw}
The teacher-student framework presented by Torrey and Taylor~\cite{torrey2013teaching} considered two homogenous agents: a teacher and a student. The teacher is regarded as an expert who has already learnt a successful policy for a certain task. On the other hand, student has no prior knowledge about the problem and requires an exploration phase. During the learning phase of the student, the teacher is available on-demand to provide action as advice, assisting the student in achieving an optimal policy. Despite the initial use case of the framework, the teacher-student paradigm has been adapted for \textit{imitation learning}, where student learns to imitate expert policy, e.g., Hester et al.~\cite{hester2018deep} and Taylor~\cite{taylor2018improving}, or to provide on-demand support to novices during exploration without providing action as advice~\cite{nair2018overcoming, liu2022, subramanian2022multi, zhu2020learning}.

The goal of teacher-student framework is to accelerate the learning of a novice agent by reducing the exploration complexity. As a result, teacher-student framework is generally coupled with action-advice, where teacher supports on-demand a novice agent by suggesting an action to take in certain states,~\cite{taylor2014reinforcement,da2020uncertainty, ilhan2019teaching,zhu2020learning,norouzi2021experience,torrey2013teaching, da2017simultaneously,taylor2019parallel,liu2022,subramanian2022multi}. 
However, instead of overriding the target policy, the action-decision process can be influenced by providing advice in form of Q-values~\cite{liu2022,zhu2021q,liang2020parallel}.
Finally, advice can also be provided in the form of an experience batch~\cite{chen2020active}.  


Within the teacher-student framework, as the communication between the two agents is generally limited by a budget, one of the initial challenges is to devise a strategy that maximises the positive impact of advice while respecting the budget. An effective strategy should focus on identifying and prioritising situations  in which the student is in pivotal need of advice. Some of the initial strategies designed to regulate the teacher-student interaction include:

\begin{itemize}
	\item \textit{Importance Advising} - aims to  identify states that are more important towards the achievement of a goal by analysing the impact of specific actions. For a state, its importance is measured as difference between the maximum and minimum impact brought by available actions. Such advising strategy has been proposed by Torrey and Taylor~\cite{torrey2013teaching} and adopted in a few approaches~\cite{taylor2018improving, da2017simultaneously};
	
	\item \textit{Reward Shaping} - teacher continuously provides a reward signal on the actions chosen by the student following its policy. This additional signal extends the reward returned by the environment and aims to promote policy imitation~\cite{taylor2018improving}; 
	
	\item \textit{Mistake Correction} - the teacher constantly monitors the student and provides advice when the action chosen by the student differs from the action expected by the teacher policy. Such technique requires the teacher to have an optimal policy~\cite{torrey2013teaching, taylor2018improving};

	\item \textit{$\epsilon$-decay Probability} - probability of giving advice is reduced as the agent explores by a small $\epsilon$ value~\cite{subramanian2022multi, zhu2020learning, norouzi2021experience};
	
	\item\textit{Early Advice} - student utilises the full allocated budget by receiving guidance during the first $n$ steps of learning where it is expected to be in the greatest need~\cite{torrey2013teaching, taylor2018improving};

\end{itemize} 

To improve the budget utility and aiming to effectively identify situations where advice can significantly impact the development of the target policy, Taylor et al.~\cite{taylor2014reinforcement} proposes \textit{Predictive Advising} aiming to lower the high communication cost of \textit{Mistake Correction} by enabling the source agent to predict the action sampled by the target agent during the learning process. Da Silva et al.~\cite{da2017simultaneously} enables transfer from an agent that has successfully completed an episode. As a result, all target agents benefit from the full trajectory followed by the source agent, which led to the accomplishment of the episode.
In addition, other sophisticated advising strategies may rely on state-visit counter~\cite{taylor2019parallel, da2017simultaneously,zhu2021q}, confidence of student and teacher~\cite{da2020uncertainty,ilhan2019teaching, liang2020parallel, wang2018interactive, chen2020active, wang2017improving} and agent performance~\cite{taylor2019parallel, gerstgrasser2022selectively}. 
Finally, when the source agent is an expert with optimal policy, the budget utility can be further increased by re-using previous advice across similar states as proposed by Zhu et al.~\cite{zhu2020learning}.

Table~\ref{tab:back_a2a-Teacher-student-framework} summarises the related work based on the Teacher-Student paradigm categorised by the following features:

\begin{itemize}

	\item \textit{Advice Source} $-$ regardless of the form of advice, advice can be provided by a human operator~($human$) or by an intelligent agent. An agent that has already encountered the task can be considered as an expert agent~($EA$). On the other hand, if the agent has not yet converged for the given task it is considered as an imperfect agent~($IA$), e.g., agents that learn simultaneously the same task. Lastly, advice can be given by a supervised model~($classifier$).
	
	\item \textit{Advice Type} $-$ indicates the form of the advice shared. This can vary between one or multiple actions~($a$), experience or demonstration~($e$), scalar reward used by target agent to complement the reward returned by the environment~($reward$) and finally network output logit, Q-values in case of \ac{td}-learning method,~($Q$).

	\item \textit{Advising Policy} $-$ indicates the advising strategy followed by the agents, which can be one of the following:  importance advising~($ia$), reward shaping~($rs$), mistake correction~($mk$),  $\epsilon$-decay probability~($\epsilon p$), early advice~($ea$), predictive advising~($pa$), sharing successful episodes~($se$), target confidence~($t$-$conf$) (evaluated on target agent), source and target confidence~($st$-$conf$) (evaluated on both target and source agent), based on state-visit counters~($svc$), based on \ac{td}-error~($td$-$error$) and performance-based metrics defined over state~($pms$).	

	Finally, for some advising policy may be not applicable~($na$), i.e., when transfer happens on initialisation of new agents and knowledge is used to pre-train a model~\cite{liu2022}.

	\item \textit{\ac{drl}} $-$ reports whether the framework is applied on a \ac{drl} algorithm or only tabular.

	\item \textit{Re-using Advice} $-$ indicates whether the target has some memory to retain the advice received to increase the budget utility.

	\item \textit{MARL} $-$ reports whether the framework is applied in a multi-agent context.
\end{itemize}

\begin{table}[hbt!]
	\small
	\renewcommand*{\arraystretch}{1.1}
	\caption{\label{tab:back_a2a-Teacher-student-framework} Summary of work based on the Teacher-Student paradigm.}
	\centering
	\begin{threeparttable}
		\begin{tabular}{|c|c|c|c|c|c|c|}
			\hline
			\multirow{2}{*}{\textbf{Reference}}&\multirow{1}{*}{\textbf{Advice}}&\multirow{1}{*}{\textbf{Advice}}&\textbf{Advising}&\multirow{2}{*}{\textbf{\acs{drl}}}&\multirow{1}{*}{\textbf{Advice}}&\multirow{2}{*}{\textbf{MARL}}\\
			&\textbf{Source}&\textbf{Type}&\textbf{Policy}&&\textbf{Reuse}&
			\\\hline

\multirow{2}{*}{Taylor$\,$et$\,$al.,$\,$2014$\,$\cite{taylor2014reinforcement}} &\multirow{2}{*}{$EA$} &\multirow{2}{*}{ $a$ }& $ea, aa, ia,$ & \multirow{2}{*}{\cmark} & \multirow{2}{*}{\xmark} & \multirow{2}{*}{\xmark} \\ 
&&& $mk, pa$&&&\\\hline
Da Silva$\,$et$\,$al.,$\,$2020$\,$\cite{da2020uncertainty}  & $EA$ & $a$ & $t$-$conf$& \cmark & \xmark& \xmark \\\hline
Zhu$\,$et$\,$al.,$\,$2020$\,$\cite{zhu2020learning}  &  $EA$ & $a$ & $\epsilon p$ & \cmark &\cmark &\xmark \\\hline

Norouzi$\,$et$\,$al.,$\,$2021$\,$\cite{norouzi2021experience} & $classifier$ &$a$&$\epsilon p$&\xmark&\xmark&\cmark\\\hline 

Torrey$\,$and$\,$Taylor,$\,$2013$\,$\cite{torrey2013teaching}& $EA$  & $a$ & $ea, ia, mk$ & \xmark & \xmark & \xmark  \\\hline

Zhu$\,$et$\,$al.,$\,$2021$\,$\cite{zhu2021q} &$IA$&$Q$&$svc$&\cmark&\xmark&\cmark\\\hline
Liang$\,$and$\,$Li,$\,$2020$\,$\cite{liang2020parallel} &$IA$&$Q$&$st$-$conf$&\cmark&\xmark&\cmark\\\hline

Da$\,$Silva$\,$et$\,$al.,$\,$2017$\,$\cite{da2017simultaneously} &$IA$&$a, e$& $ia, svc, se $&\cmark&\xmark&\cmark \\\hline
Taylor$\,$et$\,$al.,$\,$2019$\,$\cite{taylor2019parallel}  & $IA$ &$Q$ & $pms, svc$& \xmark&\xmark&\cmark\\\hline

Liu et al.,$\,$2022$\,$\cite{liu2022}  & $IA$ &$a, Q$ & $na$
& \cmark & \xmark &\cmark  \\\hline 

Subramanian$\,$et$\,$al.,$\,$2022$\,$\cite{subramanian2022multi}& $IA$ & $a$ & $\epsilon p$  & \cmark & \xmark & \cmark \\\hline

Ilhan$\,$et$\,$al.,$\,$2019$\,$\cite{ilhan2019teaching}& $IA$ & $a$ & $st$-$conf$& \cmark & \xmark & \cmark \\\hline

Taylor$\,$and$\,$Borealis,$\,$2018$\,$\cite{taylor2018improving}$\!$$\!$& $human$& $reward$ & $rs$ & \xmark & \xmark & \xmark \\\hline

Chen$\,$et$\,$al.,$\,$2020$\,$\cite{chen2020active}& $human$& $e$ & $t$-$conf$ & \cmark & \xmark& \xmark\\\hline 









\hline
		\end{tabular}
	\end{threeparttable}
\end{table}

The majority of the teacher-student algorithms, reported in Table~\ref{tab:back_a2a-Teacher-student-framework}, provides advice in form of action. Furthermore, action is the form of transfer preferred when the source is an experienced agent~\cite{torrey2013teaching, zhu2020learning, taylor2014reinforcement,da2020uncertainty}. When the source agent is not optimal, advice is provided in form of Q-values that are combined on target side to influence the action-decision process~\cite{taylor2019parallel, zhu2021q, liang2020parallel}.

The teacher-student paradigm achieves state-of-the-art performance when transferring from expert agents to novices, and the effectiveness of the transfer depends on the quality of expert policy. Advisor must not provide advice that is out of date or leads to worsening of target agent learning experience. Therefore, teacher must have an optimal policy that cannot be outperformed by other agents in any way.

Recent developments on the teacher-student framework aimed to relax the optimality constraint by establishing metrics and methodologies to identify when an exploring agent is in need of advice. 

Liu et al.~\cite{liu2022} adapt the framework to work online with non-optimal teachers and an increasing number of agents. Existing agents distil their knowledge into novel agents that join later on, resulting in less experienced agents benefiting from others. Similarly, Norouzi et al.~\cite{norouzi2021experience} use a subset of collected demonstration to train a centralised super-entity. The central-entity benefits from experience gathered by all the agents and can be queried to provide advice in specific states. Despite quality of advice being improved over time, this learning paradigm results in an additional cost to gather the \ac{rl} experience and to train the underlying model.

Da Silva et al.~\cite{da2020uncertainty}, and Chen et al.~\cite{chen2020active} rely on agent's epistemic uncertainty to decide whether an agent has adequately explored a certain state. The use of confidence has a twofold advantage, firstly it enables the agent to prevent redundant advice leading to an overall better use of the allocated budget, and secondly it enables a student agent to outperform the teacher by allowing target agent to potentially learn actions better than source of advice.

Despite the better budget utilisation by estimating the confidence on student side, these methods do not assess the advisor proficiency. Ilhan et al.~\cite{ilhan2019teaching} and Liang and Li~\cite{liang2020parallel} provide both target and sender with a confidence estimator model that allows for confidence comparison between the two agents. As a result, the target agent can still benefit from an imperfect agent as advisor while lowering the possibility of being negatively impacted by following the advice in states where the target has higher uncertainty than the teacher.

Finally, establishing standardised criteria, including confidence, state visit counters, and performance-based metrics, facilitates the comparison of learning progress among multiple agents in diverse states. Comparing the learning progress of each agent enables bi-direction advice transfer across learning agents, each agent can benefit from knowledge gathered by the other agents~\cite{ilhan2019teaching, zhu2021q, taylor2019parallel, da2017simultaneously, liang2020parallel}.

\subsection{Experience Sharing in Agent-To-Agent Transfer Learning}\label{ss:exp_share}

Feeding samples to neural network as a pre-trained stage is an established approach to boost the learning performance of a novel model across different fields, e.g., computer vision~\cite{donahue2014decaf,bengio2012deep,ghassemi2020deep} and natural language process~\cite{mikolov2013distributed,pennington2014glove}. 

\ac{rl} models have been using similar strategies to perform an addition pre-training step before letting the agent explore~\cite{pmlr-v162-seo22a, stooke2021decoupling, anderson2015faster, schwarzer2021pretraining}.

More generally, there is a recent trend in \ac{rl} on the transition from online to offline~\ac{rl}~\cite{nair2020awac, lu2022aw, lee2022offline, walke2023don}. While in online~\ac{rl} an agent collects data by interacting with the environment, agent trained in offline~\ac{rl} relies on available data collected by using an unknown policy. Although this thesis does not focus on offline \ac{rl}, there are some overlapping challenges needed to be addressed with the sharing of experience from one agent to another. For instance, predicting the impact of specific agent-environment interactions on a target agent, as well as exploring the trade-off between the quantity of tuples provided to a novel agent and its resulting impact on performance.

To follow, this thesis provides a summary of relevant work in \ac{a2a} \ac{tl} based on experience sharing in Table~\ref{tab:back_a2a-experience-sharing}. The category of \textit{Online \ac{tl}} differentiates whether the transferred experiences are used in an offline context, like a pre-training step, or online to enhance a learning process.

\begin{table}[hbt!]
	\small
	\renewcommand*{\arraystretch}{1.1}
	\caption{\label{tab:back_a2a-experience-sharing} Summary of experience sharing work in \ac{a2a} \ac{tl}.}
	\centering
	\begin{threeparttable}
		\begin{tabular}{|c|c|c|c|c|c|c|c|}
			\hline
			\multirow{2}{*}{\textbf{Reference}}&\multirow{1}{*}{\textbf{Experience}}&\textbf{Selection}&\multirow{2}{*}{\textbf{\acs{drl}}}&\multirow{1}{*}{\textbf{Advice}}&\multirow{2}{*}{\textbf{MARL}}& \multirow{1}{*}{\textbf{Online}}\\
			&\textbf{Source}&\textbf{Policy}&&\textbf{Reuse}&&\textbf{TL}
			\\\hline

			Hester$\,$et$\,$al.,$\,$2018$\,$\cite{hester2018deep} &  $human$ & $na$  & \xmark & \xmark & \xmark  & \xmark \\\hline
			
			Gerstgrasser$\,$et$\,$al.,$\,$2022$\,$\cite{gerstgrasser2022selectively}$\!$& $IA$&$td$-$error$&\cmark&\xmark&\cmark&\cmark\\	
			\hline			
			
			Wang$\,$and$\,$Taylor,$\,$2017$\,$\cite{wang2017improving}$\!$& $classifier$ &  $t$-$conf$ & \cmark & \xmark & \xmark & \xmark\\ 
			\hline
			
			Wang$\,$and$\,$Taylor,$\,$2018$\,$\cite{wang2018interactive}$\!$& $classifier$ &  $t$-$conf$& \cmark & \xmark & \xmark &\cmark \\\hline
			
			Nair$\,$et al.,$\,$2018$\,$\cite{nair2018overcoming}&  $human$  & $na$ & \cmark &\xmark &\xmark & \xmark \\ \hline
			
			Vecerik$\,$et$\,$al.,$\,$2017$\,$\cite{vecerik2017leveraging}$\!$& $human$ & $na$&\cmark & \xmark & \xmark & \xmark\\\hline 
			
			Rajeswaran$\,$ et$\,$al.,$\,$2017$\,$\cite{rajeswaran2017learning}$\!$ &$human$& $na$ & \cmark&\xmark&\xmark&\xmark\\\hline
			
			Cruz$\,$et$\,$al.,$\,$2017$\,$\cite{cruz2017pre}  & $human$ &   $na$ & \cmark& \xmark& \xmark & \xmark \\\hline
			
			Gabriel$\,$et$\,$al.,$\,$2019$\,$\cite{gabriel2019pre} & $human$&  $na$ &\cmark&\xmark&\xmark&\xmark\\\hline
			
			Gao$\,$et$\,$al.,$\,$2018$\,$\cite{gao2018reinforcement}  &$IA$ & $na$ & \cmark& \xmark &\xmark&\xmark\\\hline
			
		\end{tabular}
	\end{threeparttable}
\end{table}

Most of the \ac{tl} work based on experience sharing focuses on transferring agent-environment interactions collected from a human demonstrators to be used as a pre-training step for a \ac{rl} agent~\cite{hester2018deep,nair2018overcoming, vecerik2017leveraging, rajeswaran2017learning, cruz2017pre, gabriel2019pre}. Wang and Taylor and Borealis~\cite{wang2017improving, wang2018interactive} use pre-collected agent-environment interactions to train a classifier that provides on-demand action-advice to a learning agent. The decision to follow the advice is based on the confidence provided by the source of advice.  Wang and Taylor~\cite{wang2018interactive} also enable continuous improvement of the classifier by collecting further online demonstrations from uncertain states. 
Finally, Gerstgrasser et al.~\cite{gerstgrasser2022selectively} enable experience sharing across multiple agents by sharing agent-environment interactions with high \ac{td}-error to all the agents.

\subsection{Research Gaps and Design Challenges}
Based on the discussions in the previous two sections, the application scope of the teacher-student framework~\cite{ilhan2019teaching, zhu2021q, taylor2019parallel, da2017simultaneously, liang2020parallel} is evolving towards an online scenario where learning agents have a dual roles as both students and teachers. This bi-directional transfer allows for dynamic and effective learning process by enabling agents to fill their knowledge gap. 
In fact, when multiple agents address simultaneously a task, it is highly probable that they will quickly develop expertise in narrow and specific areas of the system. 
Therefore, by sharing their knowledge with one another, it is possible to lower the average policy training time. However, sharing knowledge is a very delicate process and negative transfer is very likely to happen when teacher is less than optimal. 

In order to design a successful online teacher-student framework in multi-agent systems requires a number of key challenges to be addressed.  First, as all agents are learning, there must be a careful balance between the communication cost associated with the transfer process and the potential influence that the shared knowledge may have on the target agent. The system must be able to identify suitable teachers, from whom the target agent will benefit by gaining new knowledge. Consequently, as the goal of transfer learning is to maximise the positive impact of external knowledge, the system must be able to filter relevant advice that is expected to bring a positive impact to target agent, and update the target policy accordingly.

Therefore, the following challenges must be taken into account:
\begin{itemize}
	\item \textit{Identify suitable teacher} - the system must be able to dynamically filter any potential candidate from which the target agent will benefit by transferring knowledge.
	
	\item \textit{Balance communication cost and transfer influence} - while ultimately the goal of transfer learning is to maximise the positive impact brought by the external knowledge, it is also essential to consider and limit the communication cost involved. In an online scenario, synchronising all of the agents to check whether the target agent will eventually benefit by an advice is unfeasible as both sender and receiver need to synchronise for each action-selection process.
	
	\item \textit{Identify relevant knowledge} - the system must select the form of shared advice and decide whether is relevant  for the target agent's current level of knowledge. As an example, a more experienced teacher that is transferring knowledge might not bring any support to a student by suggesting an action that would already be selected by the target policy anyway.
	
	\item \textit{Policy update based on received knowledge} - the impact of shared knowledge and the method of integrating it into an agent policy may vary depending on the type of advice exchanged among the agents. For instance, an action could both override the target policy or could increase the likelihood of the action being selected by the agent by influences its weight.
\end{itemize}

To address these challenges, this thesis proposes \ac{efontl} an online sharing framework based on the teacher-student paradigm where teacher is dynamically selected, based upon some fixed criteria, across available learning agents. The advice is replaced by a batch of agent-environment interactions collected by the teacher. Target agents filter the incoming batch to sample a sub-set of tuples that are expected to improve their policy.
\subsection{Baselines}\label{sec:baselines}

Based on the analysis of the \ac{tl} approaches presented in Section~\ref{ss:teacher-student-fw} and Section~\ref{ss:exp_share}, it can be concluded that expert-based teacher-student frameworks positively impact the target agent due to the reliable pool of knowledge available to be leveraged.


This thesis shows \ac{efontl} impact against two action-advice based baselines: 1) \ac{ocmas}~\cite{ilhan2019teaching} - an online confidence moderated action-advice sharing that adapts the teacher-student framework to multiple learning agents. The agents synchronise at each step and the most uncertain agent relies on others to select the appropriate action to take by majority voting. 
The majority voting along with the uncertainty comparison mechanism ensure that advice is provided in situations where the seeker is really in need of advice and reduce the budget utilisation when compared to classical teacher-student methods. 
 2) \ac{rcmp}~\cite{da2020uncertainty} - provides action-advice support to a learning agent by relying on an expert agent.
\ac{rcmp} lowers advice frequency as the target agent gradually increase its confidence by exploring the environment.
\ac{rcmp} has demonstrated state-of-the-art performance while reducing communication costs, in comparison to a traditional action-advice strategy for the teacher-student framework. The consequence of this, however, is that effectiveness of \ac{rcmp} is strictly related to the quality of teacher used to provide guidance.
Advice is moderated by target uncertainty regardless of teacher confidence in a certain state. Therefore, by overriding the actions chosen by the student policy, the teacher may restrict the student performance and prevent exploration for better solution.

\subsection{Relationship of Transfer Learning to Collaborative Multi-Agent Reinforcement Learning}\label{ss:tl_and_marl}

In an online context, multiple agents, which interact in the same environment, are enabled to transfer part of their knowledge through \ac{tl}, which enables these agents to share and leverage a portion of their accumulated knowledge.. Similarly, \ac{marl} is typically applied to multi-agent scenarios where multiple agents interact with a single instance of an environment. As a consequence, \ac{marl} introduces a collaborative framework, wherein each agent develops its individual policy to address its specific goals while taking into consideration both actions and behaviours of agents in the proximity.

To provide a more precise contextual position of~\ac{tl} and \ac{efontl}, in this section we discuss similarities and differences between \ac{tl} and \ac{marl}.
\ac{marl} is a subfield of \ac{rl} that addresses scenarios where multiple agents interact within a shared environment. The goal of~\ac{marl} is to enable multiple agents to learn concurrently by learning a policy that takes into account the behaviours of other agents.  Generally, in~\ac{marl} each agent has its own policy and \ac{rl} approaches that can be used in~\ac{marl} can be organised into three categories:
\begin{itemize}
	
	\item \ac{il}, each agent acts independently by taking decisions based on the state returned by the environment. The policy is updated based on its own experience and all the other agents are treated as part of the environment. As example, Q-learning~\cite{gaskett2002q}, \ac{dqn}~\cite{mnih2013playing} and \ac{ppo}~\cite{schulman2017proximal} can be used to learn independent policies;
	
	\item \ac{ctde}, agents' policy are trained jointly in a centralised manner using shared information. However, during execution, each agent makes independent decision based on its own policy without any need for information from other agents. 	 For instance, \ac{maddpg} is an extension of \ac{ddpg} to enable centralised learning in multi-agent context. In \ac{maddpg}, each agent selects an action based on its independent actor network, while the critic is centralised. Centralised critic allows for the evaluation of the joint actions of all agents facilitating the learning of cooperative behaviour across agents.

	
	\item networked agents, agents communicate one to the other during training to learn to take joint actions that maximise the global reward while learning an independent policy.  As example, QMIX~\cite{rashid2020monotonic} and \ac{vdn}~\cite{sunehag2017value} define a global Q-function as a composition of single agents' Q-function. 
	The global Q-function evaluates the impact of the join actions taken by the agents and is used during the training phase to induce collaborative behaviours on the single agent policies. As a result, the Q-function of a single agent is influenced by the actions of other agents.
	\end{itemize}

While the objective of \ac{marl} is to encourage agents to learn cooperative behaviours through interaction and information exchange, \ac{efontl} aims to improve the overall system performance by enabling agents to learn independent policies while sharing batch of agent-environment interactions to transfer a subset of knowledge that is expected to improve the target policy. Furthermore, while \ac{marl} allows agents to have different reward models and objectives, \ac{efontl} is not specially designed to transfer between agents with different goals. To enable the transfer across agents with different reward models, a mapping function is required to map the experiences and rewards from one agent to another.

\section{Uncertainty in Reinforcement Learning}\label{sec:back_uncertainty}

In \ac{rl} there are two types of uncertainties, \textit{aleatoric} and \textit{epistemic}.
\textit{Aleatoric} uncertainty comes from the environment and is generated by~stochasticity in observation, reward and actions, \textit{epistemic} uncertainty comes from the learning model and indicates whether the agent has adequately explored a certain state.
In \ac{rl}, uncertainty has been largely used to identify unfamiliar states and hence to enhance overall agent exploration~\cite{burda2018exploration, osband2016deep, bellemare2016unifying, tang2017exploration, fortunato2017noisy, pmlr-v70-pathak17a}.
Most recent \ac{tl} frameworks follow this insight by relying on epistemic uncertainty to determine whether an agent requires guidance or not~\cite{nair2018overcoming, da2020uncertainty, taylor2019parallel}.

One feasible way to approximate agent epistemic uncertainty within a specific task is by counting the number of visits per each state. 
In fact, Taylor et al.~\cite{taylor2019parallel}, Da Silva~\cite{da2017simultaneously} and Nourouzi et al.~\cite{norouzi2021experience} estimate uncertainty from state-visits counter. Zhu et al.~\cite{zhu2021q} improve the estimation by relying on state-action pair.
However, state-space might be continuous or very large making the naive counting unfeasible. Thus, state-visits could be approximated by \ac{rnd}~\cite{burda2018exploration}. Despite \ac{rnd} originally being introduced to encourage exploration, it has already been exploited as uncertainty estimator for \ac{tl}-\ac{rl}~\cite{ilhan2019teaching}.
RND consists of two networks, a target with unoptimised and random initialised parameters and a main predictor network. Over time, the target network is distilled within the predictor and uncertainty is defined as prediction error between the two outputs.

Other sophisticated models can be used to estimate the epistemic uncertainty. Wang et al.~\cite{wang2017improving} propose uncertainty estimation through neural network, decision tree and Gaussian process. \ac{nn} is also used by Da Silva et al.~\cite{da2020uncertainty}, where the agent learning model is expanded by replacing control-layer with an ensemble later used to approximate agent uncertainty.

Despite the different underlying technique used to estimate uncertainty, all the aforementioned methods rely uniquely on visited state and thus, might lack crucial information needed to estimate uncertainty in states where the goal is close but not yet achieved.
For instance, in sparse reward environment and tasks with continuous control space, action must be taken into account when estimating the intrinsic agent uncertainty when they are far from optimality.
While taking into account actions explored along states might not be necessary when bootstrapping expert knowledge, it is important to discriminate between sampled actions over states when transferring from learning agents.

\section{Summary of Background}


A \acf{mdp} is a mathematical framework to model tasks that require sequential decision making. \ac{mdp}s are based on  the Markovian property, which states that the probability of transitioning from one state to the next depends solely on its current state and it is independent of its past states. 

\acf{rl} is a machine learning framework that enables an intelligent agent to learn the transition function of a~\ac{mdp} through continuous interaction with the environment. \ac{rl} is a cyclic four-step process composed by: 1) Observation ; 2) Action Selection; 3) Actuation; and 4) Evaluation. This cycle enables a \ac{rl} agent to learn a policy by updating its knowledge based on the new sampled evidence.

One of the major limitations of \ac{rl} lies in the representation used to learn the mapping between states and actions. The state-action combinatorial complexity limits the applicability to domains with low-dimension state-action space. 

\acf{drl} partially overcomes \ac{rl} limitations by using a \acf{nn} to learn the high-level features that are used for the action decision process. Furthermore, the use of \ac{nn}s enables an agent to generalise across similar visited states reducing the complexity that arises from the exploration of state-action space to collect new evidence. Despite the exploration cost being mitigated by the capacity of \ac{nn}s to generalise across similar states, this along other \ac{rl} limitations remain unsolved.

\acf{tl} can partially overcome the \ac{drl} limitations as follow:
\begin{itemize}
	\item  \ac{tl} improves the sample efficiency by including external expertise within a \ac{drl} agent;
	
	\item \ac{tl} allows for policy re-use across similar tasks;
	
	\item \ac{tl} lowers the learning complexityy by partially or fully adapting previous acquired policies;
\end{itemize}

\ac{tl} enables part of knowledge to be transferred from one agent, named source, to another to improve the learning phase of a target agent. Depending on the objective of \ac{tl}, the form of knowledge transferred can vary, i.e., action, agent-environment interaction, reward and policy. Other factors that influences the designing of a \ac{tl} framework are the availability of an expert or a trained policy and the access to pre-collected experiences.

To provide support to a learning agent, the predominating framework used is the Teacher-Student paradigm, where an experienced teacher provides advice to a novel agent. In the teacher-student framework, the advice generally influences the action-decision process, i.e., action or Q-values, or the weight of learnt state-action pair, i.e., reward.

The teacher-student framework proposed by Da Silva et al.~\cite{da2020uncertainty}, \acf{rcmp}, has demonstrated state-of-the-art performance. In this approach, a target agent explores the environment and estimates the epistemic uncertainty for the states it visits. If this uncertainty falls below a predefined threshold, the target agent seeks advice from an experience teacher, which provides an action to be followed by the target.

The teacher-student framework can be extended to multiple learning agents where the roles of teacher and student are swappable over time. For instance, Ilhan et al.~\cite{ilhan2019teaching} propose \acf{ocmas}. In this approach, an uncertain agent asks other agents for an action-advice. Finally, the advice seeker follows the final action based on majority voting of the suggested actions.

Wang and Taylor~\cite{wang2018interactive} propose a \ac{tl} framework where the teacher, an external entity trained on previously stored agent-environment interactions, provide action as advice to a learning agent. Furthermore, in uncertain states, the teacher can request for additional demonstrations to improve the quality of the advice given. Gerstgrasser et al.~\cite{gerstgrasser2022selectively} propose a~\ac{tl} framework where the tuples that exhibit high \ac{td}-error within a specific agent are forwarded to all the learning agents.

This thesis focuses on the transferability of knowledge across homogeneous learning agents by introducing~\ac{efontl}. The objective of this thesis is to investigate the impact of experience sharing against the teacher-student framework. \ac{efontl} is compared against two action-advice baselines: \ac{ocmas}~\cite{ilhan2019teaching} to compare the impact of action-advice against experience sharing by \ac{efontl}; and \ac{rcmp}~\cite{da2020uncertainty} to benchmark~\ac{efontl} capability against an action-advice baseline from experienced teachers. In the \ac{rcmp} version implemented in this thesis, the single optimal teacher is replaced by multiple trained agents to prevent any bias coming from the use of a single non-optimal teacher. Consequently, the action followed by a target agent is picked by majority voting on the actions provided by the teachers.

\ac{efontl} enables agents to share collected experience autonomously to promote continuous improvement of a multi-agent system. Similarly, \acf{marl} algorithms are designed to enhance the learning of multi-agent systems by enabling the agents to learn cooperative behaviours. \ac{marl} enables collaboration among learning agents through \acf{ctde} and networked agents.
In \ac{ctde}, the agents collaborate during their training process to enable the learning of complex and collaborative policies. For instance, in the multi-agent variant of \acf{ddpg}, known as MADDPG, the critic is centralised and shared across all the agents. The centralised critic takes in input the state-action of each agent to estimate the $Q$-value of the joint actions.

In networked agents, communication among between agents plays a crucial role in coordinating their actions and making collaborative decisions. As example of networked agents, QMIX~\cite{rashid2020monotonic} and \acf{vdn}~\cite{sunehag2017value} combine the Q-values of independent agents to learn a global Q-function that considers the states and actions of each agent.

\chapter{Preliminary Studies}
\label{cpt:preliminary_studies}\acresetall%

This chapter serves as a feasibility study towards the development of~\ac{efontl}. While the objective of~\ac{efontl} is to facilitate online~\ac{tl} among multiple agents learning simultaneously, the content of this chapter focuses on \ac{tl} in the offline context. In comparison to~\ac{efontl}, the challenges addressed by experience sharing in offline~\ac{tl} are relatively reduced, as, in this context, the source of transfer is a nearly optimal agent. This situation implies that the target agent should benefit from the knowledge provided by the teacher due to the difference in the exploration time.

Through this offline evaluation, our aim is to investigate and observe the implications of offline~\ac{tl} in environments where we plan to assess the online approach. The evaluation of this preliminary stage is carried out in two distinct environments: 1) a \ac{stpp} scenario, in which experiences, selected based on their associated uncertainty, are transferred to a new agent; and 2)~\ac{rs-sumo}, a setting where experiences are transferred across agents operating with diverse demand patterns


The remainder of this chapter is structured as follows: the architecture used for the experiments is detailed in Section~\ref{ss:ppo_offline-tl}; the environments used to benchmark the transfer of experience are presented in Section~\ref{sec:preliminary-envs}. To follow, the impact of experience sharing in offline~\ac{tl} is analysed in Section~\ref{sec:eval-offline-tl}. Lastly,  Section ~\ref{sec:sarnd_eval} evaluates~\ac{rnd} as an uncertainty estimator model, highlighting its limitations when used in an online context like~\ac{efontl}, and introduces~\ac{sarnd} to overcome the identified shortcomings.

\section{\acl{ppo} Architecture}\label{ss:ppo_offline-tl}

This section introduces the architectural design for the \ac{drl} algorithm implementation used to learn the agent's policy. For the preliminary experiments on offline experience sharing, we extended the classical \ac{ppo} algorithm~\cite{schulman2017proximal} to enable learning from external experience.

\ac{ppo} clipped version, introduced earlier in Chapter~\ref{cpt:background}, has been used to carry out the preliminary experiments to assess the impact of experience sharing in \ac{pp} and \ac{rs-sumo}.

\begin{figure}[htb!]
	\centering
	\includegraphics[width=.7\columnwidth]{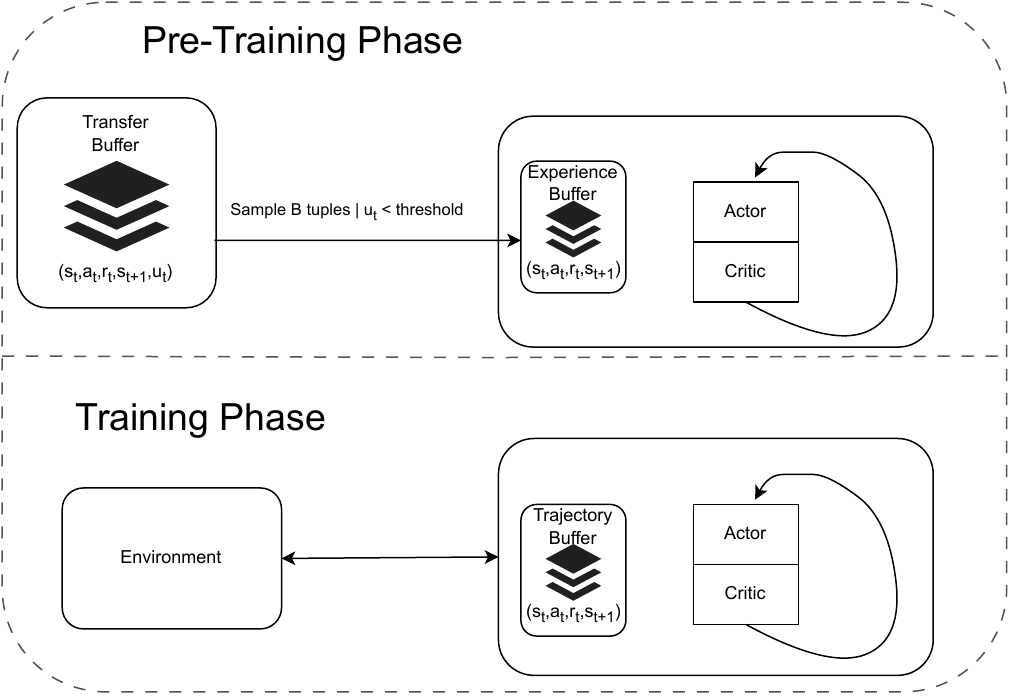}
	\caption{\ac{ppo} extended model.}
	\label{fig:ppo_for_offline-tl}
\end{figure}

Unlike off-policy algorithms that use two different policies to sample and optimise, \ac{ppo} assumes that the samples used to optimise a policy are collected by the current version of the very same policy. To enable the use of external experience in \ac{ppo}, the model needs to be expanded as depicted in Figure~\ref{fig:ppo_for_offline-tl}. 

The learning stage of the expanded~\ac{ppo} model is divided in two parts. Firstly, the learning model is optimised based on part of the transferred experiences. The agent samples $B$ tuples from the transfer buffer, ensuring that each tuple's uncertainty is below a predefined threshold. The selected tuples are then used to populate an experience buffer, which is finally used to perform a few steps of gradient descent optimisation. Secondly, the agent interacts with the environment without relying any longer on the external transferred knowledge and updates its policy based on the sampled experience.

While the expanded version of \ac{ppo} used for the preliminary experiments requires the use of a transfer buffer, there are no changes within the internal model architecture shown in Figure~\ref{fig:ppo_architecture}.  For the preliminary experiments conducted in two environments, \ac{stpp} and \ac{rs-sumo}, the \ac{ppo} models' architecture for both actor and critic networks, as well as the learning parameters, are reported in Table \ref{tab:ppo_setup}. The learning parameters specify the number of policy updates performed over each mini-batch and the clipping ratio used for the clipped \ac{ppo}.

\begin{table}[h!]
	\renewcommand*{\arraystretch}{1.1}
	\caption{\label{tab:ppo_setup} \ac{ppo} parameters setup for preliminary experiment.}
	\centering
	\begin{threeparttable}
		\begin{tabular}{|c|c|c|}
			\hline
			{\textbf{Parameter}}& \textbf{\acs{stpp}} & \textbf{\ac{rs-sumo}}\\
			\hline
			\multicolumn{3}{|c|}{\textit{actor}}\\\hline
			Input Layer & Linear(18, 64) & Linear(26, 128) \\\hline
			\multirow{4}{*}{Hidden Layer(s)} & \multirow{4}{*}{Linear(64, 64)}& Linear(128, 128)\\
			&&Linear(128, 128)\\
			&&Linear(128, 128)\\
			&&Linear(128, 64)\\\hline
			Output Layer & Linear(64, 5) & Linear(64, 5)\\\hline
			Activation & Tanh & Tanh\\
			\hline
			\multicolumn{3}{|c|}{\textit{critic}}\\\hline
			Layer & Linear(18, 64) & Linear(26, 128) \\\hline
			Output Layer & Linear(64, 1) & Linear(128, 1)\\\hline
			Activation & ReLU & ReLU\\
			\hline
			
			\hline			
			Optimiser & Adam & Adam\\\hline
			Learning Rate & 1e-3 & 1e-4\\\hline
			Betas &  (.9, .999)  & (.9, .999) \\\hline
			\#epochs & 10 & 10\\\hline
			Clipping ratio & .2 & .2\\\hline
			Mini-batch Size & 32 &32\\			
			\hline
		\end{tabular}
	\end{threeparttable}
\end{table}

However, for \ac{tl} applications in an online setting like \ac{efontl}, we transitioned from using \ac{ppo} to alternative models. This transition is necessary due to the nature of \ac{ppo} being an on-policy method, which makes it unsuitable for mixing external information with the data sampled by the current policy. On-policy methods presume that the experience used to update the policy is sampled using the current policy version.  However, in offline \ac{tl}, the optimisation steps conducted on external knowledge are carried out prior to updating the model based on the sampled experience.

\section{Benchmark Environments}\label{sec:preliminary-envs}

To evaluate the impact of experience sharing in an offline \ac{tl} scenario, we used two benchmark environments: \ac{stpp}, introduced in Section~\ref{ss:preliminary-envs-pp}
 and \ac{rs-sumo}, introduced in Section~\ref{ss:preliminary-envs-rs-sumo}.

\subsection{Predator-Prey} \label{ss:preliminary-envs-pp}

Predator-Prey is a task originating from biology where certain species hunt other individuals for survival~\cite{berryman1992orgins}. Different versions are available and this thesis uses a grid-world implementation based on an existing mini-grid environment compatible with OpenAI gym APIs~\cite{gym_minigrid}.

\begin{figure}[htb!]
	\centering
	\includegraphics[width=0.35\columnwidth]{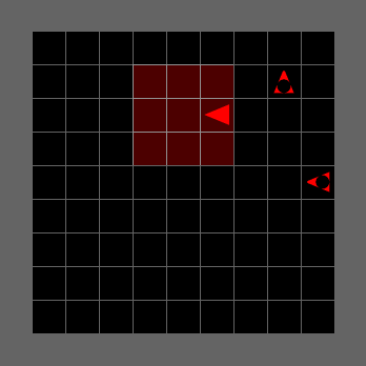}
	\caption{$9x9$ predator-prey grid with one predator and two prey.}\label{fig:preliminary-single-team-pp}
\end{figure}	

The configuration used for this study consists of a $9x9$ obstacle-free grid with a single predator and two prey, as shown in Figure~\ref{fig:preliminary-single-team-pp}, where the predator is illustrated as a filled triangle and the prey as a pierced triangles. The orientation of each entity matches the direction of the triangle. 
 
The objective of a predator is to catch all the prey available within the grid. To accomplish such a goal, a predator has $5$ possible actions to take:

\begin{enumerate}
	\item \textit{hold} $-$ the predator stays within the same cell and does not change its rotation;
	\item \textit{forward} $-$ if the facing cell is empty, the predator moves from the current cell to the next;
	\item \textit{rotate left} $-$ the predator orientation is changed by $-90^\circ$; 
	\item \textit{rotate right} $-$ the predator orientation is changed by $+90^\circ$;
	\item \textit{catch} $-$ the agent catches the object in the facing cell. A catch succeeds if and only if the object within the consecutive cell is a prey. 
\end{enumerate} 

The control level for a prey is analogous to that of the predator, with the exception of the \textit{catch} action. Prey escapes predator by sampling random actions with the following probability distribution: \textit{hold} with a probability of $10\%$, rotate $\pm 90^\circ$ with a probability of $25\%$, and \textit{forward} with a probability of $40\%$.

\subsection{\acl{rs-sumo}}\label{ss:preliminary-envs-rs-sumo}

\ac{rs-sumo} consists of a mobility on-demand scenario that enables ride-sharing while serving real-world ride-requests. 
Ride-requests are replicated by a real-world taxi request dataset obtained from the Manhattan area~\cite{NYC_data}.

While the demand pattern is replicated from the Manhattan yellow cab dataset, traffic patterns and road network infrastructures are simulated through the open-source \ac{sumo} simulator. \ac{sumo} is largely adopted because of its realistic and fully customisable simulation for urban mobility scenarios and provides high granularity in modelling urban transportation elements.

\begin{figure}[h!]
	\centering
	\begin{subfigure}[b]{1\columnwidth}
		\centering
		\includegraphics[width=1\columnwidth]{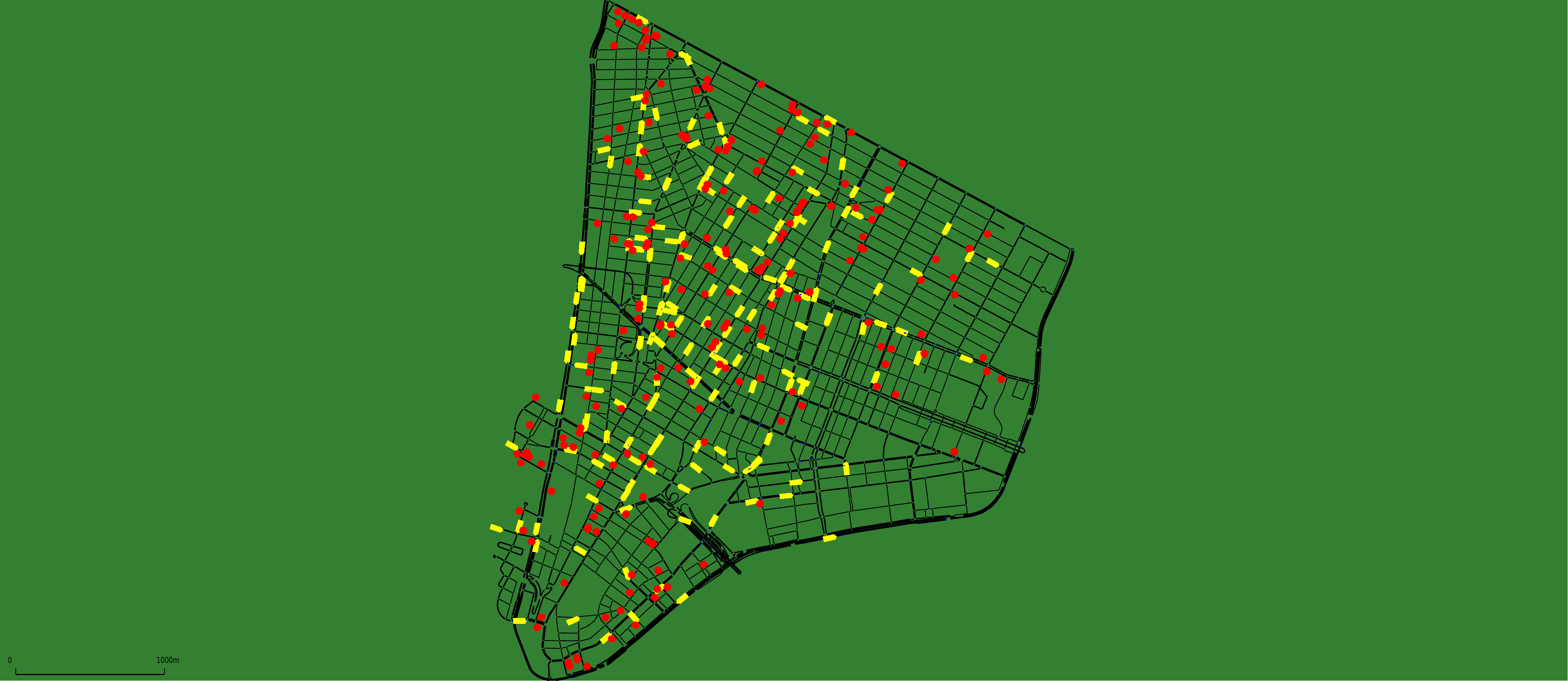}
		\caption{Ride-sharing enabled taxi cabs are depicted as yellow boxes and ride-requests as red circles.}
		\label{subfig:sumo-screenshot}
	\end{subfigure}
	\begin{subfigure}[b]{1\columnwidth}
		\centering
		\includegraphics[width=0.85\columnwidth]{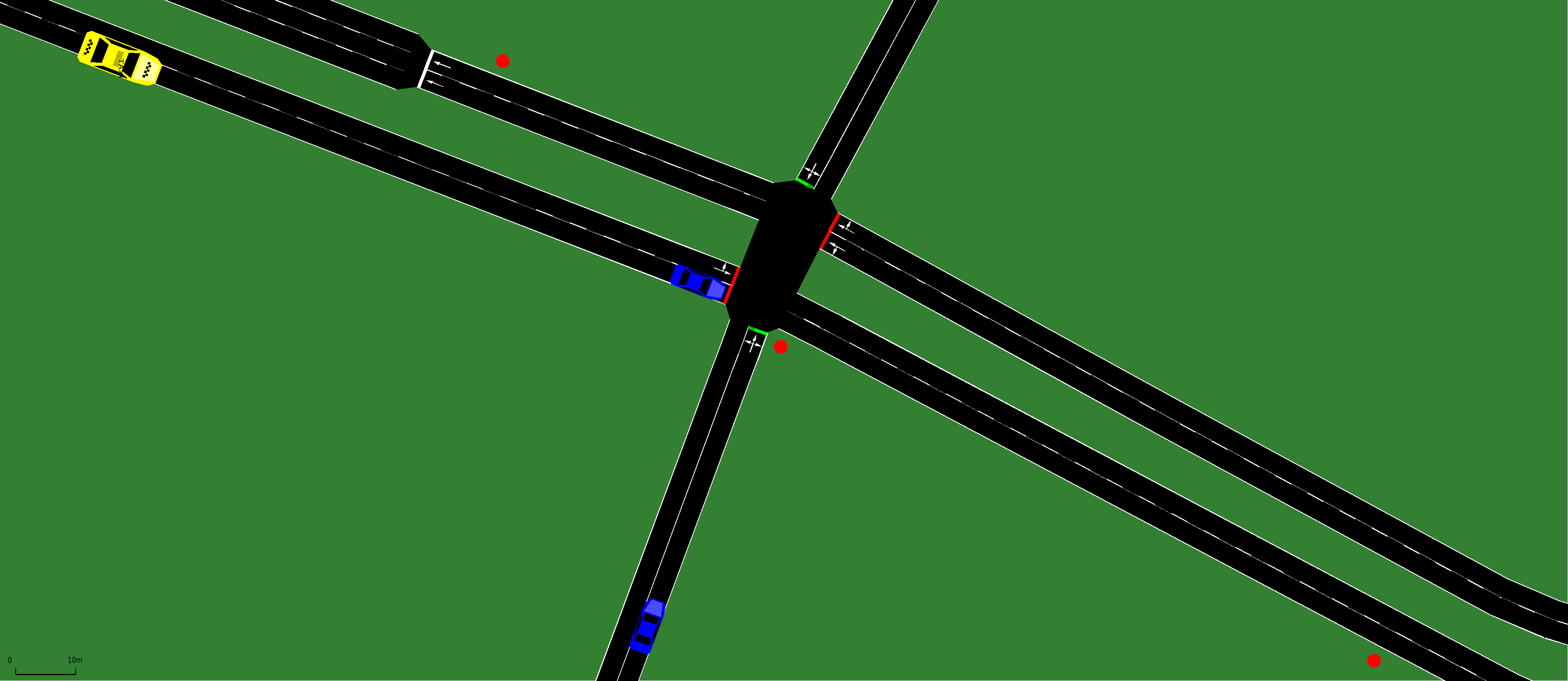}
		\caption{Close view of an intersection with its traffic lights. Taxi cabs are depicted in yellow, regular vehicles in blue, and potential passengers are shown in red.}
		\label{subfig:sumo-state_depiction}
	\end{subfigure}
	\caption{SUMO simulator screenshots.}\label{fig:sumo_screenshots}
\end{figure}

A screenshot taken from the SUMO platform is shown in Figure~\ref{fig:sumo_screenshots}. The upper image displays the Manhattan road infrastructure, while the lower one shows an intersection with a single taxi vehicle and its perceived requests that could potentially be picked-up. The red or green coloured bars intersecting the road lanes in proximity of an intersection represent the status of the traffic lights at the intersection. The vehicles are divided into two groups: traffic vehicles, represented in blue, which are travelling around to simulate traffic patterns; and taxi vehicles, which are coloured in yellow. Finally, the passenger requests are depicted as a stylised top-view human shape and coloured in red.

The problem of matching ride-requests and driver while enabling ride-sharing can be described as a fleet of vehicles travelling around a certain area to maximise the number of satisfied requests while optimising the constrained resources by enabling car sharing.  In this scenario, there is a one-to-one mapping in which each agent is uniquely associated with a single taxi vehicle, and each taxi vehicle is exclusively controlled by an agent. 

From a single agent perspective, the goal is to maximise the amount of ride-requests served throughout its life time while minimising the cumulated delay for the vehicle and any on-board passenger. Control is distributed at vehicle level and there is no communication nor coordination between agents.

In the request dispatcher system, each request is broadcast to the fleet as it occurs and remains active for $15$ minutes. If the request is not engaged within this time frame, it expires and is registered as not served. 

A request is considered engaged when a vehicle begins to drive towards the pickup point to pick up the passengers. As a constraint, the estimated time for the vehicle to reach the request point must fall within the request window time. However, the actual driving time could be longer due to varying traffic conditions compared to the initial estimation. Once the passengers are boarded, they are either travelling towards their destination or detouring to pick-up additional requests to be served in ride sharing by the vehicle.

\begin{figure}[h!]
	\centering
	\begin{subfigure}[b]{1\columnwidth}
		\centering
		\centering
		\includegraphics[width=0.85\columnwidth]{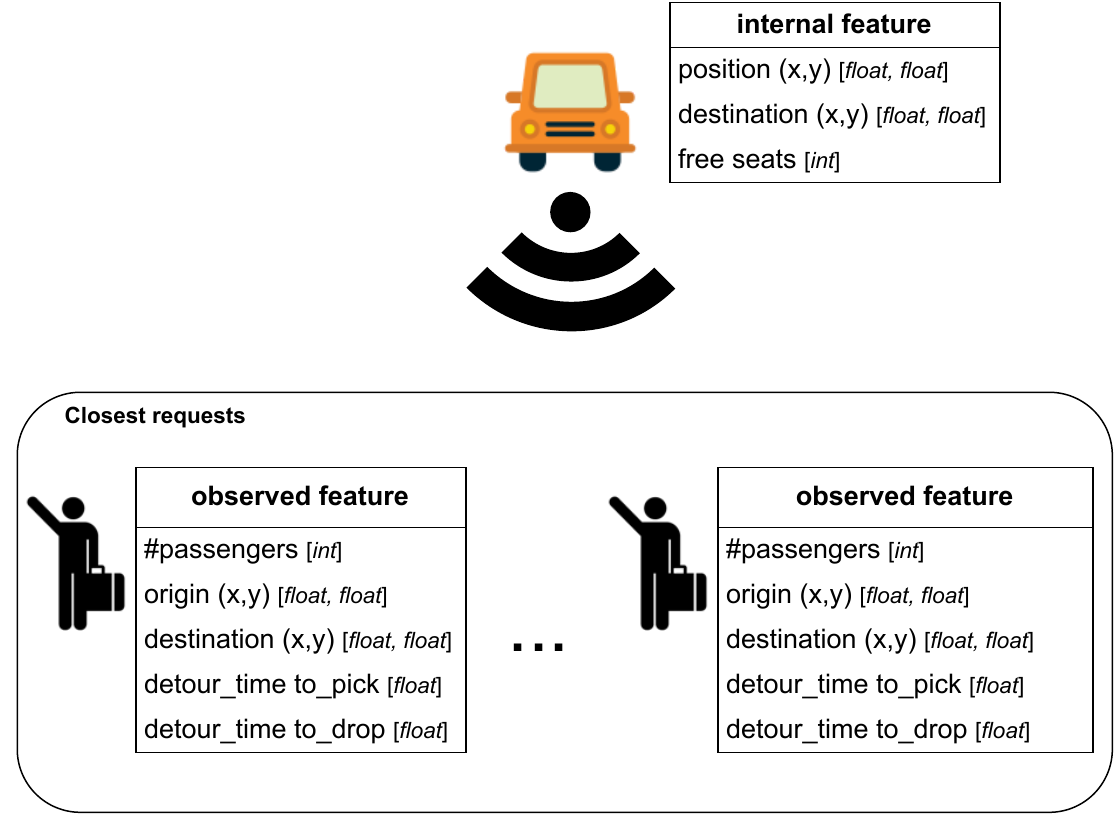}
		\caption{Perception schema \acf{rs-sumo}.}\label{subfig:sumo-rs_perception_schema}
	\end{subfigure}
	\begin{subfigure}[b]{1\columnwidth}
		\centering
		\centering
		\includegraphics[width=.82\columnwidth]{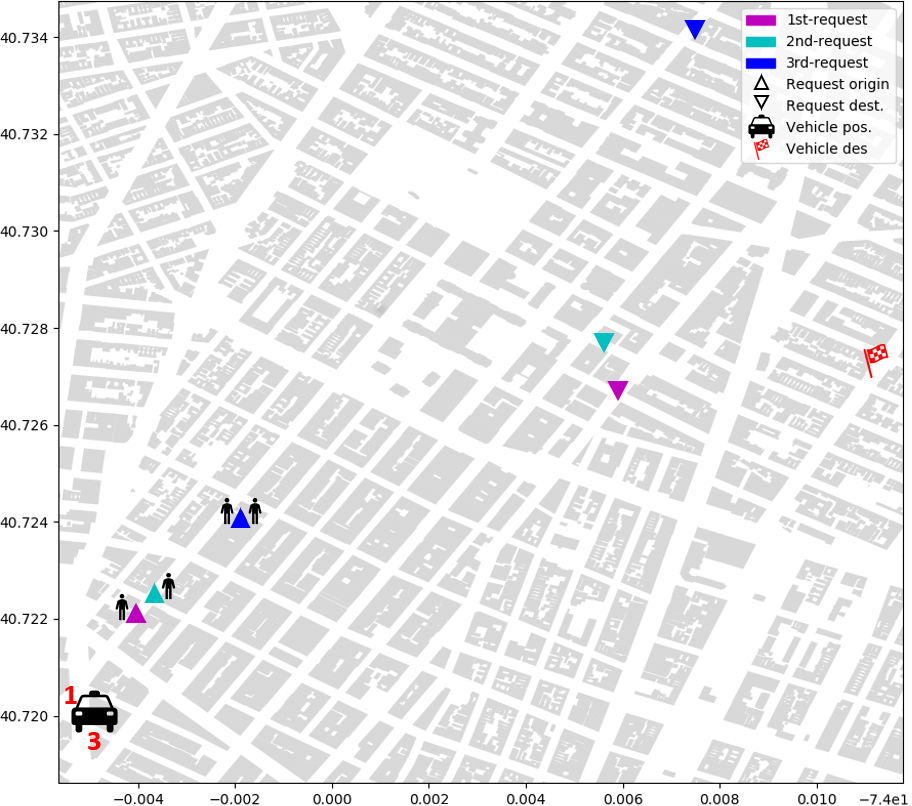}
		\caption{Internal vehicle state and agent's observation over the 3 closest requests that could be engaged by the vehicle.}\label{subfig:rs-state_depiction}
	\end{subfigure}
	\caption{Vehicle internal state and perception.}
\end{figure}

The state of an agent is composed of internal vehicle status alongside the perception of the agent on the closest requests that could potentially be served. At a specific time, given a certain vehicle and a specific request, the request is eligible to be served by the vehicle if there is enough room in the cab to accommodate the incoming passengers and the estimated travel time to reach the pick point of the request is compliant with the request expiration time. 
Internal observation is composed of current position, updated in real-time, and destination with number of empty seats that can be used to welcome new passengers. The destination of a vehicle is updated at the end of an event, i.e., vehicle arrives to pick-up or drop-off location. On top of that, a vehicle may decide to take a detour to pick additional passengers while travelling towards a certain drop-off location. This evaluation step is allowed whenever the vehicle is further from the current destination point and has at least a free seat. A representation of the state of a vehicle and its perception is reported in Figure~\ref{subfig:sumo-rs_perception_schema}.

In the experiments carried out for this research, agent perceives up to $3$ closest requests, as depicted in Figure~\ref{subfig:rs-state_depiction}. Consequently,  control space consists of $5$ actions: 1) \textit{parked} - vehicle stays idle for a predefined amount of time; 2) \textit{drop-off} - vehicle drives towards its destination until it reaches the drop-off point, and offload the passenger(s) there; 3-5) \textit{pick-up} - agent updates its destination to match the origin point of the selected request and drives towards it and passengers are on-boarded when it arrives.  Note that agent can decide to pick one of the three available requests and as a result, agent has three different pick-up options, resulting in $3$ available actions for pick-up. 
All vehicles within the fleet are synchronised under a global clock and action decision process follows a FIFO queue. Considering the nature of the task, an episode begins as vehicles enter the simulation and terminates when all taxis have left after successfully driving all the passengers on-board,  and with no pending requests left to be served. 


As introduced earlier, the reward scheme is designed to incentivise vehicles to accomplish as many requests as possible within the constraints of limited resources, while minimising the accumulated delays of vehicles and passengers.  Therefore, reward is generally based on the elapsed \ac{sumo} steps from the beginning of an action to the decision time for the subsequent action. First, when it is not possible to fulfil an action, agent receives a penalty of $-1$. Second, when agent decides to remain parked, it receives a negative reward~$R_p$ defined over the elapsed parked time as defined in Eq.~\ref{eq:reward_parked}, where $x$ is set to $1$ with empty vehicle and $5$ otherwise.

\begin{equation}
	R_p = \frac{-x}{x+elapsed\_time}
	\label{eq:reward_parked}
\end{equation}

Third, on drop-off reward~$R_d$ is defined by Eq.~\ref{eq:reward_dropoff} if the request has been served in ride-sharing. $noRStime$ is an estimation of the time that the same vehicle would have taken to fulfil the ride-request with similar traffic condition without enabling ride-sharing. $noRStime$ is the estimated travel time needed to travel from the current position to a target location based on the current state of the network. When the request is not served in ride-sharing, $R_d$ is set to $5$.

\begin{equation}
	R_d=\max \big( -1, 5 - \frac{elapsed\_time - noRStime}{noRStime}\big)
	\label{eq:reward_dropoff}
\end{equation}

Last, on pick-up, reward~$R_{up}$ is defined as reported by Eq.~\ref{eq:reward_pickup} where $x$ is set to $1$ with empty vehicle and to $2$ otherwise. $elapsed\_time$ captures the time spent to drive towards the pick location from the moment of the assignment of the request.
\begin{equation}
	R_{up} = \frac{x}{x+elapsed\_time}
	\label{eq:reward_pickup}
\end{equation}


\begin{figure}[h!]
	\centering
	\includegraphics[width=0.75\columnwidth]{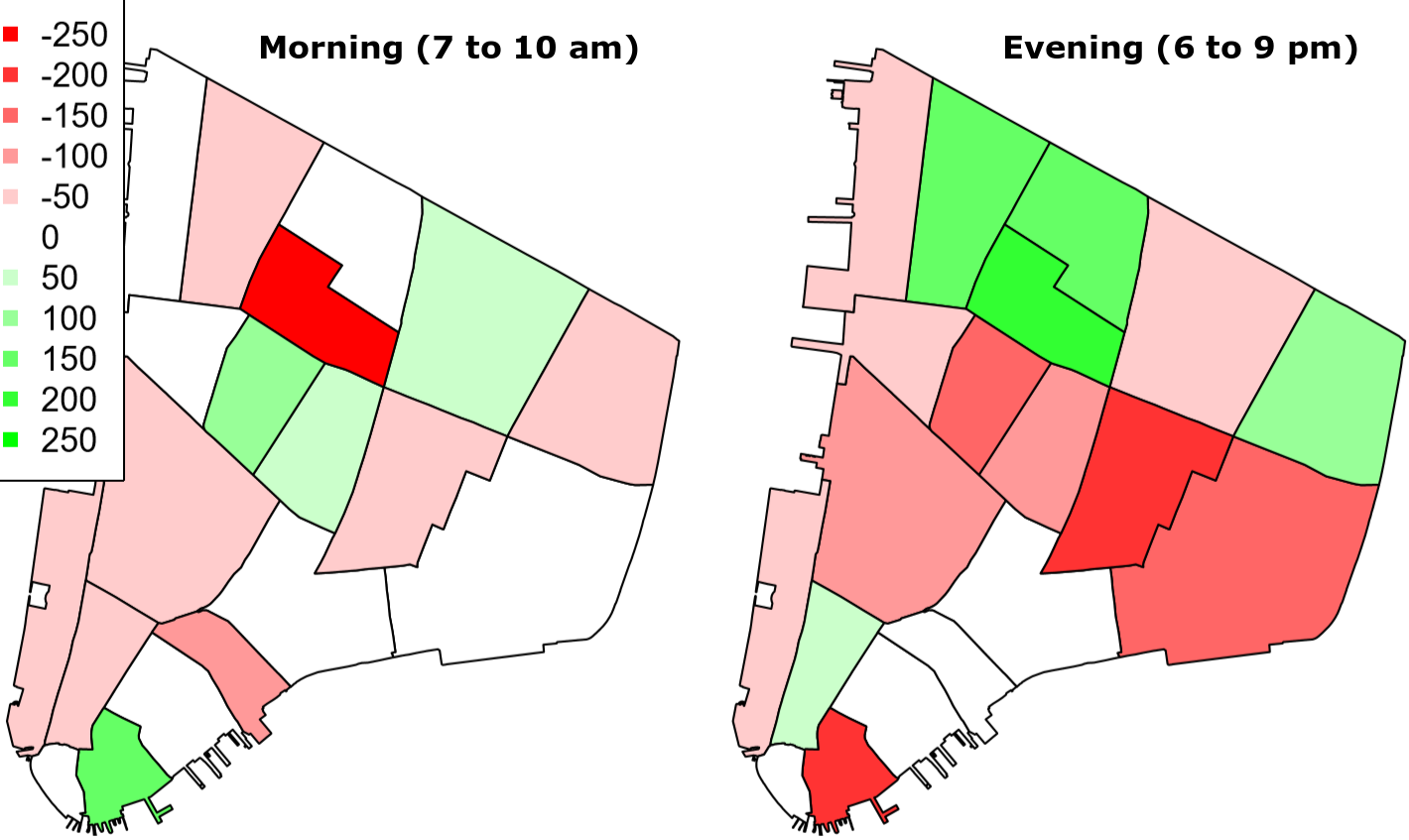}
	
	\caption{Observed demand imbalance in New York Taxi dataset~\cite{NYC_data} trips between morning (7-10am) and evening (6-9pm) peak hours in the south part of Manhattan on Tuesday, February~$2^{nd}$ 2016.}
	\label{fig:imbalance_requests_rs-sumo}
\end{figure}

To emulate a real-world scenario, this thesis utilises ride-request data generously provided by Gu\'eriau and Dusparic at~\cite{gueriau2018samod}. 
NYC taxi trips~\cite{NYC_data} from 50 consecutive Mondays between July 2015 and June 2016 in Manhattan are aggregated and divided based on time-base schedule resulting into $4$ datasets: morning, afternoon, evening and night. In this thesis, all experiments are based on the morning peak slot (from 7 to 10 am) and the evening (from 6 to 9 pm). A depiction of the different demand trend and ride-requests imbalance is reported in Figure~\ref{fig:imbalance_requests_rs-sumo}. In the~\ac{rl} context agents need to generalise across different demand patterns and thus having different demand set with imbalanced zones are crucial to evaluate the generalisation aspect.


To facilitate effective policy learning for the agents, the training is structured into $10$ epochs, each containing $10$ episodes of single-vehicle training. In a given episode, an agent can serve the unmet requests that previous vehicles could not fulfil within that specific epoch. At the beginning of a new epoch, the demand set is fully restored. This strategy emulates a scenario of concurrent exploration by multiple vehicles without any competition among them in serving customers.

All the experiences gathered during training are processed by a single learning process. Consequently, all the agents contribute to the same learning process, optimising the utilisation of accumulated knowledge and accelerating the learning process. 
In cases where a vehicle fails to perform an update or is unavailable to serve requests in a particular location, the others can continue seamlessly. Once training is completed, the acquired knowledge is propagated to all vehicles within the fleet. Furthermore, this allows new vehicles to join the fleet without requiring additional training efforts.

\section{Impact of Experience Sharing in Offline \acl{rl}}\label{sec:eval-offline-tl}

The objective of this offline~\ac{tl} study is to analyse the impact of experience sharing and comparing it against the teacher-student framework presented in Chapter~\ref{cpt:background}. This study serves as an introductory step towards the development of~\ac{efontl} in an online sharing context. In this offline experience sharing scenario, the set of challenges to be addressed is reduced as there is no source selection process as the source of transfer is a trained agent. Specifically, the aim of this study is threefold: 

\begin{enumerate}	
	\item identify and study how pivotal parameters influence the outcomes of transfer, i.e., source uncertainty and quantity of transferred tuples;
	\item assess and quantify the extent of positive transfer achieved through the sharing of external experiences;
	\item compare experience sharing influence against action-advice based teacher-student baselines that provide tailored action to take in critical situations.
\end{enumerate}

The following experiments, demonstrating the impact of experience sharing in an offline-\ac{tl} context, are based on the \ac{ppo} model presented in Section~\ref{ss:ppo_offline-tl}. The method and the baselines are applied to the \ac{stpp} environment and \ac{rs-sumo}. 

The experience transfer buffer is provided by an external agent that has already encountered and solved the task. The sharing of experience and the integration process within the target learning model happens at initialisation time. The target agent receives a buffer with tuples labelled by source agent's uncertainty and samples a subset of the available collected knowledge. To estimate confidence, source agent is equipped with a~\ac{rnd} model that enables the agent to approximate its epistemic uncertainty while learning the task. The source agent stores all the tuples visited during training along with the associated epistemic uncertainty estimated by the uncertainty model.

Compared to online~\ac{tl}, this offline experience sharing method has a reduced complexity as it assumes the existence of a pre-identified teacher agent, which is certainly more qualified than the novel target agent. To benchmark the impact of sharing experience as pre-training phase, the offline transfer approach is compared against \textit{no-transfer}, where a target agent does not utilise external knowledge, and action-advice based teacher-student framework following various advising strategies, i.e., advice at the beginning~(\textit{BS: adv\_at\_begin}), confidence based~$\epsilon$-decay~\cite{norouzi2021experience}~(\textit{BS: confidence\_based\_e-decay}) and mistake correction~(\textit{BS: mistake\_correction}). In \textit{BS: confidence\_based\_e-decay}, a target agent asks for advice following an 
$\epsilon$-decay probability and when its more uncertain than the teacher in the visited state.

\begin{figure}[h!]
	\centering
	\begin{subfigure}[b]{.5\columnwidth}
		\centering
		\includegraphics[width=\columnwidth]{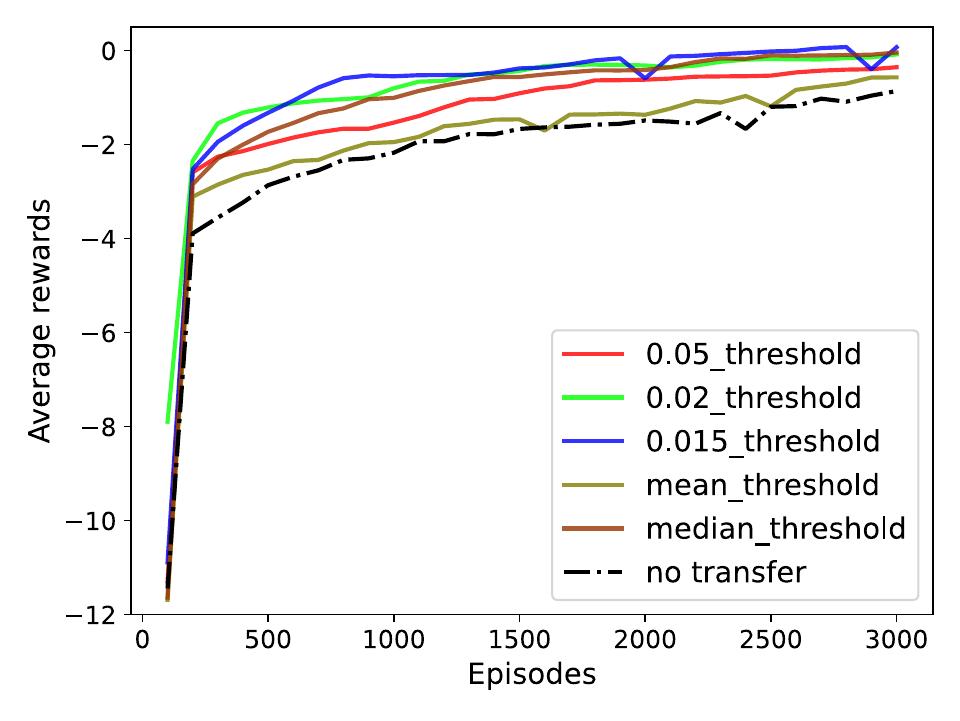}
		\caption{Average rewards per different threshold with pre-training over $5,000$ agent-environment interactions.}
	\end{subfigure}~
	\begin{subfigure}[b]{.5\columnwidth}
		\centering
		\includegraphics[width=\columnwidth]{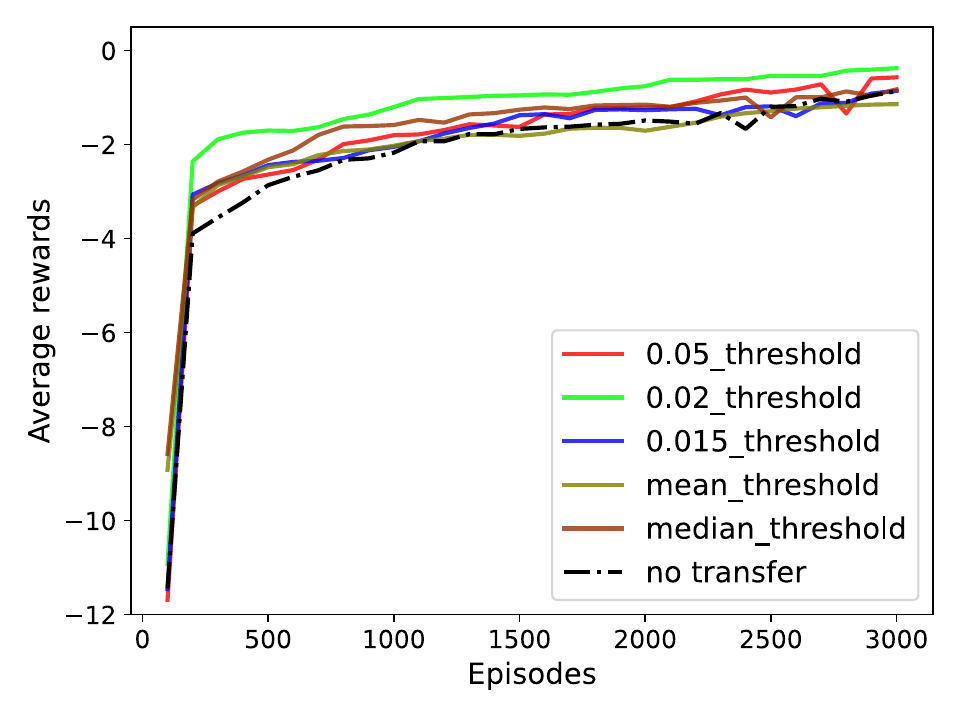}
		\caption{Average rewards per different threshold with pre-training over $10,000$ agent-environment interactions.}
	\end{subfigure}~	
	\caption{Comparison of offline experience sharing in \ac{stpp} across different filtering thresholds. Results are shown as average learning curve across 50 independent simulations.}
	\label{fig:offline-tl-pp}
\end{figure}

Figure~\ref{fig:offline-tl-pp} shows two different graphs with the average rewards over $3,000$ episodes of the pre-training enabled algorithm across different filtering thresholds and a \textit{no transfer} baseline. Figure~\ref{fig:offline-tl-pp}a enable an agent to sample $5,000$ interactions during the pre-training step, while Figure~\ref{fig:offline-tl-pp}b the pre-training budget is doubled to $10,000$. 
These experiments use three fixed thresholds: $0.05$, $0.02$ and $0.015$, and two adaptive thresholds based on the content of the buffer available: \textit{mean} and \textit{median}. The \textit{mean} threshold enables a target agent to sample tuples with an uncertainty level lower than the average uncertainty  computed across all interactions in the buffer. On the other hand, the \textit{median} threshold enables a target agent to sample tuples with an uncertainty level below the median uncertainty of all interactions.

These experiments show that pre-training a learning model with external experience coming from a trained agent successfully improves the performance of a target agent.  The positive impact appears to be unmatchable by a \textit{no transfer} agent. In fact, while jumpstart may not be consistent across the~\ac{tl} enabled agents, there seems to be an asymptotic improvement brought by the pre-training phase that enables the target agent to keep a higher level of performance from the beginning to the last episode. Finally, although there are minor differences based on the filtering threshold selected, the predominant feature, from these graphs, is the quantity of shared experiences.

Intuitively, the more information is provided to a novel agent, the better it is expected to adapt and learn the task. However, the results provided by Figure~\ref{fig:offline-tl-pp} empirically contradict this intuitive assumption by showing an opposite trend. In fact, in Figure~\ref{fig:offline-tl-pp}a, all the thresholds show an improvement against \textit{no transfer}, while in Figure~\ref{fig:offline-tl-pp}b, the impact given by the selected threshold is milder, and a few scenarios show no improvement at all or even negative transfer when compared against \textit{no transfer}.


To follow, the threshold that enabled the~\ac{tl} target agent to achieve higher overall performance and the \textit{mean} threshold are compared against the~\ac{tl} baselines to compare the impact of transferring agent-environment interactions against action-advice based baselines.

\begin{figure}[htb!]
	\centering
	\centering
	\begin{subfigure}[b]{.5\columnwidth}
		\centering
		\includegraphics[width=\columnwidth]{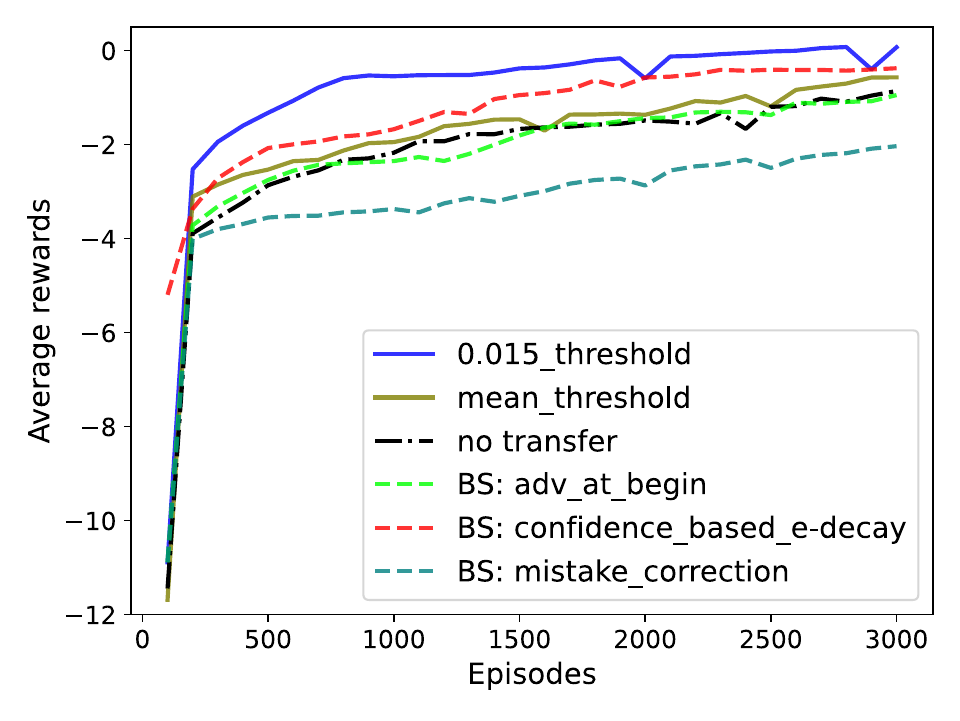}
		\caption{Average rewards per different algorithm type with pre-training over $5,000$ agent-environment interactions.}
	\end{subfigure}~
	\begin{subfigure}[b]{.5\columnwidth}
		\centering
		\includegraphics[width=\columnwidth]{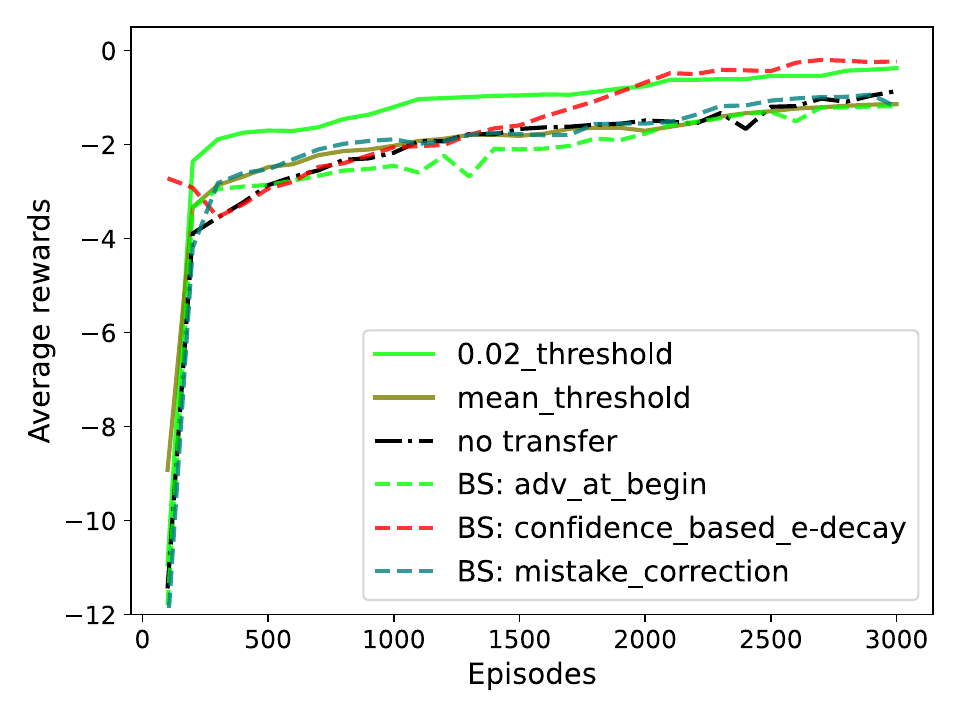}
		\caption{Average rewards per different algorithm type with pre-training over $10,000$ agent-environment interactions.}
	\end{subfigure}~	
	\caption{Comparison of offline experience sharing in \ac{stpp} against the teacher-student framework.}
	\label{fig:offline-tl-pp-vs-baseline}
\end{figure}

Figure~\ref{fig:offline-tl-pp-vs-baseline} shows two different graphs for the total reward over $3,000$ training episodes. The leftmost graph shows the~\ac{tl} algorithms with a budget up to $5,000$ while the rightmost graph depicts the experiments with a budget up to $10,000$.

These comparisons suggest that experience-based~\ac{tl} can compete against action-advice based baseline. Specifically, three main findings arise, as follows. 
Firstly, providing action as advice by taking into account the confidence measure of both source and target allows for a better budget utilisation and it leads to better performance when compared to \textit{BS: mistake\_correction} and \textit{BS: adv\_at\_begin}. Secondly, despite the remarkable jumpstart exhibited by the \textit{BS: confidence\_based\_e-decay} compared to \textit{no transfer} within the initial $50$ episodes, the \textit{0.015\_threshold} agent achieves a similar level of rewards in a shorter number of episodes. Furthermore, the interactions processed as a pre-training step are general and not tailored to overcome a specific situation. On the other hand, the advice provided by action-advice based baseline are specific to overcome a certain situation. Thirdly, while the filtering threshold plays a role into influencing the target agent performance, the predominant feature driving the transfer outcome is the amount of transferred interactions.

As second part of the initial offline evaluation phase, transfer of experience is applied to the \ac{rs-sumo} environment while varying the underlying demand pattern.
Therefore, the following scenarios are evaluated:

\begin{itemize}
	\item \textit{Train \& test 7-10am} $-$ agent is trained and tested on the same set of requests, the morning peak hours. This is considered as a baseline as in this scenario there is no transfer;
	
	\item \textit{Train \& test 6-9pm} $-$ agent is trained and tested on the evening peak hours. This is a transfer-free baseline;
	
	\item \textit{Train 7-10am test 6-9pm} $-$ agent is trained on the morning peak hours and tested on the evening demand set. In this baseline a policy trained on a demand set is tested onto another to assess the capability of a policy with no experience sharing to generalise across different demand patterns;
	
	\item \textit{\ac{tl}-enabled test 6-9pm} $-$ agent is trained and tested on the evening demand set. However, this scenario enables~\ac{tl} as the agent accesses a transfer buffer containing the experiences collected during a training of the morning peak hours to perform a pre-training step.
	
\end{itemize}

These scenarios evaluate whether and to what extent the transfer of experiences among agents, defined within the same domain but based on different underlying data, can result in positive transfer to a target agent.

\begin{table}[ht!]
	\renewcommand*{\arraystretch}{1.1}
	\caption{\label{tab:mod_results}{Performance metrics across all 4 evaluated approaches 	in \ac{rs-sumo} environment.}}
	\centering
	\begin{threeparttable}
		\begin{tabular}{|c|c|c|c|c|c|c|}
			\hline
			\multirow{1}{*}{{\textbf{Scenario}}} &\textbf{{\%Served requests}}&\multirow{1}{*}{\textbf{{\%RS}}}& \textbf{$\boldmath{\sigma}$  {pass}} & \textbf{${\overline{d_t}}$ {\textit{(km)}}} &\multirow{1}{*}{\textbf{$\boldmath{\overline{D_r}}$}}\\ 
			\hline
			Train \& test 7-10am &  \multirow{1}{*}{93} & \multirow{1}{*}{93} &\multirow{1}{*}{76.61}  &\multirow{1}{*}{140} &\multirow{1}{*}{9.4} \\
			\hline
			Train 7-10am test 6-9pm &  \multirow{1}{*}{77} & \multirow{1}{*}{99}  & \multirow{1}{*}{21.65} & \multirow{1}{*}{87}&\multirow{1}{*}{9}\\
			\hline
			Train \& test 6-9pm & \multirow{1}{*}{76} &  \multirow{1}{*}{99}  & \multirow{1}{*}{19.35} & \multirow{1}{*}{87}&\multirow{1}{*}{9.11}\\
			\hline
			{TL-enabled test 6-9pm}  &\multirow{1}{*}{{79}}  &\multirow{1}{*}{{93}} & \multirow{1}{*}{{41.12}} & \multirow{1}{*}{{82}}&\multirow{1}{*}{{5.23}}\\
			\hline
		\end{tabular}
	\end{threeparttable}
\end{table}

\begin{figure}[h!]
	\centering
	\begin{subfigure}[b]{1\columnwidth}
		\centering
		\includegraphics[width=\columnwidth]{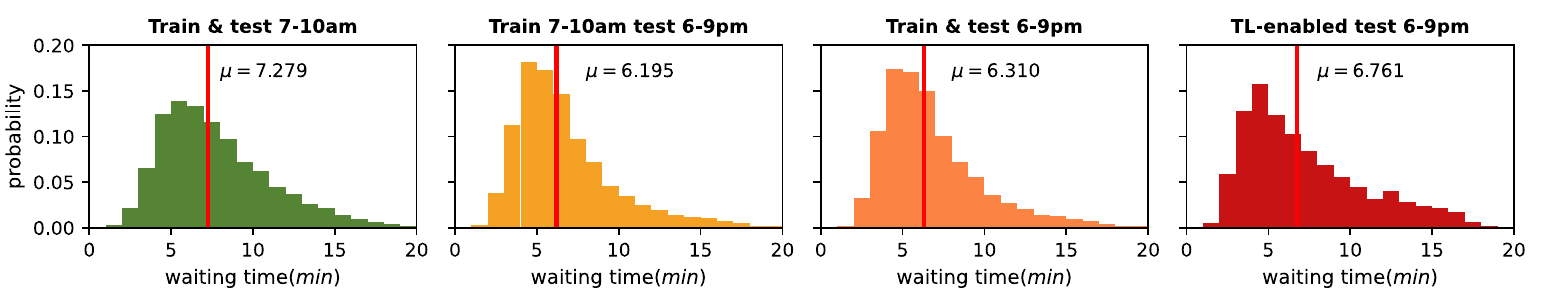}
		\caption{Distribution of waiting time over the requests.}
	\end{subfigure}
	\begin{subfigure}[b]{1\columnwidth}
		\centering
		\includegraphics[width=\columnwidth]{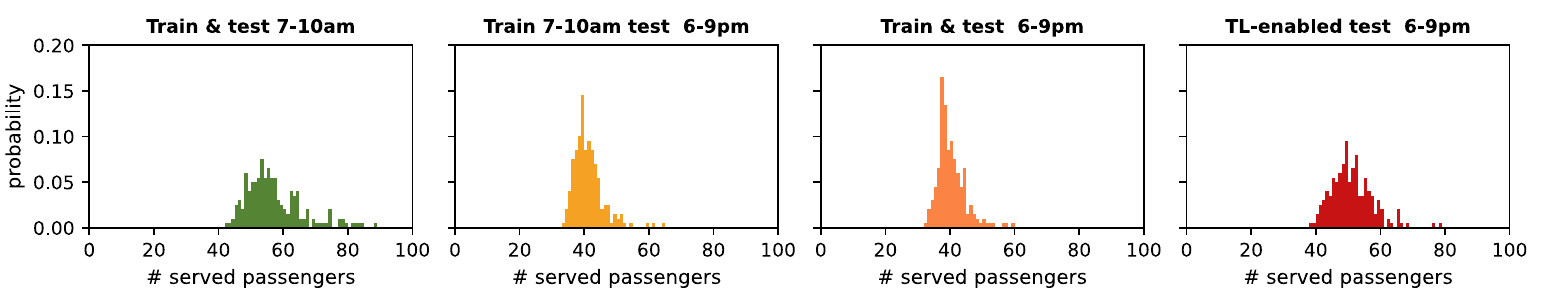}
		\caption{Distribution of passengers served per vehicle.}
	\end{subfigure}
	\begin{subfigure}[b]{1\columnwidth}
		\centering
		\includegraphics[width=\columnwidth]{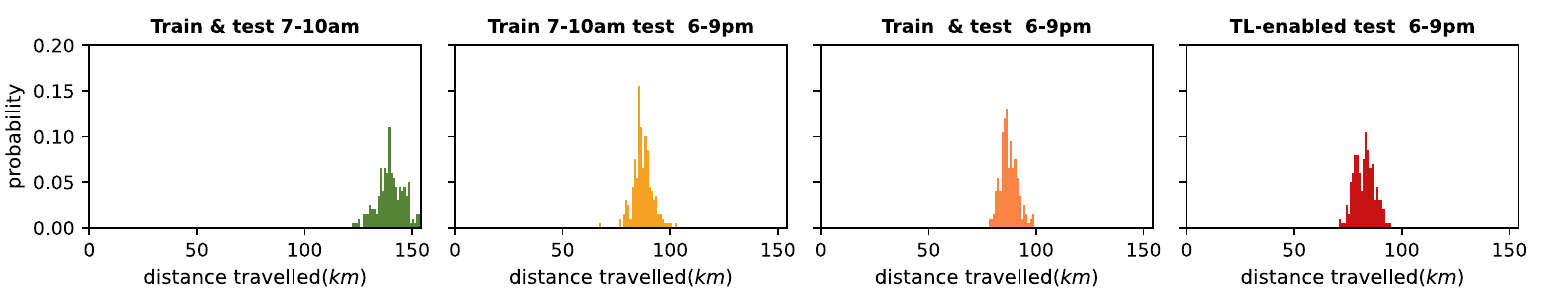}
		\caption{Distribution of distance travelled across the fleet.}
	\end{subfigure}~	
	\caption{Evaluation metrics in the \ac{rs-sumo} environment across 4 different scenarios.}
	\label{fig:offline-tl-rs_sumo}
\end{figure}

The averaged results are presented by Table~\ref{tab:mod_results} while Figure~\ref{fig:offline-tl-rs_sumo} shows the distribution across the fleet and requests. In detail, it reports the waiting time distributions for requests in \textit{(a)}, passengers distribution in \textit{(b)}, and mileage distribution in \textit{(c)} across the fleet. 

The first improvement resulting of sharing experience in the \ac{rs-sumo} environment is represented by the increase of served requests, as reported in Table~\ref{tab:mod_results}. 
When an agent samples a subset of pre-collected experiences to pre-train its learning model, (\textit{TL-enabled test 6-9pm}), it enables a fleet to satisfy an additional $3\%$ of ride-requests when compared to a similar agent that does not utilise external knowledge on same demand set, (\textit{Train \& test 6-9pm}). Despite the additional served requests, the average distance travelled~($\overline{d_t}$) is lowered by 5 km per vehicle, 82km against 87km, when compared to a no-transfer agent (\textit{Train \& test 6-9pm}). Furthermore, passenger metrics improved as well, by lowering the detour ratio~($\overline{D_r}$). $\overline{D_r}$ is defined as the ratio between the estimated driving distance from the pick-up to the drop-off location with no detour and the actual travelled distance by the vehicle while the passengers are onboard. Finally, passengers distribution~($\sigma$ pass) in the \ac{tl} enabled scenario, (\textit{TL-enabled test 6-9pm}), is unbalanced compared to no-transfer scenarios tested on the same dataset, (\textit{Train \& test 6-9pm}) and (\textit{Train 7-10am test 6-9pm}), as variance is almost doubled, 41.12 against 19.35 and 21.65, respectively.

In the \textit{TL-enabled test 6-9pm}, the average waiting time is increased by less than 1 minute, as a result of serving additional passengers. However, as shown in Figure~\ref{fig:offline-tl-rs_sumo}a, only a few requests have an increased waiting time above $15$ minutes and those are very likely the cause of the increased waiting time.

\section{\acl{rnd} Considerations}\label{sec:sarnd_eval}

The preliminary experiments have shown that scalar estimate of agent epistemic uncertainty influences the performance of a target agent when transferring external experiences. 

Therefore, this section discusses \acf{rnd}, introduced in Section~\ref{sec:back_uncertainty}. \ac{rnd} is a two-network model used to approximate agent epistemic uncertainty on visited states. Consequently, this section highlights \ac{rnd} shortcomings when used as uncertainty estimator in an online \ac{tl} scenario. Finally, this section presents~\ac{sarnd}, our proposed extension of~\ac{rnd}, to overcome the intrinsic \ac{rnd} limitations.

Before jumping into~\ac{rnd} limitations, this section evaluates~\ac{rnd} estimation capability based on the amount of encoded features compared between the two networks.  The encoder size matches the number of neurons used within the output layer.

\begin{figure}[h!]
	\centering
	\includegraphics[width=0.75\columnwidth]{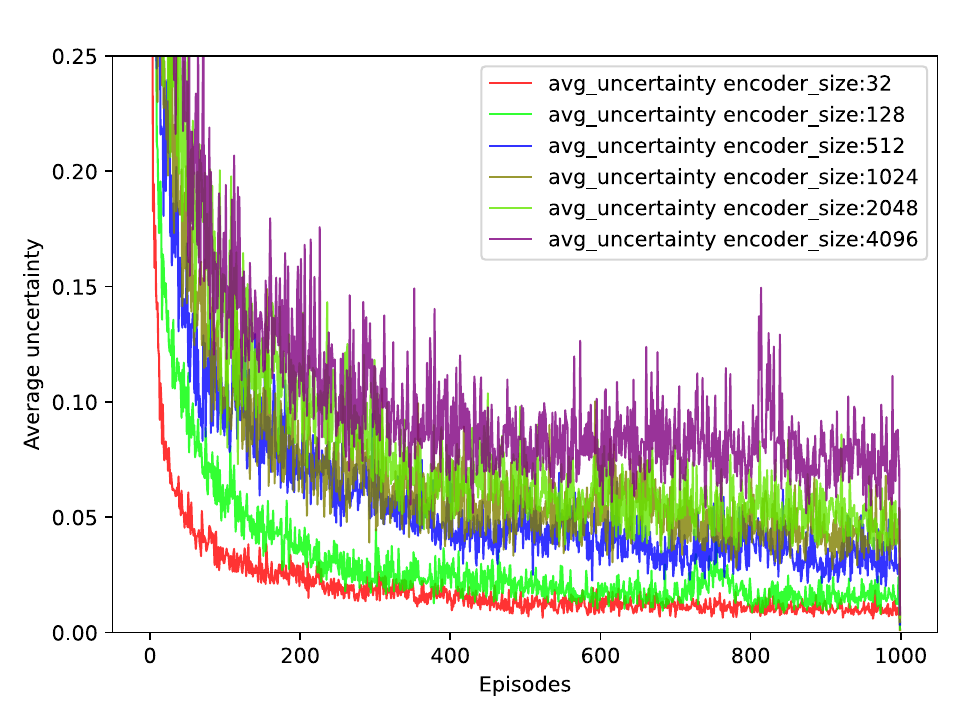}
	\caption{Uncertainty estimator with different encoder size for \ac{rnd} model in~\ac{pp}.}
	\label{fig:rnd_encoder-size}
\end{figure}

While a bigger encoder size enables the estimator to be more sophisticated, it might lead to overfitting when estimating the output of the other network and, consequently,  when estimating the uncertainty. On the other hand, a smaller encoder size does not allow for a sufficient number of features to be compared between the two networks.
To define the size used within this thesis, Figure~\ref{fig:rnd_encoder-size} presents a \ac{rnd} sensitivity analysis while varying the encoder size on the~\ac{pp} problem. In detail, the figure shows the maximum uncertainty registered within the first $1,000$ episodes. Encoder size identifies the number of features that are used to compute the \ac{mse} between the frozen random network and the optimised one. A general trend observed from the graph is that uncertainty is sharply decreased within the first episodes and thereafter anneals towards a small value close to $0$.

Regardless of the chosen encoder size, the uncertainty model presents a descending trend. A higher number of features makes the estimator more sensitive to newer states while a lower number generalises too much across different states. Therefore, for all the experiments presented in this thesis, the encoder size is set to $1,024$ as a balanced trade-off which provides sufficient granularity to generalise across similar states while allowing the model to react to new or infrequently visited states.

In Section~\ref{sec:eval-offline-tl}, we demonstrated that \ac{rnd} can be successfully used to identify meaningful agent-environment interactions that result in increased performance to a novel agent. The tuples processed by the target agent are selected based on a threshold level. Given the observed decreasing trend registered in Figure~\ref{fig:rnd_encoder-size}, the tuples processed by the target agent are very likely to be collected within the latest training episodes of the source agent. This hypothesis is supported by the experiments with a smaller transfer budget shown in Figure~\ref{fig:offline-tl-pp}a, where a lower threshold~($0.015$) increases the chances that a shared tuple is sampled from the latest part of the training process. Similarly, increasing the threshold~($0.02$ and $0.05$) reduces the probability of sampling from the latest part of training and as a result, it gradually deteriorates the performance. However, when doubling the transfer budget, from $5,000$ to $10,000$,  this hypothesis is not true as the lower threshold that was performing the best earlier is now the worst among the three. 

The deterioration of performance with an increased budget might be due to the environment that leads predator to visit very often similar states that hold little or no information while taking different actions. Given the sparse reward function and the high probability that an agent is in a state with no information, there might be many tuples transferred that are related to these situations that add no value to the agent's knowledge. For instance, in the \ac{pp} game screenshot provided by Figure~\ref{fig:preliminary-single-team-pp}, the predator is in a not useful state. The predator, filled triangle in cell ($6,3$) from the top-left corner, will transition to a same perceived situation with any of the available actions (\textit{hold}, \textit{forwards}, \textit{rotate right}, \textit{rotate left} and \textit{catch}). As an agent perceives a $3x3$ grid-based on the next consecutive cell of the controlled predator, there is no way of perceiving anything different from empty cells by taking a single step in this specific setup.

When transferring agent-environment interactions, based on uncertainty, from an experienced agent that has already completed the training phase, the redundant visits to states with little information may not be a problem.  In fact, this issue can be mitigated by selecting tuples based on uncertainty, thereby prioritising recent experiences, and by regulating the transfer budget allowing for a limited number of redundant visits. Nevertheless, when transferring in an online~\ac{tl} scenario, adjusting the budget in an ad-hoc manner will not alleviate the issue. The source agent will not have completed its training process yet, and consequently, it will not have sufficiently visited the successful state-action combinations.

\subsection{\acs{rnd} and \acs{sarnd} Comparison}\label{ss:rnd_sarnd_comparison}

\ac{rnd} has proven to be a valid uncertainty estimator model to be used within an offline~\ac{tl} context where tuples are selected from agents that have completed the full training process. However, when applied in an online context, \ac{rnd} lacks additional information that should be taken into account when selecting experiences to share from imperfect agents. This limitation arises because~\ac{rnd} estimates confidence based solely on observed states, assigning high confidence to states the agent frequently visits regardless of the action taken. Consequently, it might fail to accurately estimate the confidence for states where the agent has rarely, or even never, explored a successful state-action combination that lead to the goal fulfilment.

To overcome the limitation of \ac{rnd}, which estimates epistemic uncertainty solely from visited states, this thesis proposes~\ac{sarnd}. \ac{sarnd} is an extension of~\ac{rnd} that reduces the estimation bias by introducing additional information in the uncertainty estimation. \ac{sarnd} estimates epistemic uncertainty from a full \ac{rl} tuple $(s_t,a_t,s_{t+1}, r_{t+1})$. As a result, \ac{sarnd} is expected to estimate a higher uncertainty in never encountered states while generalising across similar agent-environment interactions.

\begin{figure}[h!]
	\centering
	\includegraphics[width=0.5\columnwidth]{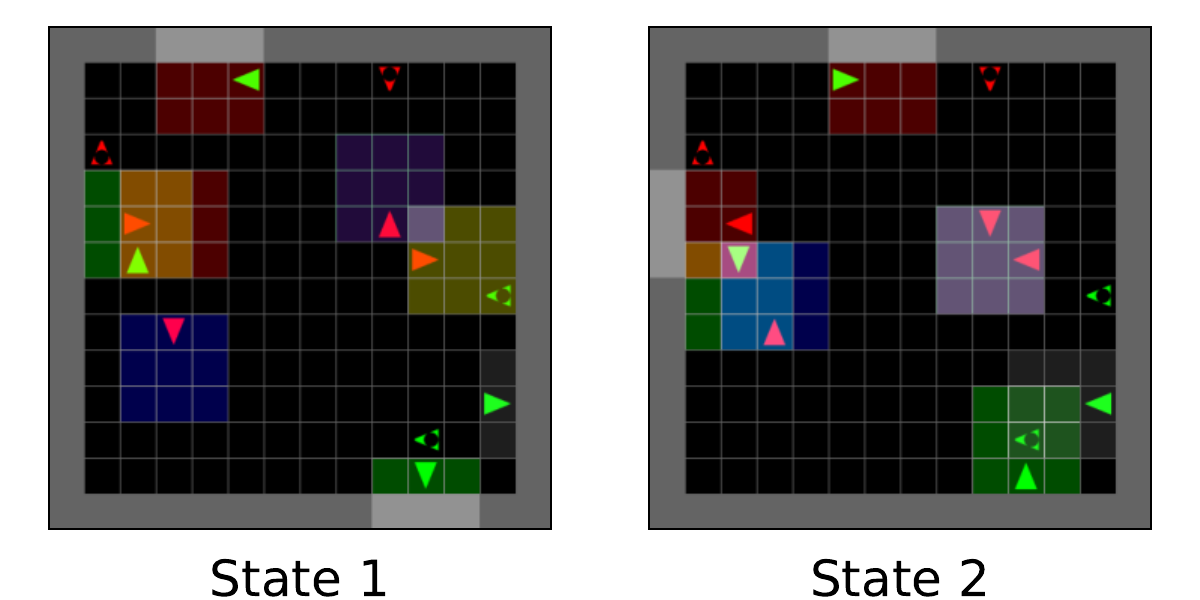}
	\caption{Two states in \ac{pp} environments.}
	\label{fig:pp-states-for-uncertainty}
\end{figure}

To show the limitation of \ac{rnd} and demonstrate the effectiveness of~\ac{sarnd}, we test both models within the~\ac{pp} benchmark environment. In this scenario, we let the agents visiting two states, shown in Figure~\ref{fig:pp-states-for-uncertainty}, while sampling various actions to analyse the uncertainty curves generated by the two models.

\begin{figure}[htb!]
	\centering
	\includegraphics[width=0.65\columnwidth]{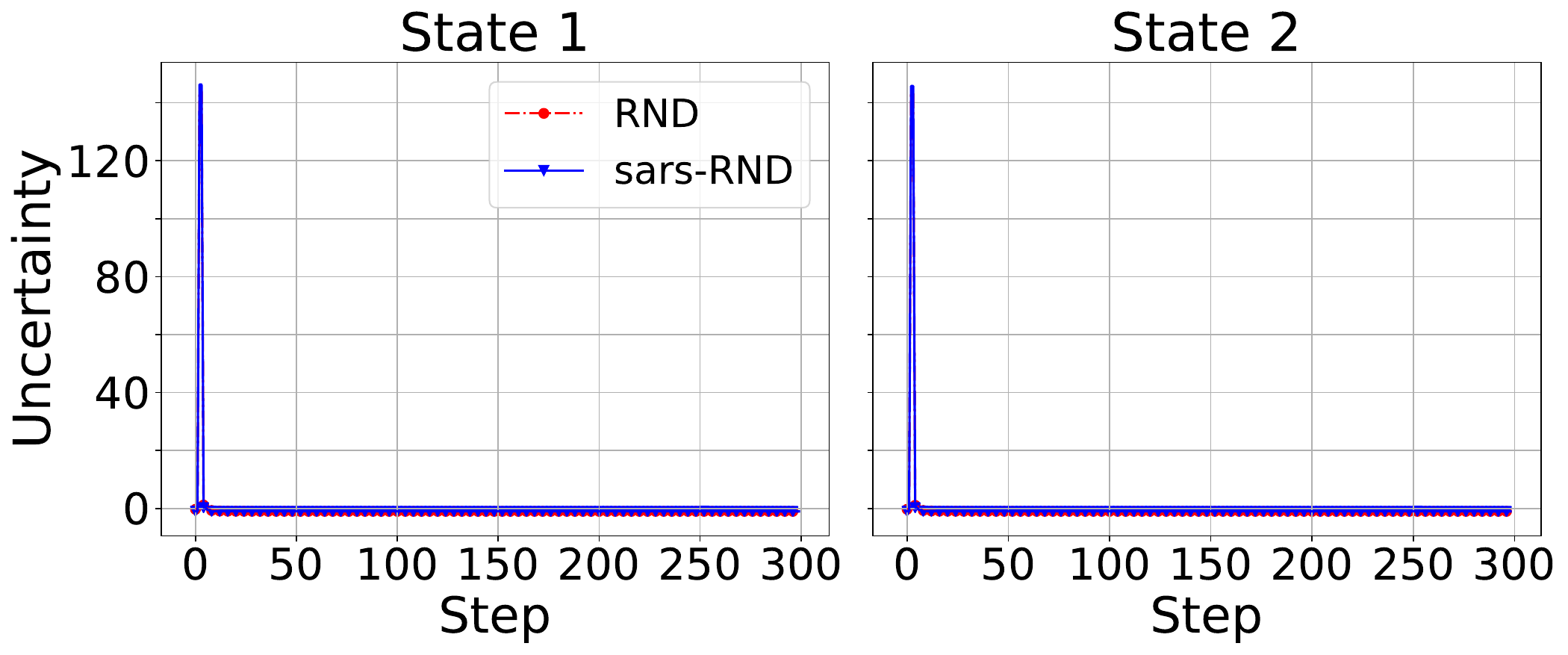}
	\caption{~\ac{rnd} and~\ac{sarnd} uncertainty curves, with their 95\% confidence interval, while sampling different actions in \ac{pp} environment.}
	\label{fig:sarnd_vs_rnd}
\end{figure}

To ensure a fair comparison, the two estimator models share the same architecture except for the input layer. \ac{sarnd} input layer is composed by a few extra neurons to accommodate action, reward and next state. To assess the uncertainties, agents take a fixed action \textit{hold} for $250$ steps to strengthen the prediction across both estimators. Afterward, action sampled is changed to a new action~\textit{rotate left} for a shorter number of steps~($25$). Finally, agents resample~\textit{hold} for another $25$ steps. The uncertainty trends are reported in Figure~\ref{fig:sarnd_vs_rnd}.

Overall, both \ac{rnd} and \ac{sarnd} uncertainty curves follow the same decreasing trend, as shown in Figure~\ref{fig:sarnd_vs_rnd}. The estimated uncertainty is high during the first steps and decreases asymptotically to $0$ as the agents visit the states. However, by evaluating uncertainty in a narrowed interval as shown in Figure~\ref{fig:sarnd_vs_rnd}, \ac{rnd} keeps a flat trend while \ac{sarnd} registers a spike in uncertainty on the change of the action, firstly, at step~($250$), when action $a_0$ is replaced by $a_1$ and secondly, at~$275$ when $a_0$ is resampled. Second spike is smother compared to first as \ac{sarnd} recognises the already seen interaction. We observe that both for states under examination, state $1$ and state $2$.

\begin{figure}[htb!]
	\centering
	\includegraphics[width=0.65\columnwidth]{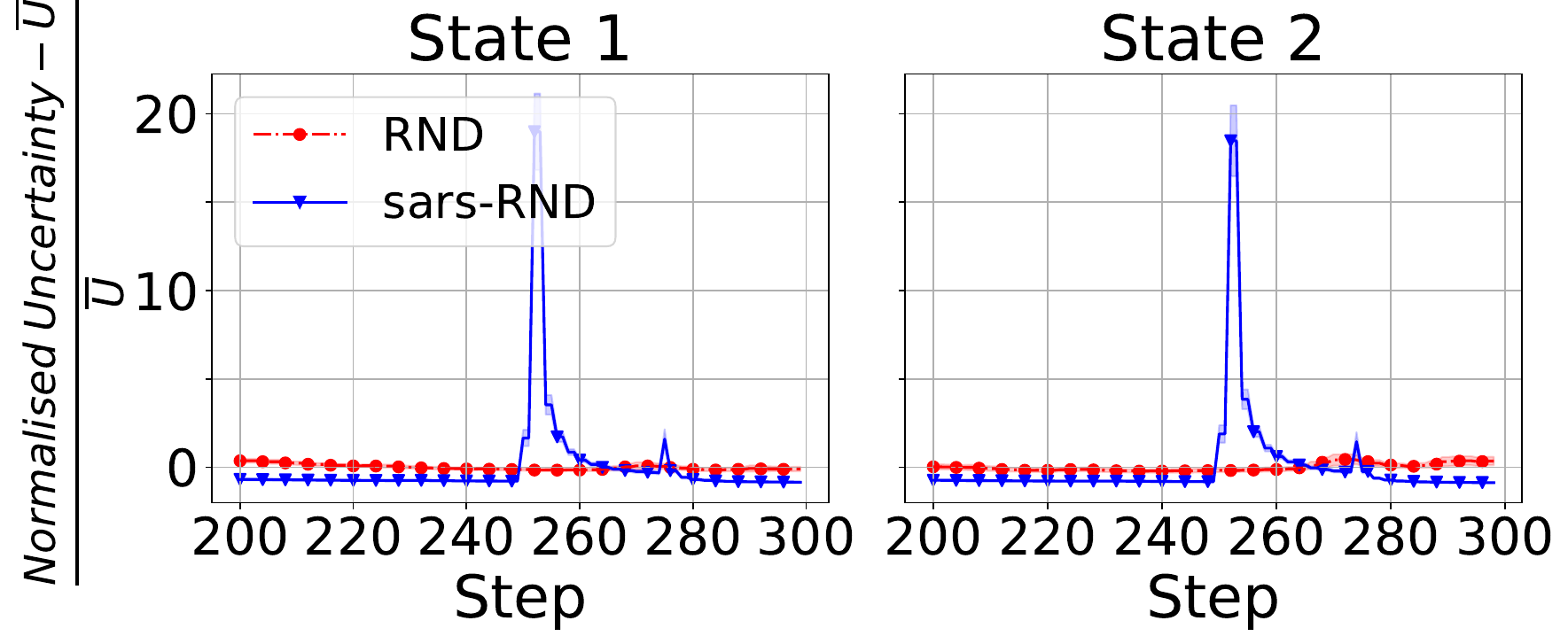}
	\caption{Zoomed in normalised uncertainties curves, with their 95\% confidence interval, while sampling different actions in \ac{pp} environment.}
	\label{fig:sarnd_vs_rnd_zoomed}
\end{figure}

This artificial study confirms that~\ac{sarnd} offers a fine-grained estimation of uncertainty by considering both the executed action and the visited state. In contrast, \ac{rnd} is not sensitive to variations in the action taken. Moreover, the observed decreasing trend, reported in Figure~\ref{fig:sarnd_vs_rnd}, suggests that~\ac{sarnd} generalises effectively, as it can recognise familiar states even when sampling different actions. Consequently, the uncertainty estimated for never encountered states is expected to be higher than that for previously seen states.

Based on the experiments presented in this section, we conclude that~\ac{sarnd} stands as a more suitable choice for online transfers in comparison to~\ac{rnd}. The ability of~\ac{sarnd} to estimate epistemic agent uncertainty from a full \ac{rl} agent-environment interaction effectively overcomes the limitation of~\ac{rnd}, while maintaining a comparable trend that gradually approaches 0 over time. Consequently, \ac{sarnd} is the model used with~\ac{efontl} in the online~\ac{tl} experiments \ac{efontl}.

\section{Conclusion}

The evaluation study, presented in Section~\ref{sec:eval-offline-tl}, on offline experience sharing based on the use of confidence measure establishes the foundation for the development of~\ac{efontl} for online experience sharing across multiple learning agents.

The evaluation results in~\ac{pp} reveal that using agent confidence as a transfer decision metric can lead a target agent to improved performance. This trend is also confirmed by the action-advice based baseline~\textit{BS: confidence\_based\_e-decay}, as this approach enabled the target agent to achieve superior performance compared to other teacher-student framework baselines. Moreover, the sharing of batches of experiences has resulted into unmatched performance improvements for a target agent, outperforming action-advice baselines.

From~\ac{pp} results it stands out that the quantity of shared information holds a stronger impact on a target agent compared to the filtering threshold used to select incoming experiences. Nevertheless, in an offline context, the threshold used for experience selection on the target side still contributes to facilitating positive transfer.

However, when transitioning from an offline to an online context and enabling the transfer across learning agents, relying on a fixed value to decide the tuples to be used by a target agent introduces a potential limitation as agents' confidence grow over time. Additionally, while tuning the transfer threshold can positively affect a target agent policy, this process is tedious, requiring numerous trials due to the high sensitivity of the outcome to this parameter, as observed in the experiments in~\ac{pp}. Consequently, a pillar requirement for \ac{efontl} is to enable the share of experience while removing the need for a hard-defined threshold.

When moving from offline to online experience sharing with~\ac{efontl}, we examine whether the counter-intuitive effect observed with the quantity of shared experiences remains valid.

Finally, given the promising outcomes observed during offline experience transfer  between agents with different underlying demand patterns in~\ac{rs-sumo}, we assess whether these findings are confirmed in an online experience sharing scenario.

\chapter{\acs{efontl} Design}
\label{cpt:efontl}\acresetall%
This chapter presents the main contribution of this thesis, \ac{efontl}, a novel online experience sharing framework that enables online transfer learning between agents in multi-agent systems with no fixed expert.

Section~\ref{sec:motivation_experience} motivates the design choices underlying \ac{efontl}. Section~\ref{sec:efontl_overview} presents the \ac{efontl} architecture. Section~\ref{ss:efontl-framework} introduces the necessary notation and presents the algorithm. 
Section~\ref{sec:trasfer_core_component} presents the transfer core engine, used to handle the knowledge transfer, by introducing both the teacher and the transfer content selection criteria. Section~\ref{sec:sardn_implementation_details} presents the uncertainty estimator model used alongside \ac{efontl} and, finally, this chapter ends with a summary in Section~\ref{sec:efontl_summary}.

\section{Motivation} \label{sec:motivation_experience}

Chapter~\ref{cpt:background} presented multiple \ac{tl} approaches differing by both the transferred object and the level of expertise possessed by the source of the transfer.

To summarise, when it is guaranteed that a dedicated expert is fully available to supervise novel agents, the most effective way to reduce the policy training time by injecting external knowledge into a novel target agent is to guide the target agent following the teacher-student framework with tailored advice, i.e., action or Q-values. The advice directly impacts the action decision process of the target agent.

On the other hand, when the provided advice is suboptimal, the target agent performance is capped by the degree of expertise of the source agent. In fact, the source's knowledge will eventually limit the capability of target agent to learn an effective policy. As a result, when incorporating suboptimal advice within the action decision process, the risk of negative transfer overweights the potential benefit of \ac{tl}.

To overcome the need of an optimal expert and enable \ac{tl} across imperfect agents, this thesis presents \ac{efontl}. 
\ac{efontl} objective is to identify a subset of gathered experiences to be transferred that supplements and enhances the target learning process rather than overwrites it.  
Experience consists of agent-environment interactions that are visited by a source agent during its exploration process. As a result, through~\ac{efontl}, target agent enhances its own local standard \ac{rl} policy with external curated experience.

\section{\acs{efontl} Notation and Definitions}
Prior to presenting in detail \ac{efontl}, this section introduces the essential notation and definitions on which~\ac{efontl} is based. These concepts simplify the framework understanding and facilitate comprehension of the remaining chapter.

\begin{itemize}
	\item $N$ $-$  number of \ac{rl}-based agents available during a simulation; 
	\item a set of \textit{Agents}~$A = \{A_1,\dots,A_N\}$;
	\item a set of \textit{Learning Processes} $LP = \{LP_1,\dots,LP_N\}$, which are mapped to agents using the following function $f_1 = {(A_i,LP_j)\in A \times LP \leftrightarrow i=j }$, such as an agent has one and only one~$LP$;
	
	\item a set of \textit{Uncertainty Estimators} $UE = \{UE_1,\dots,UE_N\}$, which are mapped to agents using the following function $f_2= {(A_i,UE_j)\in A \times UE \leftrightarrow i=j  }$, such as an agent has one and only one~$UE$;
	\item a set of \textit{Transfer Buffers} $TB =\{ TB_1,\dots,TB_N\}$, which are mapped to agents using the following function $f_3= {(A_i,TB_j)\in A \times TB \leftrightarrow i=j  }$, such as an agent has one and only one~$TB$;
	\item $B$ $-$ a scalar value \textit{Transfer Budget} that defines the exact number of experiences to be processed by a target agent during a single transfer step;
	\item \textit{TF} $-$ \textit{Transfer Frequency} defined as number of episodes that occur between two consecutive transfer steps;
	\item \textit{SS} $-$ \ac{ss} used to select source of transfer;
	\item \textit{TCS} $-$ \ac{tcs} used to filter relevant knowledge to be transferred.
\end{itemize}

The notation and definitions provided above lay the base of~\ac{efontl} framework. These elements serve as a solid foundation for the architecture presentation and algorithm description presented in the following sections.

\section{\acs{efontl}: Architecture Design }\label{sec:efontl_overview}

\ac{efontl} is a transfer learning framework for multi-agent systems that overcomes the need for an a priori determined experienced teacher by dynamically selecting a temporary expert at each transfer step based on the current performance of the agents, according to commonly defined metrics. \ac{efontl} aims to improve the performance of individual agents through experience sharing.  The selected agent is used as a source of transfer and some of its collected experience is made available to others. Subsequently, a target agent can filter and sample a batch of experiences to be integrated into its learning process and finally update its policy. The transferred batch contains agent-environment interactions sampled by the selected teacher and labelled with source's epistemic uncertainty~$(u^t)$:  $(s^t, a^t,r^t,s^{t+1},u^t)$. \ac{efontl} is independent of the underlying \ac{rl} algorithm used and thus, it can be exploited on a range of \ac{rl} methods, both tabular and neural network-based ones.

\begin{figure}[htb!]
	\centering
	\includegraphics[width=.9\columnwidth]{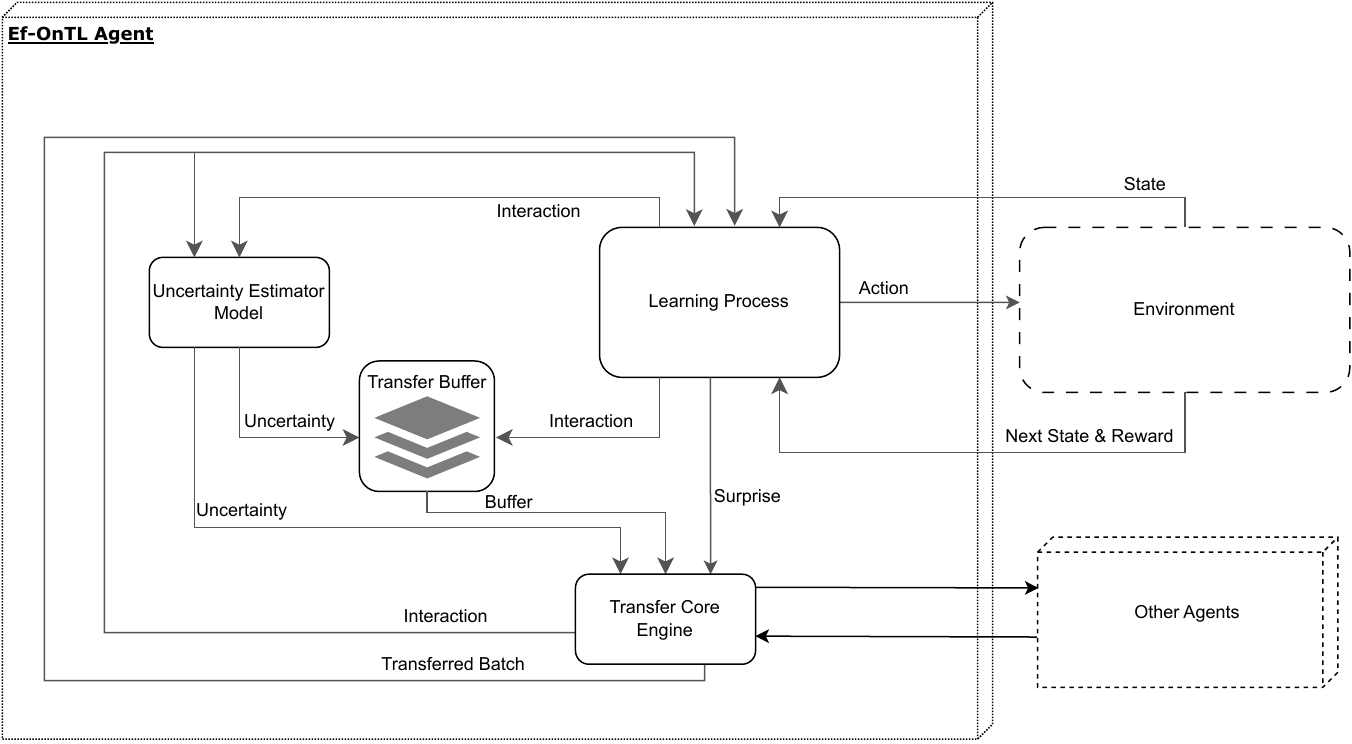}
	\caption{\ac{efontl} agent components diagram.}
	\label{fig:efontl_component_diagram}
\end{figure}

Figure~\ref{fig:efontl_component_diagram} depicts \ac{efontl} system architecture that can be divided into three main components:

\begin{itemize}
	\item \ac{rl} Agent $-$ It is the underlying \ac{rl} algorithm used by the agent to learn a policy to address a task. Such model enables the agent to sample an action from a certain state. After taking the step, the agent forwards the observation, \ac{rl} agent-environment tuple, to the learning model underneath that will be eventually updated.

	\item Transfer Core Engine $-$ The Transfer Core  is used to establish a connection with other agents and the main role is to filter the incoming experience that is tailored to the current policy of the agent. Transfer Core is detailed in Section~\ref{sec:trasfer_core_component}.

	\item Uncertainty Estimator Model $-$ The Uncertainty Estimator is used by the agent to approximate epistemic uncertainty. \ac{sarnd} is covered in Section~\ref{sec:sardn_implementation_details}.

\end{itemize}


\vcchange{
The data processed by an \ac{efontl} agent come from two distinct sources: an environment accessed online in a traditional \ac{rl} manner, and other \ac{efontl} agents. 
Beyond the traditional update of the learning process, the interactions sampled from the environment are fed into the uncertainty estimator model, which estimates the epistemic uncertainty related to each tuple. Subsequently, each tuple, along with its estimated uncertainty, is stored in an internal buffer, named \textit{Transfer Buffer}. This process is repeated iteratively until the transfer time is reached.}

\vcchange{
When transfer is initiated, the transfer core engine mutually exchanges the internal transfer buffer with other agents. Each interaction contained within the transfer buffers is then fed into the learning process, to estimate the surprise value, and the uncertainty estimator model, to estimate the uncertainty. Based on these two metrics, a source agent is selected and a subset of tuples is filtered to shape the transferred batch. Finally, the learning process is updated based on this transferred batch.
}

\section{\acs{efontl} Framework}\label{ss:efontl-framework}

This section delves into the online experience sharing workflow followed by~\ac{efontl} agents. The agent exploration and learning process is independent, except during the transfer step. When the transfer begins, a common agent is selected  as the source of transfer, and the reaming agents filter a subset of the incoming experiences to update their learning process based on the selected tuples.

Algorithm \ref{alg:efontl} introduces high level procedure followed by agents for sharing experience one to another throughout their simultaneous exploration processes.

\begin{algorithm}[h!]
	\small
	\caption{\acl{efontl}}
	\label{alg:efontl}
	\begin{algorithmic}[1]
		\STATE Given: \textit{A, LP, UE, TB, B, TF, SS, TCS, $f_1$, $f_2$, $f_3$}
		\FOR{\textit{ep} in \textit{Episodes}} 
		\FORALL[follow normal policy]{$A_i \in A$}
		\STATE get state $s_i^t$ for $A_i$ at time $t$
		\STATE sample an action $a_i^t$ based on $LP_i$
		\STATE perform a step and observe $o_i^t = (s_i^t,a_i^t,r_i^t,s_i^{t+1})$
		\STATE  estimate uncertainty  $u_i^t$ = $UE_i(o_i^t)$ 
		\STATE push ($o_i^t, u_i^t$) to $TB_i$ \algorithmiccomment{FIFO queue}
		\STATE optimise $UE_i$ on $o_i^t$
		\STATE optimise $LP_i$ 
		\ENDFOR
		\IF[start transfer-step]{\textit{ep} \% $TF$ is $0$} 
		\STATE select source agent $A_s$ by $SS$ 
		\FORALL[transfer from $A_s$ to $A_t$]{$A_t \in (A \setminus A_s)$} 
		\STATE apply \acs{tcs} over $TB_s$ and sample $B$ tuples
		\STATE optimise $LP_t$ with the sampled tuples
		\ENDFOR
		\ENDIF
		\ENDFOR
	\end{algorithmic}
\end{algorithm}

First, line~$1$ defines the \ac{efontl} parameters.
Then, agent takes a standard \ac{rl} step at lines~$3-6$, i.e, agents retrieve observation from an environment, sample an action based on their learning process and finally take a step within the environment.

 Afterward, the related $UE_i$ estimates $i$-th agent's epistemic uncertainty $u_i^t$ over the observed agent-environment tuple~$(s_i^t,a_i^t,r_i^t,s_i^{t+1})$, then $UE_i$ model is updated based on new sampled evidence and the uncertainty-labelled tuple is published to the associated transfer buffer~$TB_i$ as shown in lines~$7-9$. 
 When $TB_i$ is at full capacity, new labelled interactions replace the oldest tuples following a FIFO scheme. Learning process $LP_i$ is then updated based on the underlying \ac{rl} algorithm used, lines~$10-12$. 
 
 Lines~$14-20$ show a transfer step. Firstly, at line~$15$, source is selected among the agents w.r.t.~\acl{ss}~$SS$. Secondly, remaining agents, i.e., targets, apply a filtering function~$TCS$ over source transfer buffer and then sample a batch composed by a fixed number~$B$ of tuples, line~$17$. Finally, at line~$18$, each target agent updates its learning process based on the obtained batch.

\begin{figure}[htb!]
	\centering
	\includegraphics[width=.9\columnwidth]{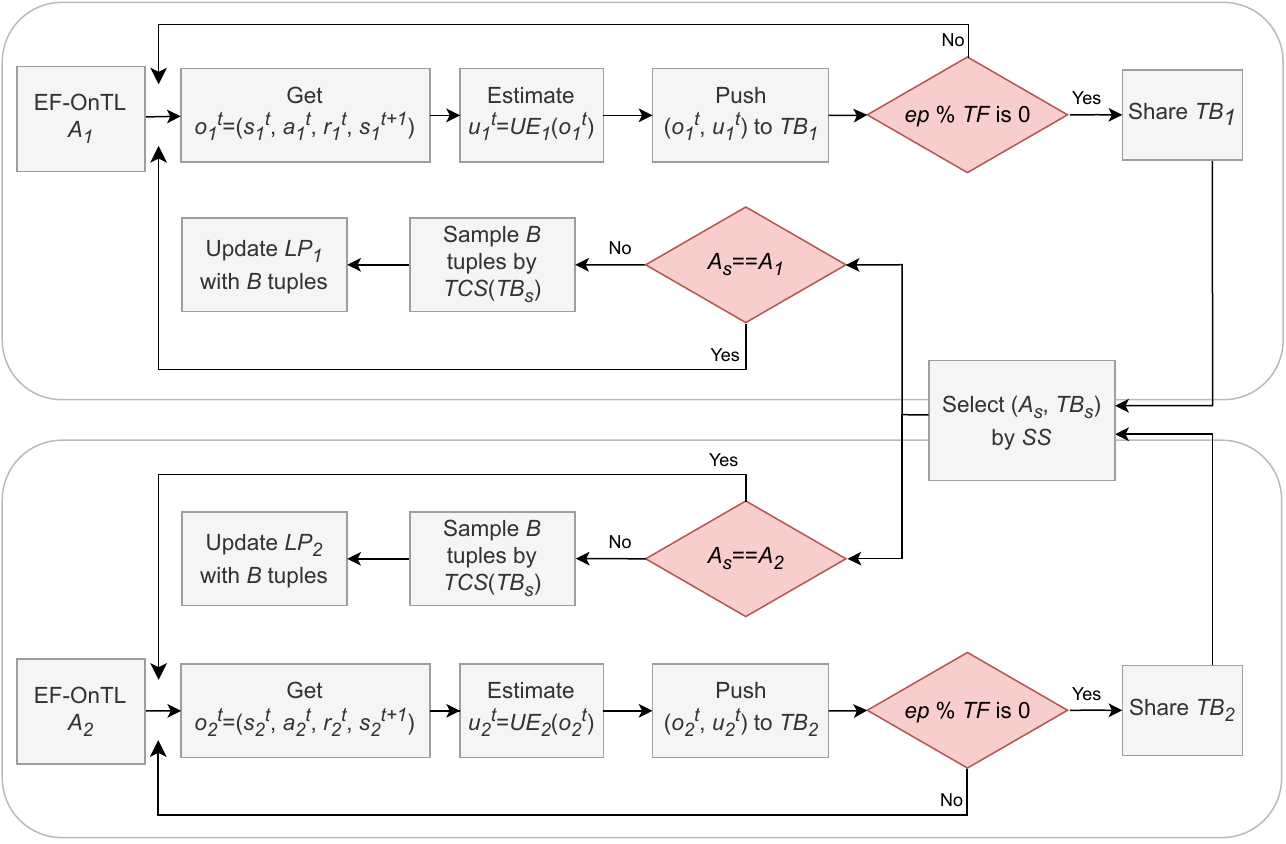}
	\caption{\ac{efontl} simplified workflow in a two agents scenario.}
	\label{fig:efontl_workflow}
\end{figure}

While the above provides a full description of the algorithm, for further clarity,~Figure~\ref{fig:efontl_workflow} summarises \ac{efontl} workflow, at a single~\ac{tl} time-step~$t$, for the simplest scenario with only 2 agents.

In detail, the source agent~$A_s$ and the related source transfer buffer~$TB_s$ are globally selected, as the $SS$ metrics, further presented in Section~\ref{ss:SourceSelection}, do not consider the target agent to identify the source of transfer. Consequently, when implementing multiple~\ac{efontl} agents, the source selection process can be either centralised or decentralised, based on specific needs. However, both implementations are equivalent given the $SS$ metrics. When \ac{efontl} is applied on more than two agents, all the agents share their transfer buffer one to another.

\begin{figure}[htb!]
	\centering
	\includegraphics[width=1\columnwidth]{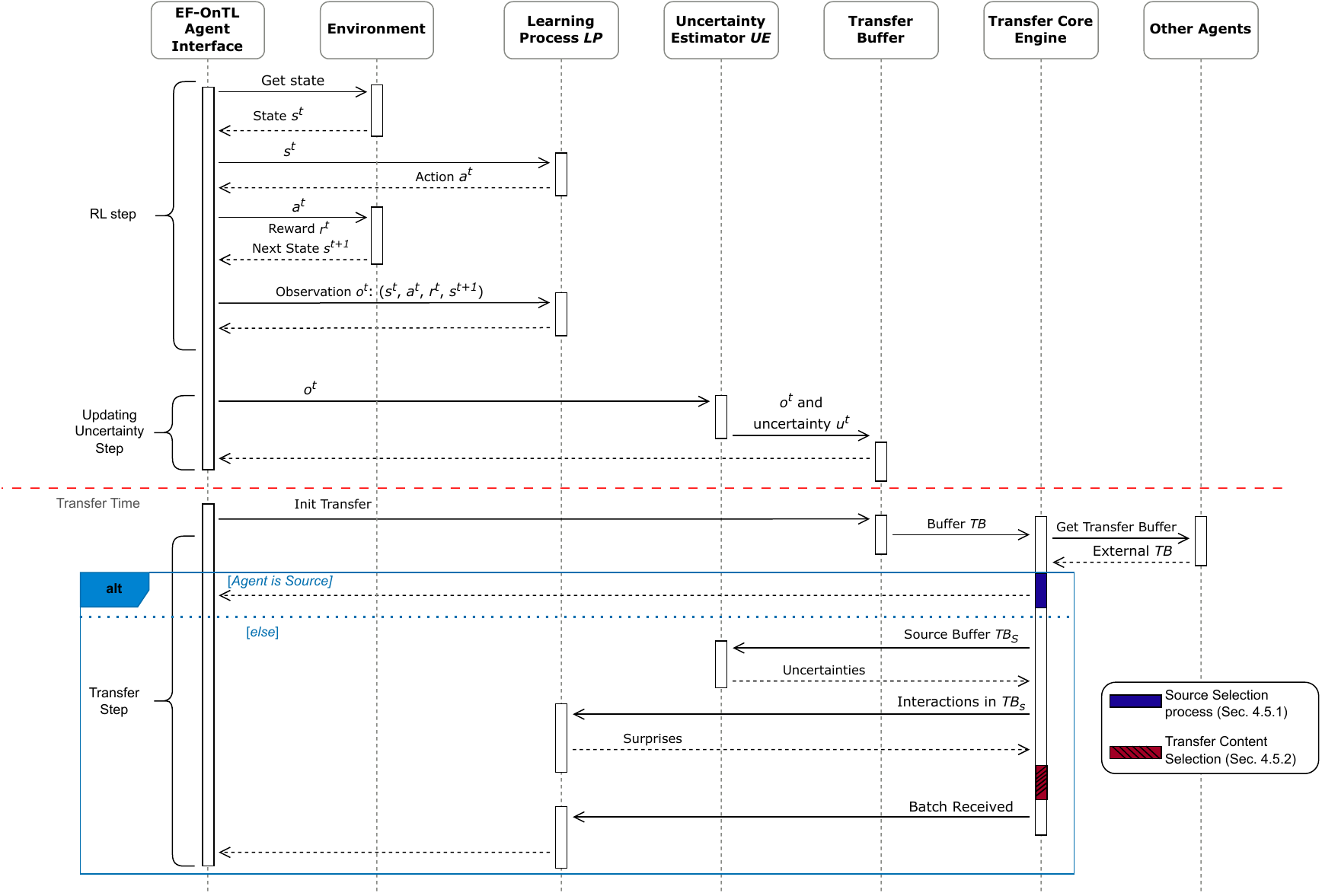}
	\caption{\ac{efontl} agent sequence diagram.}
	\label{fig:efontl_sequence_diagram}
\end{figure}

To further zoom into \ac{efontl} process, this section proceeds by showing the sequence of operations within an \ac{efontl} agent, providing insights into how an \ac{efontl} agent operates internally.
The internal agent perspective aims to provide a fine-grained analysis on how each sub-modules within~\ac{efontl} is used and interacts one to the other throughout the learning and transferring stage. To enable a detailed internal view, the diagram in Figure~\ref{fig:efontl_sequence_diagram} divides the agent into fine-grained sub-modules.

There are $3$ main sequence blocks that can be identified:

\begin{enumerate}
	\item \textit{RL step} $-$ describes the \ac{rl} agent-environment interaction, lines $4$ to $6$ in Algorithm~\ref{alg:efontl}. Agent uses its learning process~\textit{LP} to sample an action based on the observed state provided by the environment. After taking that action, agent observes the reward returned by the environment along with the next state. Finally, the full observed tuple~($o^t$), that will eventually be used to update the underlying learning policy, is passed to the Learning Process.
	
	\item \textit{Updating Uncertainty step} $-$ $o^t$ is forwarded to the \textit{Uncertainty Estimator Model}, which approximates the epistemic uncertainty~($u^t$) and updates the underlying model as shown in line $9$ of Algorithm~\ref{alg:efontl}. Finally, ($o^t, u^t$) is pushed to the associated \textit{Transfer Buffer}.

	\item \textit{Transfer Step} $-$ When the time to transfer comes, as per line $14$ in Algorithm~\ref{alg:efontl}, an agent initialises the transfer by sending its transfer buffer to the \textit{Transfer Core Engine}. The \textit{Transfer Core Engine} collects the transfer buffers from other agents and decides whose the suitable agent that should act as teacher, as shown at line~$15$ of Algorithm~\ref{alg:efontl}. 
	
	When an agent is chosen as source of transfer, the knowledge transfer process terminates for that agent, and it is not influenced by external knowledge. On the other hand, if the selected teacher is an external agent, the transfer begins. Both the \textit{Uncertainty Estimator Model} and the \textit{Learning Process} are used to estimate the uncertainties and surprises for each tuple within the \textit{Source Transfer Buffer}. Afterward, the \textit{Transfer Core Engine} filters and samples a certain number of tuples that are used to update the \textit{Learning Process}. Finally, the transfer step terminates and the control is returned to the agent.
\end{enumerate}

During a transfer step there are the two selection processes, source selection and transfer content selection, that are not captured in detail within this diagram. Next section focuses on the \textit{Transfer Core Engine} presenting these two filtering functions and the criteria used to select source of transfer and relevant knowledge.

\section{Transfer Core Engine}\label{sec:trasfer_core_component}

\begin{figure}[htb!]
	\centering
	\includegraphics[width=.8\columnwidth]{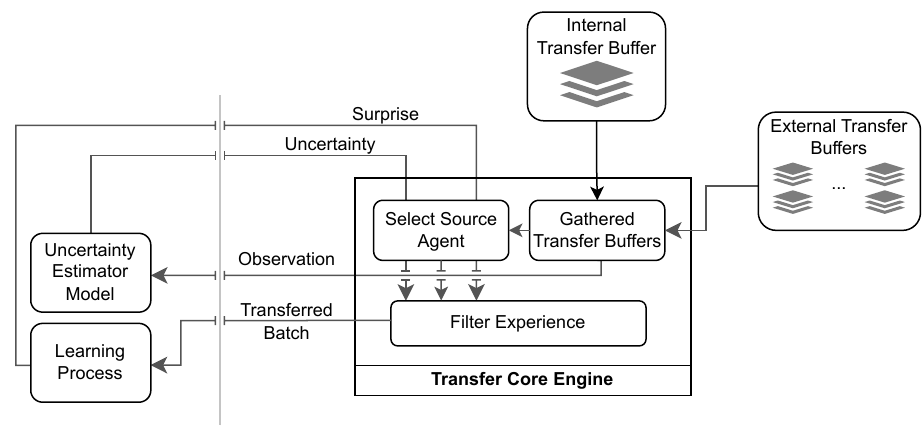}
	\caption{Detail of \ac{efontl} Transfer Core diagram.}
	\label{fig:transfer_core_detail_component}
\end{figure}

The \textit{Transfer Core Engine} is the internal part of \ac{efontl} agent that manages the transfer step. Its role is to interface with other agents to exchange the transfer buffers and to select relevant knowledge to be used to update the \ac{rl} learning process. The details of its architecture are shown in Figure~\ref{fig:transfer_core_detail_component}. 

The \textit{Transfer Core Engine} has two objectives: 1) identify the $A_s$, the agent used as source of transfer; 2) filter the relevant incoming knowledge to update the \textit{Learning Process}. The remainder of this section focuses on these two modules: Section~\ref{ss:SourceSelection} presents the metrics used to select $A_s$ alongside the related transfer buffer~$TB_s$. Then Section~\ref{ss:TransferMethods} presents~\ac{tcs}, the selection criteria applied by the target agent to select relevant knowledge.

\subsection{Source Selection}\label{ss:SourceSelection}

In \ac{tl} approaches where an expert is readily available for unidirectional knowledge transfer, as the one described in Chapter~\ref{cpt:preliminary_studies}, the source selection task is not required.  However, in \ac{efontl}, given the continuous online transfer across multiple agents, source selection becomes crucial to achieve positive transfer and to dynamically assign roles at the beginning of each transfer step.

One of the key challenges in \ac{efontl} is to identify the most suitable agent among multiple candidates with similar expertise, as shown at line~$13$ of Algorithm~\ref{alg:efontl}, which allows other agents to benefit from the sharing of the source's knowledge.

To identify the transfer source from a set of candidates, this work assumes that agents can communicate one to the other with no limitations and at a fixed cost; in reality the communication cost and range will be limited confining the communication to a subset of agents.

As discussed in Chapter~\ref{cpt:background} for \ac{a2a}~\ac{tl}, the source selection decision process can rely on confidence or uncertainty metrics defined on target and sender, performance based metrics, or can be based on completion of an episode. The ultimate choice depends upon the transfer objective.

The majority of experience-based \ac{tl} work does not consider the source selection problem, as demonstrations are assumed to be coming from agents that have already finished their training. The closest work related to \ac{efontl} are \ac{ocmas}~\cite{ilhan2019teaching} and Wang and Taylor~\cite{wang2017improving, wang2018interactive}. 
\ac{ocmas} is an online action-advice based \ac{tl} where the problem of source selection is replaced by majority voting as the final advice is the result of majority voting over a set of incoming action-advice. 
Wang and Taylor~\cite{wang2017improving, wang2018interactive} use uncertainty to select the external entity that should provide action-advice.  
In \ac{efontl}, agents use uncertainty and cumulative reward to select the source of transfer. Instead of providing a single action-advice to a target agent selected by majority voting or provided by an external entity, \ac{efontl} provides to a target agent a subset of experiences to overcome the knowledge gap between the target and the source policy. This thesis studies two different methods to select the transfer source within the Transfer Core Engine:

\begin{itemize}

	\item \ac{u} $-$ it selects source of transfer as the agent with lowest average uncertainty~$\overline{u_i}$ on the tuples stored within its transfer buffer~$TB_i$.  Formally, that means that the source agent~$A_s$ is selected accordingly to Eq.~\ref{eq:source_selection_uncertainty}.

	\begin{equation}
		\begin{split}
			A_s &\leftarrow min_{i=0}^{N}(\overline{u_i})\\
			\overline{u_i}&=\frac{\sum_{t=0}^{|TB_i|} u_i^t \in TB_i}{|TB_i|}
		\end{split}
		\label{eq:source_selection_uncertainty}
	\end{equation}

	\item \ac{bp} $-$ it relies on performance achieved by agents over a pre-defined window of most recent episodes and it is inspired by Taylor et al.~\cite{taylor2019parallel}. As performance measure, \ac{efontl} uses average cumulated reward over episodes' finite-horizon undiscounted return~($\overline{R_i}$).	 $A_s$ is selected out of average sum of rewards returned by the environment over episodes from initial to goal state with a finite number of steps. Eq.~\ref{eq:source_selection_bestPerformance} reports the formula used for the source selection, where $E$ is the number of evaluated episodes and $\tau^e_i$ represents the length of the $e$\textit{-th} episode for the $i$\textit{-th} agent.

\begin{equation}
\begin{split}
	A_s &\leftarrow max_{i=0}^{N}(\overline{R_i})\\
	\overline{R_i}&=\frac{\sum_{e=0}^{E} \sum_{t=0}^{\tau^e_i} r^{e,t}_i}{E} 
\end{split}
\label{eq:source_selection_bestPerformance}
\end{equation}

\end{itemize}

The two \ac{ss} criteria, \ac{u} and \ac{bp}, address the challenge of identifying a suitable agent to act as the source of transfer from a set of candidates, enabling dynamic roles for a transfer step. While these metrics are generally used to establish whether an agent is in need of advice in specific circumstances, \ac{efontl} uses these metrics for selecting the source of transfer, which allows the sharing of a batch of experiences aiming to overcome the gap within a target policy.

In Chapter~\ref{cpt:evaluation}, both techniques described by  Eq.~\ref{eq:source_selection_uncertainty} and Eq.~\ref{eq:source_selection_bestPerformance} are evaluated in a mutually exclusive manner. This evaluation aims to identify the performance impact of the two techniques to establish whether one prevails over the other in certain conditions or scenarios.

\subsection{Transfer Content Selection}\label{ss:TransferMethods}


Once the appropriated source of transfer is identified, the target agent has to accurately select the specific knowledge  relevant to its own learning process, as shown in line~$17$ of Algorithm~\ref{alg:efontl}. Therefore, this section introduces \ac{tcs} to prioritise certain tuples based on their score over a set of measures.

To simplify the readability, this section discusses the simplest case with two agents, which, at each transfer step, are referred to as source agent~$A_s$ and target agent~$A_t$.
Experience worth to be transferred is identified through two criteria, \textit{expected surprise} as in~\cite{gerstgrasser2022selectively} and \textit{uncertainty} as in~\cite{wang2017improving, wang2018interactive}.

Gerstgrasser et al.~\cite{gerstgrasser2022selectively} use Temporal Difference error~(TD-error) to select agent-environment tuples that surprised a certain agent, and consequently, these tuples are forwarded to all other agents by injecting the tuples into their replay buffer. However, this approach could potentially limit the capability of other agents to store and learn from more specific knowledge, as certain tuples might not have the same impact on multiple different policies. Wang and Taylor~\cite{wang2017improving,wang2018interactive} use the uncertainty to provide action as advice from a centralised entity and to decide whether to optimise the central entity with further demonstration collected online. 

In~\ac{efontl}, \textit{expected surprise}~\cite{white2014surprise}, is defined on the target \textit{Learning Process} and is approximated through TD-error. 
Despite expected surprise is not used as much as uncertainty or confidence to filter relevant knowledge in other~\ac{tl} work, we propose it as transfer criteria in this thesis based on its proven usefulness as selection criteria for prioritising experience selection in prioritised experience replay~\cite{schaul2015prioritized}.

In~\ac{efontl}, \textit{uncertainty} is estimated by an uncertainty estimator model~$UE$ with a standardised architecture. Both source and target agent have access to their own estimator, respectively $UE_s$ and $UE_t$ that are updated based on their local experiences. Consequently, their estimated uncertainties can be compared to select relevant tuples to be transferred. In detail, given ($o_s^i, u_s^i$), where $o_s^i$ is an agent-environment tuple visited by $A_s$ at time $i$ and $u_s^i$ is its associated epistemic uncertainty,  then $A_t$ estimates current uncertainty~$u_t^{o_s^i}$ over the interaction~$o_s^i$ sampled by $A_s$. Hence,  $u_t^{o_s^i} = UE_t(o_s^i)$. The discrepancy between these two estimations, delta confidence~($\Delta-$\textit{conf}), can be defined as $\Delta-$\textit{conf}$=u_t^{o_s^i} - u_s^i$. Note that $u_t^{o_s^i}$ is an epistemic estimation at the time of transfer and changes over~time.

Based on these two criteria, i.e., surprise and uncertainty, $A_t$ receives a personalised batch of experiences that aims to fill shortcomings in its policy. We define multiple filtering functions for incoming knowledge as follow:

\begin{enumerate}
	\item{\acs{rdc} $-$ \acl{rdc}.} By choosing this technique, an agent randomly samples $B$ tuples with an associated $\Delta-$\textit{conf} higher than the median value computed across all tuples.
	\item{\acs{hdc} $-$ \acl{hdc}}. Agent sorts in decreasing order the incoming tuples by the~$\Delta$\textit{-conf} and selects the top $B$ entries.
	\item{\acs{lec} $-$ \acl{lec}}. While previous filters are defined only over uncertainty, this one also considers expected surprise. To balance the different scales, uncertainty and surprise values are normalised within a $[0,1]$ interval and then weighted equally. Finally, target agent sorts the incoming tuples by the computed value and then selects the top $B$ tuples with the highest values.
\end{enumerate}

The three \ac{tcs}, \acs{rdc}, \acs{hdc} and \acs{lec}, enable a target agent to filter the incoming knowledge facilitating dynamic transfer content selection. As a result, each target agent can select a subset of the incoming experiences that is expected to improve its current learning process. 
In Chapter~\ref{cpt:evaluation}, all the three \ac{tcs} criteria are evaluated and compared in a mutually exclusive manner. The comparison aims to identify the performance impact of
filtering the incoming experience based on different selection criteria to establish whether one prevails over the other in certain conditions or scenarios.

\section{Uncertainty Estimator Model}\label{sec:sardn_implementation_details}

This section presents the uncertainty estimator model used within~\ac{efontl} to support the source selection decision process and the filtering of relevant knowledge to be integrated within the learning model of the agent.

As discussed  in Section~\ref{sec:back_uncertainty}, generally, in \ac{rl}, the epistemic uncertainty of an agent is approximated from the observed state returned by the environment. In simple scenarios a state-visit counter can be used to approximate the uncertainty. On the other hand, when it is impossible to keep track of all the different visited states, neural network-based models, as \ac{rnd}~\cite{burda2018exploration}, can be used as approximators of state-visit counter.  \ac{rnd} uses two neural networks, referred as target and predictor, to approximate the epistemic uncertainty through \ac{mse} of the networks' output. 
Over time, the predictor is optimised to predict the target output and the uncertainty is expected to decrease accordingly.


\begin{figure}[htb!]
	\centering
	\includegraphics[width=.85\columnwidth]{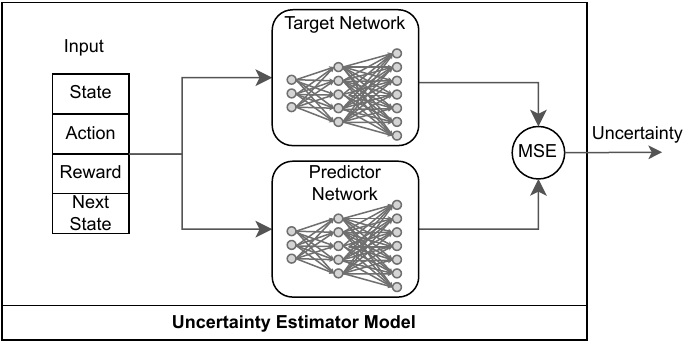}
	\caption{\acs{sarnd} architecture.}
	\label{fig:sarnd_detail_component}
\end{figure}

When used in an online \ac{tl} scenario, \ac{rnd} is expected to lack crucial informations that are necessary to estimate the epistemic uncertainty of an agent, i.e., action taken and the updated state. 
To address this, \ac{sarnd}, shown in Figure~\ref{fig:sarnd_detail_component}, an estimator model based on \ac{rnd} and expected to overcome the limitations of the base model is introduced. 
\ac{sarnd} extends \ac{rnd} by estimating uncertainty from a \ac{rl} tuple ($s^t_i, a^t_i, r^t_i, s^{t+1}_i$), composed by state, action, reward and next state.

By taking into consideration the additional features, \ac{sarnd} is expected to provide a more accurate estimation throughout the learning phase when compared to \ac{rnd}. 
The extra inputs of~\ac{sarnd} are expected to provide a more accurate estimation by enabling the agent to discriminate situations where the goal is closer but not yet explored because the action that lead to the accomplishment of the goal has never been explored.

Section~\ref{sec:sarnd_eval}, has demonstrated \ac{sarnd} superiority in estimating the agent epistemic uncertainty in an online scenario when compared against \ac{rnd}.

\section{\acs{efontl} Design Summary}\label{sec:efontl_summary}

In most \acf{tl} frameworks, knowledge is typically transferred from an optimal fixed expert to learning agents. 
However, if the source of advice is suboptimal, the capability of a target agent to learn an optimal policy might be constrained by the suboptimal policy. To overcome this limitation and enable \ac{tl} across multiple learning agents, in this chapter we presented~\ac{efontl}. \ac{efontl} is an online experience sharing framework that facilitates \ac{tl} among multiple learning agents by transferring a buffer of agent-environment interactions. 

\ac{efontl} expands the architecture of a traditional \ac{rl} agent by incorporating an uncertainty estimator model~($UE$), a transfer buffer~($TB$),  and a transfer core engine for each agent. The $UE$ is used to estimate the epistemic confidence of an agent. Subsequently, the observed agent-environment interaction labelled with uncertainty is stored in the transfer buffer. Finally, the transfer core engine handles the communication with other agents and finalises the transfer process.

To enable online experience sharing, the transfer core engine in \ac{efontl} must address two challenges: 1) identifying a suitable temporary expert to act as the source of transfer, and 2) selecting the relevant experiences that contribute to the target policy.

For source selection, \ac{efontl} defines two \acf{ss}: \acf{u} and \acf{bp}.  \ac{u} selects the source of transfer as the agent with the lowest average uncertainty within the transfer buffer, while \ac{bp} selects the source agent based on the average cumulated reward over episodes' finite horizon undiscounted return.

To filter experiences relevant to a specific target agent, \ac{efontl} defines three~\acf{tcs} based on two criteria: expected surprise and delta confidence~($\Delta$\textit{-conf}). 
Expected surprise is approximated by the Temporal Difference error, and it is used to select tuples that are expected to surprise a target agent. On the other hand, $\Delta$\textit{-conf} is defined as the difference between the uncertainty of the target agent and the uncertainty of the source agent. Consequently, the three~\ac{tcs} used in \ac{efontl} consist of \acf{rdc}, \acf{hdc}, and \acf{lec}.

The criteria used by~\ac{efontl} to select the transfer source and relevant knowledge are not new within the \ac{tl} community. For instance, uncertainty has been used to identify the source of transfer~\cite{wang2017improving, wang2018interactive} and to identify the situation where an agent is in need of advice~\cite{ilhan2019teaching}, while the idea of using performance based metrics are inspired by Taylor et al.~\cite{taylor2019parallel}. Similarly, the surprise concept has been used by Gerstgrasser et al.~\cite{gerstgrasser2022selectively}. 
However, to the best of our knowledge,  \ac{efontl} is innovative in applying these criteria to address the combined challenge of dynamically selecting the source of transfer and dynamically determining the content transferred to each individual target agent. Moreover, uncertainty has been primarily utilised to decide whether to provide an one-time advice.

To promote positive \ac{tl} outcome, \acf{sarnd} is introduced and used as $UE$ in this thesis. \ac{sarnd} is a model that enables an agent to estimate epistemic uncertainty from an agent-environment interaction. Based on the uncertainty and on other commonly defined metrics, i.e., expected surprise and cumulative reward, agents can dynamically select a source of transfer and filter the experiences from an incoming transfer buffer. This selection process aims to identify a batch of experiences that are expected to improve the target's policy.

\chapter{Implementation}
\label{cpt:implementation_details}\acresetall%

This chapter discusses the hardware and software technologies used for conducting the research presented within this thesis. Section~\ref{sec:exp_technologies} introduces the libraries and the hardware used to run the experiments, collect the data and process the results. Then, Section~\ref{sec:nn_architecture} covers the design of the \ac{rl} algorithms used for the evaluation. Finally, this chapter ends with Section~\ref{sec:imp_efontl} that discusses the implementation details of the main contribution of this thesis, \ac{efontl}.

\section{Technology for Experiments}\label{sec:exp_technologies}
This section presents the necessary libraries used to carry out the experiments presented in this work. 

All algorithms and experiments are developed in Python~3. The neural networks are implemented using PyTorch version~1.13.1+cu117 to enable the models run on a GPU with CUDA.
Performance and results are stored as CSV files and visualised in real time through TensorBoard V.~2.12.0. Finally, data are processed and analysed using Pandas V.~1.5.3, NumPy V.~1.23.0 and graphically displayed through Matplotlib V.~3.4.0.

\subsection{Hardware}
The results presented in this thesis have been collected on two distinct machines. The first machine used to carry out the experiments is a consumer-oriented laptop, specifically a Dell XPS model. Its hardware configuration details are reported in Table~\ref{tab:xps_specs}.
\begin{table}[h!]
	\renewcommand*{\arraystretch}{1.1}
	\caption{\label{tab:xps_specs} Laptop hardware specs used for experiments.}
	\centering
	\begin{threeparttable}
		\begin{tabular}{|c|c|}
			\hline
			{\textbf{Parameter}}&\textbf{Value}\\\hline
			CPU & Intel i7-9750H @ 2.60GHz\\\hline
			GPU & NVIDIA GTX 1650 Mobile\\\hline
			NVIDIA-SMI Driver & 470.182.03\\\hline
			CUDA Version & 11.4\\\hline
			RAM & $2\times$8 GiB DDR4 @ 2667 MHz\\\hline
			OS & Ubuntu 20.04.5 LTS\\\hline
		\end{tabular}
	\end{threeparttable}
\end{table}

The second machine used is a custom rack setup with multiple GPUs. The hardware and software details are provided in Table~\ref{tab:hopper_specs}.
\begin{table}[h!]
	\renewcommand*{\arraystretch}{1.1}
	\caption{\label{tab:hopper_specs} Server hardware specs used for experiments.}
	\centering
	\begin{threeparttable}
		\begin{tabular}{|c|c|}
			\hline
			{\textbf{Parameter}}&\textbf{Value}\\\hline
			CPU & Intel Xeon(R) Gold 5220R @ 2.20GHz\\\hline
			GPU & $4\times$ RTX 3070\\\hline
			NVIDIA-SMI Driver & 530.30.02 \\\hline
			CUDA Version & 12.1.1\\\hline
			RAM & 1535 GiB\\\hline
			OS & Ubuntu 20.04.6 LTS\\\hline
		\end{tabular}
	\end{threeparttable}
\end{table}

Finally, Table~\ref{tab:sims_run_device} specifies which device has been used for each step of the experimentation pipeline on each benchmark environment. First column, \textit{Benchmark}, indicates the environment, second column, \textit{Debug}, indicates the hardware used to implement and perform preliminary experiments and, last column, \textit{train \& test}, reports the machine used to collect the presented results.  
The machine used can vary between the custom rack~(\textit{server}),  and the consumer-oriented laptop~(\textit{XPS}).

\begin{table}[h!]
	\renewcommand*{\arraystretch}{1.1}
	\caption{\label{tab:sims_run_device} Details of simulations.}
	\centering
	\begin{threeparttable}
		\begin{tabular}{|c|c|c|}
			\hline
			{\textbf{Benchmark}}& \textbf{Debug} & \textbf{Train \& Test}\\
			\hline
			Cart-Pole& XPS & server\\\hline
			\acs{pp} & XPS & server\\\hline
			\acs{hfo} & XPS & XPS \\\hline
			\acs{rs-sumo} & XPS & server\\
			
			 \hline
		\end{tabular}
	\end{threeparttable}
\end{table}

The neural networks implemented are relatively simple, further details are provided in Section~\ref{sec:nn_architecture}, the computation cost mainly results from sampling the agent-environment transitions and by back-propagating the losses on the multiple networks. As such, to replicate the experiments presented within this thesis, GPU is not mandatory as the true bottleneck are the CPU capabilities, although, having a configuration similar to the server mentioned above helps to parallelise the computation and lower the overall time needed to collect the results.
 For instance, to complete a training and testing phase in the \ac{pp} environment takes around 7 hours on the XPS laptop, which is more than doubled on the server.

\section{Deep Reinforcement Learning Algorithms}\label{sec:nn_architecture}
This section outlines the architecture design for the deep~\ac{rl} algorithms used to learn the agent's policy.
For the \ac{efontl} experiments in this thesis, there are 2 \ac{rl} algorithms used: dueling~\ac{dqn}, discussed in Section~\ref{sec:ddqn}, and \ac{paddpg}, discussed in Section~\ref{sec:paddpg}. Additionally, for the \ac{pp} benchmark environment we defined an encoder to process the multichannel state. The encoder is presented as stand-alone as it has been used for both learning the agent's policy and in the uncertainty estimator model. The implementation of the encoder is detailed in Section~\ref{sec:pp_encoder}.

\subsection{Dueling \acl{dqn}}\label{sec:ddqn}

Dueling~\acl{dqn} is the model chosen to replace \ac{ppo} model, due to the on-policy limitation of \ac{ppo}, and to enable the sharing of experience. This model has been used in tasks with a discrete action space, i.e., Cart-Pole, \acf{pp} and \acf{rs-sumo}. Dueling~\ac{dqn}, depicted in Figure~\ref{fig:dqn_dueling_dqn_architecture}, decouples the learning of a state-value function and the advantage of selecting a certain action. As a result, dueling~\ac{dqn} improves the estimation of the Q-values when compared to standard~\ac{dqn}.

\begin{table}[h!]
	\renewcommand*{\arraystretch}{1.1}
	\caption{\label{tab:dueling-dqn_setup} Dueling \ac{dqn} parameters setup.}
	\centering
	\begin{threeparttable}
		\begin{tabular}{|c|c|c|c|}
			\hline
			{\textbf{Parameter}}& \textbf{Cart-Pole} & \textbf{\acs{pp}} & \textbf{\ac{rs-sumo}}\\
			\hline
			Input Layer & Linear(4, 128) & Linear(135, 256) &Linear(26, 256) \\\hline
			\multirow{2}{*}{Hidden Layer(s)} & \multirow{2}{*}{Linear(128, 64)} &\multirow{2}{*}{n.a.} & Linear(256, 128)\\
			&&& Linear(128, 64)\\\hline
			
			Advantage Layer & Linear(64, 2) & Linear(256, 5)&  Linear(64, 5) \\\hline
			Value Layer & Linear(64, 1) &  Linear(256, 1) & Linear(64, 1)\\\hline
			Activation & ReLU & ReLU & ReLU\\
			
			\hline			
			Optimiser & Adam & Adam & Adam\\\hline
			Learning Rate & 1e-4&1e-5&1e-4\\\hline
			Betas &  (.9, .999)  & (.9, .999) & (.9, .999)\\\hline
			Gamma & .999& .999& .999\\\hline
			Mini-batch Size & 32& 32&32\\\hline
			Policy Update Step & 1,000& 10,000&2,500\\
			\hline
		\end{tabular}
	\end{threeparttable}
\end{table}

Table~\ref{tab:dueling-dqn_setup} reports the setup used in Cart-Pole, \ac{pp} and \ac{rs-sumo}. Betas is a decaying parameter used within the Adam optimiser and the Policy Update Step defines the quantity of mini-batches used to update the online network before copying the weights of the online network into the target network.

Dueling~\ac{dqn} is designed for environments with discrete action spaces and is not suitable to handle continuous action space. Therefore, to enable the learning of a policy in environment with continuous control space, i.e., \ac{hfo}, in this thesis we use an off-policy based method described in the following section.

\subsection{\acl{paddpg}}\label{sec:paddpg}
\acf{paddpg}~\cite{hausknecht2015deep} is an extension of \ac{ddpg} and is an off-policy based \ac{rl} algorithm. \ac{paddpg} enables an agent to learn a policy in an environment with parametrised action space, i.e. \acf{hfo}, where an agent selects a discrete action along with the parameters required to parametrise the chosen action. 

The \ac{paddpg} architecture shown in Figure~\ref{fig:paddpg_architecture} consists of an actor with a 2-stream output to select both discrete action and parameters, and a critic network to estimate the $Q$-value for a state and an action.

\begin{table}[h!]
	\small
	\renewcommand*{\arraystretch}{1.1}
	\caption{\label{tab:paddpg_setup} \ac{paddpg} parameters for~\ac{hfo}}
	\centering
\begin{threeparttable}
	\begin{tabular}{|c|c|}
		\hline
		{\textbf{Parameter}}& \textbf{\acs{hfo}} \\
		\hline
		\multicolumn{2}{|c|}{\textit{actor}}\\\hline
		Input Layer & Linear(95, 1024) \\\hline
		\multirow{5}{*}{Hidden Layers} & Linear(1024, 512)\\
		  & Linear(512, 256)\\
		  & Linear(256, 256)\\
		  & Linear(256, 128)\\
		  & Linear(128, 128)\\
		  \hline
		  
		  \hline
		Output Discrete Action Layer& Linear(128, 4)\\\hline
		Output Parameters Action Layer& Linear(128, 7)\\\hline
		Activation &  leaky relu\\
		\hline
		\multicolumn{2}{|c|}{\textit{critic}}\\\hline
		Input Layer & Linear(106, 1024) \\\hline
		\multirow{5}{*}{Hidden Layers} & Linear(1024, 512)\\
		& Linear(512, 256)\\
		& Linear(256, 256)\\
		& Linear(256, 128)\\
		& Linear(128, 128)\\
		\hline
		
		Output Layer& Linear(128, 1)\\\hline
		Activation & leaky relu\\
		\hline
		
		\hline			
		Optimiser & Adam \\\hline
		Learning Rate & 1e-3\\\hline
		Betas &  (0.9, 0.999)\\\hline
		Target Policy Update & soft \\\hline
		Policy Update $\tau$ & 1e-3 \\\hline
		Mini-batch Size & 32\\			
		\hline
	\end{tabular}
\end{threeparttable}
\end{table}

The details of the \ac{paddpg} network used in the~\ac{hfo} experiments are reported in Table~\ref{tab:paddpg_setup}. Both the actor and the critic share the same configuration for the hidden layers. The input layer of the critic consists of the aggregation of three components: $95$ neurons for the state, $4$ neurons for the discrete action selection, and $7$ neurons for the parameters of the actions. In total, the input layer processes $106$ values.

\subsubsection{Multi-Agent~\acl{paddpg}}

In the multi-agent implementation of~\ac{paddpg}, the architecture is similar to the single-agent~\ac{paddpg} with once crucial difference, the critic is centralised. The centralised critic takes as input the state-action pairs of each agent and estimates the state-value of the joint actions.

While the hidden and output layers of the critic remain unchanged from the parameters previously presented in Table~\ref{tab:paddpg_setup} for the single-agent \ac{paddpg}, the input size is multiplied by the number of agents that share the centralised critic.

\subsection{\acl{pp} Encoder}\label{sec:pp_encoder}
For \acf{pp}, the observation of an agent is defined as a 3-channel array encoding the type of object within a cell, the orientation and the team-membership. The orientation of an object and its team membership are meaningful only for predator and prey.

\begin{table}[h!]
	\renewcommand*{\arraystretch}{1.1}
	\caption{\label{tab:preprocess_pp} Preprocess encoder for \ac{pp}.}
	\centering
	\begin{threeparttable}
		\begin{tabular}{|c|c|}
			\hline
			\textbf{Parameter}& \textbf{Value}\\
			\hline
			$1^{st} layer$ & Conv1d(3, 7, k=1)\\\hline
			$2^{nd} layer$ & Conv1d(7, 15, k=1)\\\hline
			$3^{rd} layer$ & Flatten(out\_dim=-2)\\
			\hline
		\end{tabular}
	\end{threeparttable}
\end{table}

In this thesis, the experiments provided are based on an encoder to pre-process the multi-channel observation and to flatten it into a single dimension array, Table~\ref{tab:preprocess_pp} reports its configuration.
The encoder uses a 1D convolution with a kernel size of 1 to capture local patterns and dependencies within the input. The choice of a kernel size of 1 is because there are no relationships between neighbouring cells in the grid.

\section{\acl{efontl} Implementation Details}\label{sec:imp_efontl}

\ac{efontl} is extensively detailed in Chapter~\ref{cpt:efontl}, where both the architecture and the workflow are thoroughly described to facilitate the transferring of experiences among agents. The \ac{efontl} code is accessible at the following link: \href{https://github.com/acastagn/EF-OnTL}{https://github.com/acastagn/EF-OnTL}. In summary, beyond the core \ac{rl} process, \ac{efontl} incorporates three additional components:

\begin{enumerate}
 	\item  a copy of~\ac{sarnd}, class \textit{UE} in the \textit{SARSrnd} file available in the repository, to estimate the agent's epistemic uncertainty. Each agent has associated a dedicated copy of~\ac{sarnd} that needs to be updated solely on the experiences coming from the related agent;
 	
 	\item a transfer buffer, class \textit{buffer} in the \textit{transferBuffer} file available in the repository, to store the uncertainty labelled experiences in form of a seven-element tuple $(s^t,a^t,r^t,s^{t+1},d^{t},u^t)$;
 	
 	\item Transfer Core Engine, file \textit{transferCoreEngine} in the repository, to enable the sharing of experience and the execution of the two processes needed to select the source and the content to be transferred.
\end{enumerate}

\ac{efontl} is designed to enable multiple agents to transfer experience in a distributed manner. However, for sake of simplicity, some operations can be executed in a centralised manner to reduce the implementation complexity, i.e., \textit{Source Selection Process}. The goal of this task is to select a source of transfer which is globally selected for all the agents. As the selection of source is not influenced by any external factor, the execution of this task can be executed in a centralised manner by an external entity to reduce the communication across agents. 

Regardless of the \textit{Source Selection process} implementation, the following step, \textit{Transfer Content Selection}, must be distributed on the agent side as this step is based on the target agent. To estimate the surprise values for the \textit{Transfer Content Selection}, the learning process of an agent must enable the estimation of the \ac{td}-error for an agent-environment interaction. In addition, the learning process has to allow for taking a  training step based on an external buffer passed as parameter.

\chapter{Evaluation}
\label{cpt:evaluation}\acresetall%

This chapter presents the evaluation study performed on \ac{efontl}. \ac{efontl} enables knowledge transfer across multiple agents by sharing experience as a batch of agent-environment interactions. \ac{efontl} enables a target agent to receive, filter and process a subset of the transferred experiences which is expected to expedite the convergence of its policy. 

The remainder of this chapter is organised as follows. The evaluation objectives for \ac{efontl} are introduced in Section~\ref{sec:eval-obj}, along with the baseline approaches and associated evaluation metrics. The benchmark environments are detailed in Section~\ref{sec:simulators}. Subsequently, the evaluation is organised into three sections and is presented in Section~\ref{sec:efontl_evaluation}. Section~\ref{ss:eval_efontl_vs_baselines} compares~\ac{efontl} with the baseline methods. Section~\ref{ss:eval_efontl_transfer_criteria} examines the impact of different transfer settings in~\ac{efontl}. 
Section~\ref{ss:efotnl-rs-sumo-expanded} assesses the adaptation capabilities of \ac{efontl} when transferring across tasks with heterogeneous dynamics. Lastly, Section~\ref{sec:summary_eval} summarises the evaluations presented in the chapter.

\section{Evaluation Objectives}\label{sec:eval-obj}

This section, outlines the core objectives, denoted as \textit{O1} to \textit{O9} underlying the experiments conducted and presented within this thesis:

\begin{objective-list}
	\item The first objective of evaluation is to assess whether~\ac{efontl} improves convergence time compared to a no-transfer baseline, in which agents do not share any information and learn independently. Achieving quicker convergence is the fundamental goal for any experience-based offline~\ac{tl} framework aiming to outperform a transfer-free baseline. To assess speed convergence, in all benchmark environments we also implement a no-transfer baseline where agents learn independently, more details are provided in Section~\ref{ss:eval_baseline}, which will be used to compare \ac{efontl}. The convergence time will be analysed through the comparison of the learning curves defined on the cumulated reward over an episode or a specified interval;

	\item  The second objective evaluates whether and to what extent~\ac{efontl} improves the agents' performance when compared to a no-transfer baseline, in which agents do not share any information and learn independently, therefore the baseline used in this objective is the same as in \textit{O1}, but we compare the performance against a different metric.  
	When imperfect agents are used as a source of transfer, the convergence speed alone may not be adequate to analyse the transfer impact. Furthermore, by transferring agent-environments interactions, the effect of transfer on a target agent may be delayed. 
	This objective aims to determine whether~\ac{efontl}, through online experience sharing, enables agents to achieve overall better performance after convergence when compared to  transfer-free agents. The performance improvements will be analysed by comparing the performance achieved by agents at a fixed episode, while following their policies.

	\item The third objective evaluates whether~\ac{efontl} achieves comparable performance when compared to a~\ac{tl} scenario following the teacher-student paradigm. In the baselines used, the source of transfer influences directly the action-decision process of a target agent by providing an action as advice. 
	The goal of this objective is to determine if~\ac{efontl} can be a valid alternative to the teacher-student framework, with the additional advantage of not having to have identified a fixed teacher a-priori.	For this comparison, we implement two action-advice based baselines, presented in Section~\ref{sec:baselines}: one based on experienced agents, and the other based on imperfect agents. 
	To compare~\ac{efontl} against the \ac{tl} baselines, we report both training performance over episodes and performance achieved at a fixed episode.

	\item  The fourth objective analyses whether and to what degree \ac{efontl}~transfer effectiveness is influenced by~\ac{ss} used to identify the source of transfer. When there are multiple agents that could act as a source of transfer there is a need for a process to identify the suitable source of knowledge. This objective aims to study the impact of selecting the source of transfer based on different criteria, as presented in \ac{efontl} design chapter in Section~\ref{ss:SourceSelection}. The experiments will be carried out across multiple scenarios in two different environments: Cart-Pole and \acf{pp}. The setup of these experiments will vary the~\ac{ss} while keeping the other criteria fixed.

	\item	The fifth objective analyses whether and to what degree \ac{efontl}~transfer effectiveness is influenced by~\ac{tcs} used to filter the knowledge on target side. While~\ac{efontl} system identifies a common source of transfer, the target agents, pulling batch of experiences from the source, will require different interactions specific to their learning status. To assess the impact of utilising various criteria for filtering incoming knowledge on a target agent side, this objective analyses the impact brought by~\ac{tcs}, as presented in \ac{efontl} design chapter in Section~\ref{ss:TransferMethods}. 
	Similarly to~\textit{O4}, the experiments will be carried out in multiple scenarios across two different environments: Cart-Pole and \ac{pp}. The setup of these experiments will vary~\ac{tcs} while keeping the other transfer parameters fixed.
	
	\item The sixth objective analyses whether and to what extent \ac{efontl} transfer effectiveness is influenced by the number of interactions selected by a target agent within a single transfer step. 
	While the number of interactions transferred within a single transfer step is not related to the transfer budget~$B$ used to filter the incoming knowledge on the target side, this objective studies if there is a correlation between the transfer outcome and the quantity of tuples sampled on a target agent side. 
	The experiments used for this analysis will be carried out in multiple scenarios across two environments: Cart-Pole and \ac{pp}. The setup of these experiments will vary the budget~$B$ while keeping \ac{ss} and \ac{tcs} fixed.

	\item The seventh objective studies the extent of applicability of~\ac{efontl}, based on complexity of the environment. \ac{rl} can be used to learn a policy in various environments of different complexity, i.e., dimensions of the combinatorial state-action space, delayed and sparse rewards, and complex underlying dynamics. The seventh objective aims to identify the capabilities and limitations of \ac{efontl} in relation to the complexity of different environments. \ac{efontl} will be analysed into four different environments of increasing complexity: Cart-Pole, \ac{pp}, \ac{hfo}, and \ac{rs-sumo}. These environments are detailed in Section~\ref{sec:simulators}.

	\item The eighth objective evaluates the adaptation capability of~\ac{efontl} to environments with heterogeneous dynamics. Note, however, that adaption to different dynamics was not and is not considered explicitly in the design of ~\ac{efontl}, any further than the adaptation capabilities of underlying \ac{rl} algorithm used, i.e., \ac{dqn} and \ac{paddpg}. However, in order to assess the boundaries of \ac{efontl} applicability, and establish directions for future work, we perform evaluation in which we use \ac{efontl} to transfer knowledge not just to the same environment dynamics as in previous experiments, but to those with different underlying dynamics. Specifically,  we analyse various transfer scenarios within the \ac{rs-sumo} environment, enabling transfers across agents specialised in different demand trends.

	\item The ninth objective evaluates performance comparing to centralised multi-agent learning.	While \ac{efontl} operates in a decentralised manner with reduced communication, making it distinct from multi-agent learning, we still intend to assess them purely in terms of performance. Such an evaluation would be useful, especially in scenarios where either approach might be suitable from an architectural point of view. This evaluation will be conducted through a comparison with~\ac{marl} baselines, more detail in Section~\ref{ss:eval_baseline}. The experiments will carried out in two different environments: \ac{pp} and \ac{hfo}, using two different \ac{rl} paradigms.

\end{objective-list}

\subsection{Baselines}\label{ss:eval_baseline}
To assess \ac{efontl} impact, this thesis compares \ac{efontl} against different baselines:
\begin{enumerate}
	\item \textit{no-transfer} $-$ each agent is equipped with an independent copy of the learning model that is updated only using its own experience;
	
	\item \textit{\ac{ocmas}}~\cite{ilhan2019teaching} $-$ this baseline is discussed in detail in Section~\ref{sec:baselines}, but here we repeat its main characteristics. In this approach, each agent owns a copy of the learning process, which is  optimised based on the experiences collected by that specific agent. Additionally, agents follow the teacher-student framework to share knowledge and rely on \ac{rnd} to estimate their confidence level within a state. 
	
	At each time step, every agent shares the state that they have visited with the rest of the agents to estimate their uncertainties. When an agent is the most uncertain in its visited state, the transfer process begins. Other agents provide an action-advice to the most uncertain agent. The final action decision is made through majority voting based on the received advice. Transfer is constrained by a budget set for each agent. This budget limits the number of times that an agent follows the received advice. Despite this limitation, agents maintain continuous interaction, sharing their explored states and evaluating their associated uncertainties.

	\item \textit{\ac{rcmp}}~\cite{da2020uncertainty} $-$ this baseline is discussed in detail in Section~\ref{sec:baselines}, but here we repeat its main characteristics. A target agent is guided by an external group of expert teachers, referred as jury, following the teacher-student framework. The advising strategy is based on comparing the uncertainty level of a target agent against a predefined threshold and the advice shared is in form of action. The communication is restricted by a budget and, in this implementation, the target agent selects the action to take as the most frequent action suggested by the jury.

	\item\textit{\ac{marl}} $-$ the \ac{marl} approaches used in this thesis are the multi-agent adaptation of~\ac{paddpg}~\cite{hausknecht2015deep} and QMIX~\cite{rashid2020monotonic}, both described in Section~\ref{ss:tl_and_marl}. The selection of the \ac{marl} algorithm used is based on the learning paradigm used by~\ac{efontl}. The multi-agent~\ac{paddpg} variant is used in~\ac{hfo}. Here, a centralised critic learns a global action-value function based on the joint actions taken by the agents. In the \ac{pp} environment QMIX is used. QMIX enables the agents to aggregate their independent Q-functions to influence the learning model of a certain agent based on other agents' actions.

\end{enumerate}

Intuitively, based on its design, \ac{efontl} is expected to improve over \textit{no-transfer} method and perform similarly as~\ac{ocmas}. \ac{rcmp} is expected to be an upper limit for~\ac{efontl} due to the availability of pre-trained agents.

\subsection{Metrics}\label{ss:eval_metrics}

The evaluation metrics commonly used within the \ac{tl} context are generally defined on the reward. These metrics assess the impact of transferred knowledge by measuring the difference in rewards between the~\ac{tl} approach and a baseline. In this thesis, we assess the following:
\begin{enumerate}
	\item asymptotic improvement $-$ it measures the performance gap that there might be on the learning curves of two agents;
	\item performance at fixed time $-$ it measures the performance achieved by an agent after a certain number of episodes. Performance are expressed through cumulated reward as well as environment-specific metrics;
	\item transfer cost $-$ it measures the number of interactions or action-advice transferred to a single agent;
\end{enumerate}

Although these metrics are widely used to assess the impact of~\ac{tl}, they offer a comprehensive understanding of the training stage of a~\ac{tl}-enabled agent versus an agent with no external support. In~\ac{efontl}, the impact of transferring external knowledge as agent-environment experiences is expected to be delayed when compared to other form of advice, i.e., action-advice. Therefore, in this thesis, where we required, we also introduce additional metrics specific to the environment we are analysing. Finally, to ensure a robust evaluation and minimise random bias, the results are presented as an aggregation of multiple independent simulations collected under different initialisation seeds.

\section{Evaluation Environments}\label{sec:simulators}

To assess \ac{efontl} performance against the baselines, this thesis evaluates the approaches across 4 different environments of increasing complexity: 1) Cart-Pole, 2) \ac{pp}, 3) \ac{hfo} and 4) \ac{rs-sumo} with real-world data. This section motivates the environments used in this research and presents their implementation details needed to reproduce the experiments.

\subsection{Cart-Pole}\label{ss:sim_Cart-Pole}
Cart-Pole is a simple and popular task widely adopted in the field.  Due to its simplicity, Cart-Pole is a good option for validating the correctness of \ac{rl} implementations and contributions.

\begin{figure}[!htb]
	\centering
	\includegraphics[width=.5\columnwidth]{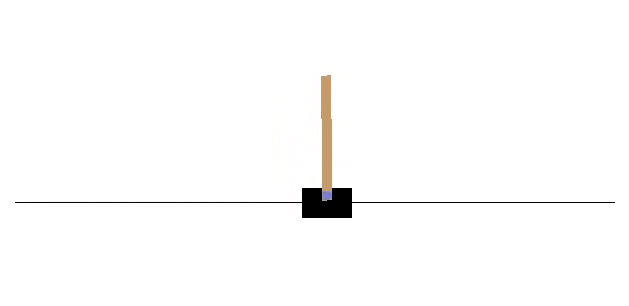}
	\caption{Cart-Pole environment~\cite{openAIgym2016}.\label{fig:cartpole_screenshot}}
\end{figure}

The Cart-Pole problem consists of an unstable pole placed vertically on top of a moving cart~\cite{sutton1998introduction}. At initialisation time, the pole is balanced but subjected to gravity forces. The goal consists of preventing the pole from falling by moving the cart along the x-axis while keeping it within certain boundaries. Figure~\ref{fig:cartpole_screenshot} shows a capture of the Cart-Pole problem used for the experiments this thesis.

Observation consists of 4 values, cart position $[-4.8, +4.8]$, cart velocity $(-inf, +inf)$, pole angle~$[-24^\circ, +24^\circ]$ and pole velocity~$(-inf, +inf)$. Cart is moved left or right by an agent through 2 discrete actions and receives a positive reward of $+1$ for every step that the pole is balanced, therefore acquiring a reward of $1$ in every time step, until the pole collapses and the episode terminates.

The Cart-Pole environment used within this research is based on the implementation by Open-AI~Gym~\cite{openAIgym2016}. To enable multiple agents visiting a Cart-Pole problem simultaneously, the environments is initialised 5 times in parallel.

As the agents interact one to the other in~\ac{ocmas} and \ac{efontl}, agents are synchronised at the end of each episode to ensure that transfer happens between agents of similar performance.

\subsection{\acl{pp}}\label{ss:sim_pp}

\begin{figure}[!htb]
	\centering
	\includegraphics[width=.5\columnwidth]{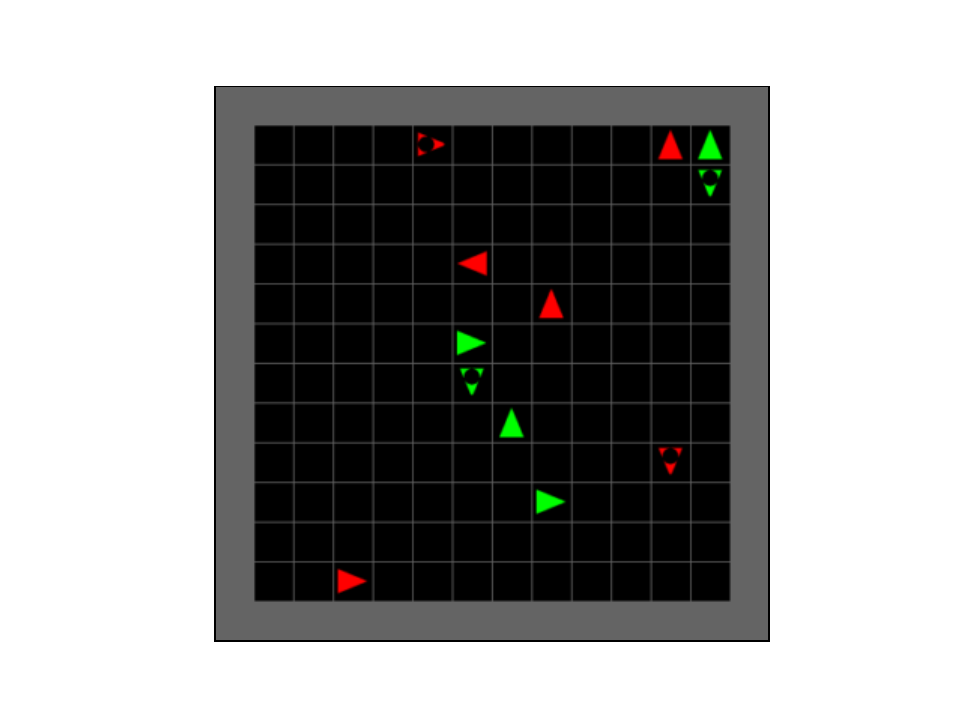}
	\caption{Screenshot of \acf{pp}.\label{fig:mtpp_screenshot}}
\end{figure}

This version of predator-prey is an advanced configuration built upon the environment introduced in Section~\ref{ss:preliminary-envs-pp}. Such a task has been widely used to study multi-agent algorithms and contributions to the field of \ac{marl}.  A screenshot of the game is available at Figure~\ref{fig:mtpp_screenshot}. There are a total of $8$ predators, shown as filled triangles, and $4$ prey, shown as pierced triangles. Predators and prey are evenly distributed in two different colour-based teams.

The environment dynamics and goal are simple, predator chases prey while prey tries to survive the longest by evading the predator. Both predator and prey can be controlled by an intelligent agent and the number of agents involved can be fully customisable. Furthermore, the grid is fully customisable in size and layout as well as the agents sensors. The results shown in this thesis assume that predators are controlled by \ac{rl}-powered agents while prey follow a random policy to escape the chasers.

\ac{pp} consists of two teams that compete to catch prey faster than the other team. At initialisation time, teams are balanced and spread over the grid, thus they have same amount of resources, predators and prey. An episode terminates when a team has no prey left to be captured. 

The predator reward model is based on the action taken by the agent and the observed outcome. This is reported in Table~\ref{tab:pp-reward_model}. The actions that change the position or orientation of the predator have an associated reward of $-0.01$. For the \textit{catch} action there are multiple possible outcomes: when a prey is caught a reward of $\pm1$ is given to the agent. This reward is positive when both predator and prey belong to the same team and negative otherwise; On the other hand, when the catch fails, a reward of $-0.5$ is given to the agent. Finally, a reward of $-0.25$ is given to an agent for the \textit{hold} action.

\begin{table}[h!]
	\renewcommand*{\arraystretch}{1.1}
	\caption{\label{tab:pp-reward_model} \acf{pp} reward model.}
	\centering
	\begin{threeparttable}
		\begin{tabular}{|c|c|}
			\hline
			{\textbf{Parameter}}&\textbf{Value}\\\hline
			catch same team prey & $+1$\\\hline
			catch opponent team prey & $-1$\\\hline
			failed catch & -.5\\\hline
			hold & -.25\\\hline
			other actions & -.01\\
			\hline
		\end{tabular}
	\end{threeparttable}
\end{table}

In the configuration used within this research, the grid is composed by $12\times12$ accessible cells shaped as a obstacle-free square delimited by a wall. Each agent perceives a $3\times3$ grid centred over next consecutive cell. Thus, the observation is composed of a 3-dimensional $3\times3$ matrix. First channel describes object type, varying between: void, wall, predator and prey. Second channel identifies the team membership: None, in case of an empty cell or wall, red and green. Third channel indicates the orientation of the perceived object. The orientation of an object is determined by a global orientation system, independent of the observing agent's orientation. It can be categorised as up, down, left, right, or none in the case of an empty or wall cell.



In the experiments reported within this thesis, predators are given precedence to select and take actions on prey. Consequently, in situations where a predator selects \textit{catch} action when facing a prey, the prey becomes unable to escape.

Moreover, the priorities among predators are randomly assigned during initialisation, regardless of team membership, and remain fixed throughout each episode. Consequently, if the actions taken by two predators are conflicting, e.g., by occupying the same empty cell that are both facing from different sides, only the predator that moves first will succeed in the subsequent step, while the other will remain in its original position.

This specific \ac{pp} setup is extremely interesting as different kind of multi-agent situations and interactions arise. As per Ferber at~\cite{10.5555/520715}, it is possible to observe competitive interactions, between the two teams as their goals are mutually incompatible, and collaborative interaction within a team, as all the agents that belong to a team aim to accomplish a common goal.

As evaluation metrics, on top of the reward, this thesis reports the observed probability of winning a match and capturing a prey assigned to the same and opposite team. 


\subsection{\acl{hfo}}\label{ss:sim_hfo}

\ac{hfo} is a multi-agent environment based on a simplified game of soccer where agents need to collaborate to score a goal.  Collaborative interaction further complicates the task that already has sparse reward function,  continuous state and parametrised control space. Thus, having a single poor performing agent within the team will likely result in capping the team performance.
Task is episodic and an episodes terminates when one of the following events occurs: 1) a goal is scored; 2) ball is out of bounds; 3) a defender or the goalkeeper gets possession of the ball; 4) episode reaches the time limit.

\begin{figure}[!htb]
	\centering
	\centering
	\includegraphics[width=.5\columnwidth]{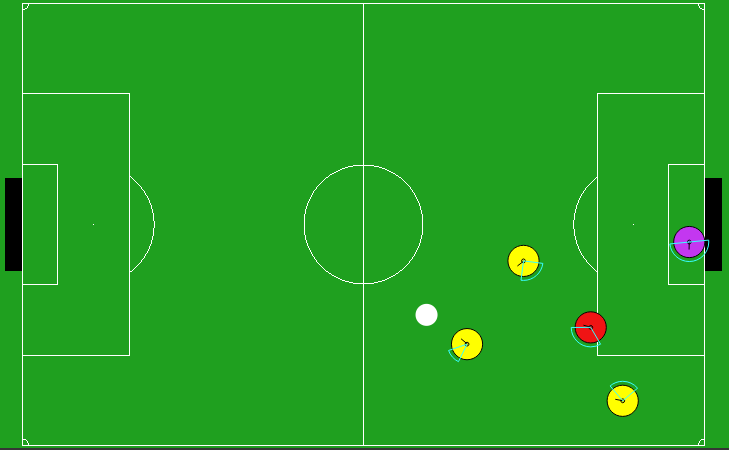}
	
	\caption{Screenshot of \ac{hfo} game.}\label{fig:hfo_screenshot}
\end{figure}

The setup used in this work is a 3 vs 2 situation consisting of 3 offense players and 2 for the defense. The defense consist of a goalie and a single defender. Figure~\ref{fig:hfo_screenshot} shows a screenshot of the \ac{hfo} setup used in this thesis. At the beginning of an episode, all players are initialised in the rightmost half of the pitch. Offense players are positioned in proximity to the midfield line, the defender is placed just outside the penalty box, and the goalie is located between the sticks. Initially, ball control is given to the offense team, while the other team tries to intercept the ball.  In this work, offense players are powered by \ac{rl} agents, while the others are played by HELIOS baseline~\cite{hausknecht2016half}.

Offense players are controlled by~\ac{rl} agents through the mid-level control space \footnote{There are three levels of control with different complexity available for an agent: high, offering only discrete actions; low, where actions are defined on continuous parameters, and mid, offering intermediate control complexity between the two.}. As a result, there are four available actions: 

\begin{enumerate}
	\item $kick\_to(x, y, speed)$ $-$ the player kicks the ball to the specified coordinates and with the provided speed;
	\item  $move\_to(x,y)$ $-$ the player goes from the current position to the specified point using the maximum running speed;
	\item $dribble\_to(x,y)$ $-$ the player dribbles the ball to the specified coordinates. If the player controls the ball it tries to avoid the opponent and tackle otherwise; 
	\item $intercept$ $-$ based on the ball velocity, the player moves to intercept the ball.
	
\end{enumerate}

$(x, y)$ coordinates used to specify a target point are normalised within $[-1,+1]$ interval while speed ranges in $[0,3]$ interval. When selecting an action, an agent must pick the discrete action alongside any relevant continuous parameters necessary to parametrise the selected action.

\ac{hfo} state provides information on the controlled player, the ball, the team mates and the opponent team players. In the setup used in this thesis, the observation for each agent consists of $95$ features. For readability, \ac{hfo} state is detailed in Appendix at Sec.~\ref{app:hfo_state}. 

The performance of players is assessed by tracking the number of goals they score within a set range of episodes. When a goal is scored, the environment returns a positive reward~($+1$) to all players on the scoring team, regardless of any individual contribution to the goal.

The high quantity of observed features and the continuous control space make the~\ac{hfo} environment an interesting benchmark for assessing~\ac{tl} frameworks, in particular~\ac{efontl}. The \ac{nn} underlying the learning process may gain significant advantages from the transferred experiences in this context.

\subsection{Multi-brain~\acl{rs-sumo}}\label{ss:multi-brain-definition}

The multi-brain version of~\ac{rs-sumo} is built upon the environment introduced in Section~\ref{ss:preliminary-envs-rs-sumo}.

While Manhattan is geographically concentrated in a limited area and could potentially be managed by a single policy replicated across all vehicles, in our environment setup, we divided the zone into different sub-areas to investigate the impact of transferring across multiple policies, each tailored to an unique non-overlapping zone. Each policy is trained on a distinct subset of data, making it interesting to observe the transfer learning effects induced by \ac{efontl}.

The multi-brain implementation of \ac{rs-sumo} consists of four brains, each trained on an independent and non-overlapping sub-area of Manhattan. The division of these zones is established by clustering the origins and destinations of ride-requests available during the morning peak hours from $7-10$am. Cluster centroids are identified using K-means~\cite{hartigan1979algorithm}, with the Manhattan distance is employed as the distance measure for estimating cluster membership.

\begin{figure}[htb!]
	\centering
	\includegraphics[width=0.75\columnwidth]{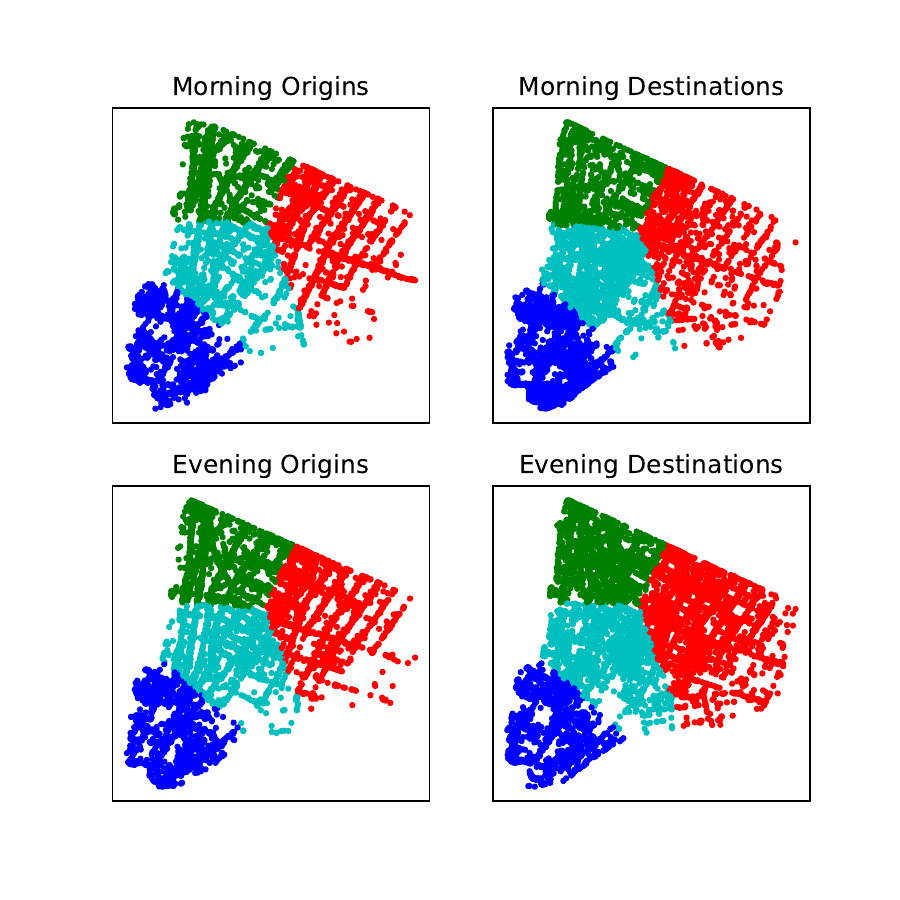}
	\caption{ Ride-requests membership over the 4 clusters identified by K-means~\cite{hartigan1979algorithm}.}
	\label{fig:clustering_manhattan}
\end{figure}

The clustering of requests origin and destination shown in Figure~\ref{fig:clustering_manhattan} demonstrates the distribution of membership across the four clusters during both morning ($7-10$am) and evening ($6-9$pm) demand sets.  Although this clustering approach is designed to achieve a balanced division of zones and mitigate demand imbalances, each brain will have different opportunities to collect training data due to the different area covered by each brain.

When an agent needs to take an action, it employs the policy from the respective brain determined by hard cluster membership based on the current vehicle's position. When enabling online~\ac{tl} across agents, despite the state-domain space visited by each agent will be unique to that agent, each brain will receive part of knowledge from the other zones.

\section{\ac{efontl} Evaluation}\label{sec:efontl_evaluation}

This section presents~\ac{efontl} evaluation experiments across the four environments to validate whether the promising results on experience sharing in an offline context, shown in Chapter~\ref{cpt:preliminary_studies}, are maintained when transferring online across learning agents. 

Firstly, Section~\ref{ss:eval_efontl_vs_baselines} compares~\ac{efontl} against the identified baselines to address the evaluation objectives \textit{O1}, \textit{O2}, \textit{O3}, \textit{O7} and \textit{O9}. The baselines used are \textit{no-transfer}, \ac{rcmp}, \ac{ocmas}, and \ac{ccl}. These evaluation objectives appraise the positive transfer impact of~\ac{efontl} and compare the improvements against the~\ac{tl} baselines and \ac{marl} \ac{ccl} to assess whether~\ac{efontl} could be a more advantageous choice for enhancing agents' performance in a multi-agent context. To provide a robust evaluation for these studies three scenarios of different complexity are utilised: Cart-Pole, \ac{pp}, and \ac{hfo}.

Secondly, Section~\ref{ss:eval_efontl_transfer_criteria} presents the \ac{efontl} studies to address the evaluation objectives \textit{O4}, \textit{O5} and \textit{O6}, presented in Section~\ref{sec:eval-obj}. These objectives aim to study the impact of various transfer parameters in~\ac{efontl} such as the \acf{ss}, introduced in Section~\ref{ss:SourceSelection}, the \ac{tcs}, introduced in Section~\ref{ss:TransferMethods}, and the transfer budget. The benchmark scenarios used for these studies are the Cart-Pole and \ac{pp} benchmark environments, presented respectively in Section~\ref{ss:sim_Cart-Pole} and Section~\ref{ss:sim_pp}.

Thirdly, Section~\ref{ss:efotnl-rs-sumo-expanded} reports the experiments in the~\ac{rs-sumo} environment with real-world demand pattern. These experiments evaluate the evaluation objective \textit{O8} to appraise the generalisation ability of~\ac{efontl} powered agents across different scenarios.

\subsection{\ac{efontl} and Baselines Evaluation}\label{ss:eval_efontl_vs_baselines}
This section answers \textbf{RQ1}, identifying if and to what extent \ac{efontl} improves the system performance. To provide a better understanding and to position \ac{efontl} performance, this section uses the evaluating objectives~\textit{O1} and \textit{O2}, by comparing the performance achieved by~\ac{efontl} against a \textit{no-transfer} baseline, \textit{O3}, by comparing \ac{efontl} against action-advice based teacher-student baselines, \ac{ocmas} and \ac{rcmp}, and \textit{O9}, by comparing the \ac{efontl} performance against the multi-agent baselines, QMIX and \ac{maddpg}. Furthermore, this section analyses the generalisation capability of \ac{efontl} by investigating its \ac{tl} effectiveness across multiple environments of varying complexities, aiming to evaluate the objective \textit{O7}.

\ac{efontl} and the baselines are tested in four different benchmark environments: Cart-Pole, discussed in Section~\ref{ss:sim_Cart-Pole}, \ac{pp}, discussed in Section~\ref{ss:sim_pp}, \ac{hfo}, discussed in Section~\ref{ss:sim_hfo}, and \ac{rs-sumo}, discussed in Section~\ref{ss:multi-brain-definition}.

\begin{table}[h!]
	\small
	\renewcommand*{\arraystretch}{1.1}
	\caption{\label{tab:sim-baselines} Summary table of the \ac{rl} approaches implemented in each benchmark scenario.}
	\centering
	\begin{threeparttable}
		\begin{tabular}{|c|c|c|c|c|c|c|}
			\hline
			\multicolumn{1}{|c|}{\textbf{Environment}}&\textbf{\ac{efontl}} &\textbf{\textit{no-transfer}} &\textbf{\ac{ocmas}} &\textbf{\ac{rcmp}} & \textbf{\textit{QMIX}} &\textbf{\textit{\acs{maddpg}}}\\\hline
			Cart-Pole &\cmark&\cmark&\cmark&\cmark&\xmark&\xmark \\\hline
			\ac{pp} &\cmark&\cmark&\cmark&\cmark&\cmark&\xmark \\\hline
			\ac{hfo} &\cmark&\cmark&\xmark&\cmark&\xmark&\cmark \\\hline
			\ac{rs-sumo} &\cmark&\cmark&\cmark&\cmark&\xmark&\xmark \\\hline
		\end{tabular}
	\end{threeparttable}
\end{table}

Table~\ref{tab:sim-baselines} reports the methods employed in each benchmark environment. Due to specific limitations of certain environments, some of the baselines were not applicable. We further discuss this matter later in the subsequent sections. Table~\ref{tab:efontl-parameters} provides a summary of the transfer settings used to evaluate~\ac{efontl}. For the \ac{rs-sumo}, epochs and episodes are separated by a period.

\begin{table}[htb!]
	\small
	\renewcommand*{\arraystretch}{1.1}
	\caption{\label{tab:efontl-parameters} \ac{efontl} parameters on benchmark environments.}
	\centering
	\begin{threeparttable}
		\begin{tabular}{|c|c|c|c|c|}
			\hline
			\multicolumn{1}{|c|}{\textbf{Parameter}}&\textbf{Cart-Pole} &\textbf{\ac{pp}}&\textbf{\ac{hfo}}&\textbf{\ac{rs-sumo}}\\\hline
			\multicolumn{1}{|c|}{{N}}
			& 5 & 4  & 3 & 4\\		\hline	
			\multicolumn{1}{|c|}{{TF}}
			& 200 & 300 & 400 & 5 \\			\hline
			\multicolumn{1}{|c|}{{TB Capacity}}
			& $10,000$ & $100,000$ & $25,000$ & $10,000$\\\hline
			\multicolumn{1}{|c|}{{SS}}
			& $\overline{U}$ & $\overline{U}$ & $\overline{U}$ & $\overline{U}$\\		\hline
			\multicolumn{1}{|c|}{{\ac{tcs}}}
			& \acs{hdc} & \acs{hdc} & \acs{hdc} & \acs{hdc}\\	\hline	
			
			\multicolumn{1}{|c|}{{Budget B}}
			& $5,000$ & $500$ & $100$  & $100$\\	\hline	
			\multicolumn{1}{|c|}{{Ep. Start Transfer}}
			& $600$ & $2,500$ & $2,400$ & $1.5$\\	\hline		
			\multicolumn{1}{|c|}{{Max Timestep}}
			& $400$ &$200$ & $500$ &	n.a.\\\hline
			\multicolumn{1}{|c|}{{Max Episode}}
			& $1,800$ &$8,000$ & $20,000$ &	$9.9$\\ \hline			
		\end{tabular}
	\end{threeparttable}
\end{table}

In the subsequent sections, we analyse the impact of~\ac{efontl} versus the baselines in each benchmark environment.

\subsubsection{Cart-Pole}\label{sss:efontl-vs-baselines_Cart-Pole}
In the Cart-Pole benchmark environment, as introduced in Section~\ref{ss:sim_Cart-Pole}, a total of $5$ agents are concurrently trained. Every agents start a new episode at a synchronised timestep, in order to facilitate the inter-agent transfer.  The sharing of experience is enabled from episode~$600$ and a batch of experiences is shared every $200$ episodes thereafter. The transfer of knowledge is postponed until the $600th$ training episode as, in the initial stage, agents lack sufficient knowledge to estimate their epistemic confidence. Furthermore, transferring knowledge in the first episodes would add little value to the target policies.

\begin{figure}[htb!]
	\centering
	\includegraphics[width=0.75\columnwidth]{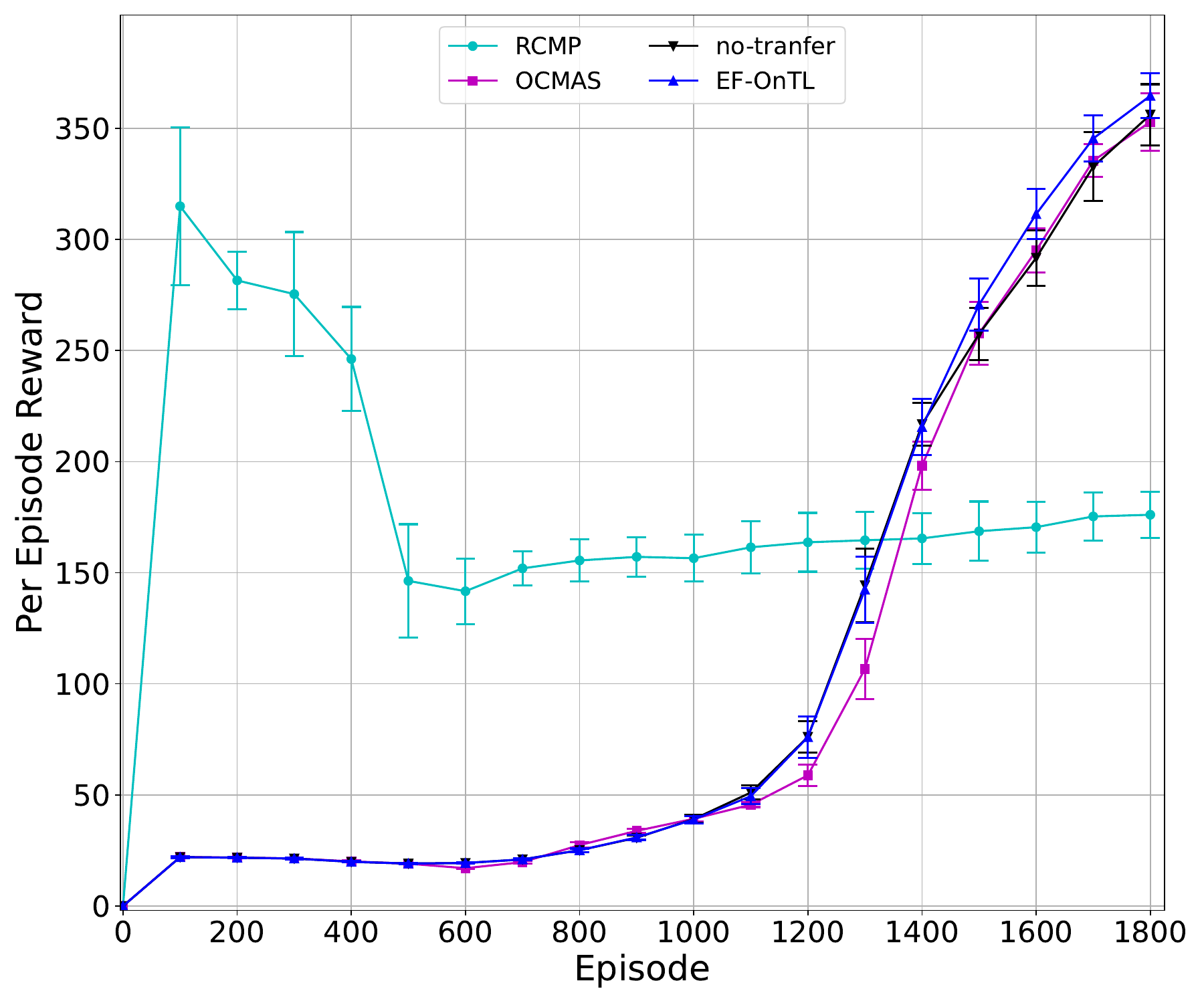}
	\caption{\ac{efontl} (\ac{hdc}, budget: $5,000$ and $SS: \overline U$)  compared against \ac{rcmp}, \ac{ocmas} and \textit{no-transfer} in Cart-Pole.}
	\label{fig:efontl_vs_baseline-CartPole}
\end{figure}

Evaluation in the Cart-Pole benchmark environment analyses the cumulative reward achieved by agents across training episodes. Figure~\ref{fig:efontl_vs_baseline-CartPole} shows episode count on the $x$-axis and the cumulative episode reward on the $y$-axis. In this benchmark environment, the multi-agent baseline is omitted as each agent interacts exclusively with its copy of the environment and therefore, there is no context of joint actions.

In Figure~\ref{fig:efontl_vs_baseline-CartPole}, \ac{efontl} demonstrates improved performance over the \textit{no-transfer} baselines starting from episode $1,400$. Prior to that episode, the learning curves of \ac{efontl} agents and the \textit{no-transfer} approach overlap, achieving, on average, similar performance level.  The \ac{rcmp} achieves very high levels of reward during the initial episodes, from $0$ to $100$. The increased reward is given by the lack of confidence on the target agents that lead to a receiving continuous guidance thorough advice. However,
following the initial spike, the learning curves exhibits a declining trend until episode $600$. Thereafter, the increased confidence of the target agents hinders advising, resulting in stationary performance with limited improvements. Lastly, \ac{ocmas} shows a noticeable performance gap compared to \textit{no-transfer} between episodes $1,100$ and $1,400$, ultimately matching the performance by the conclusion of the training episodes.

\subsubsection{\acl{pp}}\label{sss:baselines-efontl-pp}
The second environment used to benchmark \ac{efontl} against the baselines is \ac{pp}, detailed in Section~\ref{ss:sim_pp}. In \ac{pp} four agents, each controlling a predator within the same team, are concurrently trained in a multi-agent scenario. Similarly as in Cart-Pole, the sharing of experience is delayed until the $2,500$th episode and transfer happens every $300$ episodes thereafter until episode $7,500$. 

Since agents are concurrently trained to achieve their team goals, in this environment we also compare \ac{efontl} against QMIX, a \ac{marl} baseline.

\begin{figure}[htb!]
	\centering
	\includegraphics[width=.75\columnwidth]{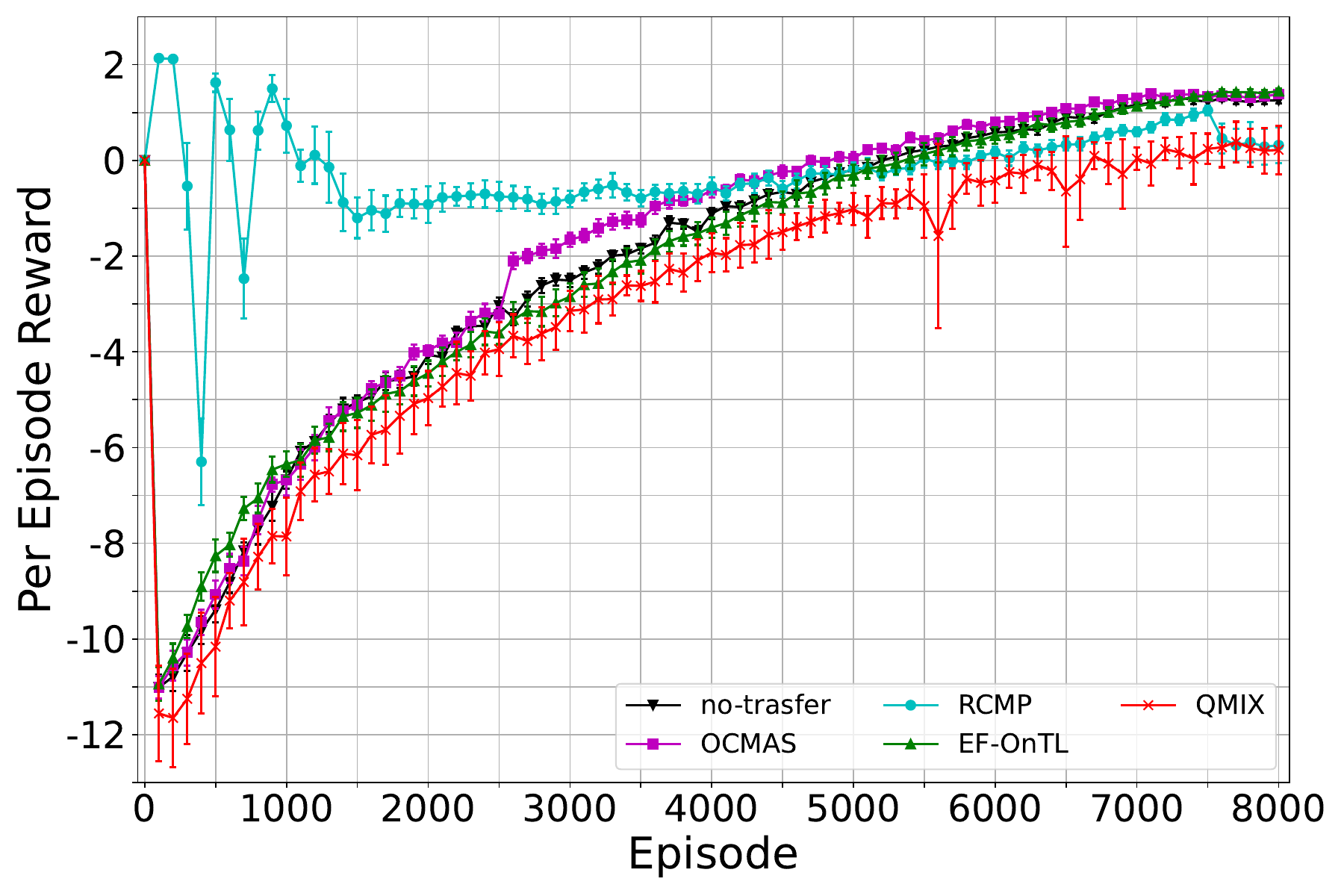}
	\caption{Learning curves of \ac{efontl} (\ac{hdc}, budget: $500$ and $SS: \overline U$) compared against \textit{no-transfer}, \ac{ocmas}, \ac{rcmp} and \textit{QMIX} in \ac{pp}.}
	\label{fig:PP-lc-efontl-vs-baselines}
\end{figure}

To evaluate the performance of the implemented methods in \ac{pp} we first report the learning curves in Figure~\ref{fig:PP-lc-efontl-vs-baselines}. The $x$-axis denotes the episode counter from $0$ to $8,000$ and the $y$-axis denotes the average reward scored by agents. The last $500$ episodes are used to test the methods as the learnt policy is sampled in a deterministic way.

Both \ac{ocmas} and \ac{rcmp} show considerable positive transfer within the first initial episodes where transfer is enabled. Specifically, \ac{ocmas} demonstrates a noticeable performance improvement between episodes $2,500$ and $2,600$ upon enabling transfer.  Similarly, \ac{rcmp}, thanks to the expert teacher, achieves positive rewards in the first $200$ episodes, while others techniques experience severe negative rewards due to random exploration. Although, \ac{rcmp} trend resembles what observed in Cart-Pole, where, after a remarkable jump-start, the learning curves assumes a flat trend. 
In contrast, \ac{efontl} does not display an immediate improvement due to the enabling of transfer; instead, the effect of transfer is distributed over time and eventually it demonstrates  a trend similar to that of \ac{ocmas}.

\vcchange{
Among all the techniques evaluated, QMIX records the lowest level of reward. Such lower performance is a consequence of the training setup, where the two teams that are benchmarked against each other are trained in a competitive scenario.}

\vcchange{
In the QMIX multi-agent learning framework, agents experience a delay in the learning phase due to the complexities involved in learning the joint state-action value function. Therefore, the opponent teams, composed by independent agents, who do not rely on joint state-action value function, learn to adapt their strategies more quickly. As a result, the opportunity for the QMIX team to develop an effective policy is lower.}

\vcchange{
While QMIX certainly encourages collaboration among agents, such joint learning can be a drawback in competitive scenarios. In contrast, \ac{efontl} decouples the learning and transfer phases. In the evaluation carried out on \ac{pp}, \ac{efontl} leads to better results. This is because the external knowledge, with the exception of cases involving negative transfer, does not delay an agent's learning process.}

\begin{figure}[htb!]
	\centering
	\includegraphics[width=1\columnwidth]{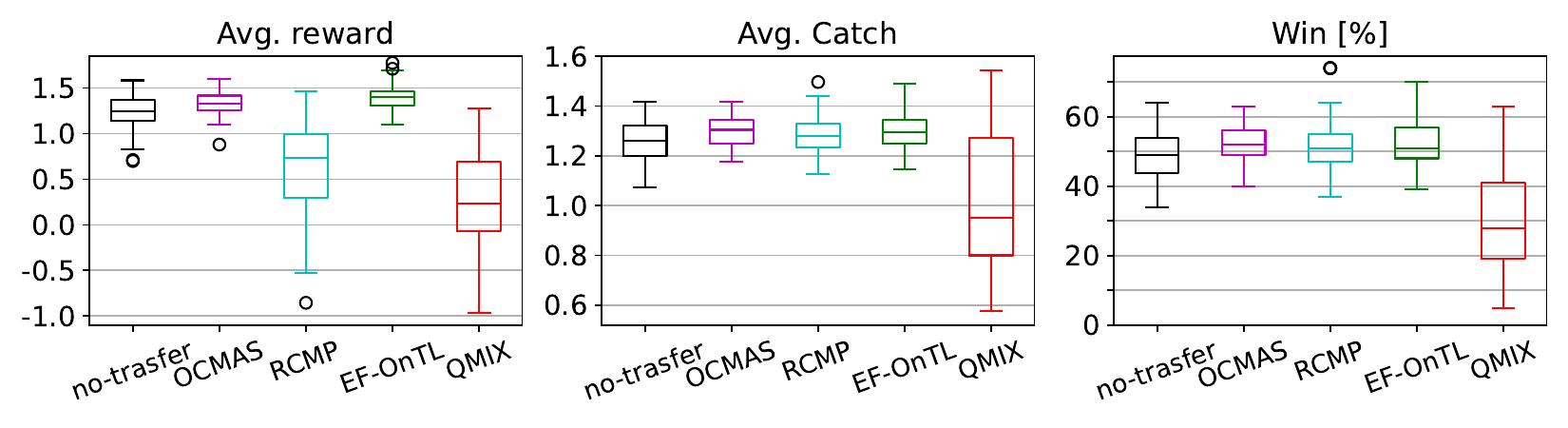}
	\caption{\ac{efontl} and baselines evaluation metrics on $500$ test episodes in \ac{pp}}
	\label{fig:PP-boxplot-efontl-vs-baselines}
\end{figure}

The \ac{pp} environment provides additional metrics that enhance the comprehension of the ongoing dynamics. In addition to the reward, this environment offers evaluation metrics such as the average number of successful prey caught in an episode~(Avg. Catch) and the win probability~(Win [\%]). The performance against these metrics are shown in Figure~\ref{fig:PP-boxplot-efontl-vs-baselines}, providing a comprehensive perspective over $500$ test episodes spanning from episode $7,500$ to $8,000$. The figure presents box plots for each assessed technique and is divided into three sections. The leftmost section shows the average reward achieved by agent in an episode, the middle section shows the average number of prey caught, and the rightmost section shows the probability of winning an episode.

Both~\ac{ocmas} and \ac{efontl} show an improvement over \textit{no-transfer} method across all the three metrics. On the contrary, \ac{rcmp} agents score a lower average reward in the test episodes while having a winning probability and an average catch aligned with the other two \ac{tl} techniques. The negative reward is given by the different strategy learnt by \ac{rcmp} agents. In fact, while \ac{efontl} and \ac{ocmas} agents learnt to pursue prey across the grid, \ac{rcmp} agents prioritised the catch action which leads to higher cost when not successful, and sometimes results in incorrectly catching other team's prey. Finally, these results confirm that QMIX agents are penalised, in this benchmark environment, by the additional step to evaluate the joint actions.

\subsubsection{\acl{hfo}}

The third environment used to benchmark \ac{efontl} against the baselines is \ac{hfo}, detailed in Section~\ref{ss:sim_hfo}. In \ac{hfo} the offense players, powered by \ac{rl} agents, are those whose performance we are evaluating. The defense player and the goalkeeper are played by \textit{HELIOS} baseline~\cite{hausknecht2016half}.  

In \ac{hfo}, \ac{efontl} target agents are enabled to receive $100$ selected interactions from episode $2,400$ to episode $20,000$. Transfer occurs every $400$ episodes and the interactions on target side are filtered by \ac{hdc} while the source of transfer is selected by \ac{u}.

All the \ac{rl} agents are based on the \ac{paddpg} model presented in Section~\ref{sec:paddpg}. \ac{efontl} is compared against a \textit{no-transfer} baseline, where agents act independently, \ac{maddpg}, and the expert action-advice based baseline~\ac{rcmp}. The \ac{tl} method~\ac{ocmas} is not implemented due to the client-server architecture of \ac{hfo}, which delays the training of the agents due to continuous communication to establish whether to provide an advice or not. \ac{ocmas}, fully discussed in Section~\ref{sec:baselines}, requires each agent to forward the visited state to all the other agents.  
As a consequence, the maximum number of intra-agent communications, calculated as the product of the number of agents ($3$), the number of episodes ($20,000$), and the maximum timestep ($500$), totals to $30,000,000$.

To evaluate the performance of the offense team, which is controlled by \ac{rl} agents, this thesis assesses the goal probability recorded within a specific number of episodes. The probability is determined by counting the number of goals scored in a fixed interval.  Whenever the offense team scores a goal, the episode terminates. During the learning phase, the goal probability is estimated using the most recent $100$ episodes. On the other hand, the goal probability during evaluation is estimated over a span of $500$ episodes, during which players take deterministic actions and no transfer is allowed. The deterministic evaluation is conducted every $500$ episodes, where agents leverage their latest knowledge.

\begin{figure}[htb!]
	\centering
	\includegraphics[width=1\columnwidth]{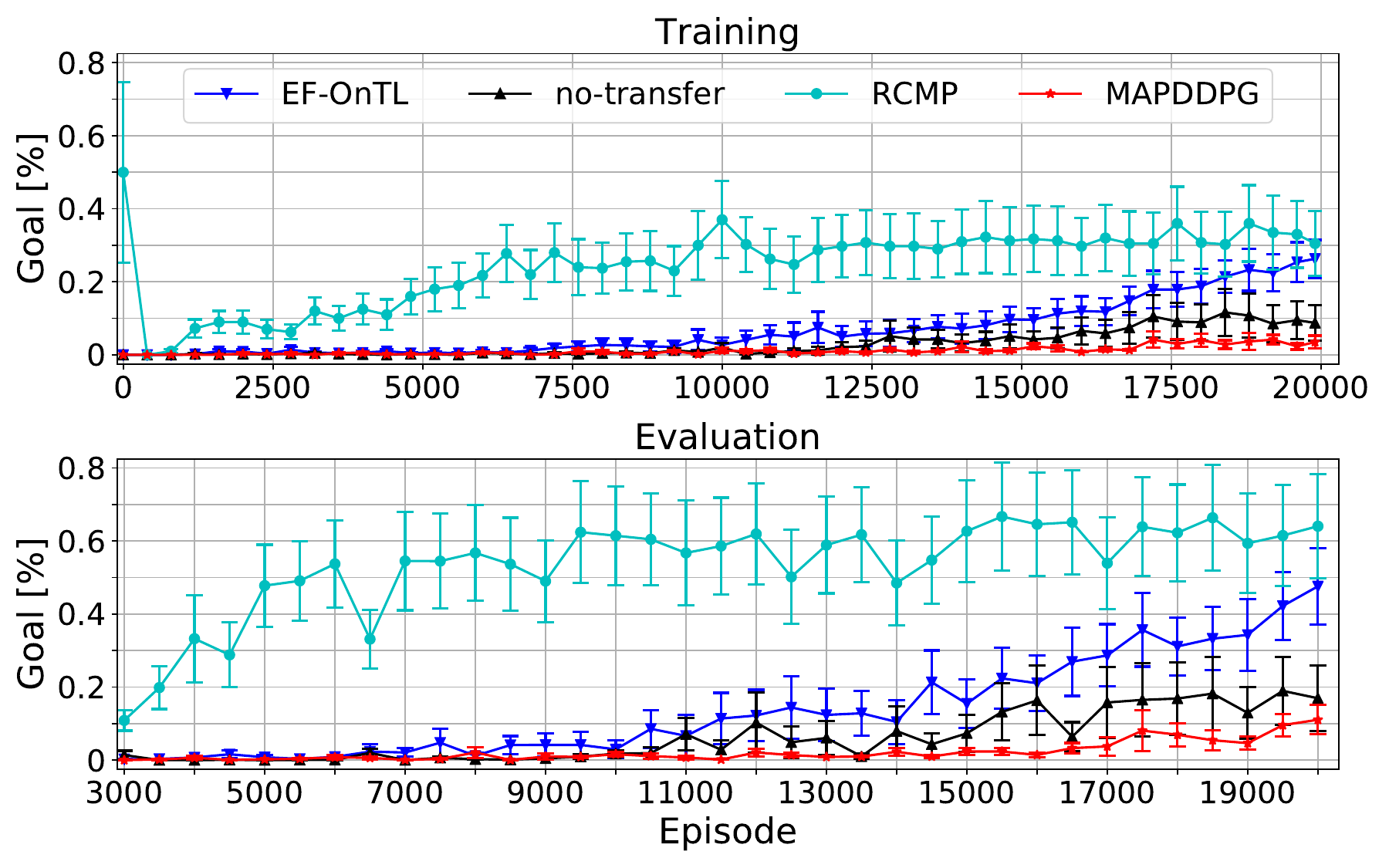}
	\caption{\ac{efontl}(\ac{hdc}, budget: $100$ and $SS: \overline U$)  compared against \textit{no-transfer}, \ac{rcmp} and multi-agent \ac{paddpg} in \ac{hfo}.}
	\label{fig:efontl_vs_baseline-HFO}
\end{figure}

Figure~\ref{fig:efontl_vs_baseline-HFO} reports the results. The $x$-axis denotes the training step and the $y$-axis denotes the goal probability, in the range of $[0,1]$. During evaluation, the training step of the $x$-axis indicates the version of the policy used by each agent. The results are organised into two graphs, the upper row reports the goal probability during training, while the lower row shows the goal probability on the test set.

Similarly to what we observed previously for the other benchmarks, \ac{efontl} successfully improves the performance of the offense team compared to the \textit{no-transfer} method. Furthermore, the positive effect of~\ac{efontl} becomes noticeable during the second-half of the training phase. 
Sharing a restrained number of selected experiences across agents driven by~\ac{efontl}, it enables the offense team to achieve up to $2.5$ times higher scores compared to \textit{no-transfer} agents. In particular, from episode $14,000$, the goal probability in training and testing for \ac{efontl} diverges from \textit{no-transfer} method and \ac{maddpg}, which requires nearly the double amount of episodes to achieve comparable performance. 

In \ac{hfo} the exploitation of expert knowledge plays a crucial role and \ac{rcmp} achieves outstanding results since an early stage of training. In training, during the first $500$ episodes, the target agents receive substantial advice being fully driven by the experts. The advice has a positive impact even on evaluation as agents powered by~\ac{rcmp} have a goal probability of $50\%$ from episode $7,000$ onward. Certainly, the exploitation of external expertise is unmatchable by~\ac{efontl} method with the transfer setting used within this study.

\begin{figure}[htb!]
	\centering
	\includegraphics[width=.8\columnwidth]{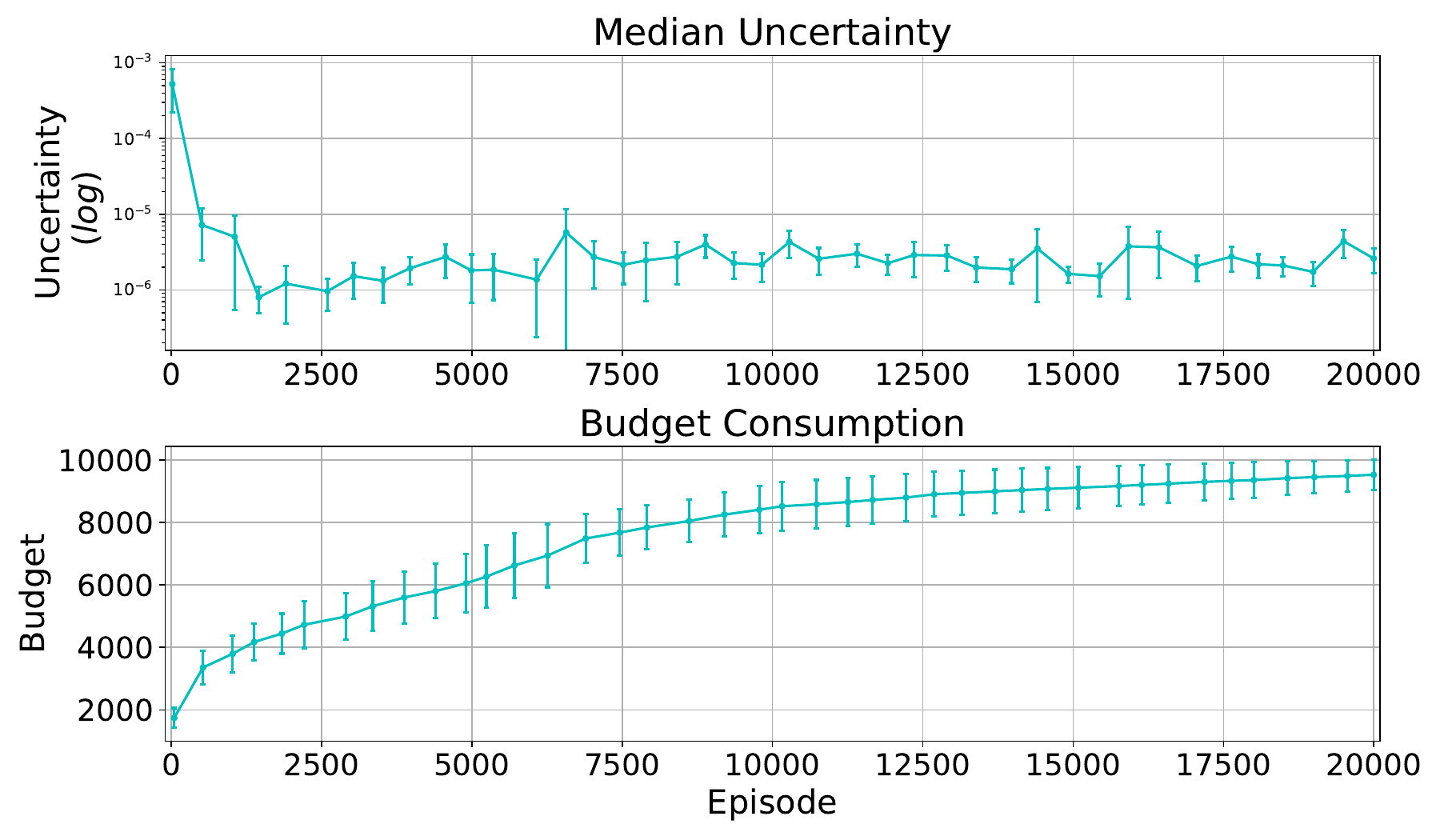}
	\caption{\ac{rcmp} analysis in \ac{hfo} with uncertainty trend and budget consumption.}
	\label{fig:rcmp-budget-consumption_HFO}
\end{figure}

Figure~\ref{fig:rcmp-budget-consumption_HFO} reports the average budget used by \ac{rcmp} powered agents, with a threshold set to $10^{-4}$, along their median uncertainties registered throughout the training phase.

The ensemble architecture used to approximate uncertainty for \ac{paddpg} is on the critic network which outputs a single value and as such, it is impossible to predict a range and hence to normalise the estimated uncertainty as done previously in \textit{Cart-Pole} and \ac{pp}. In fact, the \ac{rcmp} uncertainty curve decreases sharply within the first hundreds episodes and then ranges within a narrow interval making it not trivial to find the right threshold given the tight median uncertainty interval.
In fact, around $60\%$ of times, the budget is used up to  $10,000$, by the end of the simulation.

To conclude, even though \ac{rcmp} demonstrates remarkable performance from the early stages of training, by the end of the $20,000$ episode,  the performance gap between \ac{rcmp} and \ac{efontl} is drastically reduced. The growing trend of~\ac{efontl} suggests that eventually it could potentially catch up to the performance of \ac{rcmp}. However, it is evident that when an expert is available, \ac{rcmp} remains the preferable choice. On the other hand, \ac{efontl} remains a viable alternative in scenarios where an expert is not available up front.

\subsubsection{\acl{rs-sumo}}\label{ss:eval_efontl_in_rs-sumo}
The fourth and last environment used to benchmark \ac{efontl} against the baselines is \ac{rs-sumo} with the multiple brains configuration, detailed in Section~\ref{ss:multi-brain-definition}. In this version of the environment, the four \ac{rl}-agents control a non-overlapping zone of the Manhattan area. Consequently, each vehicle uses the policy provided by the brain of the zone where it is located to select the appropriate action. 

In the  \ac{rs-sumo} environment, as detailed in Section~\ref{ss:preliminary-envs-rs-sumo}, the training consists of $10$ epochs, with each epoch composed by $10$ episodes. In \ac{efontl}, the sharing occurs in total $17$ times. In each transfer step, target agents filter up to $100$ selected interactions. The transfer is delayed until the beginning of the second epoch and it is enabled every $5$ episodes thereafter.

Although within the environment there are multiple agents that interact with the same copy of the environment and learns concurrently, no \ac{marl} scenario has been evaluated in this setup as there is no direct collaboration between the agents. However, we further examine different settings of the environment in Section~\ref{ss:efotnl-rs-sumo-expanded}.

To assess the performance of the $200$-vehicle fleet, we analyse several commonly used measures adopted by related work~\cite{alabbasi2019deeppool, Alonso-Mora2017, gueriau2018samod, gueriau2020shared, yang2019}. These include the distribution of waiting times for passengers before they board a vehicle~(\textit{waiting time}), the total number of successfully served requests~(\textit{satisfied requests}), and specific vehicle-related metrics. These vehicle metrics report the distribution of passengers across the fleet~(\textit{passengers distribution}),  the cumulative distance travelled by the vehicles in kilometres~(\textit{mileage}), and the proportion of distance travelled by vehicles while unoccupied (\textit{\% distance travelled empty}) or engaged in ride-sharing (\textit{\% distance travelled in RS}).

\begin{figure}[htb!]
	\centering
	\includegraphics[width=1\columnwidth]{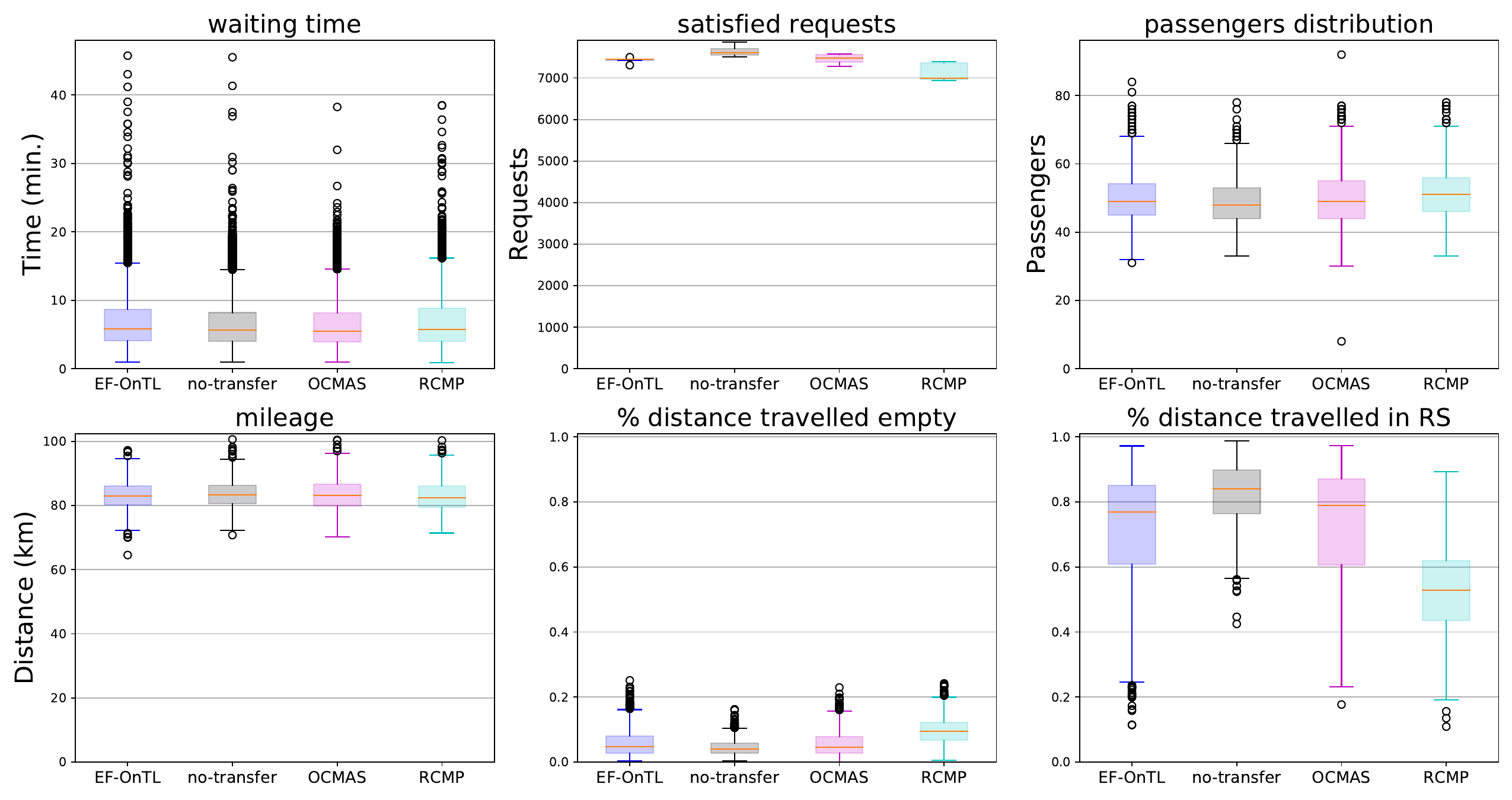}
	\caption{\ac{efontl} compared versus \textit{no-transfer}, \ac{ocmas} and \ac{rcmp} in the multi-brains \ac{rs-sumo} benchmark.}
	\label{fig:rs-sumo_multibrain}
\end{figure}

The performance of \ac{efontl} is compared against both \textit{no-transfer} and the \ac{tl} baselines, \ac{ocmas} and \ac{rcmp}. The outcomes across the evaluated metrics are visualised using box-plots, as presented in Figure~\ref{fig:rs-sumo_multibrain}.

Enabling \ac{tl} in the multi-brain \ac{rs-sumo} environment results in a slight decrease in the total number of satisfied requests by the fleet compared to the \textit{no-transfer} approach, although, the difference in the number of satisfied requests is negligible. The \ac{efontl} approach appears to be more consistent across different simulations in the total number of served request. While \ac{efontl} exhibits comparable waiting times and passenger distribution to the \textit{no-transfer} approach, the mileage recorded by vehicles is slightly reduced. This difference is likely attributable to the lower number of satisfied requests, which could suggest a reduced inclination towards ride-sharing. This affirmation is further supported by the increased distance travelled by the vehicles while being empty.

Based on the observed results from the evaluated \ac{tl}-enabled techniques, it can be concluded that \ac{tl} induces to negative transfer in this benchmark environment. However, \ac{efontl} has a lower negative impact due to the transfer of experiences compared to the other \ac{tl} baselines. Interestingly, both \ac{efontl} and \ac{ocmas} exhibit a similar performance profile in this study. In contrast, despite relying on expert agents trained within the same zone, \ac{rcmp} exhibited the most pronounced negative transfer effect. However, it is important to note that the difference in performance between these approaches is relatively moderate, and this evaluation did not consider factors like convergence time or the time required to learn an effective policy.

\subsubsection{Conclusion}



This section presented an evaluation of \ac{efontl} against the identified baselines in four different environments: Cart-Pole, \acf{pp}, \acf{hfo} and, the multi-brain implementation of \ac{rs-sumo}.


Based on the evaluation objectives \textit{O1} and \textit{O2}, the presented evaluation shows the efficacy of \ac{efontl} in contrast to \textit{no-transfer}, where agents do not share any information. Specifically, in the Cart-Pole and \ac{hfo} environments, \ac{efontl} displays a quicker convergence time, with the improvement being milder in Cart-Pole and more pronounced in \ac{hfo}. 
Additionally, \ac{efontl} agents exhibit superior performance in \ac{hfo} and \ac{pp}. While the disparity in \ac{hfo} is considerable, the enhancement is less pronounced in \ac{pp}.

With regard to the evaluation objective~\textit{O3}, \ac{efontl} has shown comparable transfer effect to those of \ac{ocmas} across the Cart-Pole, \ac{pp}, and multi-brains \ac{rs-sumo} benchmark environments. However, \ac{ocmas} requires ongoing communication between agents to assess their confidence in visited states, such process is essential for determining whether an advice is necessary for a particular agent. This requirement makes \ac{ocmas} inapplicable to the \ac{hfo} environment.

Notably, in the Cart-Pole, \ac{pp}, and multi-brains \ac{rs-sumo} scenarios, \ac{efontl} eventually surpasses the expert-based \ac{tl} baseline, \ac{rcmp}. In the case of \ac{hfo}, \ac{efontl} exhibits an upward trajectory, suggesting the potential to eventually match the performance of \ac{rcmp}.

Among the implemented \ac{tl} techniques, \ac{rcmp} demonstrates a significant jump-start in performance due to the use of expert teachers. However, as time progresses, the influence of the expert teacher tends to restrict the performance potential of the target agents. This phenomenon is particularly evident in the Cart-Pole and \ac{pp} environments, where the rewards curve shows an initial surge but then plateau over time. 
In the Cart-Pole, \ac{pp} and multi-brains \ac{rs-sumo} environments, \ac{efontl} has eventually outperformed the expert-based \ac{tl} baseline, \ac{rcmp}. In the case of \ac{hfo}, \ac{efontl} has demonstrated an uprising trend, suggesting the potential to eventually match the performance of~\ac{rcmp}.

The low performance of \ac{rcmp} has to be attributed to the threshold used for receiving advice on a target agent. Interestingly, \ac{rcmp} demonstrated exceptional performance in the \ac{hfo} environments, where the uncertainty estimator model is defined on the critic. In contrast,  when the ensemble used to approximate uncertainty is based on the network used for action selection, the uncertainty seems to be limited to certain fixed values, especially for deterministic actions. An in-depth analysis on the threshold sensitivity of \ac{rcmp} is reported in the Appendix in Section~\ref{app:rcmp_analysis}.

In the \ac{pp} and \ac{hfo} environments,  a \ac{marl} algorithm has been implemented for the evaluation objective~$O9$ and compare its performance against \ac{efontl}. Specifically, the joint state-action value is learnt through QMIX in the \ac{pp} environment and through~\ac{maddpg} in the \ac{hfo} environment. Generally, \ac{marl} has shown lower performance across the team due to the learning of joint action values, which introduces a delay. As a result, in the \ac{hfo} environment, \ac{maddpg} requires approximately twice the number of episodes to achieve a similar performance as~\ac{efontl}.

To conclude, regarding objective~\textit{O7}, \ac{efontl} has proven successful in three of the four evaluated environments, achieving a positive transfer outcome. We have observed a trend in the improvement given by the complexity of the task undertaken. However, in the~\ac{rs-sumo} environment, both \ac{efontl} and the other \ac{tl} baselines have resulted in negative transfer.

\subsection{\ac{efontl} - Transfer Criteria Evaluation}\label{ss:eval_efontl_transfer_criteria}

The studies presented in this section address the research questions \textbf{RQ2} and \textbf{RQ3}, introduced earlier in Section~\ref{sec:research_questions}. Therefore, this evaluation aims to empirically demonstrate how different selection criteria may impact the performance on \ac{efontl} algorithm.

In the following study a total of $18$~\ac{efontl} settings are compared by varying different transfer parameters within the algorithm:
\begin{itemize}
	\item \textit{\acf{ss}} $-$ to answer \textbf{RQ2}, which relates to finding a valid strategy for selecting a suitable agent to be used as a source of transfer, this section shows experiments to evaluate the evaluation objective~\textit{O4}. 
	\ac{ss} is presented in Section~\ref{ss:SourceSelection} and can vary between \textit{\ac{bp}} and \textit{\ac{u}}. \ac{bp} selects the agent that achieved better performance, measured by higher average reward, within the latest episodes, while \ac{u} selects the agent with lower average uncertainty within its transfer buffer.

	\item \textit{\acf{tcs}} $-$ to answer \textbf{RQ3}, which addresses the establishing criteria that should be used by an agent to filter the incoming knowledge to improve its performance, this section shows experiments to evaluate~\textit{O5}.	 \ac{tcs} is presented in Section~\ref{ss:TransferMethods} and can vary between \ac{rdc}, \ac{lec} and \ac{hdc}. \ac{rdc} and \ac{hdc} use only the epistemic uncertainty as filtering criteria while \ac{lec} takes into account the surprising effect of a tuple on the target side.
	\item \textit{Transfer Size} $-$ to evaluate \textit{O6}, the evaluation focuses on determining whether and to what extent the quantity of interactions transferred through~\ac{efontl} impacts the performance of the system.
\end{itemize}

The combinations of the aforementioned parameters results into $18$ different~\ac{efontl} configurations. These $18$ \ac{efontl} scenarios are tested across two benchmark environments, Cart-Pole and \ac{pp}. For each setting of \ac{efontl}, a total of $20$ independent runs are collected to obtain a robust evaluation. In the subsequent sections, we first analyse the impact of the transfer parameters in the Cart-Pole environment and then in the~\ac{pp} environment.

\subsubsection{Cart-Pole}

\begin{figure}[h!]
	\centering
	\includegraphics[width=.9\columnwidth]{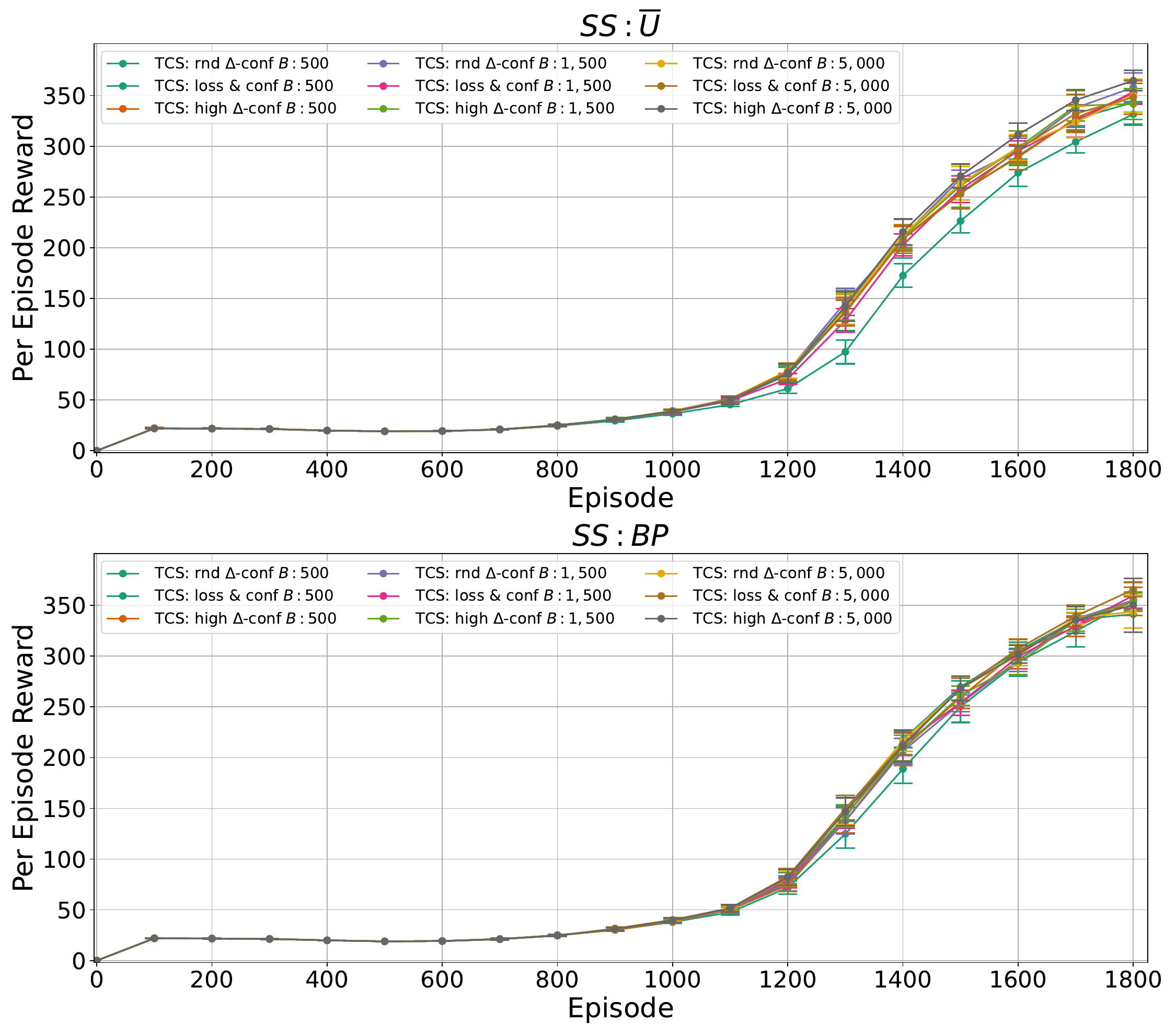}
	\caption{\ac{efontl} learning curves in Cart-Pole with different parameters. In the top chart the source of transfer is selected by \ac{u}, the agent with lower average uncertainty within the transfer buffer while, in the bottom one the source is selected by \ac{bp}, the performance achieved by the agent within the latest episodes.}
	\label{fig:efontl-CartPole}
\end{figure}

The evaluation in the Cart-Pole environment analyses the cumulative reward achieved by agents across training episodes. Figure~\ref{fig:efontl-CartPole} shows the episode count on the $x$-axis and the cumulative episode reward on the $y$-axis. To facilitate the evaluation of \textit{O4}, the curves are organised into two different graphs based on the~\ac{ss} employed within \ac{efontl} to select the source of transfer. The graph in the top row illustrates \ac{efontl} scenarios where the source of transfer is selected using~\ac{u}.Conversely, the graph in the lower row shows scenarios using~\ac{bp} for selecting the source of transfer.

\begin{figure}[htb!]
	\centering
	\begin{subfigure}[b]{.5\columnwidth}
			\includegraphics[width=1\columnwidth]{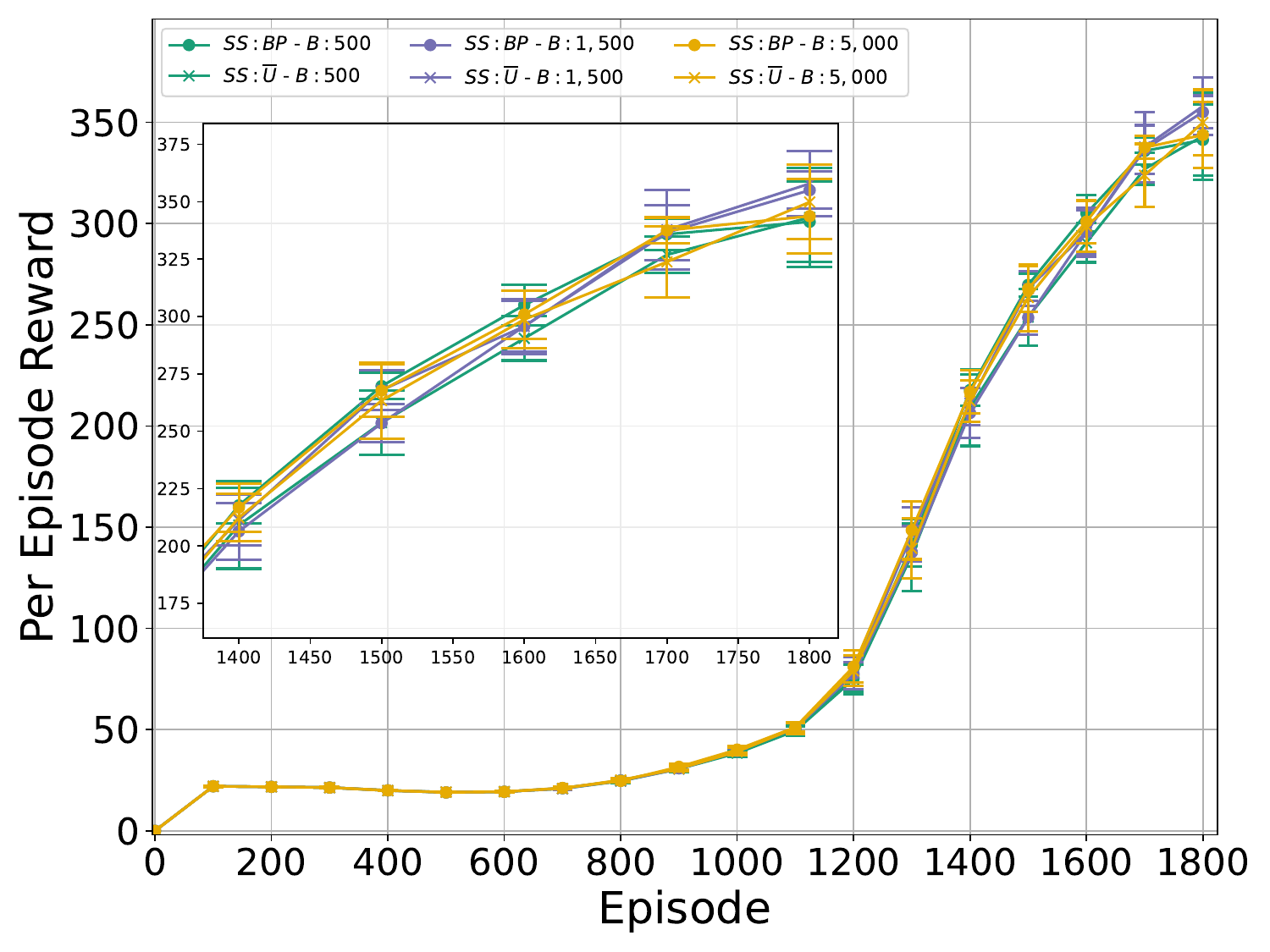}
			\caption{\ac{rdc}}
	\end{subfigure}~
	\begin{subfigure}[b]{.5\columnwidth}
		\includegraphics[width=1\columnwidth]{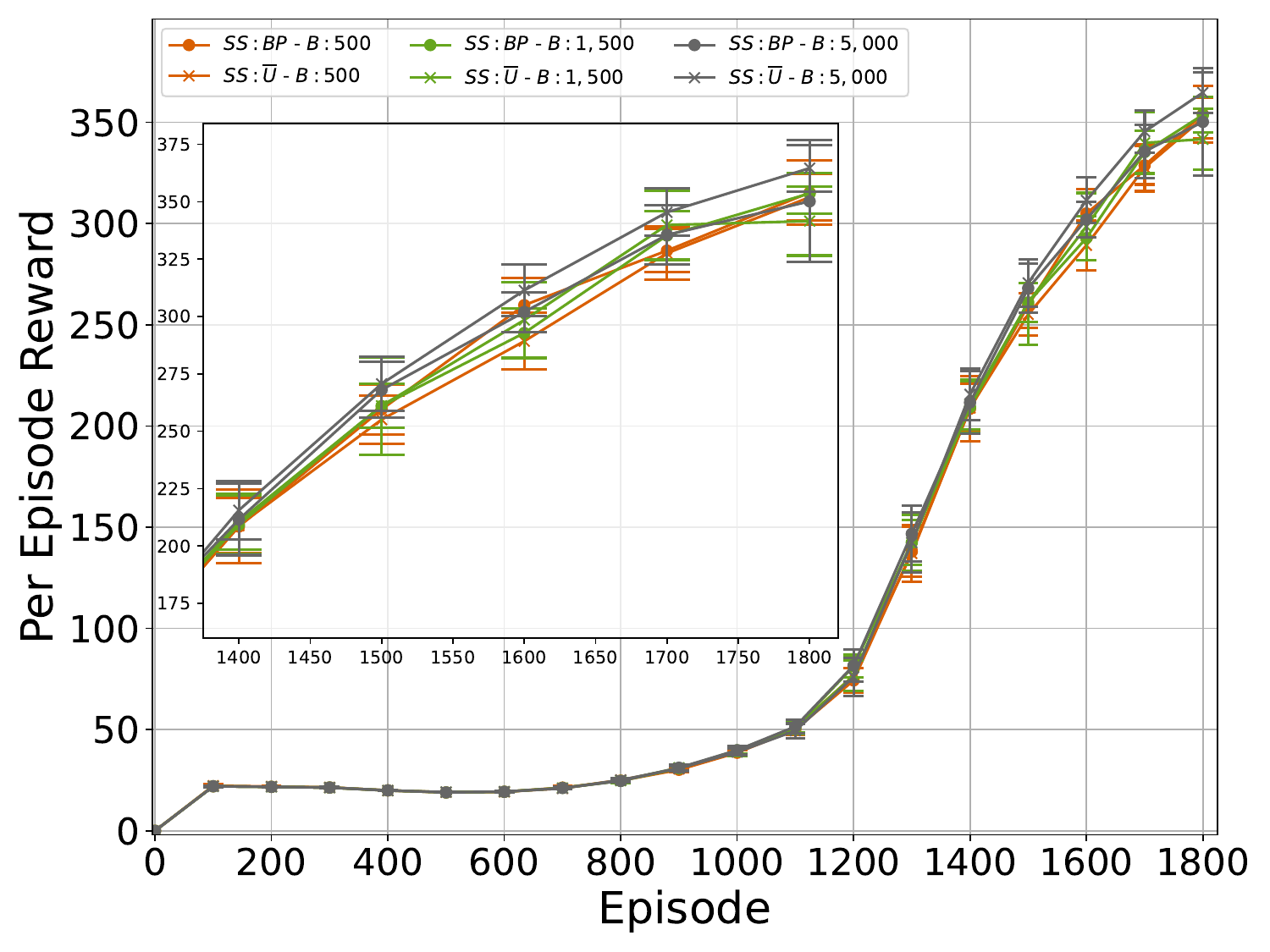}
		\caption{\ac{hdc}}
	\end{subfigure}
	\begin{subfigure}[b]{.5\columnwidth}
		\includegraphics[width=1\columnwidth]{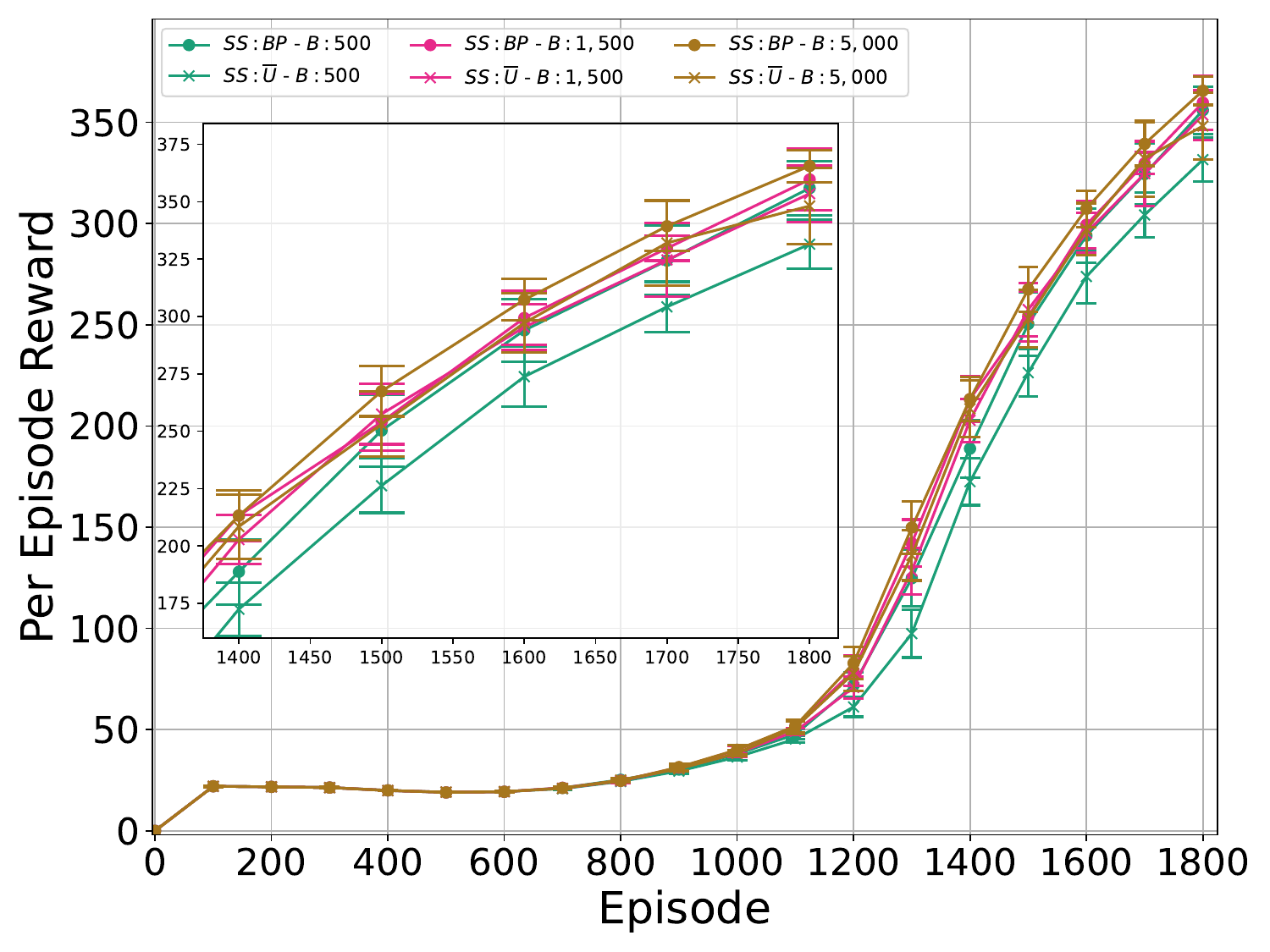}
		\caption{\ac{lec}}
	\end{subfigure}
	\caption{\ac{efontl} learning curves organised by~\ac{tcs} used by a target agent to filter incoming knowledge in Cart-Pole.}
	\label{fig:efontl-CartPole-tcs}
\end{figure}

\begin{figure}[htb!]
	\centering
	\begin{subfigure}[b]{.5\columnwidth}
		\includegraphics[width=1\columnwidth]{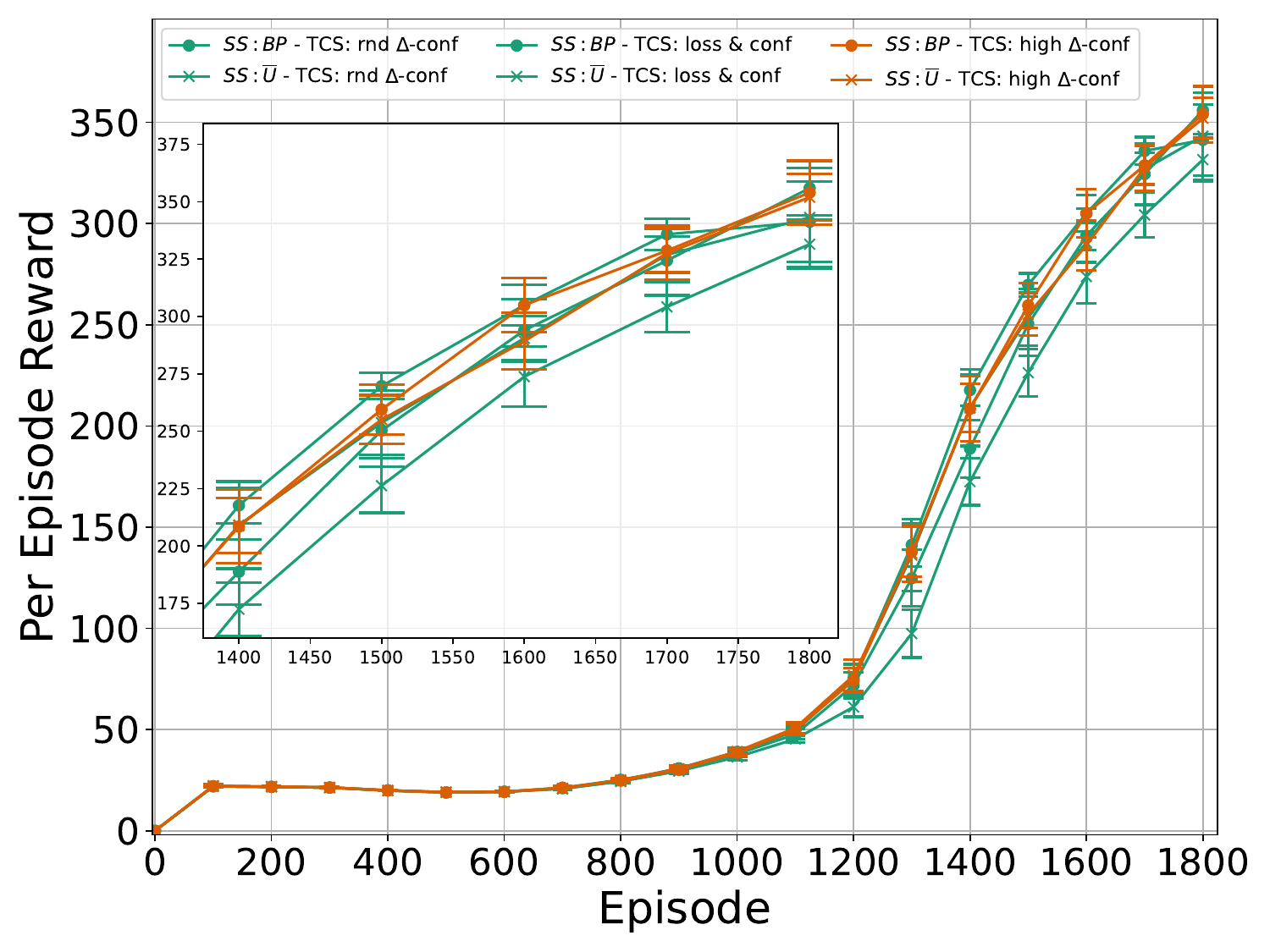}
		\caption{$B=500$}
	\end{subfigure}~
	\begin{subfigure}[b]{.5\columnwidth}
		\includegraphics[width=1\columnwidth]{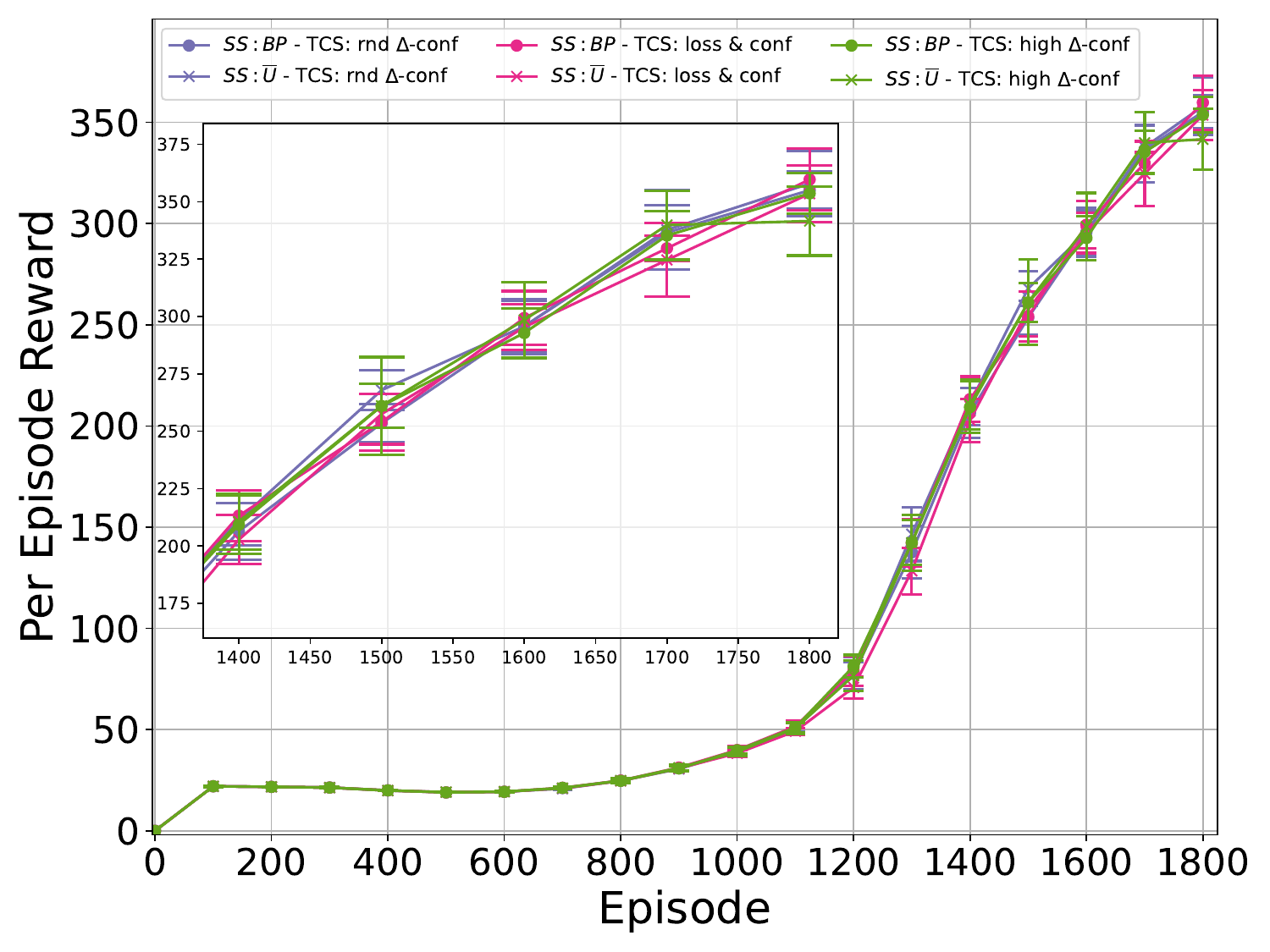}
		\caption{$B=1,500$}
	\end{subfigure}
	\begin{subfigure}[b]{.5\columnwidth}
		\includegraphics[width=1\columnwidth]{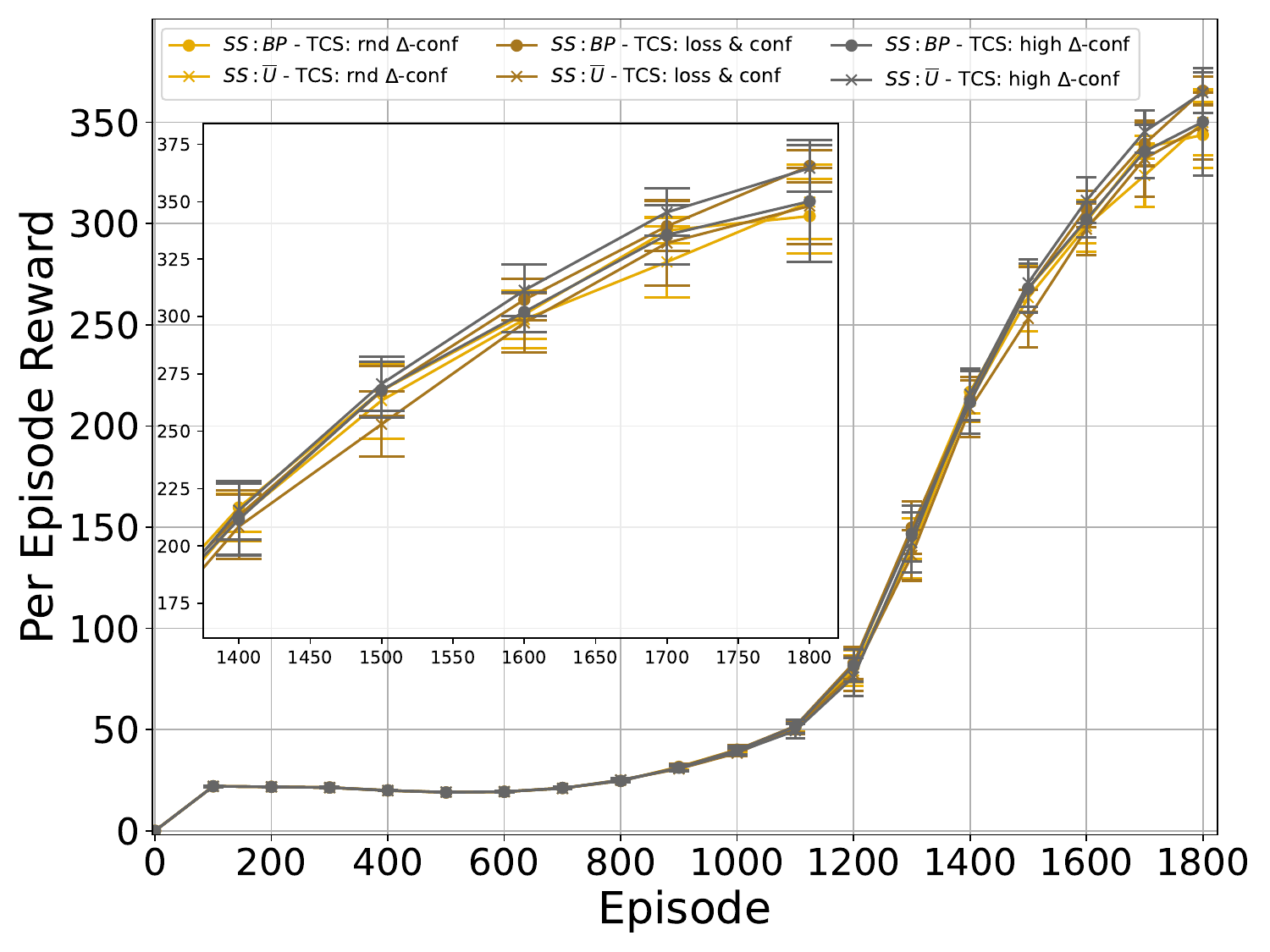}
		\caption{$B=5000$}
	\end{subfigure}
	\caption{\ac{efontl} learning curves organised by the transfer budget~$B$ in Cart-Pole.}
	\label{fig:efontl-CartPole-tbsize}
\end{figure}

To provide a comprehensive overview and to facilitate the evaluation of \textit{O4}, \textit{O5}, and \textit{O6}, the results from the Cart-Pole benchmark environment, reported in Figure~\ref{fig:efontl-CartPole}, are presented also in a further comparative manner. This includes analysing the curves according to the~\ac{tcs} used to filter the incoming experiences on a target side, as depicted in Figure~\ref{fig:efontl-CartPole-tcs}, and analysing the curves according to the employed transfer budget~$B$, as shown in Figure~\ref{fig:efontl-CartPole-tbsize}.

Across all $18$~\ac{efontl} scenarios, the observed results demonstrate agents achieving comparable performance irrespective of the transfer settings. Although the learning curves exhibit substantial similarity, slight differences emerge from episode $1,200$ onward, by which point agents have already learnt to balance the pole for a few consecutive steps.

Firstly, in Figure~\ref{fig:efontl-CartPole-tcs}a, the impact of selecting incoming knowledge by \ac{rdc} demonstrates a consistent influence on the target policy regardless of the chosen \ac{ss} and $B$. In fact, \ac{rdc} is the only \ac{tcs} method that does not benefit from an increased budget of $5,000$. Conversely, \ac{efontl} agents exhibit superior performance with~\ac{hdc}, as depicted in Figure~\ref{fig:efontl-CartPole-tcs}b, and \ac{lec}, as seen in Figure~\ref{fig:efontl-CartPole-tcs}c, when $B$ set to $5,000$.

Despite the results suggesting that filtering for $5,000$ agent-environment interactions on the target side in each transfer step leads to an overall performance improvement, an intriguing finding emerges when the lowest budget of $500$ is utilised. Specifically, the \ac{hdc} approach demonstrates lower volatility in returns when considering both sources of transfer techniques, \ac{bp} and \ac{u}, as depicted in Figure~\ref{fig:efontl-CartPole-tbsize}a. This quality is highlighted in contrast to \ac{lec}, even though the leading scenario within the most recent $500$ episodes fluctuates between the two methods.

In conclusion, none of the $18$ scenarios prevails significantly over the others. Among these scenarios, \ac{rdc} generally exhibit the mildest impact on the overall system performance. While \ac{hdc} and \ac{lec} demonstrate comparable performance using different source selection techniques, the \ac{efontl} configuration that emerges as the preferable choice is the \ac{ss}: \ac{u}, $B$: $5,000$, and selection of tuples via \ac{hdc}. This particular configuration consistently yields higher rewards from episode $1,400$ onward, as illustrated in Figure~\ref{fig:efontl-CartPole-tbsize}c.

\subsubsection{\acf{pp}}

\begin{figure}[!htb]
	\centering
		\includegraphics[width=.9\columnwidth]{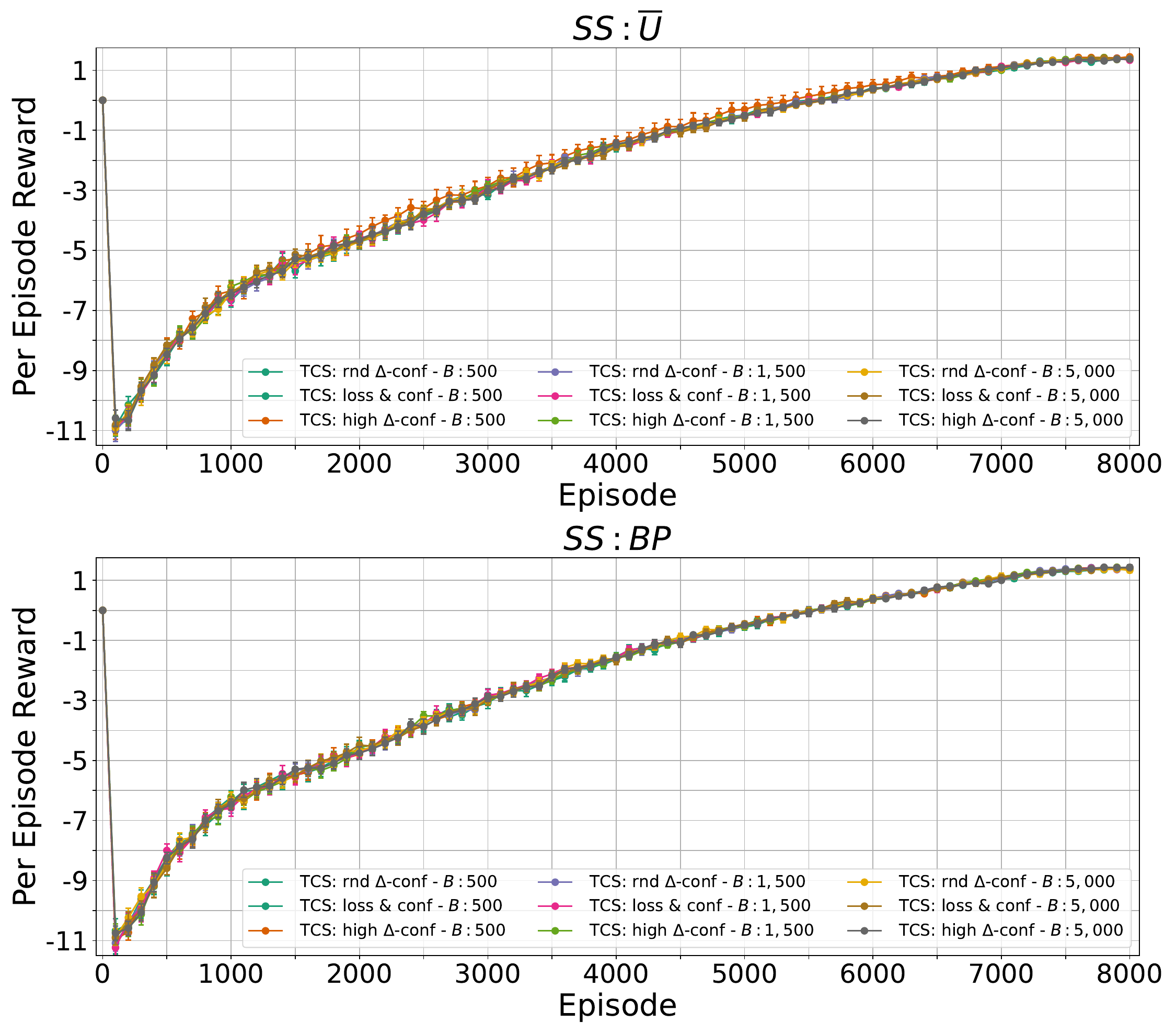}
	\caption{\ac{efontl} learning curves in \ac{pp} with different parameters. In the top chart the source of transfer is selected by $\overline U$, while, in the bottom chart the source is selected by BP.}
	\label{fig:efontl-comparison-pp}
\end{figure}

The evaluation in the \ac{pp} environment analyses both the cumulative reward achieved by agents across the training episodes as well as additional metrics specific to the environment. The overall reward trend observed in \ac{pp} is similar to what observed in the Cart-Pole environment, where despite the different transfer settings all configurations followed a similar trajectory. Therefore, to maintain the flow of the section, we present the results once, based on the \ac{ss} used for selecting the source of transfer. Additional graphs are provided in the Appendix at Section~\ref{app:expanded_pp_res}. Figure~\ref{fig:efontl-comparison-pp} displays the learning curves, the $x$-axis denotes the episode count and the $y$-axis denotes the cumulated reward. These learning curves overlap extensively, making it challenging to discern significant differences.

\begin{figure}[!htb]
	\centering
	\begin{subfigure}[b]{\columnwidth}
		\centering
		\includegraphics[width=.75\columnwidth]{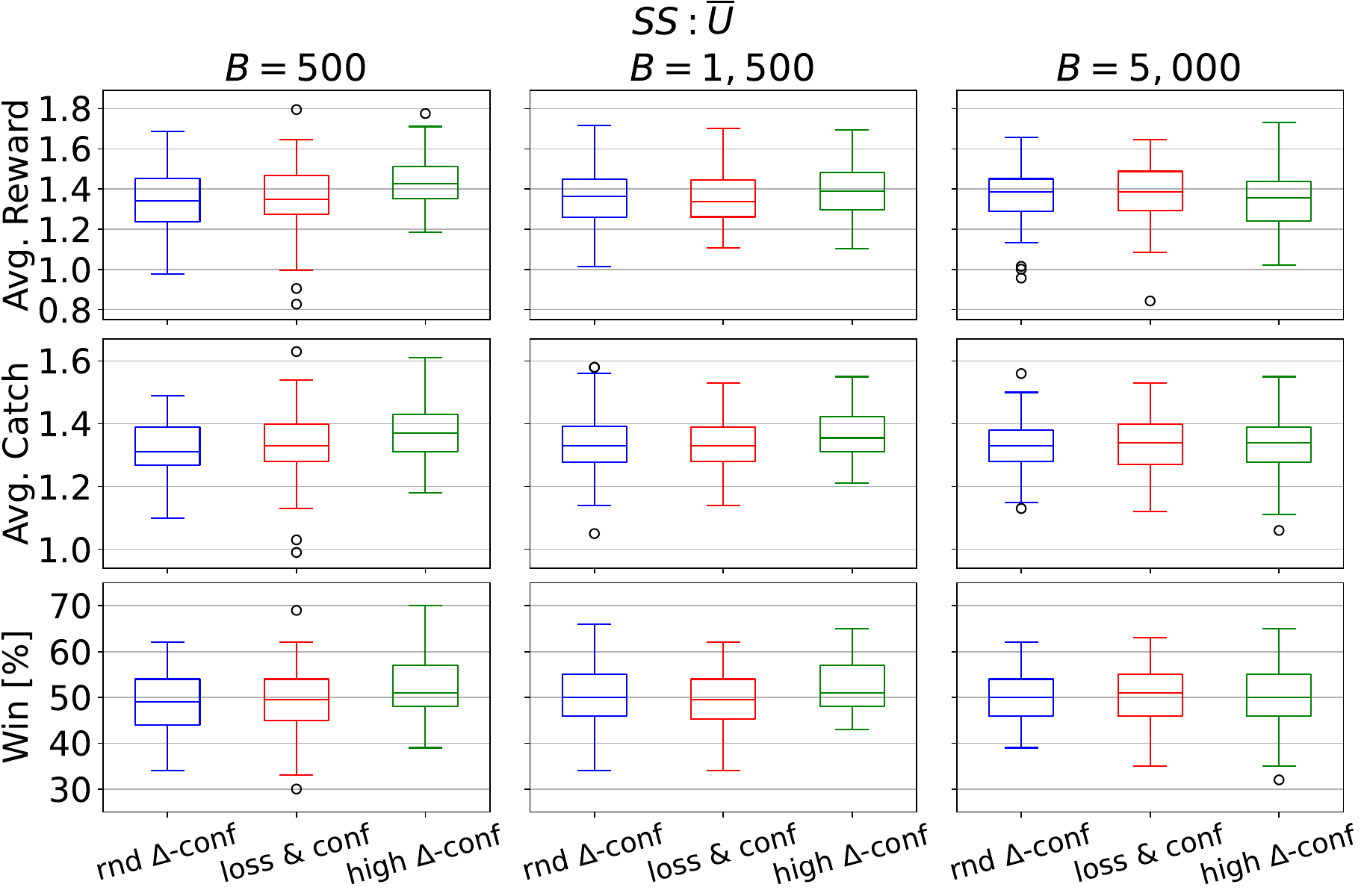}
	\end{subfigure}
	\begin{subfigure}[b]{\columnwidth}
		\centering
		\includegraphics[width=.75\columnwidth]{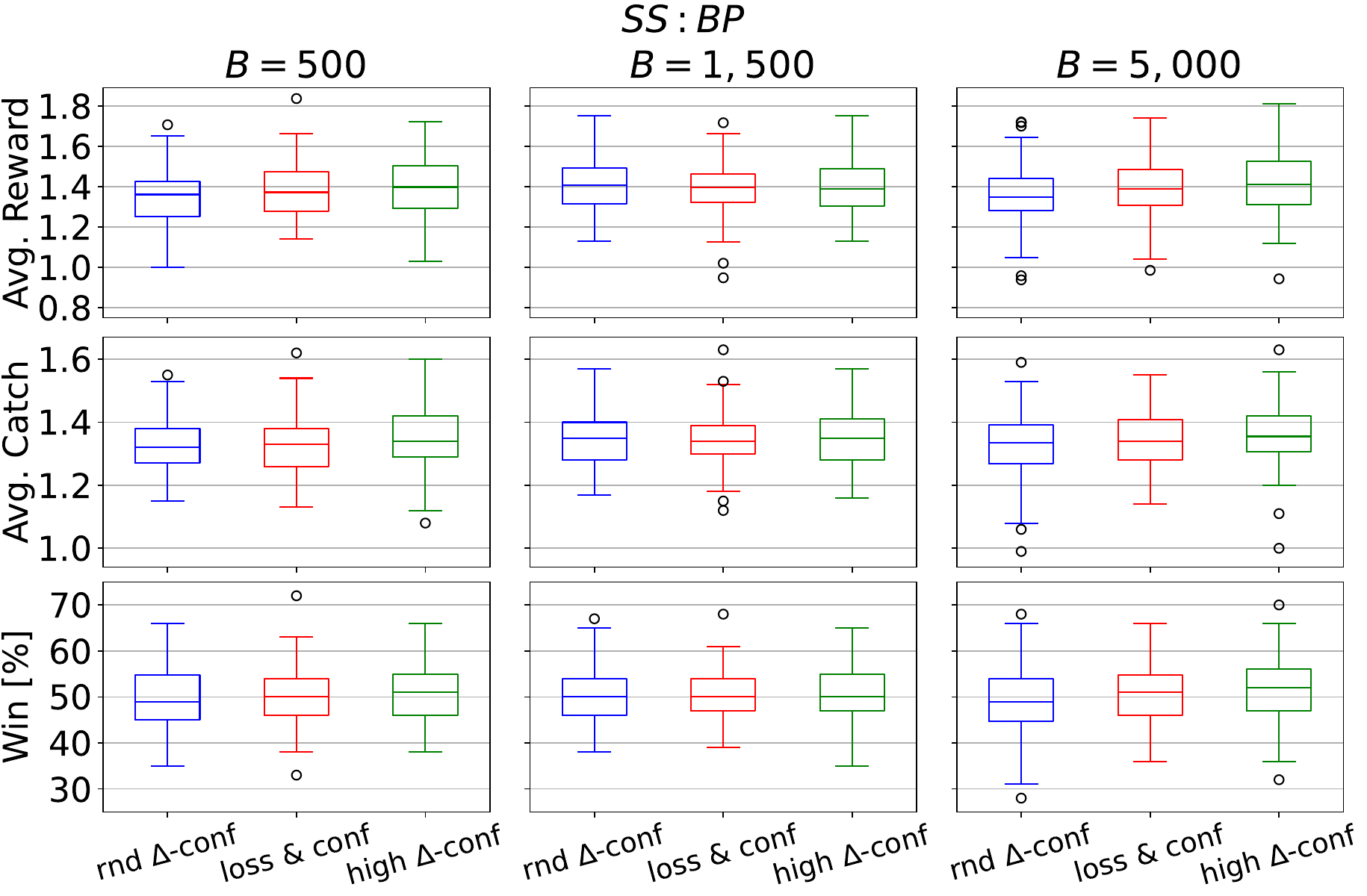}
	\end{subfigure}
	\caption{Evaluation metrics on the final $500$ episodes in the \ac{pp} environment across the $18$~\ac{efontl} configurations.}
	\label{fig:efontl-test-results-pp}
\end{figure}

On top of reward, \ac{pp} offers additional metrics that help the comprehension of the learnt behaviour.  These metrics, number of successful prey caught and the win probability, are shown in Figure~\ref{fig:efontl-test-results-pp}, which presents a comprehensive view of $500$ test episodes, $7,500$ to $8,000$. The graphs are divided into two distinct blocks. The upper block showcases \ac{efontl} configurations based on \ac{u} as the source selection criteria, while the lower block depicts \ac{efontl} configurations utilising \ac{bp} for selecting the source of transfer. Each block contains rows representing different evaluation metrics. The top row displays the average agent reward, the middle row represents the number of prey caught by the team, and the bottom row reports the win probability for the team.  The configurations utilising a transfer budget $B$ of $500$ are reported in the leftmost column, those using $B$ of $1,500$ are placed in the middle column and finally, those with a budget of $5,000$ are placed in the rightmost column.

Overall the criteria used to select the tuple clearly impacts the performance in a similar trend. \ac{hdc} allows for the better performance while \ac{lec} is slightly below and finally~\ac{rdc} is generally the lowest. The only scenarios in which such trends are not confirmed is with a budget $B$ of $5,000$ and \ac{u} as \ac{ss}. With such settings, \ac{lec} reports the highest performance across all the $3$ metrics.

Contrary to what we observed in Cart-Pole, where processing more experiences  resulted in improved performance for the target agents, in \ac{pp} transferring $5,000$ tuples within each transfer step leads to an overall decrease in performance. For instance, when using \ac{tcs} as \ac{hdc} and selecting the source of transfer using \ac{u}, the impact of transfer diminishes as the budget $B$ increases.

We believe that an increased budget could potentially lead to a deterioration in performance, similar to what we observed in the preliminary results with an elevated transfer budget. Essentially, the performance decline could be given by the dynamics of the environment, which force a learning agent to visit states with minimal or no information, regardless of the action taken. The probability of encountering these situation is reduced as the agent learns to explore the surroundings more effectively. Consequently, selecting the source of transfer based on \ac{u} is prone to selecting an agent whose transfer buffer mainly consists of interactions from these low information states. In this context, transferring a substantial volume of interactions could result into a deterioration of the system performance.

To conclude, in the \ac{pp} benchmark environment, enabling dynamic selection of the source agent using \ac{u} consistently results in improved performance for the transfer enabled team. Moreover, choosing tuples based on the difference between source and target uncertainties, denoted as \ac{hdc}, consistently leads to superior outcomes.

\subsubsection{Conclusion of \ac{efontl} - Transfer Criteria Evaluation}

This section has presented a targeted study on \ac{efontl}, aiming to evaluate the objectives \textit{O4}, \textit{O5}, and \textit{O6}. These objectives have been defined to guide the assessment of different transfer configurations within two benchmark environments: Cart-Pole and \acf{pp}. In total, $18$ distinct configurations of~\ac{efontl} have been tested. 

\textit{O4} addresses \textbf{RQ2} and it evaluates whether and to what extent \ac{efontl} is influenced by the \acf{ss}. Similarly, \textit{O5} addresses \textbf{RQ3} and it evaluates whether and to what degree \acf{tcs} influences \ac{efontl} transfer effectiveness. Finally, \textit{O6} study the relation between the transfer budget $B$ and the transfer outcome.

These evaluation study resulted into an interesting observation. From the experiments, selecting the source of transfer by \ac{u} has led to slightly better performance in the \ac{pp} environment. That cannot be confirmed by the Cart-Pole scenario, but that might be because of the reduced action space combined with non-sparse reward function. 

Contrary to the \ac{ss}, the experiments have highlighted the need of establishing a criteria to filter the incoming knowledge on a target side, and therefore have highlighted the importance of~\textbf{RQ3}. In fact, sampling by~\ac{rdc} has registered overall the lowest performance. On the contrary, \ac{lec} and \ac{hdc} has shown comparable performance. However, in the Cart-Pole environment, the \ac{hdc}  approach has demonstrated a lower volatility in return when compared to \ac{lec}.

Lastly, the transfer budget $B$ has shown a contradictory trend depending on the evaluated environment. An increased budget has led to an improvement in the Cart-Pole scenario, whereas it has yielded the opposite outcome in the \ac{pp} environment, where a lower budget has resulted in higher performance returns. The positive transfer effect observed in Cart-Pole with an increased transfer budget may be attributed to the binary action decision and the non-sparse reward function.


\subsection{\ac{efontl} Adaptation to Heterogeneous Dynamics}\label{ss:efotnl-rs-sumo-expanded}
This section evaluates objective~\textit{O8}, assessing the adaptability of~\ac{efontl} in environments with heterogeneous dynamics. To evaluate this objective, three different scenarios were designed to assess the impact of \ac{efontl} in the \ac{rs-sumo} environment. In two of these scenarios, the agents are exposed to different dynamics. 

Table~\ref{tab:rs-sumo-scenarios} presents the three scenarios used in this evaluation. The two scenarios with heterogeneous dynamics are \textit{Single-Agent Mixed Train Data} and \textit{Multi-Agent Same Train Data}. \textit{Multi-Agent Same Train Data}, where the heterogeneity arises by the non-overlapping zones, and this has already been discussed in Section~\ref{ss:eval_efontl_in_rs-sumo}. 
The heterogeneity is due by the non-overlapping zones.
On the other hand, in the \textit{Single-Agent Mixed Train Data} scenario, two agents are trained during the morning peak hours while the others on the evening peak hours. Even though agents in both scenarios operate within unique ride-request demand settings, \ac{efontl} enables the processing of experiences from agents subjected to different dynamics and therefore learning a different policy.

\begin{table}[h!]
	\small
	\renewcommand*{\arraystretch}{1.1}
	\caption{\label{tab:rs-sumo-scenarios} \ac{rs-sumo}: \ac{efontl} setup across the three evaluated scenarios.}
	\centering
	\begin{threeparttable}
		\begin{tabular}{|c|c|c|c|c|c|}
			\hline
			\multirow{2}{*}{\textbf{Scenario ID}}&\textbf{Train}&\textbf{Test} & \textbf{Number} & \multirow{2}{*}{\textbf{Multi-Agent}}\\
			&\textbf{Demand Set}& \textbf{Demand Set}& \textbf{of Agents} &\\
			\hline
			\textit{Single-Agent Same Train Data}& \multirow{2}{*}{Morning} & \multirow{2}{*}{Evening} & \multirow{2}{*}{4} &\multirow{2}{*}{\xmark} \\ ~Sec.~\ref{ss:efotnl-rs-sumo-expanded}&&&&\\\hline
			\textit{Single-Agent Mixed Train Data}& \multirow{1}{*}{Morning} & \multirow{2}{*}{Evening} & \multirow{2}{*}{4} & \multirow{2}{*}{\xmark}  \\
			~Sec.~\ref{ss:efotnl-rs-sumo-expanded} &Evening&&&\\\hline
			\textit{Multi-Agent Same Train Data}& \multirow{2}{*}{Morning} & \multirow{2}{*}{Evening} & \multirow{2}{*}{4} & \multirow{2}{*}{\cmark}  \\	
			~Sec.~\ref{ss:eval_efontl_in_rs-sumo} &&&&\\
			\hline
		\end{tabular}
	\end{threeparttable}
\end{table}

In addition to these two scenarios, \textit{Single-Agent Same Train Data} is included to assess whether \ac{efontl} can achieve positive transfer when transferring across agents in tasks with common dynamics, similarly to what was observed in the Cart-Pole scenario in Section~\ref{sss:efontl-vs-baselines_Cart-Pole}.

The \textit{Single-Agent Same Train Data} and \textit{Single-Agent Mixed Train Data} scenarios presented in this section involve four agents interacting with independent copies of the environment. These two scenarios differ in terms of the requests used by the agents to learn their policies. In \textit{Single-Agent Same Train Data}, all agents are trained using data from the morning peak hours~($7-10$am), while in \textit{Single-Agent Mixed Train Data}, two agents learn from data collected during the morning peak hours, and the other two agents learn from data collected during the evening~($6-9$pm). In both scenarios the \ac{efontl} settings matches the one used for the multi-agent scenario and are specified in Table~\ref{tab:efontl-parameters}.

In this evaluation, we included \textit{Single-Agent Mixed Train Data} to investigate the effect of transfer across agents addressing different demand sets. To provide a comprehensive analysis and assess whether transferring between agents learning on different demand sets leads to similar results as the preliminary studies presented in Chapter~\ref{cpt:preliminary_studies}, we also considered the scenario \textit{Single-Agent Same Train Data}, in which agents are enabled to transfer while learning to serve ride-requests from the same demand set.



While the ride-requests used for training the policies in the two scenarios are sourced from different datasets, the test requests are all drawn from the same dataset on the evening peak hours.

\begin{figure}[htb!]
	\centering
	\includegraphics[width=1\columnwidth]{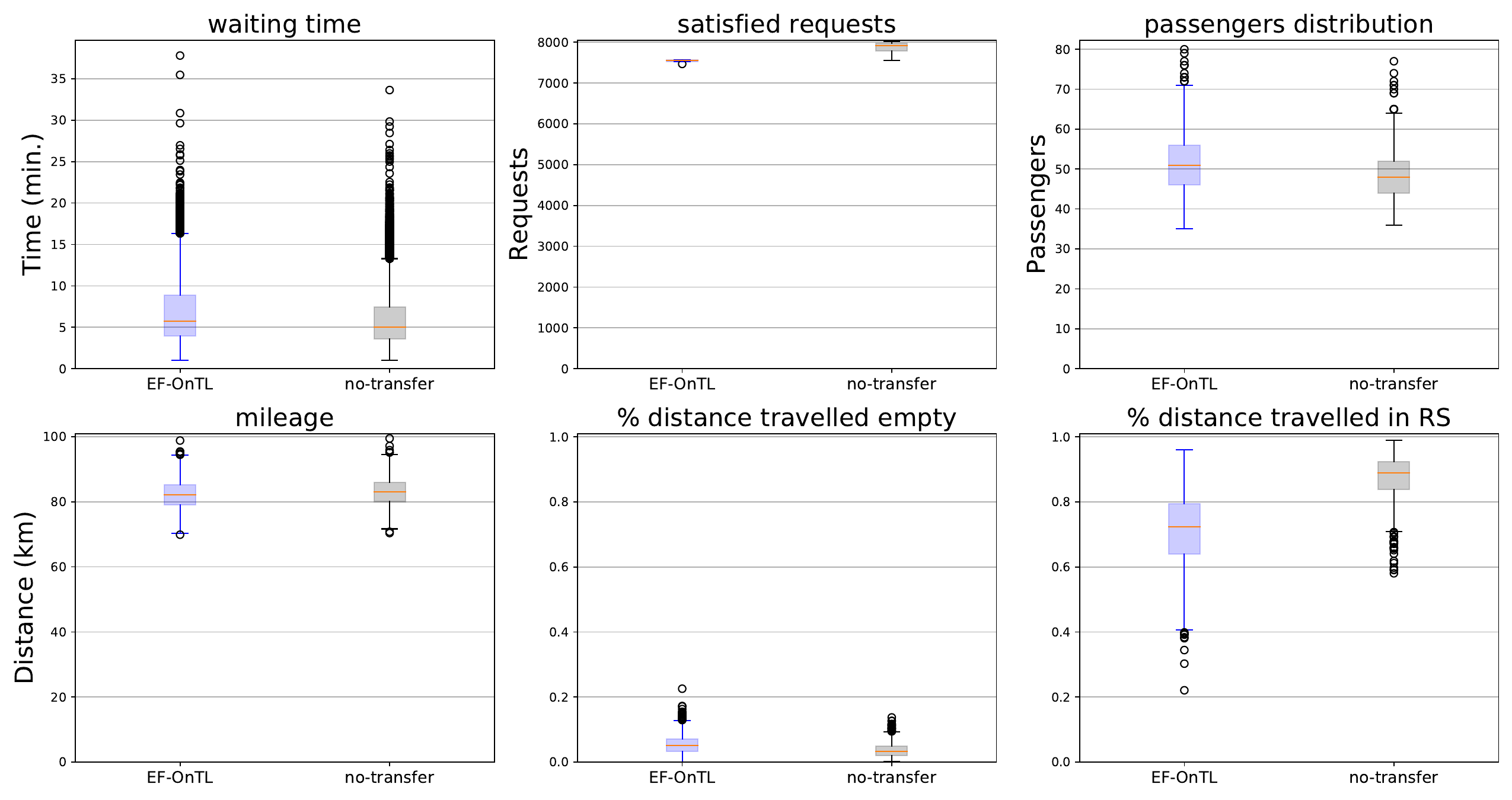}
	\caption{\textit{Single-Agent Same Train Data} - train on $7-10$am test on $6-9$pm.}
	\label{fig:rs_sumo_same-demand-set}
\end{figure}

\begin{figure}[htb!]
	\centering
	\includegraphics[width=1\columnwidth]{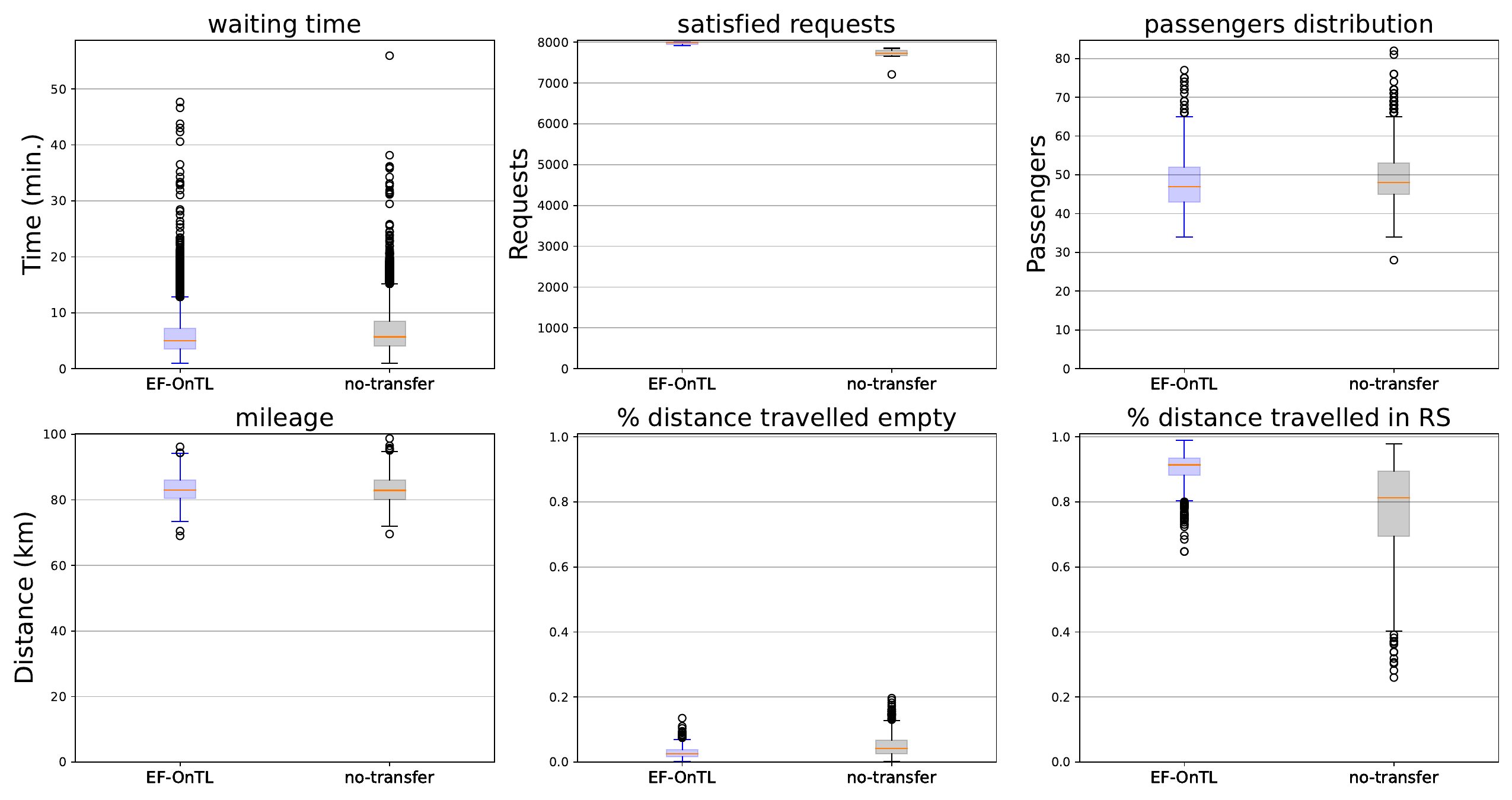}
	\caption{\textit{Single-Agent Mixed Train Data} - train on both $7-10$am and $6-9$pm and test on $6-9$pm.}
	\label{fig:rs_sumo_different-demand-set}
\end{figure}

In both scenarios, \ac{efontl} is compared against \textit{no-transfer} method. The results of \textit{Single-Agent Same Train Data} are reported in Figure~\ref{fig:rs_sumo_same-demand-set}, where agents are trained on the morning peak hours and tested on the evening. On the other hand, the results of \textit{Single-Agent Mixed Train Data} are reported in Figure~\ref{fig:rs_sumo_different-demand-set}, where the agents tested are those that are trained and tested on the evening peak hours while enabling the sharing with other agents which are trained on the morning hours.

In the \textit{Single-Agent Same Train Data} scenario, the agents exhibit a different behaviour although both \ac{efontl} and \textit{no-transfer} share the very same reward model. It appears that \textit{no-transfer} agents have likely learnt to maximise the number of requests served through ride-sharing. On the other hand, \ac{efontl} agents prioritise requests with more than one passenger, even if it means avoiding ride-sharing. As a result, the distribution of passengers per vehicle is higher for \ac{efontl} than for the \textit{no-transfer} method, while the distance travelled in ride-sharing and the number of satisfied requests are slightly lower. In conclusion, the difference in the number of served requests is negligible, and the choice of the method preferred depends on the desired behaviour.

In the \textit{Single-Agent Mixed Train Data} scenario, where transfer occurs between agents trained on different demand sets originating from the same geographical area, it appears that \ac{efontl} has enhanced the generalisation ability of the agents. This improvement has led to an increase in the number of requests served compared to the \textit{no-transfer} method.  Although the additional requests have resulted in extra distance travelled in ride-sharing, the total distance travelled has not increased compared to the \textit{no-transfer} scenario.

In relation to evaluation objective \textit{O8}, we observed that \ac{efontl} enabled a fleet of $200$ vehicles to achieve an improvement in the total number of requests served when transferring knowledge between agents serving during both the morning and evening peak hours. This improvement in total served requests aligns with what we previously observed in the preliminary results chapter, as presented in Section~\ref{ss:preliminary-envs-rs-sumo}, where the increase in terms of satisfied requests was approximately $3\%$. Furthermore, when examining the roles of the transfer sources in the \textit{Single-Agent Mixed Train Data} scenario, we noted that these roles remained relatively stable over time, with one agent being the primary source of transfer in most instances.

The \textit{Single-Agent Mixed Train Data} has demonstrated that enabling the transferring across agents that are learning from different demand set leads a fleet to serve a higher number of requests. 

While \ac{efontl} was not originally designed for enabling this type of transfer, future research should focus on adapting and facilitating transfer between agents specialised in different dynamics. \ac{sarnd} has proven unsuitable for selecting the transfer source in these scenarios, as one agent consistently serves as the source of transfer most of the time. In contrast, \ac{efontl} should promote varying roles among agents to enhance overall performance.

\section{Summary}\label{sec:summary_eval}

This chapter has presented the evaluation objectives, the baselines and the benchmark environments used to assess \ac{efontl}.  

We compared~\ac{efontl} against multiple baselines. Firstly, \textit{no-transfer}, to measure the improvement against common~\ac{rl} agents that do not share any information. Secondly, two \ac{tl} baselines, \ac{ocmas} and \ac{rcmp}, to compare \ac{efontl} versus two action-based teacher-student methods. Thirdly, we compared centralised \ac{marl} algorithms, QMIX and \ac{maddpg}, and \ac{efontl} in terms of performance.

\ac{efontl} and the baselines are evaluated across four different benchmark environments of increasingly complexity: Cart-Pole, \ac{pp}, \ac{hfo} and a real-world simulated environment~\ac{rs-sumo}. Table~\ref{tab:sim-baselines} reports the baselines implemented in each environment.

In the \ac{rs-sumo} environment we designed three different scenarios to observe \ac{efontl} effect while allowing the transfer across tasks subjected to different underlying dynamics. 

To address the research question~\textbf{RQ1}, we observed that \ac{efontl} can effectively improve the agents performance involved in a multi-agent system. However, in the \textit{Single-Agent Same Train Data} scenario of the \ac{rs-sumo} environment, \ac{efontl} has resulted into decreased performance, as shown in Figure~\ref{fig:rs-sumo_multibrain}. 

The decline in performance, however, may be attributed to the different dynamics of the studied scenario as each agent visits ride-requests originated within a restricted and unique geographical sub-area of the Manhattan borough. 
Nevertheless,  \ac{efontl} led to an increased number of requests served when enabling the transfer across agents trained on different demand sets originating from the same area as shown in Figure~\ref{fig:rs_sumo_different-demand-set}. 

When compared against \ac{ocmas} and \ac{rcmp}, \ac{efontl} has proven to be a viable alternative without disrupting the exploration process of a target agent.  However, in complex environments, if an optimal expert is available to provide on-demand advice, it should be the preferred choice.  Nevertheless, if the teacher is not optimal, \ac{efontl} should be favoured, as the target agent may be constrained by the suboptimal expertise of the teacher. On the other hand, the \ac{marl} baselines, QMIX in \ac{pp} and \ac{maddpg} in \ac{hfo}, have shown an higher convergence time, possibly because of the \ac{marl} extra step to learn the joint state-action values function.

To address the research questions~\textbf{RQ2} and \textbf{RQ3}, this thesis has evaluated multiple transfer settings in the Cart-Pole and \ac{pp} benchmark environment. The evaluation focused on the transfer impact given by \ac{ss}, \ac{tcs} and transfer budget $B$. 

We observed overall that \ac{u}, for selecting the source of transfer, and \ac{hdc}, for selecting the transfer content, led to better overall performance in both environments. However, while it was not possible to establish a single criterion for \textbf{RQ3} in Cart-Pole, the experiments have demonstrated that randomly selecting tuples to be integrated into the target learning process had no significant impact on the final performance.

From the experiments, we could see that each transfer parameter had a different effect depending on the environment encountered by the agents. In Cart-Pole, which has a binary action selection process and a dense reward model, the budget $B$ demonstrated the most significant impact on resulting performance. We observed a greater improvement with a larger budget, $B$.
On the other hand, in \ac{pp}, which features a sparse reward model and agents frequently revisiting states with no useful information, we observed better overall performance with a smaller budget.

In the end, we noticed that selecting both the source of transfer and the experiences to be integrated into the target learning process based on uncertainty estimation worked better overall. The dynamic identification of the experiences is based on the comparison between the source and target uncertainties to prevent sending redundant information that have already adequately being explored by the target agent. While using \ac{u} and \ac{hdc} gave more consistent results, how well \ac{efontl} performs ultimately depends on all the transfer parameters and the specific addressed task.

\chapter{Conclusion}
\label{cpt:conclusion}\acresetall%

This thesis has introduced \ac{efontl}, a novel online experience sharing framework designed to facilitate \ac{tl} in multi-agent systems where no fixed expert is available. In this chapter, we provide a brief summary of the achievements of this thesis in Section~\ref{sec:achievements} and discuss its limitations in Section~\ref{sec:limitations}. Finally, in Section~\ref{sec:future_work}, we conclude this dissertation by proposing future research directions.

\section{Thesis Contribution}\label{sec:achievements}
This thesis has focused on the challenging problem of enabling dynamic transfer learning in online scenarios, particularly in situations where traditional fixed expert models are absent to advise \ac{rl} agents. The main goal of this thesis has been to study and design an online transfer learning framework based on experience sharing. 
The main contribution of this thesis is \ac{efontl}, an expert-free transfer learning framework designed for online experience sharing among agents in a multi-agent environment.  \ac{efontl} enables dynamic teacher selection, allowing the system to choose the most suitable agent to be used as source of transfer at every transfer step, and dynamic transfer content selection, enabling target agents to select the most valuable experience to enhance their policies. 

To enable these dynamic selection processes, this thesis also presents \ac{sarnd}, a method designed to enhance \ac{rnd} as an uncertainty estimator in online contexts. 


In the design of the \ac{tl} framework, this thesis has identified and addressed a set of research questions.

Firstly, \textbf{RQ1} - \vcchange{can, and if so to what extent, online transfer learning through sharing of experience across homogeneous agents with no fixed expert contribute to improving the system performance?} Secondly, \textbf{RQ2} - what criteria can agents use to identify the suitable agent to be used a source of transfer? Thirdly, \textbf{RQ3} - what criteria should an agent use to filter incoming knowledge?


This thesis began with a feasibility study to analyse the impact of experience sharing in a reduced set of challenges given by an offline transfer learning context. Consequently, acknowledging the early results, this thesis introduced \acf{efontl}. \ac{efontl} stands as a transfer learning framework tailored for multi-agent systems that effectively eliminates the need for a pre-defined expert teacher by employing a dynamic selection process. This process chooses a temporary expert at each transfer step based on the real-time performance metrics of the agents, such as average cumulated rewards in recent episodes and average uncertainty in recent experiences.

The chosen temporary expert serves as the source of transfer, sharing a portion of its collected experience with the other agents, which are designated as target agents. These target agents filter and sample a batch of experiences from the transferred buffer to be integrated into their own learning processes. To prioritise certain experiences, a target agent relies on metrics like uncertainty and expected surprise. Uncertainty is compared against the source agent's uncertainty, while the expected surprise is estimated as the loss of the underlying deep learning model.

The state-of-the-art approaches enabling transfer across homogeneous agents are mostly based on fixed expert agents, where already trained agents provide advice to novel learning agents. Furthermore, the criteria used to decide whether to provide advice or not are often based on a single-side, either the source or target of transfer. 

The teacher-student framework can be adapted to enable the transfer across learning agents, as demonstrated by Ilhan et al.~\cite{ilhan2019teaching}. However, agents need continuous interaction to establish whether an advice is needed or not. 

\ac{efontl} overcomes the need for a fixed expert agent and the need for continuous interaction by transferring a batch of experiences aimed at improving the learning model of a target agent. Furthermore, the experiences transferred are sampled on criteria defined for both source and target agents.

In conclusion, the answers to our posed research questions are as follows:

\begin{itemize}

	\item \textbf{\vcchange{Can, and if so to what extent, online transfer learning through sharing of experience across homogeneous agents with no fixed expert contribute to improving the system performance?}} $-$ Compared to a traditional \ac{rl} scenario where agents do not share any information, \ac{efontl} has successfully shown the capability to improve the agents' performance when applied to multiple homogeneous agents or multi-agent systems. 	While expert-based action-advice \ac{tl} approaches can potentially limit the performance of target agents, \ac{efontl} prevents such negative impacts on a target agent's performance. This is achieved by eliminating the requirement for an expert agent and enabling dynamic teacher selection at each transfer step. 
		
	The extent of improvement shown by \ac{efontl} is closely associated to the complexity of the task being undertaken by the agents, a trend that has been similarly observed among other \ac{tl} baselines. In the \ac{hfo} benchmark environment, \ac{efontl} enables the offensive team to double the number of goals scored when compared to independent \ac{rl} agents. Additionally, \ac{efontl} achieves comparable performance to expert-based \ac{tl} systems.  However, it is worth noting that \ac{efontl} typically requires nearly double the amount of time to converge to a similar level of performance.

	\item \textbf{What criteria can agents use to identify the suitable agent to be used as source of transfer?} $-$ in this thesis we have investigated two different criteria to select the source of transfer for \ac{efontl}, \ac{u} and \ac{bp}, that agents can use to identify a suitable agent to be used as source of transfer within a transfer step. \ac{u} selects as source of transfer the agent with lowest average uncertainty across the experiences stored in its transfer buffer. On the other hand,	\ac{bp} selects as source of transfer the agent with highest cumulated reward within the most recent episodes interval. 	
	
	Based on the experiments presented in this work, the most effective criterion used by agents in \ac{efontl} to dynamically select the source of transfer is \ac{u}. 
	Nevertheless, we have observed that the criteria used for selecting the source of transfer may have minimal impact on the final performance in certain tasks, such as Cart-Pole. Further investigation is necessary to determine whether this limited impact is attributed to the simplicity of the Cart-Pole task.

	\item \textbf{What criteria should an agent use to filter incoming knowledge?} $-$ in this thesis we have presented three different criteria that agents could use to sample relevant experiences to be integrated into their learning processes, \ac{rdc}, \ac{hdc} and \ac{lec}. These criteria are based on $\Delta$\textit{-conf}, which expresses the difference between the target and source uncertainty, and expected surprise, defined solely over target agent and approximated through temporal difference error. 	
	\ac{rdc} enables an agent to randomly select the experiences from tuples with a $\Delta$\textit{-conf} higher than the median value computed across all tuples. 
	\ac{hdc} enables an agent to sort in decreasing order the incoming experiences by the $\Delta$\textit{-conf} and to select the top entries to be integrated within its learning model. 
	\ac{lec} enables an agent to balance expected surprise and uncertainty. To balance the different scales, uncertainty and surprise values are normalised within a common interval and weighted equally. Finally, agent selects the tuples with the highest values to be integrated within its learning model.

	 Selecting the experience to be integrated by \ac{rdc} has demonstrated the lowest overall impact on the system performance. On the other hand, \ac{lec} and \ac{hdc} have both shown a better performance improvement. However, selecting the experiences by \ac{hdc}, and thereby comparing the uncertainties of sender and receiver agent, has consistently led to a performance improvement. 
\end{itemize}

Given the pivotal role of uncertainty, the final performance of the presented framework depends on the capacity of the estimator model used to approximate the agent's epistemic uncertainty. 
In this thesis, we have presented \ac{sarnd}, an extension of \ac{rnd}, designed to overcome the limitations of \ac{rnd} in estimating the epistemic uncertainty of a learning agent. 
To summarise, \ac{rnd} estimates uncertainty based solely on the visited state. We believe that the state alone is insufficient for estimating the uncertainty in a learning agent since it does not provide information about the action taken. 
\ac{sarnd} extends \ac{rnd} by including into the estimation the action, the next state and the reward observed. Based on the comparison presented in Section~\ref{ss:rnd_sarnd_comparison}, \ac{sarnd} successfully identifies different actions while recognising previously visited states and allowing for an uncertainty trend similar to that of \ac{rnd}.

Finally, while sampling the experiences by~\ac{hdc} certainly provided more consistent results, the final performance given by \ac{efontl} is dependent of the combination of the transfer settings parameters, which need to be carefully evaluated based on the addressed task. 


\vcchange{In the current state of the art, experience sharing is mostly used in offline transfer learning as a phase prior to the exploration. However, this thesis has demonstrated that combining independently collected experience with external sources may be beneficial to the learning agents.}

\vcchange{When an agent with deep expertise, either human or artificial, is available to guide the exploration of learning agents, action-advice transfer methods may be preferred as these ensure a higher return over a reduced time frame. However, having an expert dedicated to this mentoring task is expensive. Thus, combining both expert-based and online experience sharing may allow to keep the same level of performance while lowering the expert-related costs.}

\vcchange{
However, given the impracticality of having an oracle agent, the expert agent, specifically human, may limit its choices to a subset of familiar actions. This constraint narrows the target agent's learning opportunities. In contrast, sharing raw experience is less likely to invalidate the learning phase of an agent. In fact, in the event of negative transfer, the target agent may take slightly longer to converge, but its performance will not be compromised.}

\section{Limitations}\label{sec:limitations}

While the results of this thesis are promising, there are several limitations to consider. The training time for policies is increased when compared to agents that do not transfer any information. This increase in training time is not caused by the dynamic selection of source and experiences  during each transfer step but rather by the optimisation step of the uncertainty estimator model. 

The \ac{sarnd} model is updated after each visited tuple, which involves back-propagation of gradients. While a single back-propagation step is relatively fast, over millions of interactions, it introduces a noticeable delay. To address this, back-propagation could be delayed and aggregated over a small batch of samples, but this would result in a delay in updating uncertainties.

The impact of experience sharing in~\ac{efontl} requires substantial time to be observed when compared to action as advice. When sharing action-advice it is possible to assess whether the transfer is leading a target agent towards positive transfer or not. This assessment requires the monitoring of the behaviour exhibited by the target agent.  On the other hand, in \ac{efontl}, with the sharing of experiences, it is hard to discern whether a certain behaviour is due to the transferred knowledge or self-exploration of an agent. From a human perspective, the sharing of experience is a black box where it is nearly impossible, or at least more complicated, to appreciate the value of the advice when compared to one-shot action.

In addition, our results have shown that \ac{efontl} can occasionally lead to negative transfer with certain transfer settings. When implementing \ac{efontl} from scratch, it can be challenging to determine whether the transfer will result in a positive or negative outcome until the simulation is completed. In contrast, when transferring from an experienced agent through the teacher-student framework, which relies on action-advice, it allows for faster estimation of whether the transfer will lead to a positive or negative outcome.


\section{Future Work}\label{sec:future_work}

When considering transfer learning among homogeneous agents, experience sharing is often not the primary choice. Instead, many \ac{tl} frameworks, following the teacher-student paradigm, opt to enable agents to share actions as advice. In this context, \ac{efontl} has demonstrated that enabling dynamic sharing of selected batches of experiences is a viable alternative to the traditional action-advice based teacher-student framework.

The effectiveness of the teacher-student framework is often dependent on the availability of an expert agent to provide guidance, and it often necessitates extensive experimentation to fine-tune the algorithm parameters.

In contrast, \ac{efontl} does not rely on a fixed expert agent to be effective. It has been designed with a forward-looking perspective, envisioning a future where transfer is self-regulated, aiming to eliminate the need for intricate parameter adjusting.

With this direction in mind and considering the observed results, we believe that having fixed criteria for selecting the source of transfer and determining the interactions to be integrated into the learning process may not be optimal. As a future work direction, we propose enabling dynamic criteria selection to choose both the source of transfer and the interactions to be integrated into a target agent's learning process. We believe that using multiple criteria based on the system's status could further enhance the system performance. 

An additional research direction concerns the investigation of  dynamic adjustment of the transfer budget $B$ and its impact on the final performance, when compared to a fixed budget. Furthermore, our experiments did not consider the frequency of transfer, so a necessary step for the future development of~\ac{efontl} is to assess the effect of increased transfer frequency with a smaller budget against lower transfer frequency with an increased budget.

In addition to \ac{efontl}, this thesis introduced \ac{sarnd} to extend the \ac{rnd} model used for estimating epistemic uncertainty. In the future, we plan to refine \ac{sarnd} by reducing its input to state-action pairs and assess the impact that this adjustment might have on both the estimation of epistemic uncertainty and the transfer effect.

\vcchange{Finally, the findings of this thesis could be integrated into other \ac{rl} methodologies that necessitate the prioritisation of specific samples. An intriguing application would be the adaptation of \ac{efontl} to shape a dataset for offline \ac{rl} tasks. Subsequently, it would be insightful to train an agent prioritising certain tuples, deemed to be the most beneficial for learning. Comparing the performance of this approach against other sampling methods from the dataset may result in enhancing the learning outcomes.}


\setcounter{section}{0}
\chapstar{Appendix} \label{cpt:appendix}


\renewcommand*{\theHsection}{chY.\the\value{section}}
\renewcommand{\thesection}{\Alph{section}}
\section{\ac{hfo} State}\label{app:hfo_state}
For the experiments carried out in this research, perceived states consists of $95$ values. From the manual available online~\cite{hfo_manual}:
\begin{itemize}
	\item 0 Self\_Pos\_Valid [Valid] Indicates if self position is valid.
	\item 1 Self\_Vel\_Valid [Valid] Indicates if the agent's velocity is valid.
	\item 2-3 Self\_Vel\_Ang [Angle] Angle of agent's velocity.
	\item 4 Self\_Vel\_Mag [Other] Magnitude of the agent's velocity.
	\item 5-6 Self\_Ang [Angle] Agent's Global Body Angle.
	\item 7 Stamina [Other] Agent's Stamina: Low stamina slows movement.
	\item 8 Frozen [Boolean] Indicates if the agent is Frozen. Frozen status can happen when tackling or being tackled by another player.
	\item 9 Colliding\_with\_ball [Boolean] Indicates the agent is colliding with the ball.
	\item 10 Colliding\_with\_player [Boolean] Indicates the agent is colliding with another player.
	\item 11 Colliding\_with\_post [Boolean] Indicates the agent is colliding with a goal post.
	\item 12 Kickable [Boolean] Indicates the agent is able to kick the ball.
	\item 13-15 Goal Center [Landmark] Center point between the goal posts.
	\item 16-18 Goal Post Top [Landmark] Top goal post.
	\item 19-21 Goal Post Bot [Landmark] Bottom goal post.
	\item 22-24 Penalty Box Center [Landmark] Center of the penalty box line.
	\item 25-27 Penalty Box Top [Landmark] Top corner of the penalty box.
	\item 28-30 Penalty Box Bot [Landmark] Bottom corner of the penalty box.
	\item 31-33 Center Field [Landmark] The left middle point of the RoboCup field (note that this is not the center of the HFO play area).
	\item 34-36 Corner Top Left [Landmark] Top left corner HFO Playfield.
	\item 37-39 Corner Top Right [Landmark] Top right corner HFO Playfield.
	\item 40-42 Corner Bot Right [Landmark] Bottom right corner HFO Playfield.
	\item 43-45 Corner Bot Left [Landmark] Bottom left corner HFO Playfield.
	\item 46 OOB Left Dist [Proximity] Proximity to the nearest point of the left side of the HFO playable
	area. E.g. distance remaining before the agent goes out of bounds in left field.
	\item 47 OOB Right Dist [Proximity] Proximity to the right field line.
	\item 48 OOB Top Dist [Proximity] Proximity to the top field line.
	\item 49 OOB Bot Dist [Proximity] Proximity to the bottom field line.
	\item 50 Ball Pos Valid [Valid] Indicates the ball position estimate is valid.
	\item 51-52 Ball Angle [Angle] Agent's angle to the ball.
	\item 53 Ball Dist [Proximity] Proximity to the ball.
	\item 54 Ball Vel Valid [Valid] Indicates the ball velocity estimate is valid.
	\item 55 Ball Vel Mag [Other] Magnitude of the ball's velocity.
	\item 56-57 Ball Vel Ang [Angle] Global angle of ball velocity.
	\item 8$T$ Teammate Features [Player] One teammate feature set (8 features) for each teammate active in HFO, sorted by proximity to the agent.
	\item 8$O$ Opponent Features [Player] One opponent feature set (8 features) for each opponent active in HFO, sorted by proximity to the player.
	\item 1$T$ Teammate Uniform Nums [Unum] One uniform number for each teammate active in HFO,
	sorted by proximity to the agent.
	\item 1$O$ Opponent Uniform Nums [Unum] One uniform number for each opponent active in HFO,
	sorted by proximity to the player.
	\item +1 Last\_Action\_Success\_Possible [Boolean] Whether there is any chance the last action taken was successful, either in accomplishing the usual intent of the action or (primarily for the offense) in some other way such as getting out of a goal-collision state.
\end{itemize}

Where $xT$ means that there are $x$ value for each teammate and $xO$ $x$ values for each opponent.

\section{\acs{efontl} - Transfer Criteria Evaluation in \acf{pp}}\label{app:expanded_pp_res}

\begin{figure}[htb!]
	\centering
	\begin{subfigure}[b]{.5\columnwidth}
		\centering
			\includegraphics[width=1\columnwidth]{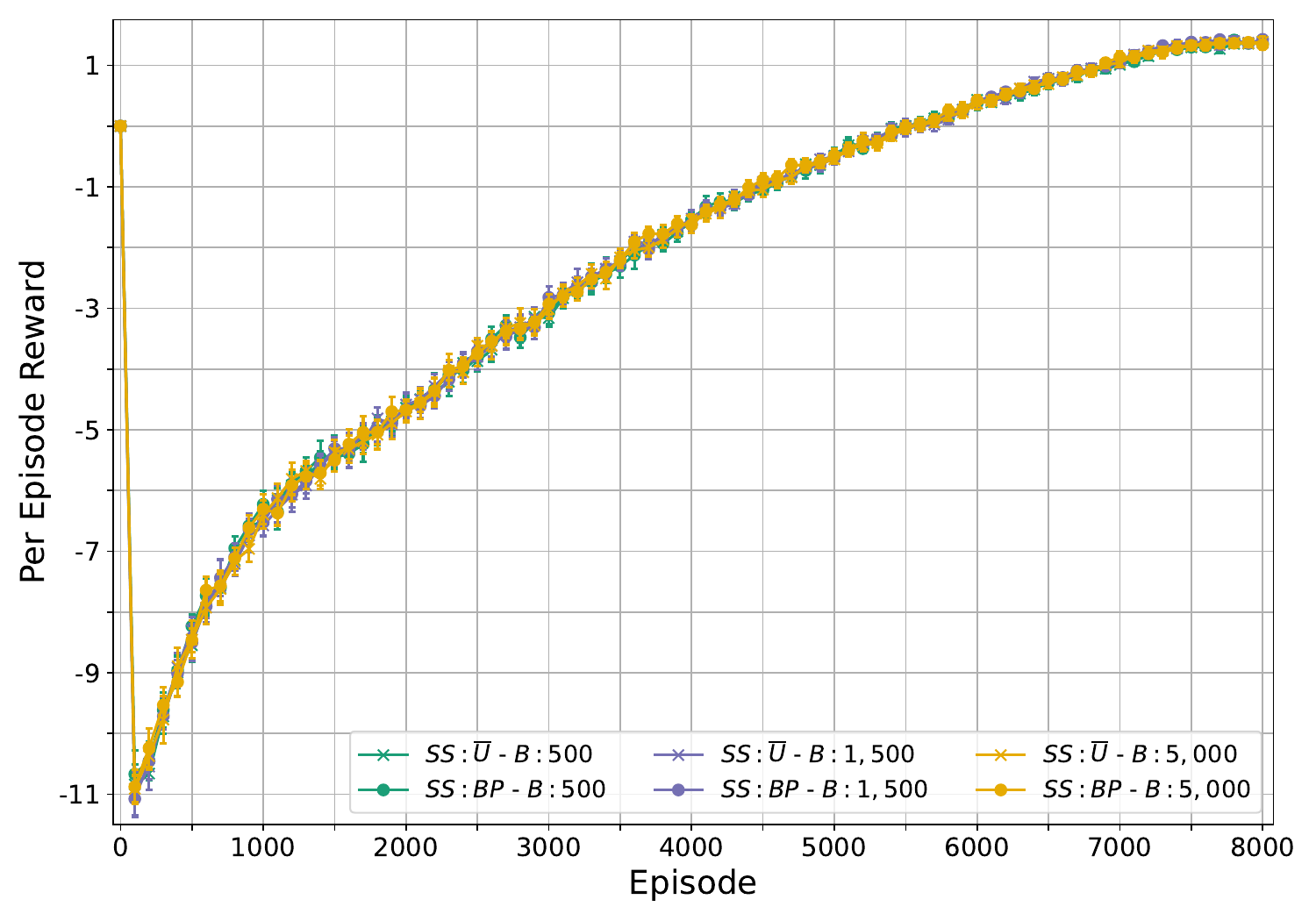}
			\caption{\acs{rdc}}
	\end{subfigure}~
	\begin{subfigure}[b]{.5\columnwidth}
		\centering
		\includegraphics[width=1\columnwidth]{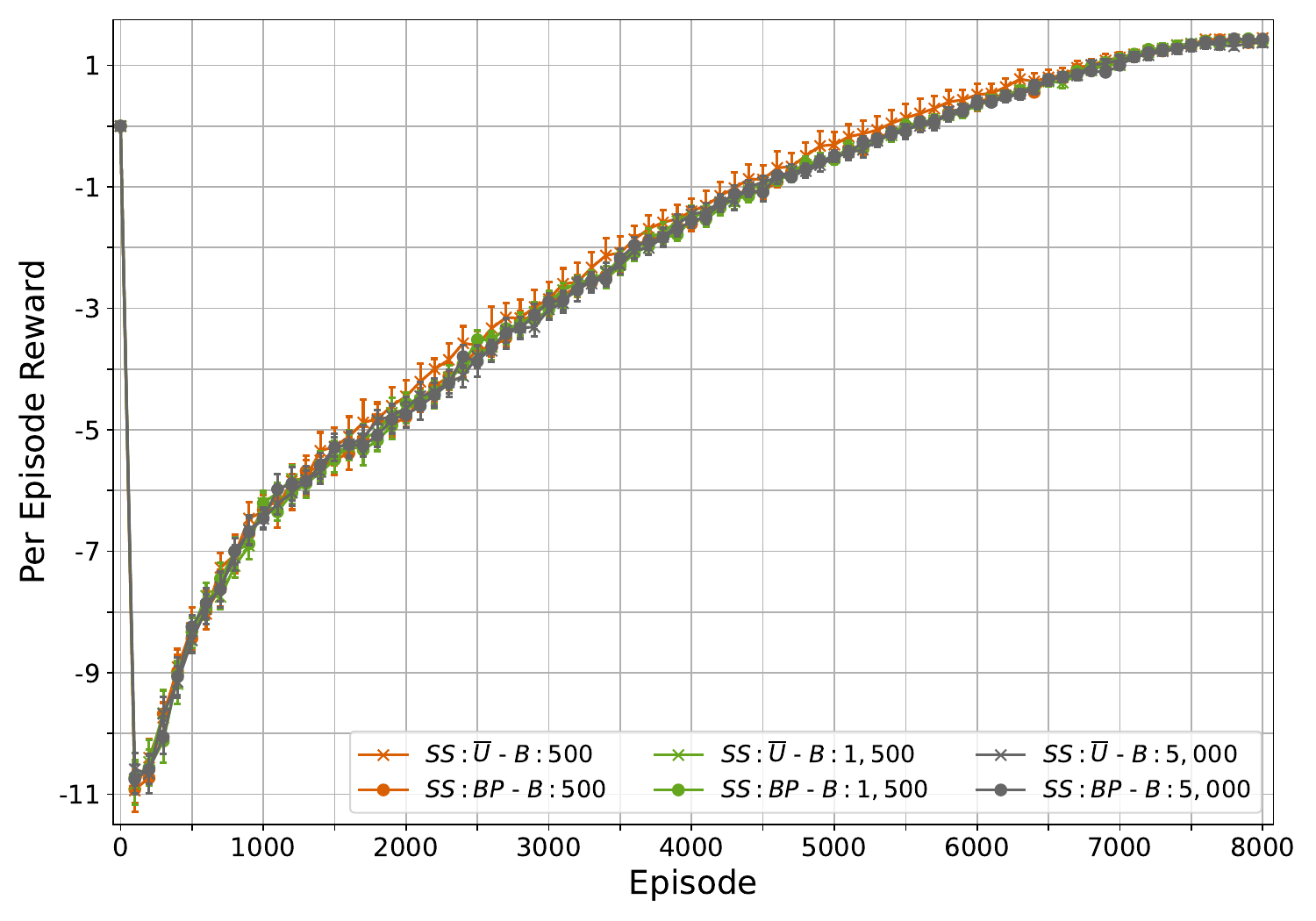}
		\caption{\acs{hdc}}
	\end{subfigure}
	\begin{subfigure}[b]{.5\columnwidth}
		\centering
		\includegraphics[width=1\columnwidth]{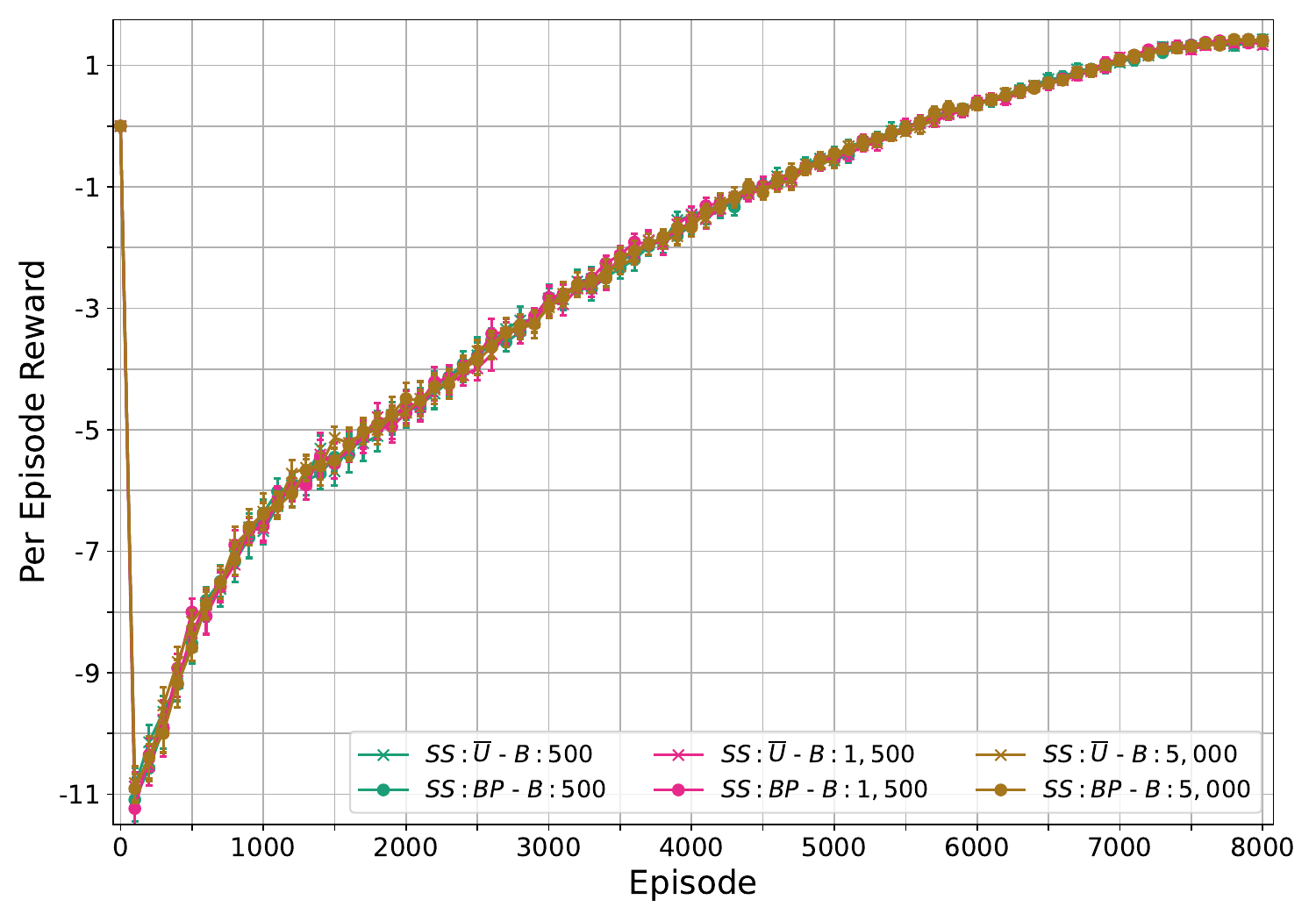}
		\caption{\acs{lec}}
	\end{subfigure}
	\caption{\ac{efontl} learning curves organised by~\ac{tcs} used by a target agent to filter incoming knowledge.}
	\label{fig:efontl-pp-tcs}
\end{figure}

\begin{figure}[htb!]
	\centering
	\begin{subfigure}[b]{.5\columnwidth}
		\centering
		\includegraphics[width=1\columnwidth]{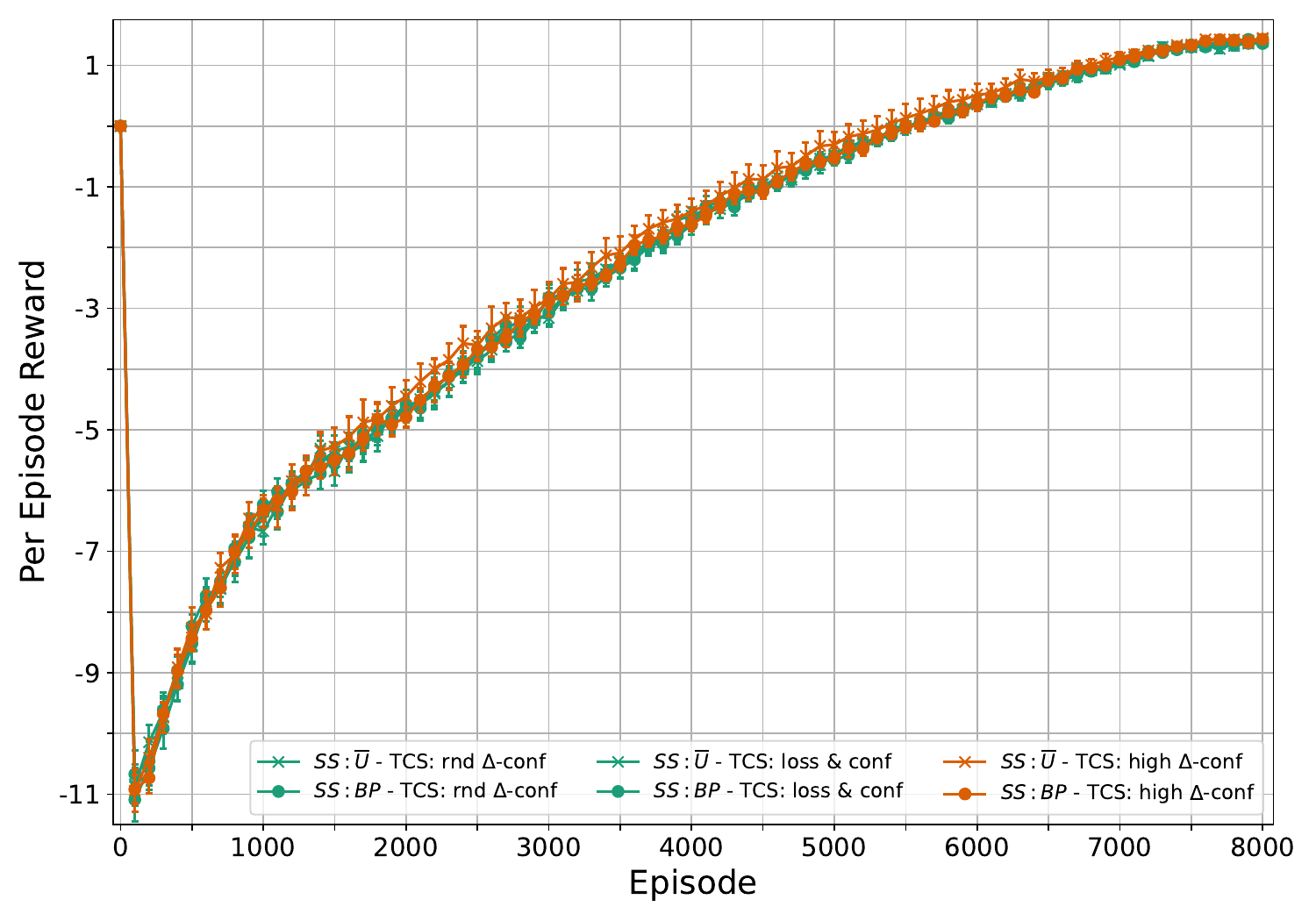}
		\caption{$B=500$}
	\end{subfigure}~
	\begin{subfigure}[b]{.5\columnwidth}
		\centering
		\includegraphics[width=1\columnwidth]{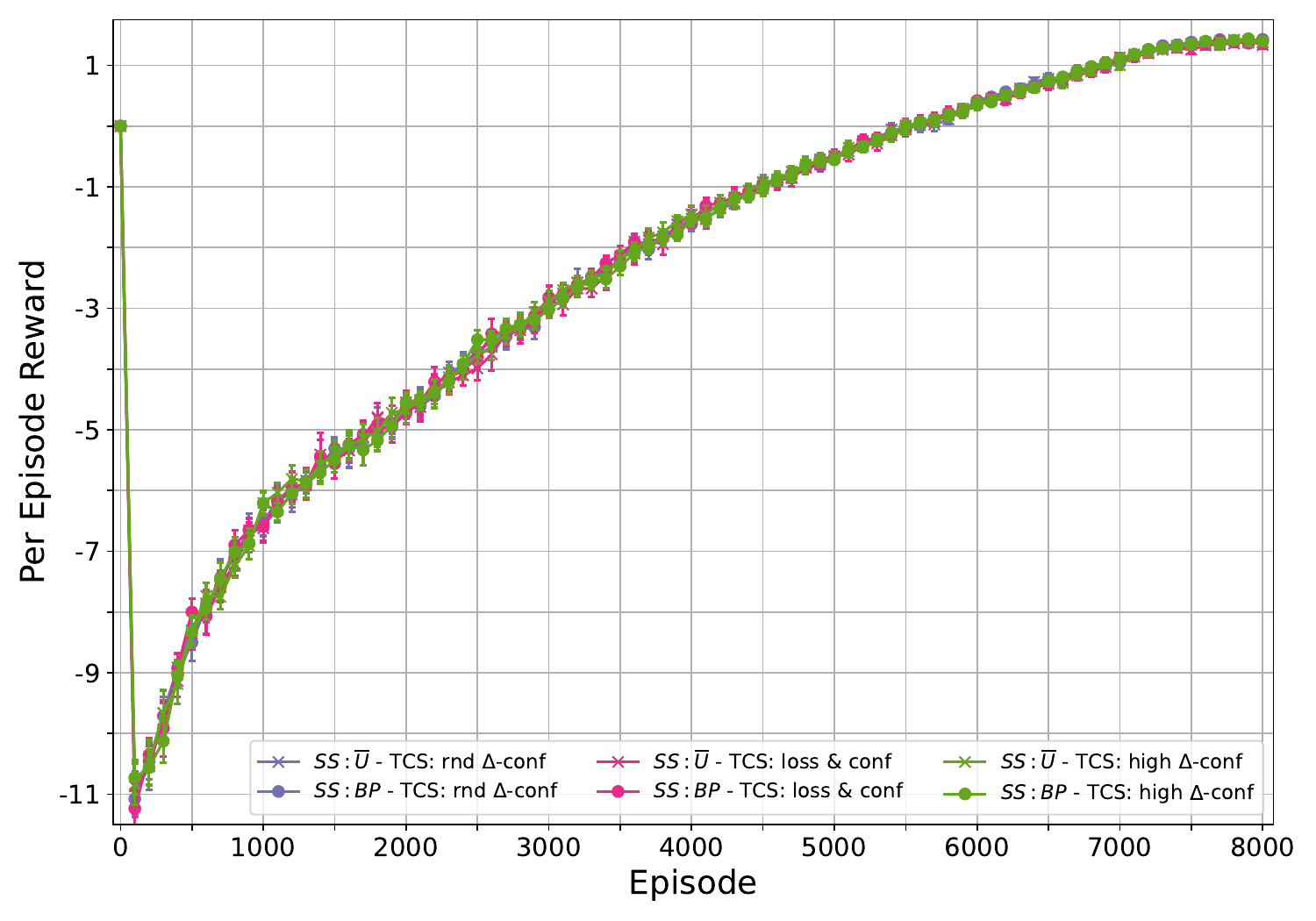}
		\caption{$B=1,500$}
	\end{subfigure}
	\begin{subfigure}[b]{.5\columnwidth}
		\centering
		\includegraphics[width=1\columnwidth]{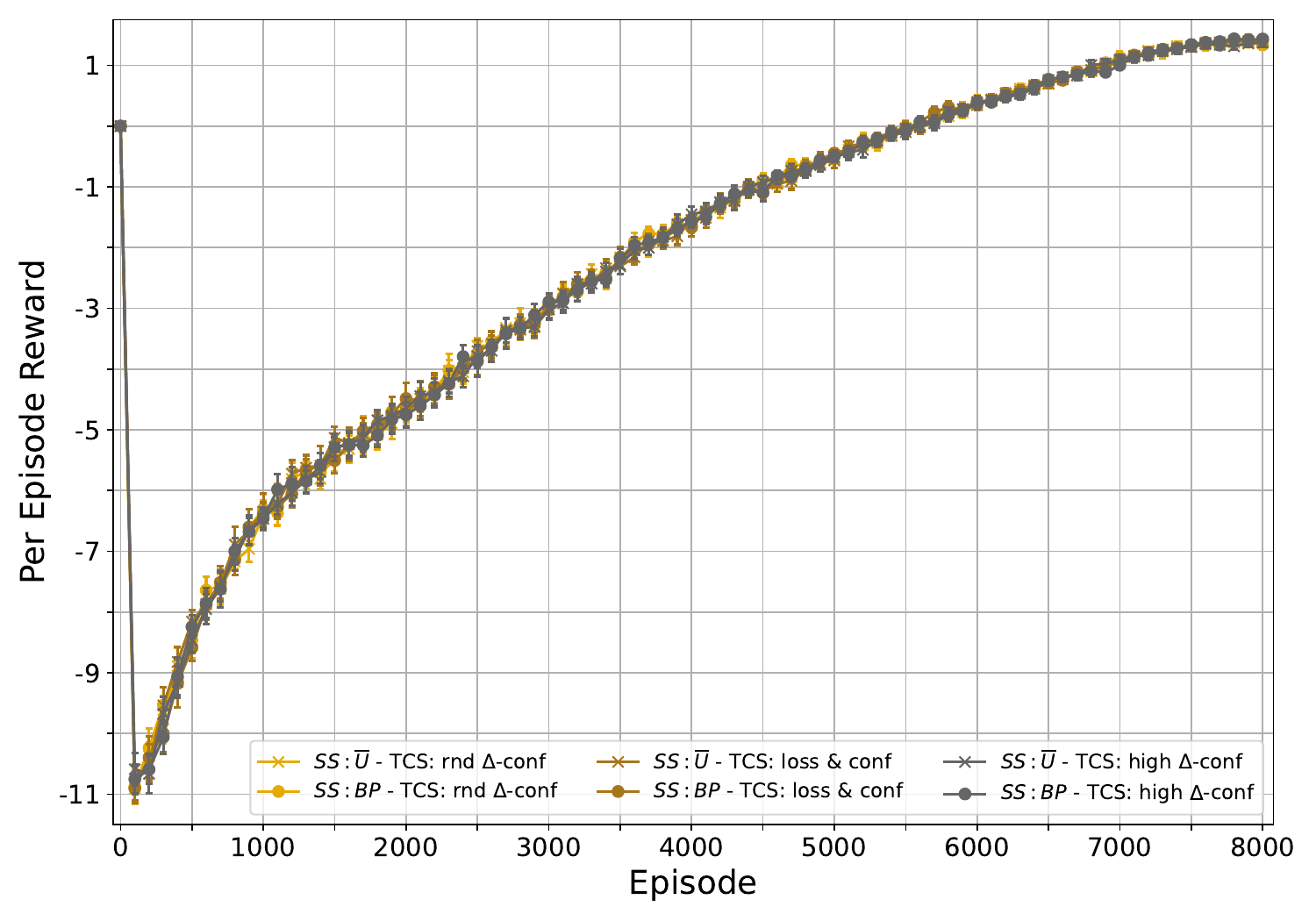}
		\caption{$B=5000$}
	\end{subfigure}
	\caption{\ac{efontl} learning curves organised by the transfer budget~$B$ in \ac{pp}.}
	\label{fig:efontl-pp-tbsize}
\end{figure}

This section presents the expanded results obtained by~\ac{efontl} while varying the transfer settings in the \ac{pp}. The evaluation matches the results shown in Figure~\ref{fig:efontl-comparison-pp}. However, here we show the learning curves compared by the~\ac{tcs}, shown in Figure~\ref{fig:efontl-pp-tcs}, used to filter the experiences on a target agent and by the budget $B$ used within a transfer step, shown in Figure~\ref{fig:efontl-pp-tbsize}.

\clearpage
\section{\acl{rcmp} - Analysis}\label{app:rcmp_analysis}
This section investigates the low performance registered by the \ac{rcmp} method. In particular,  it explores the motivations behind the negative transfer in Cart-Pole and includes an analysis of threshold sensitivity in the \ac{pp} environment.

\subsection{Cart-Pole}

In the Cart-Pole benchmark environment, as shown in the evaluation graph in Figure~\ref{fig:efontl_vs_baseline-CartPole}, \ac{rcmp} demonstrates an initial jump-start in performance from the beginning up to episode $200$. Subsequently, there is a brief decline in performance, followed by a gradual increase in the reward starting from episode $500$ onward.

\begin{figure}[htb!]
	\centering
	\includegraphics[width=0.75\columnwidth]{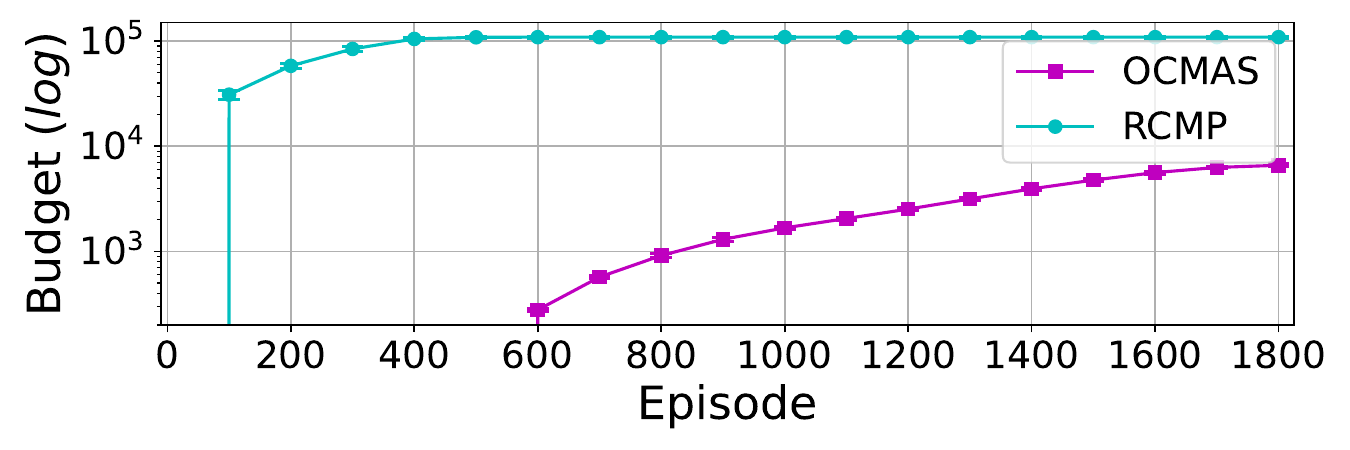}
	\caption{Budget utilisation of \acs{ocmas} and \ac{rcmp} in the Cart-Pole environment. This graph complements Figure~\ref{fig:efontl_vs_baseline-CartPole}.}
	\label{fig:cart-pole-budget-baselines}
\end{figure}

The notable jump-start is a result of the continuous provision of advice from the expert teachers. However, the subsequent drop in performance can be attributed to a lack of advice caused by an inadequate allocation of the budget, as shown in Figure~\ref{fig:cart-pole-budget-baselines}. This issue arises due to the threshold used to trigger the advice process.

The normalisation process applied to maintain a consistent scale for uncertainty, along with the binary action decision, results in the ensemble used as an uncertainty estimator showing low granularity in uncertainty estimation.  Uncertainty falls into three possibilities based on the agreement among the five heads: $0$ when all heads agree, $0.2$ when a single head disagrees, and $0.3$ when two heads disagree with the others.

\subsection{\acl{pp}}
\begin{figure}[htb!]
	\centering
	\includegraphics[width=0.65\columnwidth]{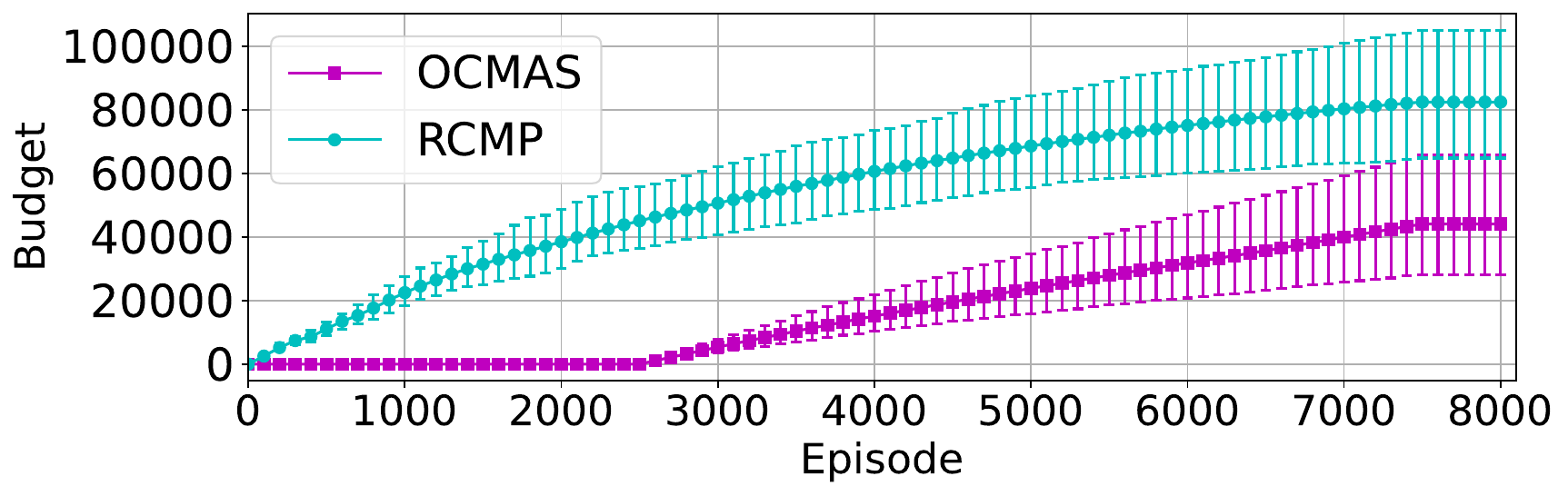}
	\caption{Budget utilisation of \acs{ocmas} and \ac{rcmp} in the \ac{pp} environment. This graph complements Figure~\ref{fig:PP-lc-efontl-vs-baselines}.}
	\label{fig:pp-budget-baselines}
\end{figure}

In the \ac{pp} benchmark environment, we observed a more balanced budget utilisation throughout the episodes, as shown in Figure~\ref{fig:pp-budget-baselines}. Due to the discrete action decision process involving multiple actions, the uncertainty estimated by the ensemble allowed for a sufficient range of values to ensure a fair budget allocation.

In the \ac{pp} environment, we have studied the performance of \ac{rcmp} in relation to the selected threshold that initialise the advising process. For comparison, we present the learning curves, the \ac{pp} metrics, including the average number of preys caught from the opposite team, and the budget utilisation. 

\begin{figure}[h!]
	\centering
	\begin{subfigure}[b]{.85\columnwidth}
		\centering
		\includegraphics[width=\columnwidth]{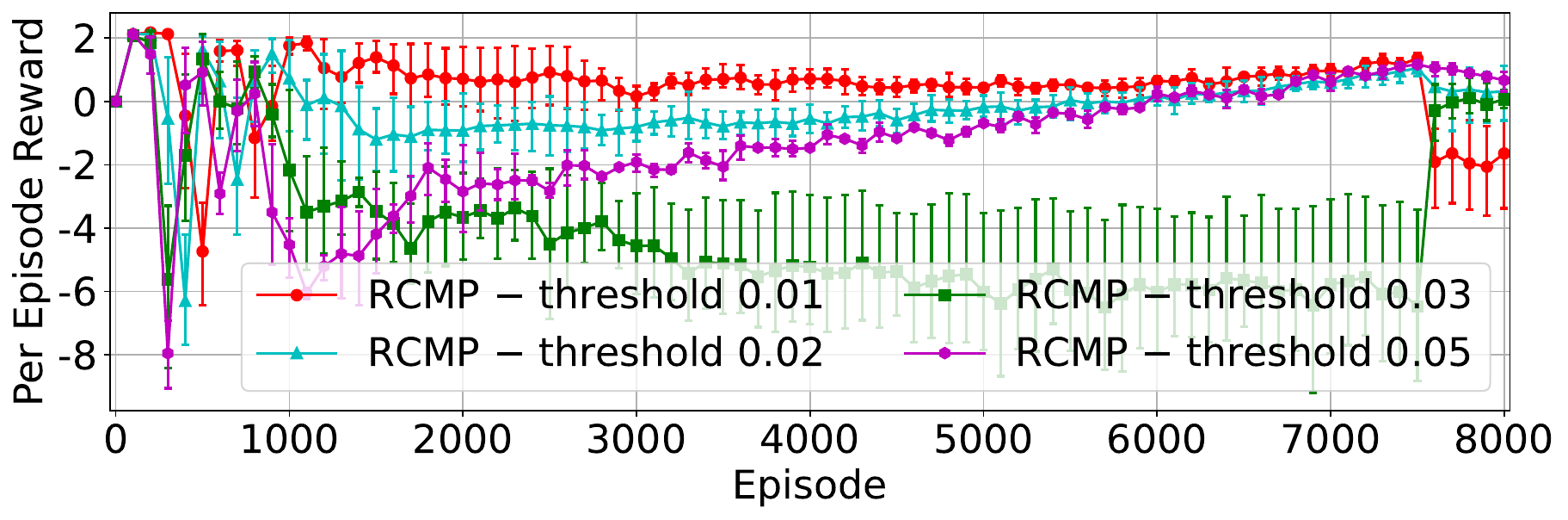}
		\caption{Learning curves.}
	\end{subfigure}
	\begin{subfigure}[b]{.85\columnwidth}
		\centering
		\includegraphics[width=1\columnwidth]{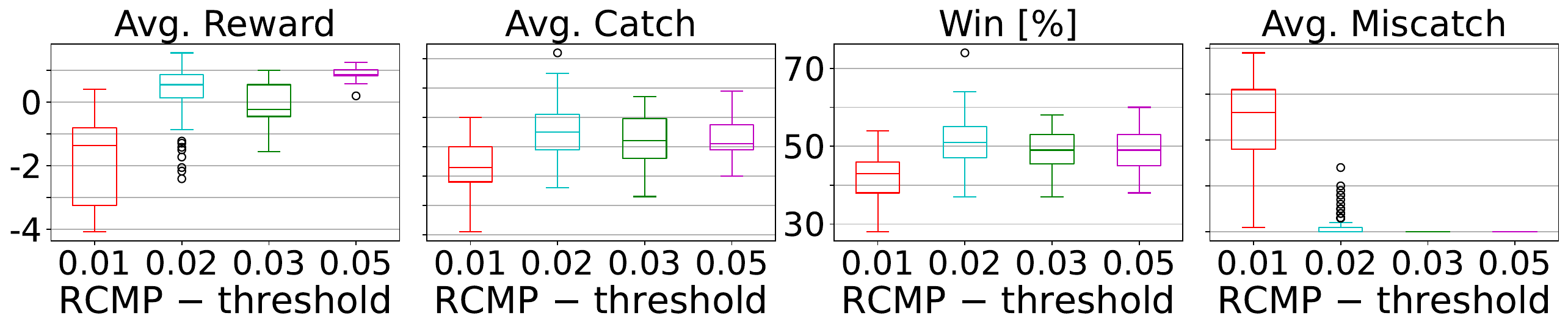}
		\caption{Evaluation metrics.}
	\end{subfigure}
	\begin{subfigure}[b]{.8\columnwidth}
		\centering
		\includegraphics[width=1\columnwidth]{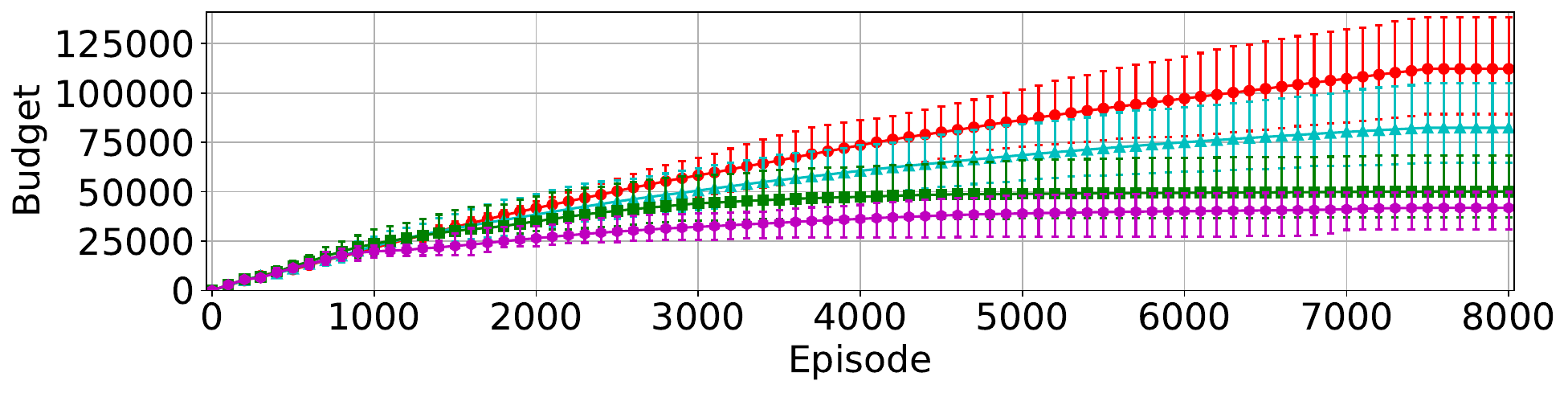}
		\caption{Budget consumption.}
	\end{subfigure}
	\caption{\ac{rcmp} performance in the \ac{pp} benchmark environment with different thresholds.}
	\label{fig:pp-rcmp-thresholds}
\end{figure}

The results are reported in Figure~\ref{fig:pp-rcmp-thresholds}, where we evaluated five different threshold levels: $0.01$, $0.02$, $0.03$, and $0.05$. The results presented in the thesis, and that are shown in Figure~\ref{fig:PP-lc-efontl-vs-baselines}, are based on a threshold of $0.02$.

A lower threshold leads the agent to be guided for longer, resulting in a higher cumulative reward during the training. However, the final performance in the tracked metrics are significantly lower. For instance, with a threshold of $0.01$ we observed an overall higher cumulative reward until episode $7,000$, as shown in Figure~\ref{fig:pp-rcmp-thresholds}a. Eventually, this lower threshold led to a deterioration of performance in the tracked metrics, as shown in Figure~\ref{fig:pp-rcmp-thresholds}b. It is possible that the target agents have learned to prioritise the catch action, which leads to catching the wrong prey and negatively impacted their overall performance.

Increasing the threshold to $0.05$, results in lower budget utilisation, as shown in Figure~\ref{fig:pp-rcmp-thresholds}c. Consequently, the learning curve during the initial episodes, up to episode $4,000$, is lower compared to a more balanced threshold like $0.02$. Nevertheless, the final tracked metrics remain comparable between the two thresholds. The main difference is that a higher reduces the prioritisation of the catch action and therefore prevents the capturing of adversarial prey.

From the analysis presented in this section, it is evident that \ac{rcmp} necessitates significant experimentation to investigate the correlation between transfer outcomes and the threshold used by target agents.


\chapstar{Acronyms} \label{cpt:acronyms}\acresetall

\begin{acronym}[JSONP] 
	\acro{rs-sumo}[3R2S]{Ride-Sharing Ride-Requests Simulator}
	\acro{a2a}[A2A]{Agent-to-Agent}
	\acro{bp}[BP]{Best Performance}
	\acro{ccl}[CCL]{Cooperative Centralised Learning}
	\acro{ctde}[CTDE]{Centralised Training with Decentralised Execution}
	\acro{ddpg}[DDPG]{Deep Deterministic Policy Gradient}
	\acro{dqn}[DQN]{Deep Q-Network}
	\acro{drl}[DRL]{Deep Reinforcement Learning}
	\acro{efontl}[\textit{EF-OnTL}]{Expert-Free Online Transfer Learning}
	\acro{hfo}[HFO]{Half Field Offense}
	\acro{hdc}[\textit{high $\Delta$-conf}]{transfer based on Higher Delta Confidence}
	\acro{horizon}[$H$]{Horizon}
	\acro{il}[IL]{Independent Learners}
	\acro{lec}[\textit{loss~\&~conf}]{transfer based on Higher Loss and Confidence combined}
	\acro{maddpg}[MADDPG]{Multi-Agent Deep Deterministic Policy Gradient~\cite{lowe2017multi}}
	\acro{marl}[MARL]{Multi-Agent Reinforcement Learning}
	\acro{mas}[MAS]{Multi-Agent System}
	\acro{ml}[ML]{Machine Learning}
	\acro{mdp}[MDP]{Markov Decision Process}
	\acro{mse}[MSE]{Mean Squared Error}
	\acro{pp}[MT-PP]{Multi-Team Predator-Prey}
	\acro{nn}[NN]{Neural Network}
	\acro{ocmas}[\textit{OCMAS}]{Online Confidence-Moderated Advice Sharing}
	
	\acro{pomdp}[POMDP]{Partially-Observable Markov Decision Process}
	\acro{paddpg}[PA-DDPG]{Parameterised Action Deep Deterministic Policy Gradient}
	\acro{ppo}[PPO]{Proximal Policy Optimisation}
	\acro{rcmp}[\textit{RCMP}]{Requesting Confidence-Moderated Policy}
	\acro{rl}[RL]{Reinforcement Learning}
	\acro{rnd}[RND]{Random Network Distillation}
	\acro{rdc}[\textit{rnd $\Delta$-conf}]{transfer Randomly based on threshold determined by Delta Confidence}
	\acro{stpp}[ST-PP]{Single-Team Predator-Prey}
	\acro{sarnd}[\textit{sars-RND}]{State Action Reward Next-state Random Network Distillation}
	\acro{ss}[SS]{Source Selection Criteria}
	\acro{sumo}[SUMO]{Simulation of Urban MObility~\cite{SUMO2018}}
	\acro{vdn}[VDN]{Value Decomposition Network}
	\acro{td}[TD]{Temporal-Difference}
	\acro{tcs}[TCS]{Transfer Content Selection Criteria}
	\acro{tl}[TL]{Transfer Learning}
	\acro{t2t}[T2T]{Task-to-Task}
	\acro{u}[$\overline{U}$]{Average Uncertainty}
	\end{acronym}

%
\fancypagestyle{empty}{\pagestyle{bibmine}}

\begin{singlespacing}
    \cleardoublepage\pagestyle{bibmine}%
    \phantomsection%
    \addcontentsline{toc}{chapter}{Bibliography}%
    \bibliographystyle{IEEEtran}
    \bibliography{bib/thesis}
\end{singlespacing}


\cleardoublepage%
\pagestyle{mine}%
\fancypagestyle{empty}{\pagestyle{mine}}

\begin{singlespacing}
\small
\printindex
\end{singlespacing}

\end{document}